\documentclass[11pt]{article}

\usepackage[preprint]{acl}

\usepackage{times}
\usepackage{latexsym}
\usepackage{amssymb}
\usepackage[T1]{fontenc}

\usepackage[utf8]{inputenc}

\usepackage{microtype}

\usepackage{inconsolata}
\usepackage{amsmath}
\usepackage{svg}
\usepackage{graphicx}
\usepackage[table]{xcolor}
\usepackage{booktabs}
\usepackage{cleveref}
\usepackage{subcaption}
\usepackage{multirow}
\usepackage{enumitem}

\newcommand{\ecoparagraph}[1]{\noindent\textbf{#1}\;}

\title{Extending LLM Context via Associative Recurrent Memory}

\author{\bf Gleb Kuzmin\textsuperscript{1,4,8} \quad
Ivan Rodkin\textsuperscript{2,6} \quad
Aydar Bulatov\textsuperscript{3,6} \quad
Yuri Kuratov\textsuperscript{3,6} \\
\bf Lyudmila Rvanova\textsuperscript{1} \quad
Mikhail Katkov\textsuperscript{9,10} \quad
Ilia Sochenkov\textsuperscript{7} \quad
Misha Tsodyks\textsuperscript{9,10} \\
\bf Timothy Baldwin\textsuperscript{2} \quad
Mikhail Burtsev\textsuperscript{5} \quad
Artem Shelmanov\textsuperscript{2} \\
\textsuperscript{1}FusionBrain Lab ~
\textsuperscript{2}MBZUAI ~
\textsuperscript{3}Cognitive AI Systems Lab ~
\textsuperscript{4}RUDN \\
\textsuperscript{5}London Institute for Mathematical Sciences ~
\textsuperscript{6}MIRAI ~
\textsuperscript{7}Lomonosov Moscow State University \\
\textsuperscript{8}Laboratory for Analysis and Controllable Text Generation Technologies RAS \\ 
\textsuperscript{9}School of Natural Sciences, Institute for Advanced Study, Princeton \\
\textsuperscript{10}Department of Brain Sciences, Weizmann Institute of Science \\
\href{mailto:kuzmin.gyu@gmail.com}{kuzmin.gyu@gmail.com} ~~
\href{mailto:artem.shelmanov@mbzuai.ac.ae}{artem.shelmanov@mbzuai.ac.ae}
}

\begin{document}
\maketitle
\begin{abstract}

Extending the context length of large language models (LLMs) is critical for many real-world applications, yet standard transformers remain constrained by quadratic compute and linear memory scaling. In this work, we investigate the Associative Recurrent Memory Transformer (ARMT) as a practical approach for enabling long-context processing in LLMs, constant memory scaling, and better efficiency. We make three main contributions. First, we construct two domain-specific long-context datasets designed to evaluate realistic workloads, focusing on narrow-domain fine-tuning scenarios. Second, we propose a comprehensive training recipe for ARMT-based context extension, combining continued pre-training, synthetic long-context data generation, curriculum learning, and selective integration of associative memory into chosen model layers. Third, we present an extensive experimental study demonstrating that ARMT-augmented models: (i) process inputs well beyond their original context limits without degrading performance relative to in-limit baselines; (ii) generalize more effectively to out-of-distribution context lengths; and (iii) need 30\% less FLOPs while preserving baseline performance within the original context window. 
\end{abstract}

\section{Introduction}

Long-context understanding is crucial for many tasks, such as processing and understanding technical and financial reports, software development, and multi-document reasoning in scientific and legal domains. These scenarios often require models to integrate information distributed across hundreds of thousands or even millions of tokens. However, standard transformer architectures~\citep{vaswani2017attention} struggle to scale to such contexts, as the computational and memory costs of self-attention grow quadratically with sequence length. Moreover, transformer performance degrades as the context length increases~\citep{liu2023lost, kuratov2024babilong}.
Therefore, since the introduction of the transformer architecture, long-context processing has emerged as a central and rapidly-evolving research direction~\citep{beltagy2020longformer, katharopoulos2020transformers, bulatov2022rmt}.  Traditionally, efficient long-context approaches have been built using recurrent architectures~\citep{gu2023mamba,peng-etal-2023-rwkv}; however, such models must typically be trained from scratch, limiting the ability to leverage existing pre-trained LLMs. Moreover, fully-recurrent LMs update the memory at each time step, which complicates high-level information processing in tasks such as structured copying~\citep{jelassi2024repeat} and instruction following~\citep{park2024can}.

Recent studies~\citep{bulatov2024beyond, rodkin2024associative} have explored enhancing transformers with segment-wise context processing and recurrent memory mechanisms. Using human memory as an analogy \cite{cowan2008differences}, full attention within a segment models short-term/working memory, while the module that recurrently propagates crucial information from segment to segment can be viewed as long-term memory.
These approaches preserve strong intra-segment modeling performance while enabling linear scaling with respect to context length. 

In this work, we focus on the Associative Recurrent Memory Transformer (ARMT: \citet{rodkin2024associative}), which introduces a capacious segment-level associative memory and features strong scaling to extremely long input sizes. Prior work on ARMT-based models has been limited to scales below 200M parameters and evaluated on a narrow set of tasks~\citep{rodkin2024associative}, leaving their behavior at larger model sizes mostly unexplored. Models at this scale typically struggle to handle complex real-world workloads. In this work, we extend ARMT to small- and medium-sized LMs (1B parameters), which are substantially more capable in practical settings. These models provide a practical middle ground, enabling linear-compute, constant-memory long-context processing while maintaining strong performance on real-world tasks.

\begin{figure*}[t]
    \centering
    \includegraphics[width=0.75\textwidth]{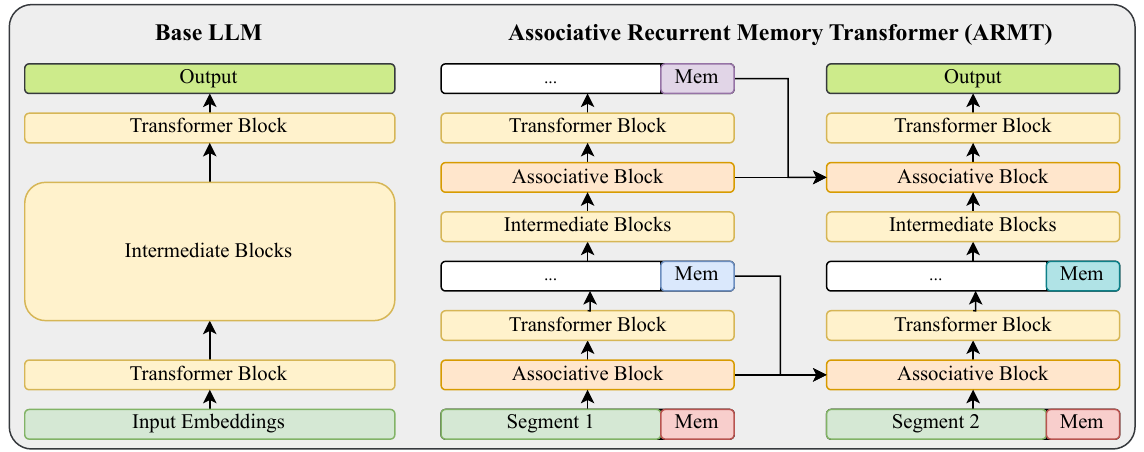}
    \caption{Base LLM architecture (left) and ARMT architecture (right). ARMT divides the input text into segments and processes them sequentially, allowing the model to handle long contexts.}
    \label{fig:armt_model}
\end{figure*}

Our contributions are as follows:
\begin{itemize}[noitemsep, topsep=0pt]
    \item We construct two new datasets over real-world tasks to train and evaluate the long-context performance of LLMs, focusing on narrow domain fine-tuning.
    \item We propose a training recipe for extending the effective context of LLMs using ARMT. This approach incorporates continued pre-training, synthetic generation of training data, curriculum learning, and the strategic integration of the associative memory mechanism into selected LLM layers.
    \item We present an experimental study on extending the context windows of state-of-the-art LLMs using ARMT. Our findings demonstrate that: (1) ARMT-augmented models process inputs \textbf{well beyond their original context limits} without degrading performance relative to in-limit baselines; (2) they exhibit \textbf{superior generalization} to out-of-distribution context lengths compared to base models; and (3) they \textbf{need 30\% less FLOPs} while maintaining baseline performance \textbf{within the original context limits}.
\end{itemize}

\section{Related Work and Background}

In recent years, the demand for long-context task handling has grown as the problem-solving capabilities of LLMs have evolved~\citep{openai2023gpt4_turbo,reid2024gemini,anthropic2024claude3}. However, even for narrow tasks, both small and large models experience sharp performance degradation when the context size increases. Additionally, scaling the transformer's context is computationally expensive due to the quadratic computational complexity of attention. This problem has been addressed by sparse attention~\citep{child2019generating,zaheer2020big}, attention linearization~\citep{wang2020linformer, katharopoulos2020transformers}, and recurrent approaches. Modern RNNs are increasing in popularity due to their linear scaling with length: xLSTM~\citep{beck2024xlstm} builds on traditional recurrent architectures to achieve efficient scaling, RWKV~\citep{peng-etal-2023-rwkv} and state space models such as Mamba~\citep{gu2023mamba} and Gated DeltaNet~\citep{yang2024gated} utilize simple linear recurrence to achieve faster parallel training. 
Fully-recurrent models exhibit strong performance on some long-context tasks; however, they still face significant limitations compared to transformers in complex algorithmic tasks~\citep{jelassi2024repeat, merrill2024illusion}, instruction following~\citep{park2024can}, and long-context reasoning~\citep{kuratov2024babilong}. The ability to efficiently address the context is limited by the bottleneck in the fixed-size recurrent state. Hybrid models with interleaving attention and recurrent layers~\citep{lieber2024jamba, funemotron, qwen35blog} partially address these performance issues, but they still face expensive quadratic compute scaling with context size and require training from scratch. 

Models with segment-level recurrence present a promising middle ground: they retain linear scaling on longer sequences and maintain the strong short-context performance of transformers with full attention~\citep{dai2019transformerxl, raecompressive}. ARMT~\citep{rodkin2024associative} has been shown not only to scale linearly with context size, but also to efficiently process tens of millions of tokens on selected tasks. Scaling ARMT opens up possibilities to develop LLMs that are compute-efficient, potentially scalable to extremely long contexts, and maintain all the strengths of full attention within each segment.

An overview of the ARMT architecture is presented in~\Cref{fig:armt_model}. The backbone LLM is passed to a wrapper that enables segment-wise processing of input segments, splitting long-context inputs into non-overlapping segments of fixed length. Each segment is also augmented with trainable memory embeddings (``Mem''), which are processed throughout the model. 
At the core of ARMT lies the associative block, a layerwise memory module that retrieves and updates key-value associations from previous segments and injects the recalled information into the current segment representation.
The associative memory mechanism consists of three parts:
\begin{itemize}[noitemsep, topsep=0pt]
    \item \textbf{Memory extraction}: each transformer layer compresses an input segment into memory embeddings.
    \item \textbf{Memory consolidation}: memory embeddings are then consolidated in a per-layer associative matrix as key-value pairs.
    \item \textbf{Association}: every embedding of the following segment is transformed into a query vector and multiplied by the associative matrix. 
\end{itemize}
Appendix \ref{app:formal} presents the formal definition of the associative block.

ARMT combines the best of both worlds: the ability of recurrent architectures to propagate information across arbitrarily long contexts (in principle) and the strong performance of full self-attention within a limited context window.
In contrast to Mamba \citep{gu2023mamba} and RWKV \citep{peng-etal-2023-rwkv}, ARMT’s computational depth scales with sequence length, enabling more effective multi-step reasoning by combining precise local attention with deep, high-capacity long-term memory \citep{rodkin2024associative, kuratov2024babilong}.

\section{LLM Context Extension via ARMT}\label{sec:methods}

To enable LLMs to handle long contexts, we developed a set of techniques essential for effective ARMT training: continued pre-training, synthetic long context supervision, curriculum learning, and pruning of ARMT layers.

\ecoparagraph{Continued pre-training.}
When associative memory is introduced into a pre-trained LLM, the newly added parameters remain uninitialized. As a result, although the backbone LLM is already pre-trained, ARMT still requires additional adaptation to learn how to use the memory mechanism effectively. We hypothesize that continued unsupervised language-model pre-training can properly initialize the associative memory parameters and simplify subsequent task-specific fine-tuning. This stage should be performed on sufficiently long contexts so that the memory mechanism is actively used to propagate salient information across multiple segments. In our experiments, we used 8 segments. To make such continued pre-training feasible, we developed an optimized ARMT implementation compatible with DeepSpeed ZeRO Stage 3 \citep{rasley2020deepspeed,rajbhandari2020zero}.

\ecoparagraph{Curriculum learning.} 
When associative memory layers are trained from scratch, ARMT initially performs near-randomly on challenging tasks that require propagating salient information across multiple segments, thereby yielding little useful learning signals from such instances.
We argue that, much like human memory, machine memory should be trained through a gradual increase in task complexity, i.e.\ curriculum learning \cite{bengio2009curriculum}.
Accordingly, we apply curriculum learning during fine-tuning by progressively increasing task difficulty. We control task difficulty via the maximum context length, defined by the number of ARMT segments. Specifically, the model first learns how to propagate important information across 2 segments, then 4, and finally 8 segments. 
In addition, we anneal the learning rate across the curriculum stages, gradually reducing it in the later stages.

\ecoparagraph{Generating synthetic long context training data.} 
Long context instances might be scarce in the original training dataset; therefore, fine-tuning ARMT on such data might be challenging. Moreover, in order to use curriculum learning, we need relatively large bins of instances that have contexts of specific lengths. To mitigate this problem, we suggest generating synthetic long-context training instances (see Section~\ref{sec:long_context_datasets} for further details). We first sample multiple short passages (paragraphs) from a long document and generate a question-answer pair for each passage.
These passages are then concatenated into a single long context, which is used as input during training while preserving the original QA supervision. 
To increase diversity and reduce model-specific biases in the generated data, we produce QA pairs using multiple LLMs from different model families.

\ecoparagraph{Associative memory layers pruning.} 
We further investigate whether associative memory is needed in every layer for effective context processing. Similar to modern LLMs that apply sliding-window attention only in a subset of layers \cite{team2025gemma}, we hypothesize that associative memory can likewise be retained only in selected layers.
Therefore, as an optional efficiency step, we suggest removing redundant associative layers after training. 
We also propose a recipe for the pre-selection of layers with associative memory, showing that one can train the ARMT model with only a few associative layers and achieve performance comparable to that of the full ARMT model.

\section{Long-Context Datasets}
\label{sec:long_context_datasets}

Approaches to LLM context extension rely either on continued LM pre-training~\citep{gao2025train} or on long-context supervised fine-tuning (SFT)~\citep{xuchatqa}. Continued LM training is prohibitively computationally expensive, requiring hundreds of GPU hours and is limited in effectiveness. Existing long-context SFT dataset selection is scarce and mostly targets selected domains. In our experiments, we aim to extend the context in a compute-efficient way by carefully crafting long-context datasets with sufficient diversity, complexity, and size, as well as controllable sample length suitable for curriculum learning.

We construct training and evaluation datasets using the ManyTypes4Py~\citep{mt4py2021} and GovReport-QS~\citep{cao2022hibrids} datasets. We focus on these datasets to demonstrate the possibility of context extension with the ARMT model on domain-specific data, even with limited compute resources, while avoiding more compute-demanding instruction following datasets.

\ecoparagraph{ManyTypes-long (MT)} targets variable type prediction in code with long context. We split the original dataset into non-overlapping repositories for training, validation, and test splits, and stacked the code scripts from each repository to obtain a long text with the desired length of up to 64k tokens. We reused the original labels for the variable types from ManyTypes4Py~\citep{mt4py2021}.

\ecoparagraph{GovReport-long (GR)} focuses on long-document question answering. The original GovReport-QS consists of triplets \{report, question, answer\} with ground-truth paragraphs in the report for each question. We used ground-truth paragraphs from the report and mixed them with non-relevant paragraphs, keeping the paragraph order to build datasets of up to 64k in length.
Due to the limited size of the original dataset, we augment the training split with synthetic examples to create GR-100+, while keeping the validation and test sets unchanged.

For both datasets, we place the question before the context to reduce reliance on parametric knowledge.
Examples of MT and GR datasets are presented in~\Cref{tab:data_example_mt,tab:data_example_gr} in Appendix~\ref{app:dataset_stats}. Dataset statistics are provided in~\Cref{tab:data_stats} in Appendix~\ref{app:dataset_stats}.

\begin{figure*}[ht!]
\centering
\begin{subfigure}{1.\columnwidth}
  \centering
  \includegraphics[width=1\columnwidth]{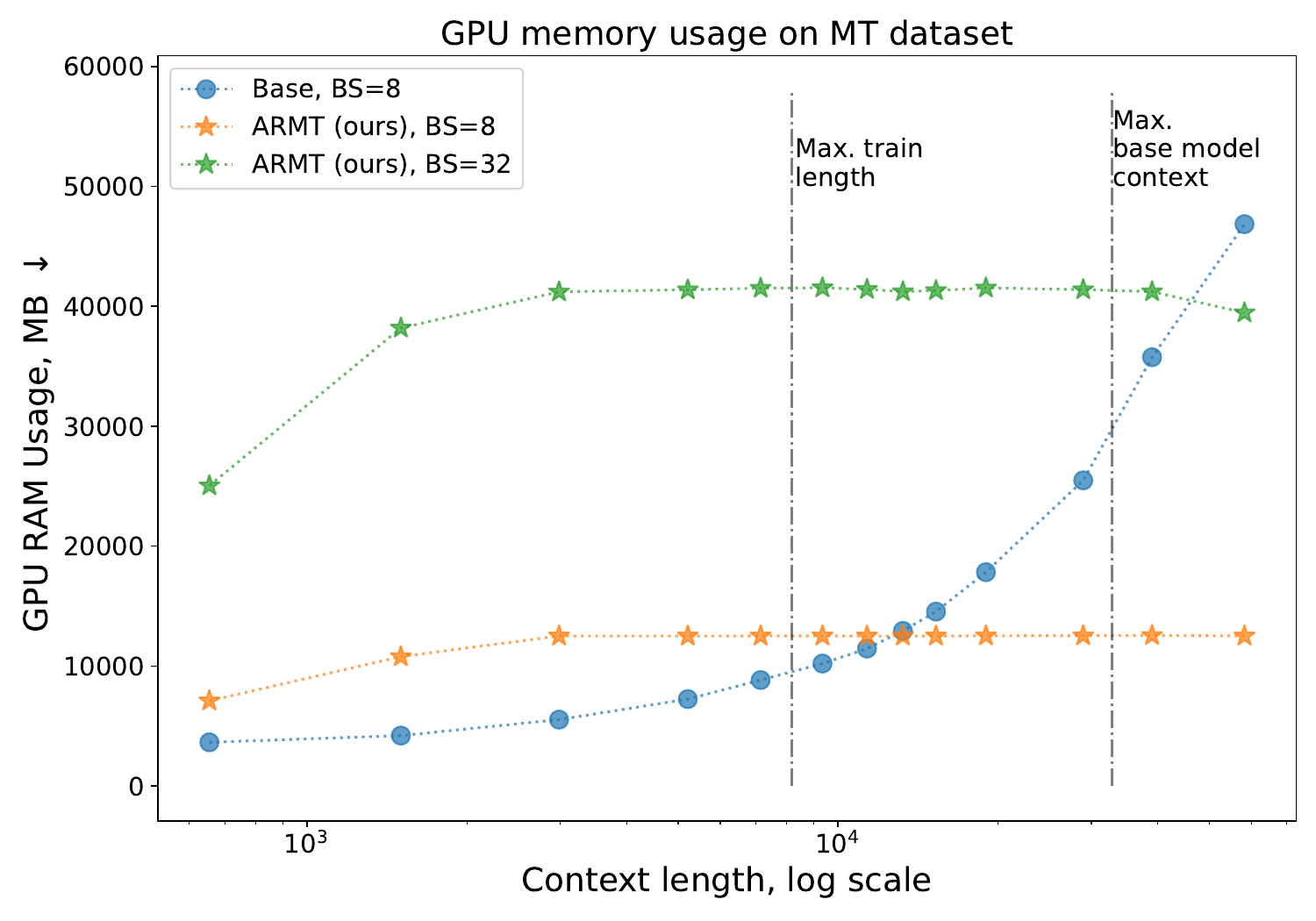}
  \caption{GPU memory usage.}
  \label{fig:main_res_mem_usage}
\end{subfigure}%
\begin{subfigure}{1.\columnwidth}
  \centering
  \includegraphics[width=1.\columnwidth]{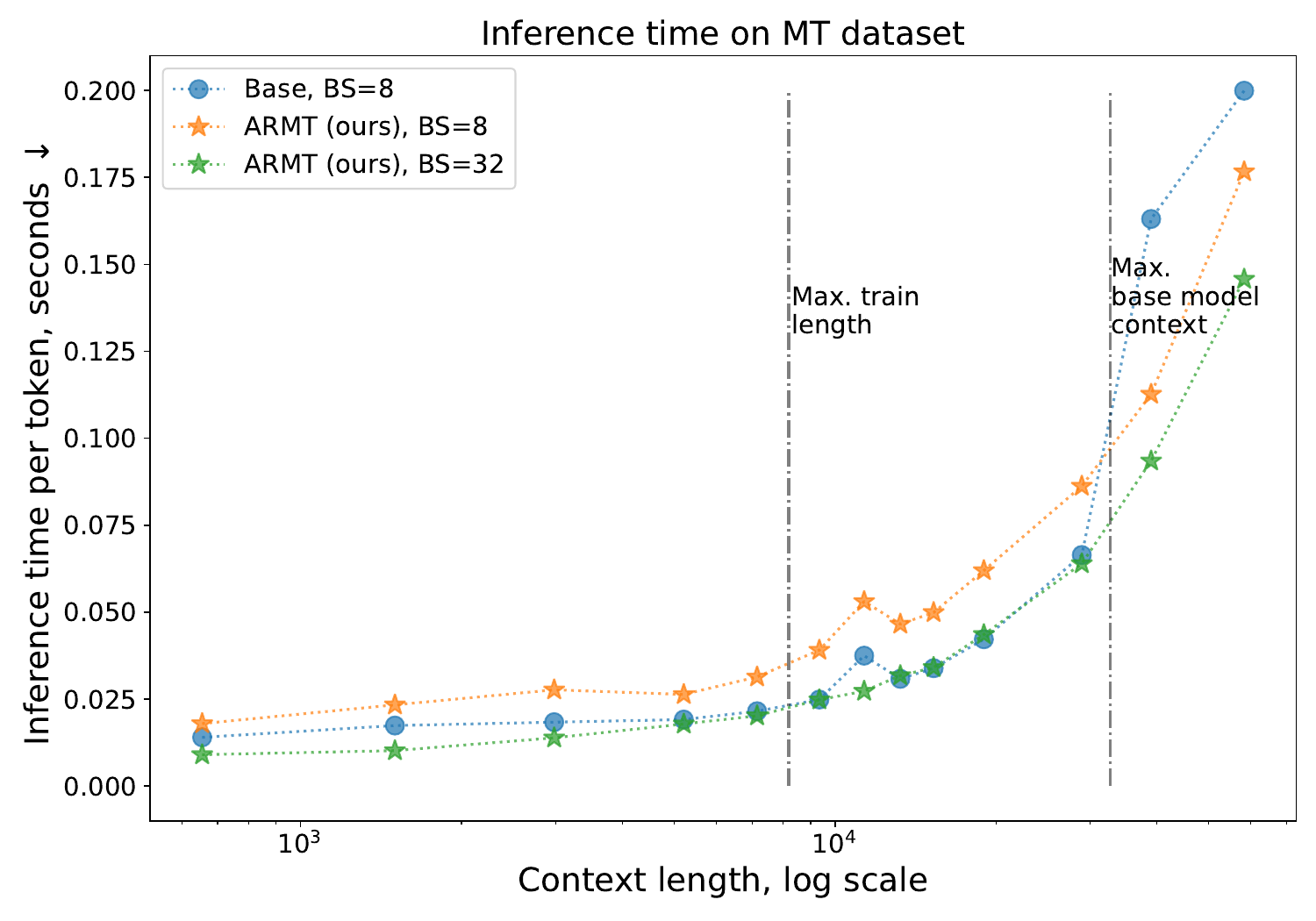}
  \caption{Inference time.}
  \label{fig:main_res_inf_time}
\end{subfigure}
\caption{ARMT performance in terms of inference time and memory usage. \textbf{(1)} Gemma-3-1B-IT (Base) with batch size of 8 uses approximately 40 GB of GPU RAM on 32k length, while ARMT model uses only 12 GB GPU RAM (left) while having approximately the same inference time (right). \textbf{(2)} ARMT shows constant memory usage with respect to context length, improving base model performance and allowing for long-context data processing.}
\label{fig:main_res_speedup}
\end{figure*}

\section{Experiments}

\subsection{Experimental Setup}

The main objective of our experiments is to demonstrate that ARMT, together with the proposed training techniques, can: (1) effectively extend the context capacity of the base LLM; and (2) replace full quadratic self-attention with a piece-wise attention mechanism that remains quadratic only within a small local window, while preserving the performance of the original LLM through the recurrent memory mechanism. To this end, we conducted an experiment in which we compared the base LLM and an ARMT-augmented LLM in which the native context window is drastically reduced relative to the base model (to 1,024 tokens), and the primary burden of long-context processing is placed on the recurrent memory mechanism instead of full self-attention. This means, e.g., that if we process a 32k size context, we need to split it into 32 segments, in each of which the full-attention only spans 1k tokens, and important information is propagated through memory.

\ecoparagraph{Base LLMs.}
We experimented with Gemma-3-1B-IT~\citep{team2025gemma} and SmolLM-2-360M-IT~\citep{allal2025smollm2smolgoesbig} as base LLMs. 

Gemma-3-1B-IT was selected because it is a strong state-of-the-art model in the 1B-parameter category, while still having a relatively limited maximum context length of 32k tokens. Moreover, it exhibits substantial performance degradation in long-context settings, even within this nominal context window \citep{team2025gemma}. We hypothesize that this arises from the use of interleaving sliding-window attention in its architecture instead of monolithic full attention. Therefore, this model is a good candidate for context extension.
SmolLM-2-360M-IT is also well suited for our experiments: it is small enough to train under a limited GPU memory budget, yet remains highly capable for its size. Its limited 8k-token context window makes it a natural candidate for context extension.

\ecoparagraph{Evaluation and metrics.}
We fine-tuned the models on the training sets and evaluated their performance on the test sets of MT and GR.
We used the exact match (EM) metric for the MT dataset, as the target answers are short and require precise variable type prediction. For the GR and GR-100+ datasets, we use the ROUGE-L~\citep{lin-2004-rouge} metric, as the ground-truth answers in GR are free-form and relatively long.

\ecoparagraph{Training details.}
We fine-tune the LLMs using LoRA \citep{lora} for GPU memory efficiency. 
We used the training setup from~\Cref{sec:methods} for both the GR and MT datasets. Continued pre-training of the Gemma-3 ARMT model was conducted on 19B tokens from FineWeb-Edu~\citep{10.5555/3737916.3738886}. We also experimented with additional ARMT pre-training on a synthetic long-context QA task.
Training hyperparameters are provided in Appendix~\ref{app:hyperpars}.

\subsection{Analysis of Computational Efficiency}
\label{sec:flops}

\ecoparagraph{GPU memory consumption} is the main advantage of ARMT, as it remains constant while the sequence length increases, whereas even the most efficient implementations of full self-attention scale at least linearly. This is illustrated in \Cref{fig:main_res_speedup}(a): for Gemma, memory consumption grows as the context length increases and can eventually exceed memory limits. In contrast, ARMT does not require additional memory to process longer sequences, so it can use larger batch sizes. For example, under the same memory budget required by Gemma to process a 65k-token sequence with a batch size of 8, ARMT can process sequences of the same length using a batch size of 32.

\ecoparagraph{FLOPs and token throughput.}
Scaling the batch size does not automatically increase token throughput, since GPU cores may already be saturated. However, because self-attention in ARMT is quadratic only within short, fixed-size segments, ARMT also requires substantially fewer self-attention FLOPs than full-attention-based models. For a sequence of length $T$ and segment size $S$, ARMT reduces global-attention FLOPs by approximately $T/S$; for $T{=}32{,}768$ and $S{=}1{,}024$, this corresponds to a $\mathbf{32\times}$ reduction in global-attention FLOPs. When feed-forward and projection layers are taken into account, ARMT theoretically requires roughly one-third fewer FLOPs in total than Gemma under our approximation; see \Cref{app:flops} for the full derivation.

\Cref{fig:main_res_speedup}(b) empirically validates this reduction by illustrating the average token inference time. For a batch size of 8, ARMT is slower than the base model due to the sequential processing of segments and the under-utilization of GPU cores. However, when we increase the batch size to 32, ARMT substantially outperforms the base Gemma model. Optimized ARMT implementations could further improve efficiency \cite{sivtsov2025diagonal}.

\subsection{Analysis of Model Performance}
\label{sec:main_results}

\begin{figure*}[ht!]
\centering
\begin{subfigure}{1.\columnwidth}
  \centering
  \includegraphics[width=1\columnwidth]{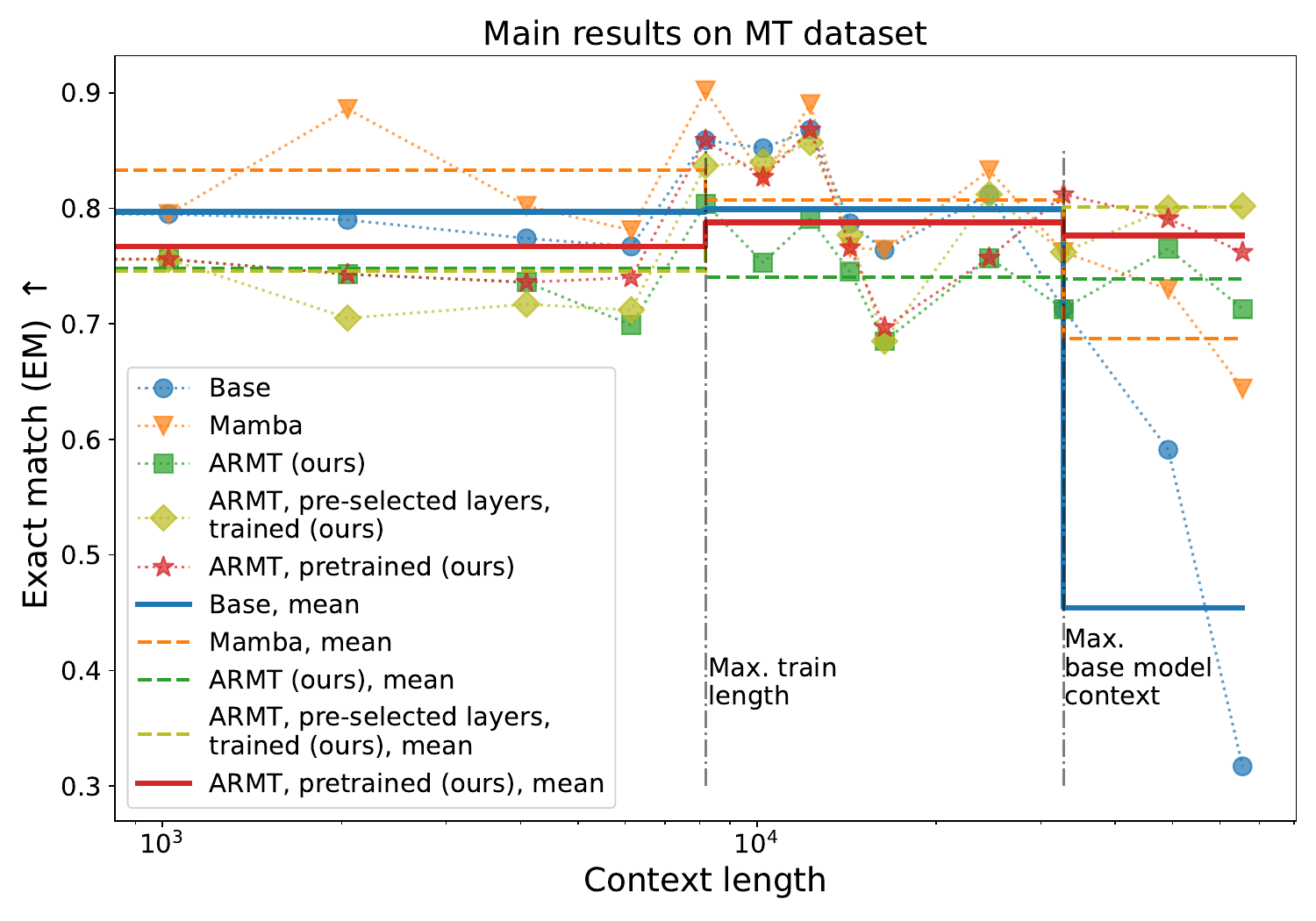}
  \caption{MT results.}
  \label{fig:main_res_mt}
\end{subfigure}%
\begin{subfigure}{1.\columnwidth}
  \centering
  \includegraphics[width=1.\columnwidth]{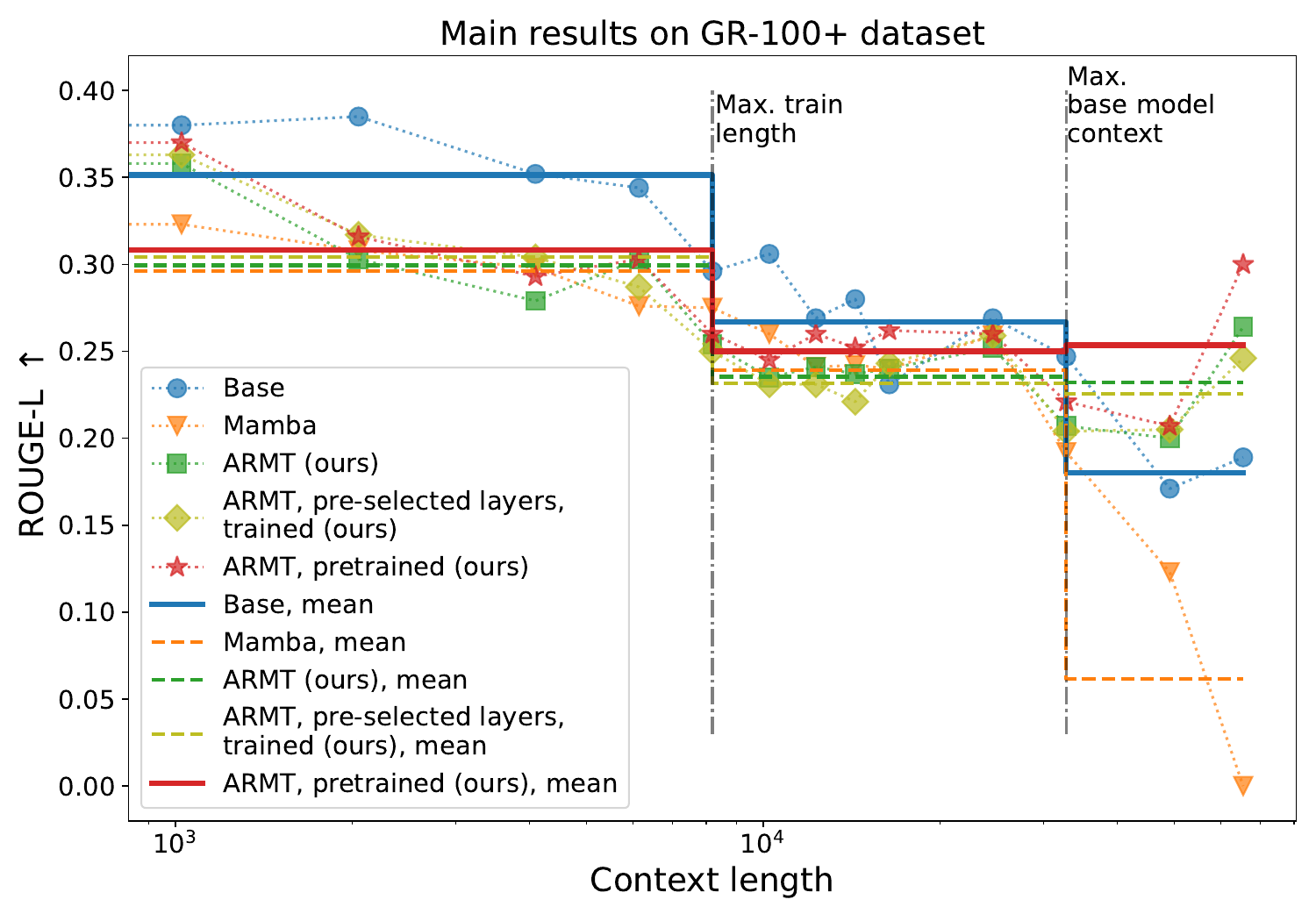}
  \caption{GR-100+ results.}
  \label{fig:main_res_gr100}
\end{subfigure}
\caption{Main results. \textbf{(1)} Gemma-3-1B-IT (Base) with full attention and Mamba-2 trained with a context length of 8,192 show a sharp performance drop beyond 32k tokens. \textbf{(2)} ARMT model with Gemma-3-1B-IT backbone (ARMT) maintains stable performance across all evaluated context lengths and outperforms other models in the long out-of-distribution (Long-OOD) regimen (>32k). \textbf{(3)} ARMT with continued pre-training followed by fine-tuning consistently outperforms the variant trained with fine-tuning only. \textbf{(4)} ARMT fine-tuned with only five pre-selected associative layers out of 26 achieves comparable or better performance than the full ARMT model.}
\label{fig:main_res}
\end{figure*}

\begin{table}[t]

\begin{center}
\resizebox{1.\linewidth}{!}{
\begin{tabular}{lccccc}
\toprule
\begin{tabular}[c]{@{}l@{}}\textbf{Model\//}\\ \textbf{Lengths}\end{tabular} & \begin{tabular}[c]{@{}l@{}}\textbf{Base,}\\ \textbf{No Fine-}\\ \textbf{Tuning}\end{tabular} & \begin{tabular}[c]{@{}l@{}}\textbf{Base,}\\ \textbf{MT, 8k}\end{tabular} & \begin{tabular}[c]{@{}l@{}}\textbf{ARMT,}\\ \textbf{MT, 8k}\end{tabular} & \begin{tabular}[c]{@{}l@{}}\textbf{ARMT, MT, 8k}\\ \textbf{pre-selected layers,}\\ \textbf{trained}\end{tabular} & \begin{tabular}[c]{@{}l@{}}\textbf{ARMT,}\\ \textbf{MT, 8k,}\\ \textbf{pretrain}\end{tabular} \\
\midrule
0k-1k & {\cellcolor[HTML]{FF0000}} \color[HTML]{F1F1F1} 0.000 & {\cellcolor[HTML]{2A9500}} \color[HTML]{F1F1F1} 0.795 & {\cellcolor[HTML]{42A100}} \color[HTML]{F1F1F1} 0.756 & {\cellcolor[HTML]{42A100}} \color[HTML]{F1F1F1} 0.756 & {\cellcolor[HTML]{42A100}} \color[HTML]{F1F1F1} 0.756 \\
1k-2k & {\cellcolor[HTML]{FF0000}} \color[HTML]{F1F1F1} 0.000 & {\cellcolor[HTML]{2E9700}} \color[HTML]{F1F1F1} 0.790 & {\cellcolor[HTML]{48A400}} \color[HTML]{F1F1F1} 0.743 & {\cellcolor[HTML]{60B000}} \color[HTML]{F1F1F1} 0.705 & {\cellcolor[HTML]{48A400}} \color[HTML]{F1F1F1} 0.743 \\
2k-4k & {\cellcolor[HTML]{FF0000}} \color[HTML]{F1F1F1} 0.000 & {\cellcolor[HTML]{369B00}} \color[HTML]{F1F1F1} 0.774 & {\cellcolor[HTML]{4CA600}} \color[HTML]{F1F1F1} 0.736 & {\cellcolor[HTML]{58AC00}} \color[HTML]{F1F1F1} 0.717 & {\cellcolor[HTML]{4CA600}} \color[HTML]{F1F1F1} 0.736 \\
4k-6k & {\cellcolor[HTML]{FF0000}} \color[HTML]{F1F1F1} 0.000 & {\cellcolor[HTML]{3A9D00}} \color[HTML]{F1F1F1} 0.767 & {\cellcolor[HTML]{62B100}} \color[HTML]{F1F1F1} 0.699 & {\cellcolor[HTML]{5CAE00}} \color[HTML]{F1F1F1} 0.712 & {\cellcolor[HTML]{4AA500}} \color[HTML]{F1F1F1} 0.740 \\
6k-8k & {\cellcolor[HTML]{FF0000}} \color[HTML]{F1F1F1} 0.000 & {\cellcolor[HTML]{048200}} \color[HTML]{F1F1F1} 0.859 & {\cellcolor[HTML]{249200}} \color[HTML]{F1F1F1} 0.804 & {\cellcolor[HTML]{128900}} \color[HTML]{F1F1F1} 0.837 & {\cellcolor[HTML]{048200}} \color[HTML]{F1F1F1} 0.859 \\
8k-10k & {\cellcolor[HTML]{FF0000}} \color[HTML]{F1F1F1} 0.000 & {\cellcolor[HTML]{088400}} \color[HTML]{F1F1F1} 0.852 & {\cellcolor[HTML]{42A100}} \color[HTML]{F1F1F1} 0.753 & {\cellcolor[HTML]{108800}} \color[HTML]{F1F1F1} 0.840 & {\cellcolor[HTML]{188C00}} \color[HTML]{F1F1F1} 0.827 \\
10k-12k & {\cellcolor[HTML]{FF0600}} \color[HTML]{F1F1F1} 0.011 & {\cellcolor[HTML]{008000}} \color[HTML]{F1F1F1} 0.868 & {\cellcolor[HTML]{2C9600}} \color[HTML]{F1F1F1} 0.791 & {\cellcolor[HTML]{068300}} \color[HTML]{F1F1F1} 0.857 & {\cellcolor[HTML]{008000}} \color[HTML]{F1F1F1} 0.868 \\
12k-14k & {\cellcolor[HTML]{FF0000}} \color[HTML]{F1F1F1} 0.000 & {\cellcolor[HTML]{2E9700}} \color[HTML]{F1F1F1} 0.787 & {\cellcolor[HTML]{48A400}} \color[HTML]{F1F1F1} 0.745 & {\cellcolor[HTML]{349A00}} \color[HTML]{F1F1F1} 0.777 & {\cellcolor[HTML]{3C9E00}} \color[HTML]{F1F1F1} 0.766 \\
14k-16k & {\cellcolor[HTML]{FF0000}} \color[HTML]{F1F1F1} 0.000 & {\cellcolor[HTML]{3C9E00}} \color[HTML]{F1F1F1} 0.764 & {\cellcolor[HTML]{6AB500}} \color[HTML]{F1F1F1} 0.685 & {\cellcolor[HTML]{6AB500}} \color[HTML]{F1F1F1} 0.685 & {\cellcolor[HTML]{64B200}} \color[HTML]{F1F1F1} 0.697 \\
16k-24k & {\cellcolor[HTML]{FF0000}} \color[HTML]{F1F1F1} 0.000 & {\cellcolor[HTML]{209000}} \color[HTML]{F1F1F1} 0.812 & {\cellcolor[HTML]{40A000}} \color[HTML]{F1F1F1} 0.757 & {\cellcolor[HTML]{209000}} \color[HTML]{F1F1F1} 0.812 & {\cellcolor[HTML]{40A000}} \color[HTML]{F1F1F1} 0.757 \\
24k-32k & {\cellcolor[HTML]{FF0000}} \color[HTML]{F1F1F1} 0.000 & {\cellcolor[HTML]{5AAD00}} \color[HTML]{F1F1F1} 0.713 & {\cellcolor[HTML]{5AAD00}} \color[HTML]{F1F1F1} 0.713 & {\cellcolor[HTML]{3E9F00}} \color[HTML]{F1F1F1} 0.762 & {\cellcolor[HTML]{209000}} \color[HTML]{F1F1F1} 0.812 \\
32k-49k & {\cellcolor[HTML]{FF0000}} \color[HTML]{F1F1F1} 0.000 & {\cellcolor[HTML]{A2D100}} \color[HTML]{000000} 0.591 & {\cellcolor[HTML]{3C9E00}} \color[HTML]{F1F1F1} 0.765 & {\cellcolor[HTML]{289400}} \color[HTML]{F1F1F1} 0.800 & {\cellcolor[HTML]{2C9600}} \color[HTML]{F1F1F1} 0.791 \\
49k-65k & {\cellcolor[HTML]{FF0000}} \color[HTML]{F1F1F1} 0.000 & {\cellcolor[HTML]{FFBA00}} \color[HTML]{000000} 0.317 & {\cellcolor[HTML]{5AAD00}} \color[HTML]{F1F1F1} 0.713 & {\cellcolor[HTML]{269300}} \color[HTML]{F1F1F1} 0.802 & {\cellcolor[HTML]{3E9F00}} \color[HTML]{F1F1F1} 0.762 \\
\hline
In-Domain (0k-8k) & {\cellcolor[HTML]{FF0000}} \color[HTML]{F1F1F1} 0.000 & {\cellcolor[HTML]{289400}} \color[HTML]{F1F1F1} 0.797 & {\cellcolor[HTML]{46A300}} \color[HTML]{F1F1F1} 0.749 & {\cellcolor[HTML]{48A400}} \color[HTML]{F1F1F1} 0.744 & {\cellcolor[HTML]{3A9D00}} \color[HTML]{F1F1F1} 0.767 \\
OOD (8k-65k) & {\cellcolor[HTML]{FF0000}} \color[HTML]{F1F1F1} 0.001 & {\cellcolor[HTML]{5CAE00}} \color[HTML]{F1F1F1} 0.709 & {\cellcolor[HTML]{4AA500}} \color[HTML]{F1F1F1} 0.741 & {\cellcolor[HTML]{2C9600}} \color[HTML]{F1F1F1} 0.793 & {\cellcolor[HTML]{329900}} \color[HTML]{F1F1F1} 0.783 \\
Long-OOD (32k-65k) & {\cellcolor[HTML]{FF0000}} \color[HTML]{F1F1F1} 0.000 & {\cellcolor[HTML]{EEF700}} \color[HTML]{000000} 0.463 & {\cellcolor[HTML]{4AA500}} \color[HTML]{F1F1F1} 0.741 & {\cellcolor[HTML]{269300}} \color[HTML]{F1F1F1} 0.801 & {\cellcolor[HTML]{349A00}} \color[HTML]{F1F1F1} 0.777 \\
Full (0k-65k) & {\cellcolor[HTML]{FF0000}} \color[HTML]{F1F1F1} 0.001 & {\cellcolor[HTML]{4AA500}} \color[HTML]{F1F1F1} 0.741 & {\cellcolor[HTML]{48A400}} \color[HTML]{F1F1F1} 0.744 & {\cellcolor[HTML]{369B00}} \color[HTML]{F1F1F1} 0.776 & {\cellcolor[HTML]{349A00}} \color[HTML]{F1F1F1} 0.777 \\
\bottomrule
\end{tabular}}

\end{center}
\caption{Best results on the MT dataset for Gemma-3-1B-IT model, metric = EM. ARMT performs comparably with the base model overall, and outperforms it on OOD and Long-OOD.}
\label{tab:mt_main_res}
\end{table}
\begin{table}[t]

\begin{center}
\resizebox{1.\linewidth}{!}{
\begin{tabular}{lccccc}
\toprule
\begin{tabular}[c]{@{}l@{}}\textbf{Model\//}\\ \textbf{Lengths}\end{tabular} & \begin{tabular}[c]{@{}l@{}}\textbf{Base,}\\ \textbf{No Fine-}\\ \textbf{Tuning}\end{tabular} & \begin{tabular}[c]{@{}l@{}}\textbf{Base,}\\ \textbf{GR-100+, 8k}\end{tabular} & \begin{tabular}[c]{@{}l@{}}\textbf{ARMT,}\\ \textbf{GR-100+, 8k}\end{tabular} & \begin{tabular}[c]{@{}l@{}}\textbf{ARMT, GR-100+, 8k}\\ \textbf{pre-selected layers,}\\ \textbf{trained}\end{tabular} & \begin{tabular}[c]{@{}l@{}}\textbf{ARMT,}\\ \textbf{GR-100+, 8k,}\\ \textbf{pretrain}\end{tabular} \\
\midrule
0k-1k & {\cellcolor[HTML]{FFB600}} \color[HTML]{000000} 0.188 & {\cellcolor[HTML]{088400}} \color[HTML]{F1F1F1} 0.380 & {\cellcolor[HTML]{2C9600}} \color[HTML]{F1F1F1} 0.358 & {\cellcolor[HTML]{249200}} \color[HTML]{F1F1F1} 0.363 & {\cellcolor[HTML]{188C00}} \color[HTML]{F1F1F1} 0.370 \\
1k-2k & {\cellcolor[HTML]{FF8A00}} \color[HTML]{F1F1F1} 0.161 & {\cellcolor[HTML]{008000}} \color[HTML]{F1F1F1} 0.385 & {\cellcolor[HTML]{88C400}} \color[HTML]{000000} 0.303 & {\cellcolor[HTML]{70B800}} \color[HTML]{F1F1F1} 0.317 & {\cellcolor[HTML]{72B900}} \color[HTML]{F1F1F1} 0.316 \\
2k-4k & {\cellcolor[HTML]{FF6600}} \color[HTML]{F1F1F1} 0.140 & {\cellcolor[HTML]{369B00}} \color[HTML]{F1F1F1} 0.352 & {\cellcolor[HTML]{B0D800}} \color[HTML]{000000} 0.279 & {\cellcolor[HTML]{86C300}} \color[HTML]{000000} 0.304 & {\cellcolor[HTML]{98CC00}} \color[HTML]{000000} 0.293 \\
4k-6k & {\cellcolor[HTML]{FF3C00}} \color[HTML]{F1F1F1} 0.114 & {\cellcolor[HTML]{44A200}} \color[HTML]{F1F1F1} 0.344 & {\cellcolor[HTML]{88C400}} \color[HTML]{000000} 0.303 & {\cellcolor[HTML]{A2D100}} \color[HTML]{000000} 0.287 & {\cellcolor[HTML]{88C400}} \color[HTML]{000000} 0.303 \\
6k-8k & {\cellcolor[HTML]{FF3A00}} \color[HTML]{F1F1F1} 0.113 & {\cellcolor[HTML]{94CA00}} \color[HTML]{000000} 0.296 & {\cellcolor[HTML]{DAED00}} \color[HTML]{000000} 0.254 & {\cellcolor[HTML]{E0F000}} \color[HTML]{000000} 0.250 & {\cellcolor[HTML]{D0E800}} \color[HTML]{000000} 0.260 \\
8k-10k & {\cellcolor[HTML]{FF2E00}} \color[HTML]{F1F1F1} 0.106 & {\cellcolor[HTML]{82C100}} \color[HTML]{000000} 0.306 & {\cellcolor[HTML]{FAFD00}} \color[HTML]{000000} 0.235 & {\cellcolor[HTML]{FFFE00}} \color[HTML]{000000} 0.231 & {\cellcolor[HTML]{E8F400}} \color[HTML]{000000} 0.245 \\
10k-12k & {\cellcolor[HTML]{FF2600}} \color[HTML]{F1F1F1} 0.101 & {\cellcolor[HTML]{C0E000}} \color[HTML]{000000} 0.269 & {\cellcolor[HTML]{F0F800}} \color[HTML]{000000} 0.241 & {\cellcolor[HTML]{FFFE00}} \color[HTML]{000000} 0.231 & {\cellcolor[HTML]{D0E800}} \color[HTML]{000000} 0.260 \\
12k-14k & {\cellcolor[HTML]{FF1400}} \color[HTML]{F1F1F1} 0.091 & {\cellcolor[HTML]{AED700}} \color[HTML]{000000} 0.280 & {\cellcolor[HTML]{F6FB00}} \color[HTML]{000000} 0.237 & {\cellcolor[HTML]{FFEE00}} \color[HTML]{000000} 0.221 & {\cellcolor[HTML]{DCEE00}} \color[HTML]{000000} 0.252 \\
14k-16k & {\cellcolor[HTML]{FF1000}} \color[HTML]{F1F1F1} 0.088 & {\cellcolor[HTML]{FFFE00}} \color[HTML]{000000} 0.231 & {\cellcolor[HTML]{F0F800}} \color[HTML]{000000} 0.240 & {\cellcolor[HTML]{ECF600}} \color[HTML]{000000} 0.243 & {\cellcolor[HTML]{CCE600}} \color[HTML]{000000} 0.262 \\
16k-24k & {\cellcolor[HTML]{FF0800}} \color[HTML]{F1F1F1} 0.083 & {\cellcolor[HTML]{C0E000}} \color[HTML]{000000} 0.269 & {\cellcolor[HTML]{DCEE00}} \color[HTML]{000000} 0.252 & {\cellcolor[HTML]{D2E900}} \color[HTML]{000000} 0.259 & {\cellcolor[HTML]{D0E800}} \color[HTML]{000000} 0.260 \\
24k-32k & {\cellcolor[HTML]{FF0000}} \color[HTML]{F1F1F1} 0.078 & {\cellcolor[HTML]{E6F300}} \color[HTML]{000000} 0.247 & {\cellcolor[HTML]{FFD600}} \color[HTML]{000000} 0.207 & {\cellcolor[HTML]{FFD200}} \color[HTML]{000000} 0.204 & {\cellcolor[HTML]{FFEE00}} \color[HTML]{000000} 0.221 \\
32k-49k & {\cellcolor[HTML]{FF0000}} \color[HTML]{F1F1F1} 0.078 & {\cellcolor[HTML]{FF9A00}} \color[HTML]{000000} 0.171 & {\cellcolor[HTML]{FFCA00}} \color[HTML]{000000} 0.200 & {\cellcolor[HTML]{FFD200}} \color[HTML]{000000} 0.205 & {\cellcolor[HTML]{FFD600}} \color[HTML]{000000} 0.207 \\
49k-65k & {\cellcolor[HTML]{FF2800}} \color[HTML]{F1F1F1} 0.103 & {\cellcolor[HTML]{FFB800}} \color[HTML]{000000} 0.189 & {\cellcolor[HTML]{C8E400}} \color[HTML]{000000} 0.264 & {\cellcolor[HTML]{E6F300}} \color[HTML]{000000} 0.246 & {\cellcolor[HTML]{8CC600}} \color[HTML]{000000} 0.300 \\
\hline
\begin{tabular}[c]{@{}l@{}}{In-Domain}\\ {(0k-8k)}\end{tabular} & {\cellcolor[HTML]{FF6C00}} \color[HTML]{F1F1F1} 0.143 & {\cellcolor[HTML]{389C00}} \color[HTML]{F1F1F1} 0.351 & {\cellcolor[HTML]{8EC700}} \color[HTML]{000000} 0.299 & {\cellcolor[HTML]{86C300}} \color[HTML]{000000} 0.304 & {\cellcolor[HTML]{80C000}} \color[HTML]{000000} 0.308 \\
\begin{tabular}[c]{@{}l@{}}{OOD}\\ {(8k-65k)}\end{tabular} & {\cellcolor[HTML]{FF1600}} \color[HTML]{F1F1F1} 0.092 & {\cellcolor[HTML]{C6E300}} \color[HTML]{000000} 0.266 & {\cellcolor[HTML]{F4FA00}} \color[HTML]{000000} 0.238 & {\cellcolor[HTML]{FCFE00}} \color[HTML]{000000} 0.234 & {\cellcolor[HTML]{DCEE00}} \color[HTML]{000000} 0.252 \\
\begin{tabular}[c]{@{}l@{}}{Long-OOD}\\ {(32k-65k)}\end{tabular} & {\cellcolor[HTML]{FF0800}} \color[HTML]{F1F1F1} 0.084 & {\cellcolor[HTML]{FFA200}} \color[HTML]{000000} 0.175 & {\cellcolor[HTML]{FFE400}} \color[HTML]{000000} 0.215 & {\cellcolor[HTML]{FFE200}} \color[HTML]{000000} 0.214 & {\cellcolor[HTML]{FFFA00}} \color[HTML]{000000} 0.228 \\
\begin{tabular}[c]{@{}l@{}}{Full}\\ {(0k-65k)}\end{tabular} & {\cellcolor[HTML]{FF3E00}} \color[HTML]{F1F1F1} 0.116 & {\cellcolor[HTML]{84C200}} \color[HTML]{000000} 0.306 & {\cellcolor[HTML]{C4E200}} \color[HTML]{000000} 0.266 & {\cellcolor[HTML]{C4E200}} \color[HTML]{000000} 0.267 & {\cellcolor[HTML]{B0D800}} \color[HTML]{000000} 0.278 \\
\bottomrule
\end{tabular}}
\end{center}
\caption{Best results on the GR-100+ dataset for Gemma-3-1B-IT model, metric = ROUGE-L. ARMT performs comparably to the base model overall, and outperforms it on OOD and Long-OOD.}
\label{tab:gov_report_main_res}
\end{table}

\ecoparagraph{Extending the context window of base LLM.}
The main results with Gemma-3 on the MT dataset are shown in~\Cref{fig:main_res_mt} and \Cref{tab:mt_main_res} (see also~\Cref{tab:mt_best} in Appendix~\ref{app:armt_base_setup}). The base model without fine-tuning achieves near-zero performance across all context lengths, highlighting the necessity of task-specific adaptation. After fine-tuning on 8k contexts (Base, MT, 8k), the model performs well up to its maximum supported length of 32k tokens, but degrades sharply beyond this limit due to context truncation.

In contrast, ARMT shows no degradation beyond 32k tokens. Notably, after curriculum training with only 4k contexts, ARMT already generalizes to inputs as long as 65k, demonstrating the effectiveness of the recurrent memory mechanism. The final ARMT model (ARMT, MT, 8k) matches the fine-tuned base model on in-domain inputs (less than 8k) while substantially improving performance on out-of-distribution (OOD) context lengths (larger than 8k).

The results with Gemma-3 on GovReport-long-100+ are presented in~\Cref{fig:main_res_gr100}, \Cref{tab:gov_report_main_res} and in~\Cref{tab:gov_report_best} in Appendix~\ref{app:armt_base_setup}. They exhibit a similar trend. The base model achieves low zero-shot in-domain performance (ROUGE-L 0.116) and improves to 0.306 after fine-tuning on GR-100+. Its performance drops substantially on Long-OOD inputs (0.175). ARMT partially mitigates this degradation, improving the Long-OOD performance to 0.215, with only a minor trade-off on shorter contexts.

Experiments with SmolLM show a similar pattern  (\Cref{tab:gov_report_best_smollm,tab:mt_best_smollm} in Appendix~\ref{app:smollm_ablation}). While the fine-tuned base model degrades beyond its native context length (8k), ARMT maintains competitive performance up to 32k tokens.

Overall, these results highlight two main findings: (1) ARMT, together with the proposed training techniques, effectively extends the ability of the base model to handle longer inputs; and (2) for contexts that fall within the original LLM's native context window, our piecewise attention mechanism with recurrent memory 
largely preserves the performance of the original full-attention model.
Combined with constant memory consumption and the estimated 30\% reduction in FLOPs derived in \Cref{sec:flops}, these results demonstrate the strong practical applicability of ARMT.

\ecoparagraph{Comparison to other baselines.}
We fine-tuned a 1.3B-parameter Mamba-2 model~\citep{dao2024transformers} on the MT and GR-100+ datasets (\Cref{fig:main_res}). Mamba-2 is a variant of a state space model that shows comparable performance with transformers and linear scaling in sequence length, making it feasible for long-context tasks.
Although Mamba-2 handles long contexts (32--64k) better than the base full-attention LLM (on MT), it still substantially underperforms ARMT in this setting. However, in shorter contexts (on MT), Mamba-2 performs better than ARMT and the base full-attention model, which can be attributed to the larger model size. Overall, ARMT shows better length generalization than other baselines. Additional results for Mamba~\citep{gu2023mamba}, Mamba-2~\citep{dao2024transformers}, DeltaNet~\citep{yang2024parallelizing}, and xLSTM~\citep{beck2024xlstm} are reported in~\Cref{tab:mt_best_all_baselines} (Appendix~\ref{app:baselines}). We also compared the ARMT model with the simpler RMT model~\citep{bulatov2022rmt}, which does not use associative memory layers (\Cref{tab:gov_report_rmt_ablation} in the Appendix~\ref{app:rmt_ablations}). Results show that the associative memory mechanism introduced in the advanced ARMT architecture substantially improves performance and generalization compared to RMT.

\ecoparagraph{Other long-context benchmarks.} We further trained and evaluated ARMT on BABILong~\citep{kuratov2024babilong} and ContractNLI from SCROLLS~\citep{shaham-etal-2022-scrolls} to validate the training recipe and model. The results are reported in~\Cref{tab:babilong_ablation_qa1_qa3,tab:babilong_ablation_qa4_qa5} (Appendix~\ref{app:babilong}) and~\Cref{tab:cnli_best_with_pretrain} (Appendix~\ref{app:cnli}). On both benchmarks, ARMT with the proposed training procedure outperforms the base full-attention model.

\begin{table}[t]

\begin{center}
\resizebox{1.\linewidth}{!}{
\begin{tabular}{lccccc}
\toprule
\begin{tabular}[c]{@{}l@{}}\textbf{Model\//}\\ \textbf{Lengths}\end{tabular} & \begin{tabular}[c]{@{}l@{}}\textbf{Base,}\\ \textbf{MT, 8k}\end{tabular} & \begin{tabular}[c]{@{}l@{}}\textbf{ARMT,}\\ \textbf{MT, 8k}\end{tabular} &  \begin{tabular}[c]{@{}l@{}}\textbf{ARMT,}\\ \textbf{ MT, 8k}\\ \textbf{only top-4 layers}\end{tabular} &  \begin{tabular}[c]{@{}l@{}}\textbf{ARMT, MT, 8k}\\ \textbf{only top-4 layers,} \\\textbf{trained}\end{tabular} &  \begin{tabular}[c]{@{}l@{}}\textbf{ARMT, MT, 8k}\\ \textbf{pre-selected layers,} \\\textbf{trained}\end{tabular} \\
\midrule
0k-1k & {\cellcolor[HTML]{42A100}} \color[HTML]{F1F1F1} 0.795 & {\cellcolor[HTML]{68B400}} \color[HTML]{F1F1F1} 0.756 & {\cellcolor[HTML]{68B400}} \color[HTML]{F1F1F1} 0.756 & {\cellcolor[HTML]{68B400}} \color[HTML]{F1F1F1} 0.756 & {\cellcolor[HTML]{68B400}} \color[HTML]{F1F1F1} 0.756 \\
1k-2k & {\cellcolor[HTML]{48A400}} \color[HTML]{F1F1F1} 0.790 & {\cellcolor[HTML]{74BA00}} \color[HTML]{F1F1F1} 0.743 & {\cellcolor[HTML]{84C200}} \color[HTML]{000000} 0.724 & {\cellcolor[HTML]{8EC700}} \color[HTML]{000000} 0.714 & {\cellcolor[HTML]{96CB00}} \color[HTML]{000000} 0.705 \\
2k-4k & {\cellcolor[HTML]{56AB00}} \color[HTML]{F1F1F1} 0.774 & {\cellcolor[HTML]{7ABD00}} \color[HTML]{F1F1F1} 0.736 & {\cellcolor[HTML]{94CA00}} \color[HTML]{000000} 0.708 & {\cellcolor[HTML]{7ABD00}} \color[HTML]{F1F1F1} 0.736 & {\cellcolor[HTML]{8CC600}} \color[HTML]{000000} 0.717 \\
4k-6k & {\cellcolor[HTML]{5CAE00}} \color[HTML]{F1F1F1} 0.767 & {\cellcolor[HTML]{9CCE00}} \color[HTML]{000000} 0.699 & {\cellcolor[HTML]{90C800}} \color[HTML]{000000} 0.712 & {\cellcolor[HTML]{9CCE00}} \color[HTML]{000000} 0.699 & {\cellcolor[HTML]{90C800}} \color[HTML]{000000} 0.712 \\
6k-8k & {\cellcolor[HTML]{088400}} \color[HTML]{F1F1F1} 0.859 & {\cellcolor[HTML]{3A9D00}} \color[HTML]{F1F1F1} 0.804 & {\cellcolor[HTML]{44A200}} \color[HTML]{F1F1F1} 0.793 & {\cellcolor[HTML]{309800}} \color[HTML]{F1F1F1} 0.815 & {\cellcolor[HTML]{1C8E00}} \color[HTML]{F1F1F1} 0.837 \\
8k-10k & {\cellcolor[HTML]{0E8700}} \color[HTML]{F1F1F1} 0.852 & {\cellcolor[HTML]{6AB500}} \color[HTML]{F1F1F1} 0.753 & {\cellcolor[HTML]{5EAF00}} \color[HTML]{F1F1F1} 0.765 & {\cellcolor[HTML]{269300}} \color[HTML]{F1F1F1} 0.827 & {\cellcolor[HTML]{1A8D00}} \color[HTML]{F1F1F1} 0.840 \\
10k-12k & {\cellcolor[HTML]{008000}} \color[HTML]{F1F1F1} 0.868 & {\cellcolor[HTML]{46A300}} \color[HTML]{F1F1F1} 0.791 & {\cellcolor[HTML]{66B300}} \color[HTML]{F1F1F1} 0.758 & {\cellcolor[HTML]{3C9E00}} \color[HTML]{F1F1F1} 0.802 & {\cellcolor[HTML]{0A8500}} \color[HTML]{F1F1F1} 0.857 \\
12k-14k & {\cellcolor[HTML]{4AA500}} \color[HTML]{F1F1F1} 0.787 & {\cellcolor[HTML]{72B900}} \color[HTML]{F1F1F1} 0.745 & {\cellcolor[HTML]{68B400}} \color[HTML]{F1F1F1} 0.755 & {\cellcolor[HTML]{54AA00}} \color[HTML]{F1F1F1} 0.777 & {\cellcolor[HTML]{54AA00}} \color[HTML]{F1F1F1} 0.777 \\
14k-16k & {\cellcolor[HTML]{60B000}} \color[HTML]{F1F1F1} 0.764 & {\cellcolor[HTML]{AAD500}} \color[HTML]{000000} 0.685 & {\cellcolor[HTML]{BEDF00}} \color[HTML]{000000} 0.663 & {\cellcolor[HTML]{94CA00}} \color[HTML]{000000} 0.708 & {\cellcolor[HTML]{AAD500}} \color[HTML]{000000} 0.685 \\
16k-24k & {\cellcolor[HTML]{349A00}} \color[HTML]{F1F1F1} 0.812 & {\cellcolor[HTML]{66B300}} \color[HTML]{F1F1F1} 0.757 & {\cellcolor[HTML]{80C000}} \color[HTML]{000000} 0.729 & {\cellcolor[HTML]{66B300}} \color[HTML]{F1F1F1} 0.757 & {\cellcolor[HTML]{349A00}} \color[HTML]{F1F1F1} 0.812 \\
24k-32k & {\cellcolor[HTML]{90C800}} \color[HTML]{000000} 0.713 & {\cellcolor[HTML]{90C800}} \color[HTML]{000000} 0.713 & {\cellcolor[HTML]{A2D100}} \color[HTML]{000000} 0.693 & {\cellcolor[HTML]{62B100}} \color[HTML]{F1F1F1} 0.762 & {\cellcolor[HTML]{62B100}} \color[HTML]{F1F1F1} 0.762 \\
32k-49k & {\cellcolor[HTML]{FFFE00}} \color[HTML]{000000} 0.591 & {\cellcolor[HTML]{5EAF00}} \color[HTML]{F1F1F1} 0.765 & {\cellcolor[HTML]{5EAF00}} \color[HTML]{F1F1F1} 0.765 & {\cellcolor[HTML]{5EAF00}} \color[HTML]{F1F1F1} 0.765 & {\cellcolor[HTML]{3E9F00}} \color[HTML]{F1F1F1} 0.800 \\
49k-65k & {\cellcolor[HTML]{FF0000}} \color[HTML]{F1F1F1} 0.317 & {\cellcolor[HTML]{90C800}} \color[HTML]{000000} 0.713 & {\cellcolor[HTML]{90C800}} \color[HTML]{000000} 0.713 & {\cellcolor[HTML]{6AB500}} \color[HTML]{F1F1F1} 0.752 & {\cellcolor[HTML]{3C9E00}} \color[HTML]{F1F1F1} 0.802 \\
\hline
\begin{tabular}[c]{@{}l@{}}{In-Domain}\\ {(0k-8k)}\end{tabular} & {\cellcolor[HTML]{40A000}} \color[HTML]{F1F1F1} 0.797 & {\cellcolor[HTML]{6EB700}} \color[HTML]{F1F1F1} 0.749 & {\cellcolor[HTML]{78BC00}} \color[HTML]{F1F1F1} 0.738 & {\cellcolor[HTML]{72B900}} \color[HTML]{F1F1F1} 0.744 & {\cellcolor[HTML]{72B900}} \color[HTML]{F1F1F1} 0.744 \\
\begin{tabular}[c]{@{}l@{}}{OOD}\\ {(8k-65k)}\end{tabular} & {\cellcolor[HTML]{92C900}} \color[HTML]{000000} 0.709 & {\cellcolor[HTML]{74BA00}} \color[HTML]{F1F1F1} 0.741 & {\cellcolor[HTML]{80C000}} \color[HTML]{000000} 0.730 & {\cellcolor[HTML]{5CAE00}} \color[HTML]{F1F1F1} 0.767 & {\cellcolor[HTML]{44A200}} \color[HTML]{F1F1F1} 0.793 \\
\begin{tabular}[c]{@{}l@{}}{Long-OOD}\\ {(32k-65k)}\end{tabular} & {\cellcolor[HTML]{FF8600}} \color[HTML]{F1F1F1} 0.463 & {\cellcolor[HTML]{76BB00}} \color[HTML]{F1F1F1} 0.741 & {\cellcolor[HTML]{76BB00}} \color[HTML]{F1F1F1} 0.741 & {\cellcolor[HTML]{64B200}} \color[HTML]{F1F1F1} 0.759 & {\cellcolor[HTML]{3E9F00}} \color[HTML]{F1F1F1} 0.801 \\
\begin{tabular}[c]{@{}l@{}}{Full}\\ {(0k-65k)}\end{tabular} & {\cellcolor[HTML]{76BB00}} \color[HTML]{F1F1F1} 0.741 & {\cellcolor[HTML]{72B900}} \color[HTML]{F1F1F1} 0.744 & {\cellcolor[HTML]{7CBE00}} \color[HTML]{000000} 0.733 & {\cellcolor[HTML]{64B200}} \color[HTML]{F1F1F1} 0.759 & {\cellcolor[HTML]{54AA00}} \color[HTML]{F1F1F1} 0.776 \\
\bottomrule

\end{tabular}}
\caption{Associative layers ablation on the MT dataset for Gemma-3-1B-IT model, metric = EM. ARMT with only top-4 associative blocks retains almost the same performance as the full ARMT without training. The model with 5 associative blocks (approximately 20\%) achieves the same performance as the full ARMT.}
\label{tab:mt_best_assoc_ablation_with_train}
\end{center}
\end{table}
\begin{table}[t]

\begin{center}
\resizebox{1.\linewidth}{!}{
\begin{tabular}{llccc}
\toprule
\begin{tabular}[c]{@{}l@{}}\textbf{Num. of}\\ \textbf{Segments}\end{tabular} & \textbf{Model} & \begin{tabular}[c]{@{}l@{}}\textbf{Num. of}\\ \textbf{Trainable}\\ \textbf{Parameters}\end{tabular} & \begin{tabular}[c]{@{}l@{}}\textbf{GPU RAM}\\ \textbf{Usage $\downarrow$}\end{tabular} & \begin{tabular}[c]{@{}l@{}}\textbf{Steps per}\\ \textbf{Second $\uparrow$}\end{tabular} \\
\midrule
\multirow{ 2}{*}{2} & ARMT  & 90.5M &  45.4GB & 2.86 \\
 & ARMT, pre-selected layers & 59.6M & 44.5GB & 3.64 \\
\hline
\multirow{ 2}{*}{4} & ARMT & 90.5M &  71.8GB & 1.62 \\
 & ARMT, pre-selected layers & 59.6M & 69.8GB & 2.14 \\

\bottomrule
\end{tabular}}
\caption{ARMT with pre-selected layers saves 30\% of trainable parameters and speed-ups training up to 30\%, while reducing GPU memory usage. Results for ARMT with Gemma-3-1B-IT backbone; only five associative layers out of 26 are used in the pre-selected setup.}
\label{tab:armt_preselected_speedup}
\end{center}
\end{table}

\ecoparagraph{Ablation: synthetic data.}
To study the impact of synthetic data, we train ARMT on GR variants with different proportions of synthetic samples (Appendix~\ref{app:gr_ablation}, \Cref{fig:synth_scaling}). Adding synthetic long-context data consistently improves performance up to a moderate scale, after which the gains saturate. The best results are achieved with GR-100+, corresponding to a synthetic-to-real ratio of 5.5, which we use in subsequent experiments.

\ecoparagraph{Ablation: continued pre-training.}
We analyze the effect of continued pre-training on ARMT prior to task-specific fine-tuning in~\Cref{fig:main_res} and in~\Cref{tab:gov_report_best_with_pretrain,tab:mt_best_with_pretrain} in Appendix~\ref{app:armt_ft_pretrain}. Starting from Gemma-3-1B-IT, we add an uninitialized associative memory with 1024-token segments and continue pre-training on 19B FineWeb-Edu tokens~\citep{10.5555/3737916.3738886} using 8-segment sequences.
The loss curves of the pre-trained ARMT model are presented in~\Cref{fig:armt_pretrain} in Appendix~\ref{app:armt_lm_pretrain}. After pre-training, the model is fine-tuned on the GR-100+ dataset using curriculum learning.

Continued pre-training consistently improves performance for both in-domain and out-of-domain context lengths. It effectively initializes the associative memory, enabling the model to propagate information across segments and making this capability easier to exploit during fine-tuning.

\ecoparagraph{Ablation: curriculum learning.}
To further examine the role of curriculum learning, we also fine-tune the pre-trained ARMT model without curriculum scheduling (ARMT w/o CL). As shown in~\Cref{tab:gov_report_best_with_pretrain} in Appendix~\ref{app:armt_ft_pretrain}, this simpler training procedure achieves performance comparable to the curriculum-based variant, indicating that continued pre-training alone is sufficient to effectively initialize associative memory.
However, according to our results, for models without continued pre-training, curriculum learning is necessary.

\ecoparagraph{Associative layer pruning and selection.}
Adding associative memory layers to a pre-trained LLM increases the number of trainable parameters and training cost, and adds the need for re-adaptation as inserted associative memory layers alter the  internal representations of LLM.
This motivates the following RQ: \emph{what is the minimal architectural transformation needed for effective associative memory integration, and which layers of the LLM are most important for introducing associative memory?} To investigate this, we conduct two sets of experiments: (1) after fine-tuning, we replace a subset of ARMT layers with an identity function; (2) we introduce associative memory only into a subset of LLM layers before fine-tuning.

The results of the first (pruning) experiment are provided in~\Cref{tab:mt_best_assoc_ablation,tab:gov_report_best_assoc_ablation} in Appendix~\ref{app:assoc_ablation} for the MT and GR-100+ datasets. For both datasets, the most important layers are concentrated in the third quarter of the network (layers 13--18), while the early layers (0--6) contribute the least and can be removed after training without performance loss. 

Fine-grained ablations further show that only a few layers are critical, with layer 14 consistently being the most important. Retaining only the top-1 or top-4 layers preserves performance, reducing the number of associative layers by up to six times.
Detailed layer-wise ablation results are provided in~\Cref{tab:gov_report_best_all_layers_assoc_ablation_0_12,tab:gov_report_best_all_layers_assoc_ablation_13_26} in  Appendix~\ref{app:assoc_ablation}.

In the second experiment, we investigate whether we can add the associative memory only to a subset of LLM layers before fine-tuning. We first train ARMT models on GR-100+ and MT using only the 4 most important layers according to the previous analysis (see~\Cref{tab:gov_report_best_assoc_ablation_with_train} in Appendix~\ref{app:assoc_ablation} and~\Cref{tab:mt_best_assoc_ablation_with_train}). On GR-100+, the model with the most important ARMT layers achieves slightly lower overall performance than the full ARMT model but performs better on the Long-OOD subset. On the MT dataset, such a model outperforms the full ARMT model. 
The drawback is that identifying important layers requires training a full ARMT model first, which is not convenient.

To address this limitation, we propose a \emph{universal pre-selection strategy}: two middle layers, the final layer, and one layer in each of the first and last quarters. For ARMT with Gemma-3-1B-IT (26 layers), this corresponds to layers 7, 13, 14, 19, and 25. This configuration outperforms the full ARMT model on both MT and GR-100+ (\Cref{fig:main_res}, \Cref{tab:gov_report_best_assoc_ablation_with_train} in Appendix~\ref{app:assoc_ablation} and~\Cref{tab:mt_best_assoc_ablation_with_train}), while eliminating the need for prior layer importance analysis. It also reduces the number of trainable parameters and speeds up fine-tuning by 30\% (see~\Cref{tab:armt_preselected_speedup}).

\ecoparagraph{Other ablations.} We also evaluated the supervised pre-training of ARMT on synthetic QA data prior to training on GR-100+ (\Cref{tab:gov_report_best_with_synth_pretrain} in Appendix~\ref{app:pretrain_ablation}). Although this improves in-domain performance, it degrades long-context generalization, likely due to limited sequence lengths in the synthetic data. In contrast, continued language model pre-training is more effective for initializing associative memory.

\section{Conclusion}

We present a practical recipe for extending the context length of LLMs on domain-specific tasks using ARMT. The proposed approach combines several techniques: (i) continued pre-training, (ii) synthetic long-context data generation, (iii) curriculum learning, and (iv) optional pruning of associative layers. Together, these components enable small local models to achieve strong long-context understanding, helping to close an important gap for privacy-preserving applications that cannot rely on remote API-based LLMs. ARMT also enables constant GPU memory utilization for arbitrarily long contexts and a 30\% reduction of TFLOPs.
We introduced two datasets for the training and testing of LLMs in long-context tasks: ManyTypes-long and GovReport-long. 
Finally, we show that only a few of the most important layers need to be augmented with ARMT.
Based on this observation, we propose a universal strategy for pre-selecting layers for augmentation prior to fine-tuning. Our experiments demonstrate that fine-tuning ARMT with only 20\% of the associative layers is sufficient to match the performance of the full ARMT model while reducing the training time by approximately 30\%.

\section*{Limitations}
We experimented only with relatively small LLMs (up to 1B parameters) due to the chosen scope of the paper and computational constraints. However, we argue that this setting is still important, especially for practical domain-specific deployments.

Although we trained and evaluated ARMT on multiple long-context benchmarks that covered code, question answering, and document understanding, our study still covers only a subset of possible long-context tasks and modalities. Future work should evaluate the proposed approach across a broader range of tasks and domains.

Finally, our analysis of associative memory focuses primarily on layer-level ablations within the ARMT architecture. Although we identify a subset of associative layers that contribute most to long-context performance, the underlying mechanisms of how associative memory interacts with transformer representations remain only partially understood.

We used AI assistants in a limited and controlled manner for two purposes: (i) generating synthetic training data for fine-tuning of the models and (ii) improving the grammar and clarity of the manuscript. The AI tools were used according to their intended purpose and with careful consideration of responsible and ethical research practices.

\section*{Ethical Considerations}
This work focuses on improving long-context capabilities of small LLMs intended for local deployment in privacy-sensitive environments. By enabling efficient processing of long documents without reliance on large external services, the proposed approach can help reduce the risks of data leakage when handling confidential or proprietary information.

The datasets used in this work are derived from previously published sources. We do not introduce new personal or sensitive data in presented datasets.

As with other LLM technologies, improved long-context processing can enable more effective analysis of large collections of text, which could potentially be misused for large-scale information extraction or surveillance. 

We therefore encourage careful evaluation of downstream deployments, particularly in settings involving sensitive data, and recommend transparency about dataset construction and model limitations when releasing derived resources.

\bibliography{custom}

\clearpage
\appendix

\section{Formal Description of the ARMT Architecture}
\label{app:formal}

For the hidden states of segment $s$ in layer $l$ $H_s^l$, memory tokens from the previous layer $M_s^{l-1}$, the associative matrix $A_s^l$, and normalization $z_s^l$ are updated as follows:
\begin{gather}
M_s^l = \{m_i\}, \quad M_s^l = \text{TransLayer}(H^{l-1}_s, M_s^{l-1}) \\
k_{i},v_{i} =W_K m_{i},W_V m_{i}; \quad \beta_{i} = \sigma(W_\beta m_i); \\
A_0^l = \vec{0}; \quad  z_0^l = \vec{0};\\
\overline{v}_i = \frac{A_{s-1}^l \phi(k_i)}{(z_{s-1})^T \phi(k_i)}; \quad \gamma_i = 1 - \frac{(z_{s-1})^T\phi(k_i)}{\|\phi(k_i)\|^2}; \\
A_s^l = A_{s-1}^l + \sum_i \beta_i (v_i - \overline{v}_i) \otimes \phi(k_i); \\
z^l_{s} = z^l_{s-1} + \sum_i\gamma_i \phi (k_i).
\end{gather}
Reading from memory in the following segments for embedding $x_j$ from $H_{s+1}^{l-1}$:
 \begin{align}
q_j = W_Q x_j; \quad 
y_j = \frac{A_s^l \phi(q_j)}{(z^l_s)^T \phi(q_j)},
\end{align}
where $y_j$ is an association for $x_j$.

\subsection{FLOP Analysis}
\label{app:flops}

We provide the FLOP calculation used in \Cref{sec:flops}. We compare Gemma-3 and ARMT on a sequence of length $T$, with segment size $S$, $M$ memory tokens, $N$ layers, $N_g$ global-attention layers, $N_l$ local-attention layers, hidden dimension $d$, head dimension $d_h$, and $H$ attention heads.

\paragraph{Gemma-3.}
For a full sequence of length $T$, the attention cost per layer is $4HT^2d_h$
for global attention and $4HTWd_h$ for local attention with window size $W$. For long sequences, the sequence-length-dependent attention cost is dominated by global attention, giving
\[
    F_\text{Gemma\_Attn}
    \sim
    4N_gHd_hT^2.
\]

Feed-forward and projection layers scale linearly with the sequence length and quadratically with the hidden dimension. Using the same coarse accounting as in the main text, the dense-layer cost is
\[
    F_\text{FFN}
    =
    (2\times2\times6 + 2\times3)NTd^2
    =
    30NTd^2.
\]
This term is independent of the segmentation and is therefore approximately equal for Gemma and ARMT when the memory-token overhead is neglected.

\paragraph{ARMT.}
ARMT splits the sequence into $\frac{T}{S}$ segments. Each segment is processed with attention context length $S+M$, where $M$ is the number of memory tokens. Therefore, the global-attention FLOP cost becomes
\[
    F_\text{ARMT\_Attn}
    \sim
    \frac{T}{S}
    4N_gHd_h(S+M)^2.
\]
When $M \ll S$, this simplifies to
\[
    F_\text{ARMT\_Attn}
    \approx
    4N_gHd_hST.
\]

Thus, for global attention alone, the reduction factor is approximately
\[
    \frac{F_\text{Gemma\_Attn}}{F_\text{ARMT\_Attn}}
    \approx
    \frac{T}{S}.
\]
For $T{=}32{,}768$ and $S{=}1{,}024$, this gives
\[
    \frac{T}{S}
    =
    32,
\]
i.e., ARMT achieves a $\mathbf{32\times}$ reduction in global-attention FLOPs when memory-token overhead is neglected.

\paragraph{Overall FLOP ratio.}
To estimate the total FLOP ratio, we combine the global-attention term with the dense-layer term. Ignoring local attention and memory-token overhead, we have
\[
    F_\text{Gemma}
    \approx
    4N_gHd_hT^2 + 30NTd^2,
\]
and
\[
    F_\text{ARMT}
    \approx
    4N_gHd_hST + 30NTd^2.
\]
Dividing both numerator and denominator by $2T$ gives
\[
    \frac{F_\text{ARMT}}{F_\text{Gemma}}
    \approx
    \frac{2N_gHd_hS + 15Nd^2}
         {2N_gHd_hT + 15Nd^2}.
\]

For Gemma-3-1B-IT, substituting the model constants into this expression gives an overall FLOP ratio of approximately $0.67$. Thus, although ARMT reduces global-attention FLOPs by $32\times$ for $T{=}32{,}768$ and $S{=}1{,}024$, the total FLOP reduction is more modest because feed-forward and projection layers remain unchanged. Under this approximation, ARMT uses roughly one third fewer total FLOPs than Gemma.

\section{Datasets Statistics and Ablations}

\subsection{Dataset Statistics and Examples}\label{app:dataset_stats}

In this section, we present statistics for all dataset versions and splits used in our experiments. The dataset overview is summarized in~\Cref{tab:data_stats}. We also provide short illustrative examples from the MT and GR datasets in~\Cref{tab:data_example_mt,tab:data_example_gr}.
\begin{table}[h]

\footnotesize
\begin{center}
\begin{tabular}{lccc}
\toprule
\begin{tabular}[c]{@{}l@{}}\textbf{Dataset\//}\\ \textbf{Split}\end{tabular} & \begin{tabular}[c]{@{}l@{}}\textbf{Train}\end{tabular} & \begin{tabular}[c]{@{}l@{}}\textbf{Validation}\end{tabular} & \begin{tabular}[c]{@{}l@{}}\textbf{Test}\end{tabular} \\
\midrule
MT, 2k & 229.4k &  3.1k &  1.3k \\
MT, 4k & 289.2k &  4.5k &  1.3k \\
MT, 8k & 328.3k &  5.1k &  1.3k \\
GR, 2k & 9.9k &  0.5k &  1.0k \\
GR, 4k & 17.4k &  0.9k &  1.0k \\
GR, 8k & 19.7k &  1.0k &  1.0k \\
GR-100+, 2k & 90.5k &  0.5k &  1.0k \\
GR-100+, 4k & 121.7k &  0.9k &  1.0k \\
GR-100+, 8k & 128.4k &  1.0k &  1.0k \\
ContractNLI (SCROLLS) & 7.2k & 1.0k & 2.1k \\
\bottomrule
\end{tabular}
\caption{Dataset statistics - number of samples for each dataset and for each split.}
\label{tab:data_stats}
\end{center}
\end{table}
\begin{table*}[h]

\begin{center}
\begin{tabular}{lp{14cm}}
\toprule
\textbf{Example} & \begin{tabular}[c]{@{}l@{}}\textbf{Text}\end{tabular} \\
\midrule
Context & from typing import Any \newline import typing \newline [docstring] \newline from alembic import op \newline import sqlalchemy as sa \newline \newline \newline from sqlalchemy . ext . declarative import declarative\_base \newline from sqlalchemy . orm import sessionmaker , relationship \newline \newline [comment] \newline revision = [string] \newline down\_revision = [string] \newline \newline Base = declarative\_base ( ) \newline db = sa \newline db . Model = Base \newline db . relationship = relationship \newline \newline \newline def create\_session ( ) : \newline connection = op . get\_bind ( ) \newline session\_maker = sa . orm . sessionmaker ( ) \newline session = session\_maker ( bind = connection ) \newline db . session = session \newline \newline \newline def upgrade ( ) : \newline create\_session ( ) \newline \newline [comment] \newline op . alter\_column ( [string] , [string] , type\_ = sa . Text , existing\_type = sa . String ) \newline \newline [comment] \newline \newline def downgrade ( ) : \newline create\_session ( ) \newline \newline [comment] \newline op . alter\_column ( [string] , [string] , type\_ = sa . String , existing\_type = sa . Text ) \newline [comment] \newline \newline \newline [comment] \newline \\
Question &  What is the type of variable down\_revision? \\
Answer &  builtins.str \\

\bottomrule
\end{tabular}
\caption{Example from MT dataset. The question during training and evaluation is placed before context (i. e. at the start of the prompt).}
\label{tab:data_example_mt}

\end{center}
\end{table*}
\begin{table*}[t]
\begin{center}
\begin{tabular}{lp{12cm}}
\toprule
\textbf{Example} & \begin{tabular}[c]{@{}l@{}}\textbf{Text}\end{tabular} \\
\midrule
Context & Economic Significance of Intellectual Property Protection and Theft\newline As we reported in April 2010, IP is an important component of the U.S. economy and IP-related industries pay higher wages and contribute a significant percentage to the U.S. economy. However, the U.S. economy as a whole may grow at a slower pace than it otherwise would because of counterfeiting and piracy’s effect on U.S. industries, government, and consumers.\newline Quantifying Economic Impacts Is Difficult, However Industry Research Suggests the Impacts Are Sizable\newline Generally, as we reported in April 2010, the illicit nature of counterfeiting and piracy makes estimating the economic impact of IP infringements extremely difficult, so assumptions must be used to offset the lack of data. Efforts to estimate losses involve assumptions such as the rate at which consumers would substitute counterfeit for legitimate products, which can have enormous impacts on the resulting estimates. Because of the significant differences in types of counterfeited and pirated goods and industries involved, no single method can be used to develop estimates. Each method has limitations, and most experts observed that it is difficult, if not impossible, to quantify the economy-wide impacts. Nonetheless, research in specific industries suggests that the problem is sizeable. \\
Question & What makes cost-estimates of IP infringements difficult to calculate? \\
Answer & Generally, as GAO reported in April 2010, the illicit nature of counterfeiting and piracy makes estimating the economic impact of IP infringements extremely difficult. \\

\bottomrule
\end{tabular}
\caption{Example from GR dataset. The question during training and evaluation is placed before context (i. e. at the start of the prompt).}
\label{tab:data_example_gr}
\end{center}
\end{table*}

\section{ARMT Training Dynamics}\label{app:armt_base_setup}
\Cref{tab:mt_best,tab:gov_report_best} presents results obtained using curriculum learning for the ARMT model on the MT and GR-100+ datasets. The performance of the ARMT models gradually increases during curriculum learning.
\begin{table*}[t]
\begin{center}
\resizebox{0.7\linewidth}{!}{
\begin{tabular}{lccccc}
\toprule
\begin{tabular}[c]{@{}l@{}}\textbf{Model\//}\\ \textbf{Lengths}\end{tabular} & \begin{tabular}[c]{@{}l@{}}\textbf{Base,}\\ \textbf{No Fine-}\\ \textbf{Tuning}\end{tabular} & \begin{tabular}[c]{@{}l@{}}\textbf{Base,}\\ \textbf{MT, 8k}\end{tabular} & \begin{tabular}[c]{@{}l@{}}\textbf{ARMT,}\\ \textbf{MT, 2k}\end{tabular} & \begin{tabular}[c]{@{}l@{}}\textbf{ARMT,}\\ \textbf{MT, 4k}\end{tabular} & \begin{tabular}[c]{@{}l@{}}\textbf{ARMT,}\\ \textbf{MT, 8k}\end{tabular} \\
\midrule
0k-1k & {\cellcolor[HTML]{FF0000}} \color[HTML]{F1F1F1} 0.000 & {\cellcolor[HTML]{2A9500}} \color[HTML]{F1F1F1} 0.795 & {\cellcolor[HTML]{329900}} \color[HTML]{F1F1F1} 0.782 & {\cellcolor[HTML]{329900}} \color[HTML]{F1F1F1} 0.782 & {\cellcolor[HTML]{42A100}} \color[HTML]{F1F1F1} 0.756 \\
1k-2k & {\cellcolor[HTML]{FF0000}} \color[HTML]{F1F1F1} 0.000 & {\cellcolor[HTML]{2E9700}} \color[HTML]{F1F1F1} 0.790 & {\cellcolor[HTML]{54AA00}} \color[HTML]{F1F1F1} 0.724 & {\cellcolor[HTML]{5AAD00}} \color[HTML]{F1F1F1} 0.714 & {\cellcolor[HTML]{48A400}} \color[HTML]{F1F1F1} 0.743 \\
2k-4k & {\cellcolor[HTML]{FF0000}} \color[HTML]{F1F1F1} 0.000 & {\cellcolor[HTML]{369B00}} \color[HTML]{F1F1F1} 0.774 & {\cellcolor[HTML]{68B400}} \color[HTML]{F1F1F1} 0.689 & {\cellcolor[HTML]{64B200}} \color[HTML]{F1F1F1} 0.698 & {\cellcolor[HTML]{4CA600}} \color[HTML]{F1F1F1} 0.736 \\
4k-6k & {\cellcolor[HTML]{FF0000}} \color[HTML]{F1F1F1} 0.000 & {\cellcolor[HTML]{3A9D00}} \color[HTML]{F1F1F1} 0.767 & {\cellcolor[HTML]{94CA00}} \color[HTML]{000000} 0.616 & {\cellcolor[HTML]{84C200}} \color[HTML]{000000} 0.644 & {\cellcolor[HTML]{62B100}} \color[HTML]{F1F1F1} 0.699 \\
6k-8k & {\cellcolor[HTML]{FF0000}} \color[HTML]{F1F1F1} 0.000 & {\cellcolor[HTML]{048200}} \color[HTML]{F1F1F1} 0.859 & {\cellcolor[HTML]{98CC00}} \color[HTML]{000000} 0.609 & {\cellcolor[HTML]{58AC00}} \color[HTML]{F1F1F1} 0.717 & {\cellcolor[HTML]{249200}} \color[HTML]{F1F1F1} 0.804 \\
8k-10k & {\cellcolor[HTML]{FF0000}} \color[HTML]{F1F1F1} 0.000 & {\cellcolor[HTML]{088400}} \color[HTML]{F1F1F1} 0.852 & {\cellcolor[HTML]{D4EA00}} \color[HTML]{000000} 0.506 & {\cellcolor[HTML]{42A100}} \color[HTML]{F1F1F1} 0.753 & {\cellcolor[HTML]{42A100}} \color[HTML]{F1F1F1} 0.753 \\
10k-12k & {\cellcolor[HTML]{FF0600}} \color[HTML]{F1F1F1} 0.011 & {\cellcolor[HTML]{008000}} \color[HTML]{F1F1F1} 0.868 & {\cellcolor[HTML]{FFDC00}} \color[HTML]{000000} 0.374 & {\cellcolor[HTML]{74BA00}} \color[HTML]{F1F1F1} 0.670 & {\cellcolor[HTML]{2C9600}} \color[HTML]{F1F1F1} 0.791 \\
12k-14k & {\cellcolor[HTML]{FF0000}} \color[HTML]{F1F1F1} 0.000 & {\cellcolor[HTML]{2E9700}} \color[HTML]{F1F1F1} 0.787 & {\cellcolor[HTML]{FFEE00}} \color[HTML]{000000} 0.404 & {\cellcolor[HTML]{60B000}} \color[HTML]{F1F1F1} 0.702 & {\cellcolor[HTML]{48A400}} \color[HTML]{F1F1F1} 0.745 \\
14k-16k & {\cellcolor[HTML]{FF0000}} \color[HTML]{F1F1F1} 0.000 & {\cellcolor[HTML]{3C9E00}} \color[HTML]{F1F1F1} 0.764 & {\cellcolor[HTML]{FFC600}} \color[HTML]{000000} 0.337 & {\cellcolor[HTML]{86C300}} \color[HTML]{000000} 0.640 & {\cellcolor[HTML]{6AB500}} \color[HTML]{F1F1F1} 0.685 \\
16k-24k & {\cellcolor[HTML]{FF0000}} \color[HTML]{F1F1F1} 0.000 & {\cellcolor[HTML]{209000}} \color[HTML]{F1F1F1} 0.812 & {\cellcolor[HTML]{FF8600}} \color[HTML]{F1F1F1} 0.229 & {\cellcolor[HTML]{82C100}} \color[HTML]{000000} 0.646 & {\cellcolor[HTML]{40A000}} \color[HTML]{F1F1F1} 0.757 \\
24k-32k & {\cellcolor[HTML]{FF0000}} \color[HTML]{F1F1F1} 0.000 & {\cellcolor[HTML]{5AAD00}} \color[HTML]{F1F1F1} 0.713 & {\cellcolor[HTML]{FF7400}} \color[HTML]{F1F1F1} 0.198 & {\cellcolor[HTML]{A6D300}} \color[HTML]{000000} 0.584 & {\cellcolor[HTML]{5AAD00}} \color[HTML]{F1F1F1} 0.713 \\
32k-49k & {\cellcolor[HTML]{FF0000}} \color[HTML]{F1F1F1} 0.000 & {\cellcolor[HTML]{A2D100}} \color[HTML]{000000} 0.591 & {\cellcolor[HTML]{FF5000}} \color[HTML]{F1F1F1} 0.139 & {\cellcolor[HTML]{98CC00}} \color[HTML]{000000} 0.609 & {\cellcolor[HTML]{3C9E00}} \color[HTML]{F1F1F1} 0.765 \\
49k-65k & {\cellcolor[HTML]{FF0000}} \color[HTML]{F1F1F1} 0.000 & {\cellcolor[HTML]{FFBA00}} \color[HTML]{000000} 0.317 & {\cellcolor[HTML]{FF3400}} \color[HTML]{F1F1F1} 0.089 & {\cellcolor[HTML]{9ACD00}} \color[HTML]{000000} 0.604 & {\cellcolor[HTML]{5AAD00}} \color[HTML]{F1F1F1} 0.713 \\
\hline
In-Domain (0k-8k) & {\cellcolor[HTML]{FF0000}} \color[HTML]{F1F1F1} 0.000 & {\cellcolor[HTML]{289400}} \color[HTML]{F1F1F1} 0.797 & {\cellcolor[HTML]{6AB500}} \color[HTML]{F1F1F1} 0.685 & {\cellcolor[HTML]{5CAE00}} \color[HTML]{F1F1F1} 0.711 & {\cellcolor[HTML]{46A300}} \color[HTML]{F1F1F1} 0.749 \\
OOD (8k-65k) & {\cellcolor[HTML]{FF0000}} \color[HTML]{F1F1F1} 0.001 & {\cellcolor[HTML]{5CAE00}} \color[HTML]{F1F1F1} 0.709 & {\cellcolor[HTML]{FF9E00}} \color[HTML]{000000} 0.271 & {\cellcolor[HTML]{82C100}} \color[HTML]{000000} 0.647 & {\cellcolor[HTML]{4AA500}} \color[HTML]{F1F1F1} 0.741 \\
Long-OOD (32k-65k) & {\cellcolor[HTML]{FF0000}} \color[HTML]{F1F1F1} 0.000 & {\cellcolor[HTML]{EEF700}} \color[HTML]{000000} 0.463 & {\cellcolor[HTML]{FF4400}} \color[HTML]{F1F1F1} 0.116 & {\cellcolor[HTML]{9ACD00}} \color[HTML]{000000} 0.607 & {\cellcolor[HTML]{4AA500}} \color[HTML]{F1F1F1} 0.741 \\
Full (0k-65k) & {\cellcolor[HTML]{FF0000}} \color[HTML]{F1F1F1} 0.001 & {\cellcolor[HTML]{4AA500}} \color[HTML]{F1F1F1} 0.741 & {\cellcolor[HTML]{FFF600}} \color[HTML]{000000} 0.419 & {\cellcolor[HTML]{74BA00}} \color[HTML]{F1F1F1} 0.670 & {\cellcolor[HTML]{48A400}} \color[HTML]{F1F1F1} 0.744 \\
\bottomrule
\end{tabular}}

\end{center}
\caption{Best results on the MT dataset for Gemma-3-1B-IT model, metric - EM. ARMT shows comparable with the base model overall performance, and outperforms it on OOD and Long-OOD.}
\label{tab:mt_best}
\end{table*}
\begin{table*}[t]

\begin{center}
\resizebox{0.7\linewidth}{!}{
\begin{tabular}{lccccc}
\toprule
\begin{tabular}[c]{@{}l@{}}\textbf{Model\//}\\ \textbf{Lengths}\end{tabular} & 
\begin{tabular}[c]{@{}l@{}}\textbf{Base,}\\ \textbf{No Fine-}\\ \textbf{Tuning}\end{tabular} & \begin{tabular}[c]{@{}l@{}}\textbf{Base,}\\ \textbf{GR-100+, 8k}\end{tabular} & \begin{tabular}[c]{@{}l@{}}\textbf{ARMT,}\\ \textbf{GR-100+, 2k}\end{tabular} & \begin{tabular}[c]{@{}l@{}}\textbf{ARMT,}\\ \textbf{GR-100+, 4k}\end{tabular} & \begin{tabular}[c]{@{}l@{}}\textbf{ARMT,}\\ \textbf{GR-100+, 8k}\end{tabular} \\
\midrule
0k-1k & {\cellcolor[HTML]{FFFA00}} \color[HTML]{000000} 0.188 & {\cellcolor[HTML]{068300}} \color[HTML]{F1F1F1} 0.380 & {\cellcolor[HTML]{249200}} \color[HTML]{F1F1F1} 0.357 & {\cellcolor[HTML]{249200}} \color[HTML]{F1F1F1} 0.357 & {\cellcolor[HTML]{229100}} \color[HTML]{F1F1F1} 0.358 \\
1k-2k & {\cellcolor[HTML]{FFD600}} \color[HTML]{000000} 0.161 & {\cellcolor[HTML]{008000}} \color[HTML]{F1F1F1} 0.385 & {\cellcolor[HTML]{76BB00}} \color[HTML]{F1F1F1} 0.296 & {\cellcolor[HTML]{78BC00}} \color[HTML]{F1F1F1} 0.294 & {\cellcolor[HTML]{6CB600}} \color[HTML]{F1F1F1} 0.303 \\
2k-4k & {\cellcolor[HTML]{FFBA00}} \color[HTML]{000000} 0.140 & {\cellcolor[HTML]{2A9500}} \color[HTML]{F1F1F1} 0.352 & {\cellcolor[HTML]{A0D000}} \color[HTML]{000000} 0.264 & {\cellcolor[HTML]{92C900}} \color[HTML]{000000} 0.275 & {\cellcolor[HTML]{8CC600}} \color[HTML]{000000} 0.279 \\
4k-6k & {\cellcolor[HTML]{FF9600}} \color[HTML]{000000} 0.114 & {\cellcolor[HTML]{369B00}} \color[HTML]{F1F1F1} 0.344 & {\cellcolor[HTML]{C0E000}} \color[HTML]{000000} 0.240 & {\cellcolor[HTML]{82C100}} \color[HTML]{000000} 0.286 & {\cellcolor[HTML]{6CB600}} \color[HTML]{F1F1F1} 0.303 \\
6k-8k & {\cellcolor[HTML]{FF9600}} \color[HTML]{000000} 0.113 & {\cellcolor[HTML]{76BB00}} \color[HTML]{F1F1F1} 0.296 & {\cellcolor[HTML]{DCEE00}} \color[HTML]{000000} 0.219 & {\cellcolor[HTML]{C2E100}} \color[HTML]{000000} 0.238 & {\cellcolor[HTML]{AED700}} \color[HTML]{000000} 0.254 \\
8k-10k & {\cellcolor[HTML]{FF8C00}} \color[HTML]{F1F1F1} 0.106 & {\cellcolor[HTML]{68B400}} \color[HTML]{F1F1F1} 0.306 & {\cellcolor[HTML]{EEF700}} \color[HTML]{000000} 0.205 & {\cellcolor[HTML]{CAE500}} \color[HTML]{000000} 0.233 & {\cellcolor[HTML]{C6E300}} \color[HTML]{000000} 0.235 \\
10k-12k & {\cellcolor[HTML]{FF8600}} \color[HTML]{F1F1F1} 0.101 & {\cellcolor[HTML]{9ACD00}} \color[HTML]{000000} 0.269 & {\cellcolor[HTML]{E4F200}} \color[HTML]{000000} 0.213 & {\cellcolor[HTML]{DAED00}} \color[HTML]{000000} 0.221 & {\cellcolor[HTML]{BEDF00}} \color[HTML]{000000} 0.241 \\
12k-14k & {\cellcolor[HTML]{FF7800}} \color[HTML]{F1F1F1} 0.091 & {\cellcolor[HTML]{8AC500}} \color[HTML]{000000} 0.280 & {\cellcolor[HTML]{FEFF00}} \color[HTML]{000000} 0.193 & {\cellcolor[HTML]{D4EA00}} \color[HTML]{000000} 0.225 & {\cellcolor[HTML]{C4E200}} \color[HTML]{000000} 0.237 \\
14k-16k & {\cellcolor[HTML]{FF7400}} \color[HTML]{F1F1F1} 0.088 & {\cellcolor[HTML]{CCE600}} \color[HTML]{000000} 0.231 & {\cellcolor[HTML]{FFEE00}} \color[HTML]{000000} 0.180 & {\cellcolor[HTML]{DEEF00}} \color[HTML]{000000} 0.217 & {\cellcolor[HTML]{C0E000}} \color[HTML]{000000} 0.240 \\
16k-24k & {\cellcolor[HTML]{FF6E00}} \color[HTML]{F1F1F1} 0.083 & {\cellcolor[HTML]{9ACD00}} \color[HTML]{000000} 0.269 & {\cellcolor[HTML]{FF9800}} \color[HTML]{000000} 0.115 & {\cellcolor[HTML]{F6FB00}} \color[HTML]{000000} 0.200 & {\cellcolor[HTML]{B0D800}} \color[HTML]{000000} 0.252 \\
24k-32k & {\cellcolor[HTML]{FF6600}} \color[HTML]{F1F1F1} 0.078 & {\cellcolor[HTML]{B6DB00}} \color[HTML]{000000} 0.247 & {\cellcolor[HTML]{FF1400}} \color[HTML]{F1F1F1} 0.016 & {\cellcolor[HTML]{FFDE00}} \color[HTML]{000000} 0.168 & {\cellcolor[HTML]{ECF600}} \color[HTML]{000000} 0.207 \\
32k-49k & {\cellcolor[HTML]{FF6600}} \color[HTML]{F1F1F1} 0.078 & {\cellcolor[HTML]{FFE200}} \color[HTML]{000000} 0.171 & {\cellcolor[HTML]{FF0200}} \color[HTML]{F1F1F1} 0.002 & {\cellcolor[HTML]{FFAA00}} \color[HTML]{000000} 0.128 & {\cellcolor[HTML]{F6FB00}} \color[HTML]{000000} 0.200 \\
49k-65k & {\cellcolor[HTML]{FF8800}} \color[HTML]{F1F1F1} 0.103 & {\cellcolor[HTML]{FFFA00}} \color[HTML]{000000} 0.189 & {\cellcolor[HTML]{FF0000}} \color[HTML]{F1F1F1} 0.000 & {\cellcolor[HTML]{FF3400}} \color[HTML]{F1F1F1} 0.040 & {\cellcolor[HTML]{A0D000}} \color[HTML]{000000} 0.264 \\
\hline
\begin{tabular}[c]{@{}l@{}}{In-Domain}\\ {(0k-8k)}\end{tabular} & {\cellcolor[HTML]{FFBE00}} \color[HTML]{000000} 0.143 & {\cellcolor[HTML]{2C9600}} \color[HTML]{F1F1F1} 0.351 & {\cellcolor[HTML]{92C900}} \color[HTML]{000000} 0.275 & {\cellcolor[HTML]{7EBF00}} \color[HTML]{000000} 0.290 & {\cellcolor[HTML]{72B900}} \color[HTML]{F1F1F1} 0.299 \\
\begin{tabular}[c]{@{}l@{}}{OOD}\\ {(8k-65k)}\end{tabular} & {\cellcolor[HTML]{FF7A00}} \color[HTML]{F1F1F1} 0.092 & {\cellcolor[HTML]{9ECF00}} \color[HTML]{000000} 0.266 & {\cellcolor[HTML]{FFD600}} \color[HTML]{000000} 0.162 & {\cellcolor[HTML]{E6F300}} \color[HTML]{000000} 0.211 & {\cellcolor[HTML]{C2E100}} \color[HTML]{000000} 0.238 \\
\begin{tabular}[c]{@{}l@{}}{Long-OOD}\\ {(32k-65k)}\end{tabular}  & {\cellcolor[HTML]{FF6E00}} \color[HTML]{F1F1F1} 0.084 & {\cellcolor[HTML]{FFE800}} \color[HTML]{000000} 0.175 & {\cellcolor[HTML]{FF0200}} \color[HTML]{F1F1F1} 0.002 & {\cellcolor[HTML]{FF8E00}} \color[HTML]{F1F1F1} 0.108 & {\cellcolor[HTML]{E2F100}} \color[HTML]{000000} 0.215 \\
\begin{tabular}[c]{@{}l@{}}{Full}\\ {(0k-65k)}\end{tabular} & {\cellcolor[HTML]{FF9A00}} \color[HTML]{000000} 0.116 & {\cellcolor[HTML]{68B400}} \color[HTML]{F1F1F1} 0.306 & {\cellcolor[HTML]{E2F100}} \color[HTML]{000000} 0.215 & {\cellcolor[HTML]{B6DB00}} \color[HTML]{000000} 0.248 & {\cellcolor[HTML]{9CCE00}} \color[HTML]{000000} 0.266 \\
\bottomrule
\end{tabular}}

\end{center}
\caption{Best results on the GR-100+ dataset for Gemma-3-1B-IT model, ROUGE-L. ARMT shows slightly lower overall performance than the base model, but outperforms it on Long-OOD.}
\label{tab:gov_report_best}
\end{table*}

\section{Additional Experimental Results}

\subsection{ARMT language modeling pre-training}\label{app:armt_lm_pretrain}
For a fairer comparison between ARMT fine-tuning and the fine-tuning of other pre-trained models, we also conducted a small-scale pre-training of Gemma-3-1B-IT augmented with ARMT. The model was trained on 19B tokens from the FineWeb-Edu dataset, using concatenated text samples of length 8,192 tokens, which were split into 8 segments of 1,024 tokens each.

\begin{figure}[t]

    \hspace{-1.2cm}
    \includegraphics[width=1.25\linewidth]{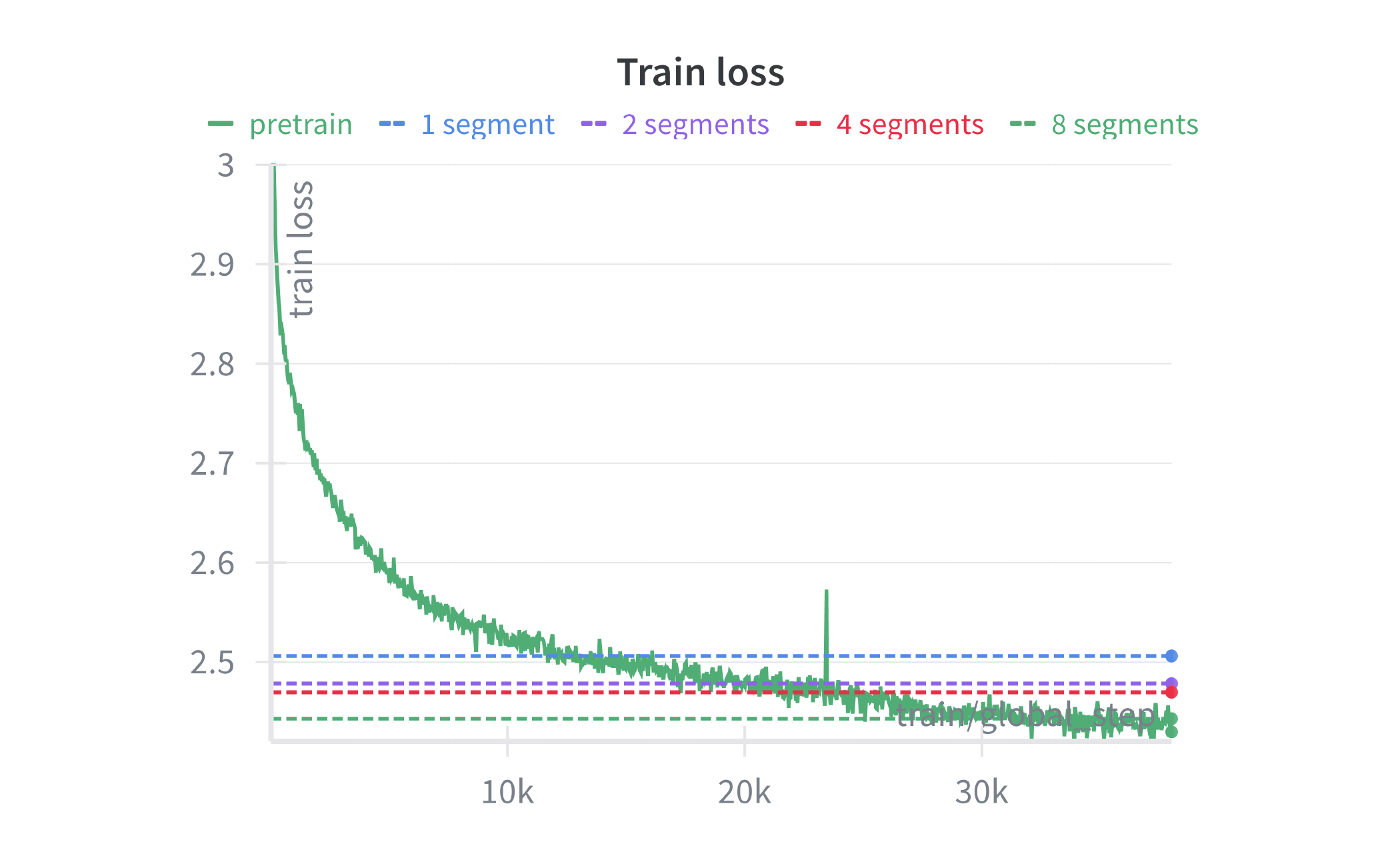}
    \caption{ARMT-augmented Gemma-3-1B-IT pre-training on 19B tokens from the FineWeb-Edu dataset. Pre-training was performed with 8,192-token sequences, divided into 8 segments of 1024 tokens each. We additionally report the final checkpoint loss on the train subset, averaged over the first 1, 2, 4, and 8 segments.}
    \label{fig:armt_pretrain}
\end{figure}

Both ARMT and Gemma parameters were fully trained. We used a batch size of 0.5M tokens with a learning rate of 1e-5 and a linear warmup scheduler with 5k warmup steps. In~\Cref{fig:armt_pretrain}, we show the ARMT pre-training convergence compared to the loss values computed from 1, 2, 4, and 8 segments using the final checkpoint.

The pre-training took approximately 100 hours on an 8×H100-80GB GPU cluster.

\subsection{ARMT Fine-Tuning After Pre-Training}\label{app:armt_ft_pretrain}

\begin{table*}[t]

\begin{center}
\resizebox{0.9\textwidth}{!}{
\begin{tabular}{lcccccccc}
\toprule
\begin{tabular}[c]{@{}l@{}}\textbf{Model\//}\\ \textbf{Lengths}\end{tabular} & \begin{tabular}[c]{@{}l@{}}\textbf{Base,}\\ \textbf{GR-100+,}\\ \textbf{8k}\end{tabular} & \begin{tabular}[c]{@{}l@{}}\textbf{ARMT,}\\ \textbf{GR-100+, 2k}\end{tabular} & \begin{tabular}[c]{@{}l@{}}\textbf{ARMT,}\\ \textbf{GR-100+, 4k}\end{tabular} & \begin{tabular}[c]{@{}l@{}}\textbf{ARMT,}\\ \textbf{GR-100+, 8k}\end{tabular} & \begin{tabular}[c]{@{}l@{}}\textbf{ARMT,}\\ \textbf{GR-100+, 2k,}\\ \textbf{pretrain}\end{tabular} & \begin{tabular}[c]{@{}l@{}}\textbf{ARMT,}\\ \textbf{GR-100+, 4k,}\\ \textbf{pretrain}\end{tabular} & \begin{tabular}[c]{@{}l@{}}\textbf{ARMT,}\\ \textbf{GR-100+, 8k,}\\ \textbf{pretrain}\end{tabular} & \begin{tabular}[c]{@{}l@{}}\textbf{ARMT w/o CL,}\\ \textbf{GR-100+, 8k,}\\ \textbf{pretrain}\end{tabular} \\
\midrule
0k-1k & {\cellcolor[HTML]{068300}} \color[HTML]{F1F1F1} 0.380 & {\cellcolor[HTML]{249200}} \color[HTML]{F1F1F1} 0.357 & {\cellcolor[HTML]{249200}} \color[HTML]{F1F1F1} 0.357 & {\cellcolor[HTML]{229100}} \color[HTML]{F1F1F1} 0.358 & {\cellcolor[HTML]{128900}} \color[HTML]{F1F1F1} 0.371 & {\cellcolor[HTML]{008000}} \color[HTML]{F1F1F1} 0.384 & {\cellcolor[HTML]{128900}} \color[HTML]{F1F1F1} 0.370 & {\cellcolor[HTML]{028100}} \color[HTML]{F1F1F1} 0.382 \\
1k-2k & {\cellcolor[HTML]{008000}} \color[HTML]{F1F1F1} 0.385 & {\cellcolor[HTML]{76BB00}} \color[HTML]{F1F1F1} 0.296 & {\cellcolor[HTML]{78BC00}} \color[HTML]{F1F1F1} 0.294 & {\cellcolor[HTML]{6CB600}} \color[HTML]{F1F1F1} 0.303 & {\cellcolor[HTML]{62B100}} \color[HTML]{F1F1F1} 0.310 & {\cellcolor[HTML]{56AB00}} \color[HTML]{F1F1F1} 0.320 & {\cellcolor[HTML]{5AAD00}} \color[HTML]{F1F1F1} 0.316 & {\cellcolor[HTML]{76BB00}} \color[HTML]{F1F1F1} 0.295 \\
2k-4k & {\cellcolor[HTML]{2A9500}} \color[HTML]{F1F1F1} 0.352 & {\cellcolor[HTML]{A0D000}} \color[HTML]{000000} 0.264 & {\cellcolor[HTML]{92C900}} \color[HTML]{000000} 0.275 & {\cellcolor[HTML]{8CC600}} \color[HTML]{000000} 0.279 & {\cellcolor[HTML]{9ACD00}} \color[HTML]{000000} 0.268 & {\cellcolor[HTML]{6EB700}} \color[HTML]{F1F1F1} 0.301 & {\cellcolor[HTML]{7ABD00}} \color[HTML]{F1F1F1} 0.293 & {\cellcolor[HTML]{6EB700}} \color[HTML]{F1F1F1} 0.302 \\
4k-6k & {\cellcolor[HTML]{369B00}} \color[HTML]{F1F1F1} 0.344 & {\cellcolor[HTML]{C0E000}} \color[HTML]{000000} 0.240 & {\cellcolor[HTML]{82C100}} \color[HTML]{000000} 0.286 & {\cellcolor[HTML]{6CB600}} \color[HTML]{F1F1F1} 0.303 & {\cellcolor[HTML]{AED700}} \color[HTML]{000000} 0.254 & {\cellcolor[HTML]{7ABD00}} \color[HTML]{F1F1F1} 0.293 & {\cellcolor[HTML]{6CB600}} \color[HTML]{F1F1F1} 0.303 & {\cellcolor[HTML]{5CAE00}} \color[HTML]{F1F1F1} 0.315 \\
6k-8k & {\cellcolor[HTML]{76BB00}} \color[HTML]{F1F1F1} 0.296 & {\cellcolor[HTML]{DCEE00}} \color[HTML]{000000} 0.219 & {\cellcolor[HTML]{C2E100}} \color[HTML]{000000} 0.238 & {\cellcolor[HTML]{AED700}} \color[HTML]{000000} 0.254 & {\cellcolor[HTML]{D8EC00}} \color[HTML]{000000} 0.222 & {\cellcolor[HTML]{AAD500}} \color[HTML]{000000} 0.256 & {\cellcolor[HTML]{A6D300}} \color[HTML]{000000} 0.260 & {\cellcolor[HTML]{A4D200}} \color[HTML]{000000} 0.261 \\
8k-10k & {\cellcolor[HTML]{68B400}} \color[HTML]{F1F1F1} 0.306 & {\cellcolor[HTML]{EEF700}} \color[HTML]{000000} 0.205 & {\cellcolor[HTML]{CAE500}} \color[HTML]{000000} 0.233 & {\cellcolor[HTML]{C6E300}} \color[HTML]{000000} 0.235 & {\cellcolor[HTML]{FFB200}} \color[HTML]{000000} 0.135 & {\cellcolor[HTML]{D0E800}} \color[HTML]{000000} 0.228 & {\cellcolor[HTML]{BADD00}} \color[HTML]{000000} 0.245 & {\cellcolor[HTML]{C2E100}} \color[HTML]{000000} 0.238 \\
10k-12k & {\cellcolor[HTML]{9ACD00}} \color[HTML]{000000} 0.269 & {\cellcolor[HTML]{E4F200}} \color[HTML]{000000} 0.213 & {\cellcolor[HTML]{DAED00}} \color[HTML]{000000} 0.221 & {\cellcolor[HTML]{BEDF00}} \color[HTML]{000000} 0.241 & {\cellcolor[HTML]{FF4400}} \color[HTML]{F1F1F1} 0.052 & {\cellcolor[HTML]{CCE600}} \color[HTML]{000000} 0.231 & {\cellcolor[HTML]{A6D300}} \color[HTML]{000000} 0.260 & {\cellcolor[HTML]{A6D300}} \color[HTML]{000000} 0.260 \\
12k-14k & {\cellcolor[HTML]{8AC500}} \color[HTML]{000000} 0.280 & {\cellcolor[HTML]{FEFF00}} \color[HTML]{000000} 0.193 & {\cellcolor[HTML]{D4EA00}} \color[HTML]{000000} 0.225 & {\cellcolor[HTML]{C4E200}} \color[HTML]{000000} 0.237 & {\cellcolor[HTML]{FF2A00}} \color[HTML]{F1F1F1} 0.033 & {\cellcolor[HTML]{C4E200}} \color[HTML]{000000} 0.237 & {\cellcolor[HTML]{B0D800}} \color[HTML]{000000} 0.252 & {\cellcolor[HTML]{B2D900}} \color[HTML]{000000} 0.251 \\
14k-16k & {\cellcolor[HTML]{CCE600}} \color[HTML]{000000} 0.231 & {\cellcolor[HTML]{FFEE00}} \color[HTML]{000000} 0.180 & {\cellcolor[HTML]{DEEF00}} \color[HTML]{000000} 0.217 & {\cellcolor[HTML]{C0E000}} \color[HTML]{000000} 0.240 & {\cellcolor[HTML]{FF2200}} \color[HTML]{F1F1F1} 0.026 & {\cellcolor[HTML]{D6EB00}} \color[HTML]{000000} 0.224 & {\cellcolor[HTML]{A2D100}} \color[HTML]{000000} 0.262 & {\cellcolor[HTML]{BADD00}} \color[HTML]{000000} 0.244 \\
16k-24k & {\cellcolor[HTML]{9ACD00}} \color[HTML]{000000} 0.269 & {\cellcolor[HTML]{FF9800}} \color[HTML]{000000} 0.115 & {\cellcolor[HTML]{F6FB00}} \color[HTML]{000000} 0.200 & {\cellcolor[HTML]{B0D800}} \color[HTML]{000000} 0.252 & {\cellcolor[HTML]{FF0C00}} \color[HTML]{F1F1F1} 0.010 & {\cellcolor[HTML]{B8DC00}} \color[HTML]{000000} 0.246 & {\cellcolor[HTML]{A6D300}} \color[HTML]{000000} 0.260 & {\cellcolor[HTML]{94CA00}} \color[HTML]{000000} 0.273 \\
24k-32k & {\cellcolor[HTML]{B6DB00}} \color[HTML]{000000} 0.247 & {\cellcolor[HTML]{FF1400}} \color[HTML]{F1F1F1} 0.016 & {\cellcolor[HTML]{FFDE00}} \color[HTML]{000000} 0.168 & {\cellcolor[HTML]{ECF600}} \color[HTML]{000000} 0.207 & {\cellcolor[HTML]{FF0000}} \color[HTML]{F1F1F1} 0.001 & {\cellcolor[HTML]{FEFF00}} \color[HTML]{000000} 0.194 & {\cellcolor[HTML]{DAED00}} \color[HTML]{000000} 0.221 & {\cellcolor[HTML]{EAF500}} \color[HTML]{000000} 0.209 \\
32k-49k & {\cellcolor[HTML]{FFE200}} \color[HTML]{000000} 0.171 & {\cellcolor[HTML]{FF0200}} \color[HTML]{F1F1F1} 0.002 & {\cellcolor[HTML]{FFAA00}} \color[HTML]{000000} 0.128 & {\cellcolor[HTML]{F6FB00}} \color[HTML]{000000} 0.200 & {\cellcolor[HTML]{FF0000}} \color[HTML]{F1F1F1} 0.000 & {\cellcolor[HTML]{F6FB00}} \color[HTML]{000000} 0.199 & {\cellcolor[HTML]{ECF600}} \color[HTML]{000000} 0.207 & {\cellcolor[HTML]{D6EB00}} \color[HTML]{000000} 0.223 \\
49k-65k & {\cellcolor[HTML]{FFFA00}} \color[HTML]{000000} 0.189 & {\cellcolor[HTML]{FF0000}} \color[HTML]{F1F1F1} 0.000 & {\cellcolor[HTML]{FF3400}} \color[HTML]{F1F1F1} 0.040 & {\cellcolor[HTML]{A0D000}} \color[HTML]{000000} 0.264 & {\cellcolor[HTML]{FF0000}} \color[HTML]{F1F1F1} 0.000 & {\cellcolor[HTML]{7EBF00}} \color[HTML]{000000} 0.289 & {\cellcolor[HTML]{70B800}} \color[HTML]{F1F1F1} 0.300 & {\cellcolor[HTML]{84C200}} \color[HTML]{000000} 0.285 \\
\hline
\begin{tabular}[c]{@{}l@{}}{In-Domain}\\ {(0k-8k)}\end{tabular} & {\cellcolor[HTML]{2C9600}} \color[HTML]{F1F1F1} 0.351 & {\cellcolor[HTML]{92C900}} \color[HTML]{000000} 0.275 & {\cellcolor[HTML]{7EBF00}} \color[HTML]{000000} 0.290 & {\cellcolor[HTML]{72B900}} \color[HTML]{F1F1F1} 0.299 & {\cellcolor[HTML]{84C200}} \color[HTML]{000000} 0.285 & {\cellcolor[HTML]{62B100}} \color[HTML]{F1F1F1} 0.311 & {\cellcolor[HTML]{66B300}} \color[HTML]{F1F1F1} 0.308 & {\cellcolor[HTML]{62B100}} \color[HTML]{F1F1F1} 0.311 \\
\begin{tabular}[c]{@{}l@{}}{OOD}\\ {(8k-65k)}\end{tabular} & {\cellcolor[HTML]{9ECF00}} \color[HTML]{000000} 0.266 & {\cellcolor[HTML]{FFD600}} \color[HTML]{000000} 0.162 & {\cellcolor[HTML]{E6F300}} \color[HTML]{000000} 0.211 & {\cellcolor[HTML]{C2E100}} \color[HTML]{000000} 0.238 & {\cellcolor[HTML]{FF3C00}} \color[HTML]{F1F1F1} 0.045 & {\cellcolor[HTML]{CCE600}} \color[HTML]{000000} 0.231 & {\cellcolor[HTML]{B0D800}} \color[HTML]{000000} 0.252 & {\cellcolor[HTML]{B2D900}} \color[HTML]{000000} 0.250 \\
\begin{tabular}[c]{@{}l@{}}{Long-OOD}\\ {(32k-65k)}\end{tabular} & {\cellcolor[HTML]{FFE800}} \color[HTML]{000000} 0.175 & {\cellcolor[HTML]{FF0200}} \color[HTML]{F1F1F1} 0.002 & {\cellcolor[HTML]{FF8E00}} \color[HTML]{F1F1F1} 0.108 & {\cellcolor[HTML]{E2F100}} \color[HTML]{000000} 0.215 & {\cellcolor[HTML]{FF0000}} \color[HTML]{F1F1F1} 0.000 & {\cellcolor[HTML]{DAED00}} \color[HTML]{000000} 0.220 & {\cellcolor[HTML]{D0E800}} \color[HTML]{000000} 0.228 & {\cellcolor[HTML]{C4E200}} \color[HTML]{000000} 0.237 \\
\begin{tabular}[c]{@{}l@{}}{Full}\\ {(0k-65k)}\end{tabular} & {\cellcolor[HTML]{68B400}} \color[HTML]{F1F1F1} 0.306 & {\cellcolor[HTML]{E2F100}} \color[HTML]{000000} 0.215 & {\cellcolor[HTML]{B6DB00}} \color[HTML]{000000} 0.248 & {\cellcolor[HTML]{9CCE00}} \color[HTML]{000000} 0.266 & {\cellcolor[HTML]{FFD000}} \color[HTML]{000000} 0.157 & {\cellcolor[HTML]{9ACD00}} \color[HTML]{000000} 0.268 & {\cellcolor[HTML]{8CC600}} \color[HTML]{000000} 0.278 & {\cellcolor[HTML]{8CC600}} \color[HTML]{000000} 0.278 \\
\bottomrule
\end{tabular}}
\caption{Results on the GR-100+ dataset for Gemma-3-1B-IT model with ARMT after continuous pretraining, ROUGE-L. ARMT after continuous pretraining shows significantly better results on all splits.}
\label{tab:gov_report_best_with_pretrain}
\end{center}
\end{table*}
\begin{table*}[t]

\begin{center}
\resizebox{0.85\textwidth}{!}{
\begin{tabular}{lcccccccc}
\toprule
\begin{tabular}[c]{@{}l@{}}\textbf{Model\//}\\ \textbf{Lengths}\end{tabular} & \begin{tabular}[c]{@{}l@{}}\textbf{Base,}\\ \textbf{MT,}\\ \textbf{8k}\end{tabular} & \begin{tabular}[c]{@{}l@{}}\textbf{ARMT,}\\ \textbf{MT, 2k}\end{tabular} & \begin{tabular}[c]{@{}l@{}}\textbf{ARMT,}\\ \textbf{MT, 4k}\end{tabular} & \begin{tabular}[c]{@{}l@{}}\textbf{ARMT,}\\ \textbf{MT, 8k}\end{tabular} & \begin{tabular}[c]{@{}l@{}}\textbf{ARMT,}\\ \textbf{MT, 2k,}\\ \textbf{pretrain}\end{tabular} & \begin{tabular}[c]{@{}l@{}}\textbf{ARMT,}\\ \textbf{MT, 4k,}\\ \textbf{pretrain}\end{tabular} & \begin{tabular}[c]{@{}l@{}}\textbf{ARMT,}\\ \textbf{MT, 8k,}\\ \textbf{pretrain}\end{tabular} \\
\midrule
0k-1k & {\cellcolor[HTML]{2E9700}} \color[HTML]{F1F1F1} 0.795 & {\cellcolor[HTML]{389C00}} \color[HTML]{F1F1F1} 0.782 & {\cellcolor[HTML]{389C00}} \color[HTML]{F1F1F1} 0.782 & {\cellcolor[HTML]{48A400}} \color[HTML]{F1F1F1} 0.756 & {\cellcolor[HTML]{40A000}} \color[HTML]{F1F1F1} 0.769 & {\cellcolor[HTML]{48A400}} \color[HTML]{F1F1F1} 0.756 & {\cellcolor[HTML]{48A400}} \color[HTML]{F1F1F1} 0.756 \\
1k-2k & {\cellcolor[HTML]{329900}} \color[HTML]{F1F1F1} 0.790 & {\cellcolor[HTML]{5EAF00}} \color[HTML]{F1F1F1} 0.724 & {\cellcolor[HTML]{64B200}} \color[HTML]{F1F1F1} 0.714 & {\cellcolor[HTML]{52A900}} \color[HTML]{F1F1F1} 0.743 & {\cellcolor[HTML]{4CA600}} \color[HTML]{F1F1F1} 0.752 & {\cellcolor[HTML]{52A900}} \color[HTML]{F1F1F1} 0.743 & {\cellcolor[HTML]{52A900}} \color[HTML]{F1F1F1} 0.743 \\
2k-4k & {\cellcolor[HTML]{3C9E00}} \color[HTML]{F1F1F1} 0.774 & {\cellcolor[HTML]{74BA00}} \color[HTML]{F1F1F1} 0.689 & {\cellcolor[HTML]{6EB700}} \color[HTML]{F1F1F1} 0.698 & {\cellcolor[HTML]{56AB00}} \color[HTML]{F1F1F1} 0.736 & {\cellcolor[HTML]{68B400}} \color[HTML]{F1F1F1} 0.708 & {\cellcolor[HTML]{68B400}} \color[HTML]{F1F1F1} 0.708 & {\cellcolor[HTML]{56AB00}} \color[HTML]{F1F1F1} 0.736 \\
4k-6k & {\cellcolor[HTML]{42A100}} \color[HTML]{F1F1F1} 0.767 & {\cellcolor[HTML]{A4D200}} \color[HTML]{000000} 0.616 & {\cellcolor[HTML]{92C900}} \color[HTML]{000000} 0.644 & {\cellcolor[HTML]{6EB700}} \color[HTML]{F1F1F1} 0.699 & {\cellcolor[HTML]{80C000}} \color[HTML]{000000} 0.671 & {\cellcolor[HTML]{80C000}} \color[HTML]{000000} 0.671 & {\cellcolor[HTML]{54AA00}} \color[HTML]{F1F1F1} 0.740 \\
6k-8k & {\cellcolor[HTML]{048200}} \color[HTML]{F1F1F1} 0.859 & {\cellcolor[HTML]{AAD500}} \color[HTML]{000000} 0.609 & {\cellcolor[HTML]{62B100}} \color[HTML]{F1F1F1} 0.717 & {\cellcolor[HTML]{2A9500}} \color[HTML]{F1F1F1} 0.804 & {\cellcolor[HTML]{309800}} \color[HTML]{F1F1F1} 0.793 & {\cellcolor[HTML]{2A9500}} \color[HTML]{F1F1F1} 0.804 & {\cellcolor[HTML]{048200}} \color[HTML]{F1F1F1} 0.859 \\
8k-10k & {\cellcolor[HTML]{0A8500}} \color[HTML]{F1F1F1} 0.852 & {\cellcolor[HTML]{ECF600}} \color[HTML]{000000} 0.506 & {\cellcolor[HTML]{4AA500}} \color[HTML]{F1F1F1} 0.753 & {\cellcolor[HTML]{4AA500}} \color[HTML]{F1F1F1} 0.753 & {\cellcolor[HTML]{4AA500}} \color[HTML]{F1F1F1} 0.753 & {\cellcolor[HTML]{329900}} \color[HTML]{F1F1F1} 0.790 & {\cellcolor[HTML]{1A8D00}} \color[HTML]{F1F1F1} 0.827 \\
10k-12k & {\cellcolor[HTML]{008000}} \color[HTML]{F1F1F1} 0.868 & {\cellcolor[HTML]{FFBA00}} \color[HTML]{000000} 0.374 & {\cellcolor[HTML]{82C100}} \color[HTML]{000000} 0.670 & {\cellcolor[HTML]{329900}} \color[HTML]{F1F1F1} 0.791 & {\cellcolor[HTML]{72B900}} \color[HTML]{F1F1F1} 0.692 & {\cellcolor[HTML]{4EA700}} \color[HTML]{F1F1F1} 0.747 & {\cellcolor[HTML]{008000}} \color[HTML]{F1F1F1} 0.868 \\
12k-14k & {\cellcolor[HTML]{349A00}} \color[HTML]{F1F1F1} 0.787 & {\cellcolor[HTML]{FFCE00}} \color[HTML]{000000} 0.404 & {\cellcolor[HTML]{6CB600}} \color[HTML]{F1F1F1} 0.702 & {\cellcolor[HTML]{50A800}} \color[HTML]{F1F1F1} 0.745 & {\cellcolor[HTML]{50A800}} \color[HTML]{F1F1F1} 0.745 & {\cellcolor[HTML]{58AC00}} \color[HTML]{F1F1F1} 0.734 & {\cellcolor[HTML]{42A100}} \color[HTML]{F1F1F1} 0.766 \\
14k-16k & {\cellcolor[HTML]{44A200}} \color[HTML]{F1F1F1} 0.764 & {\cellcolor[HTML]{FFA200}} \color[HTML]{000000} 0.337 & {\cellcolor[HTML]{94CA00}} \color[HTML]{000000} 0.640 & {\cellcolor[HTML]{78BC00}} \color[HTML]{F1F1F1} 0.685 & {\cellcolor[HTML]{9CCE00}} \color[HTML]{000000} 0.629 & {\cellcolor[HTML]{94CA00}} \color[HTML]{000000} 0.640 & {\cellcolor[HTML]{70B800}} \color[HTML]{F1F1F1} 0.697 \\
16k-24k & {\cellcolor[HTML]{249200}} \color[HTML]{F1F1F1} 0.812 & {\cellcolor[HTML]{FF5C00}} \color[HTML]{F1F1F1} 0.229 & {\cellcolor[HTML]{90C800}} \color[HTML]{000000} 0.646 & {\cellcolor[HTML]{48A400}} \color[HTML]{F1F1F1} 0.757 & {\cellcolor[HTML]{9ACD00}} \color[HTML]{000000} 0.632 & {\cellcolor[HTML]{96CB00}} \color[HTML]{000000} 0.639 & {\cellcolor[HTML]{48A400}} \color[HTML]{F1F1F1} 0.757 \\
24k-32k & {\cellcolor[HTML]{64B200}} \color[HTML]{F1F1F1} 0.713 & {\cellcolor[HTML]{FF4600}} \color[HTML]{F1F1F1} 0.198 & {\cellcolor[HTML]{BADD00}} \color[HTML]{000000} 0.584 & {\cellcolor[HTML]{64B200}} \color[HTML]{F1F1F1} 0.713 & {\cellcolor[HTML]{BADD00}} \color[HTML]{000000} 0.584 & {\cellcolor[HTML]{80C000}} \color[HTML]{000000} 0.673 & {\cellcolor[HTML]{249200}} \color[HTML]{F1F1F1} 0.812 \\
32k-49k & {\cellcolor[HTML]{B6DB00}} \color[HTML]{000000} 0.591 & {\cellcolor[HTML]{FF2000}} \color[HTML]{F1F1F1} 0.139 & {\cellcolor[HTML]{AAD500}} \color[HTML]{000000} 0.609 & {\cellcolor[HTML]{42A100}} \color[HTML]{F1F1F1} 0.765 & {\cellcolor[HTML]{DEEF00}} \color[HTML]{000000} 0.530 & {\cellcolor[HTML]{AAD500}} \color[HTML]{000000} 0.609 & {\cellcolor[HTML]{329900}} \color[HTML]{F1F1F1} 0.791 \\
49k-65k & {\cellcolor[HTML]{FF9400}} \color[HTML]{000000} 0.317 & {\cellcolor[HTML]{FF0000}} \color[HTML]{F1F1F1} 0.089 & {\cellcolor[HTML]{ACD600}} \color[HTML]{000000} 0.604 & {\cellcolor[HTML]{64B200}} \color[HTML]{F1F1F1} 0.713 & {\cellcolor[HTML]{FFDC00}} \color[HTML]{000000} 0.426 & {\cellcolor[HTML]{D4EA00}} \color[HTML]{000000} 0.545 & {\cellcolor[HTML]{44A200}} \color[HTML]{F1F1F1} 0.762 \\
\hline
In-Domain (0k-8k) & {\cellcolor[HTML]{2E9700}} \color[HTML]{F1F1F1} 0.797 & {\cellcolor[HTML]{78BC00}} \color[HTML]{F1F1F1} 0.685 & {\cellcolor[HTML]{66B300}} \color[HTML]{F1F1F1} 0.711 & {\cellcolor[HTML]{4EA700}} \color[HTML]{F1F1F1} 0.749 & {\cellcolor[HTML]{54AA00}} \color[HTML]{F1F1F1} 0.740 & {\cellcolor[HTML]{54AA00}} \color[HTML]{F1F1F1} 0.738 & {\cellcolor[HTML]{42A100}} \color[HTML]{F1F1F1} 0.767 \\
OOD (8k-65k) & {\cellcolor[HTML]{68B400}} \color[HTML]{F1F1F1} 0.709 & {\cellcolor[HTML]{FF7600}} \color[HTML]{F1F1F1} 0.271 & {\cellcolor[HTML]{90C800}} \color[HTML]{000000} 0.647 & {\cellcolor[HTML]{52A900}} \color[HTML]{F1F1F1} 0.741 & {\cellcolor[HTML]{A4D200}} \color[HTML]{000000} 0.618 & {\cellcolor[HTML]{84C200}} \color[HTML]{000000} 0.665 & {\cellcolor[HTML]{369B00}} \color[HTML]{F1F1F1} 0.783 \\
Long-OOD (32k-65k) & {\cellcolor[HTML]{FFF400}} \color[HTML]{000000} 0.463 & {\cellcolor[HTML]{FF1000}} \color[HTML]{F1F1F1} 0.116 & {\cellcolor[HTML]{AAD500}} \color[HTML]{000000} 0.607 & {\cellcolor[HTML]{52A900}} \color[HTML]{F1F1F1} 0.741 & {\cellcolor[HTML]{FEFF00}} \color[HTML]{000000} 0.481 & {\cellcolor[HTML]{BCDE00}} \color[HTML]{000000} 0.579 & {\cellcolor[HTML]{3A9D00}} \color[HTML]{F1F1F1} 0.777 \\
Full (0k-65k) & {\cellcolor[HTML]{52A900}} \color[HTML]{F1F1F1} 0.741 & {\cellcolor[HTML]{FFD800}} \color[HTML]{000000} 0.419 & {\cellcolor[HTML]{82C100}} \color[HTML]{000000} 0.670 & {\cellcolor[HTML]{50A800}} \color[HTML]{F1F1F1} 0.744 & {\cellcolor[HTML]{86C300}} \color[HTML]{000000} 0.661 & {\cellcolor[HTML]{74BA00}} \color[HTML]{F1F1F1} 0.691 & {\cellcolor[HTML]{3A9D00}} \color[HTML]{F1F1F1} 0.777 \\
\bottomrule
\end{tabular}}
\caption{Results on the MT dataset for Gemma-3-1B-IT model with ARMT after continuous pretraining, EM. ARMT after continuous pretraining shows significantly better results on all splits.}
\label{tab:mt_best_with_pretrain}
\end{center}
\end{table*}

After ARMT pre-training, we also conducted fine-tuning using the same setup as for the ARMT model without continued pre-training on GR-100+ and MT datasets. The results are shown in~\Cref{tab:gov_report_best_with_pretrain,tab:mt_best_with_pretrain}. For both datasets, pre-training substantially improves performance on downstream tasks, especially for long-context samples from 8k to 65k in length.

\subsection{Ablation Study for Associative Memory Layers}\label{app:assoc_ablation}
\begin{table*}[t]

\begin{center}
\resizebox{0.7\linewidth}{!}{
\begin{tabular}{lcccccccc}
\toprule
\begin{tabular}[c]{@{}l@{}}\textbf{Model\//}\\ \textbf{Lengths}\end{tabular} & \begin{tabular}[c]{@{}l@{}}\textbf{Base,}\\ \textbf{MT,}\\ \textbf{8k}\end{tabular} & \begin{tabular}[c]{@{}l@{}}\textbf{ARMT,}\\ \textbf{MT,}\\ \textbf{8k}\end{tabular} & \begin{tabular}[c]{@{}l@{}}\textbf{ARMT,}\\ \textbf{ MT, 8k,}\\ \textbf{w/o layers}\\ \textbf{0-6}\end{tabular} & \begin{tabular}[c]{@{}l@{}}\textbf{ARMT,}\\ \textbf{ MT, 8k,}\\ \textbf{w/o layers}\\ \textbf{7-12}\end{tabular} & \begin{tabular}[c]{@{}l@{}}\textbf{ARMT,}\\ \textbf{ MT, 8k,}\\ \textbf{w/o layers}\\ \textbf{13-18}\end{tabular} & \begin{tabular}[c]{@{}l@{}}\textbf{ARMT,}\\ \textbf{ MT, 8k,}\\ \textbf{w/o layers}\\ \textbf{19-25}\end{tabular} & \begin{tabular}[c]{@{}l@{}}\textbf{ARMT,}\\ \textbf{MT, 8k,}\\ \textbf{only top-1}\\ \textbf{layer}\end{tabular} & \begin{tabular}[c]{@{}l@{}}\textbf{ARMT,}\\ \textbf{ MT, 8k,}\\ \textbf{only top-4}\\ \textbf{layers}\end{tabular}\\
\midrule
0k-1k & {\cellcolor[HTML]{3A9D00}} \color[HTML]{F1F1F1} 0.795 & {\cellcolor[HTML]{58AC00}} \color[HTML]{F1F1F1} 0.756 & {\cellcolor[HTML]{58AC00}} \color[HTML]{F1F1F1} 0.756 & {\cellcolor[HTML]{58AC00}} \color[HTML]{F1F1F1} 0.756 & {\cellcolor[HTML]{58AC00}} \color[HTML]{F1F1F1} 0.756 & {\cellcolor[HTML]{58AC00}} \color[HTML]{F1F1F1} 0.756 & {\cellcolor[HTML]{58AC00}} \color[HTML]{F1F1F1} 0.756 & {\cellcolor[HTML]{58AC00}} \color[HTML]{F1F1F1} 0.756 \\
1k-2k & {\cellcolor[HTML]{3E9F00}} \color[HTML]{F1F1F1} 0.790 & {\cellcolor[HTML]{62B100}} \color[HTML]{F1F1F1} 0.743 & {\cellcolor[HTML]{6AB500}} \color[HTML]{F1F1F1} 0.733 & {\cellcolor[HTML]{62B100}} \color[HTML]{F1F1F1} 0.743 & {\cellcolor[HTML]{ECF600}} \color[HTML]{000000} 0.571 & {\cellcolor[HTML]{98CC00}} \color[HTML]{000000} 0.676 & {\cellcolor[HTML]{DCEE00}} \color[HTML]{000000} 0.590 & {\cellcolor[HTML]{72B900}} \color[HTML]{F1F1F1} 0.724 \\
2k-4k & {\cellcolor[HTML]{4AA500}} \color[HTML]{F1F1F1} 0.774 & {\cellcolor[HTML]{68B400}} \color[HTML]{F1F1F1} 0.736 & {\cellcolor[HTML]{68B400}} \color[HTML]{F1F1F1} 0.736 & {\cellcolor[HTML]{70B800}} \color[HTML]{F1F1F1} 0.726 & {\cellcolor[HTML]{FFAC00}} \color[HTML]{000000} 0.443 & {\cellcolor[HTML]{78BC00}} \color[HTML]{F1F1F1} 0.717 & {\cellcolor[HTML]{DAED00}} \color[HTML]{000000} 0.594 & {\cellcolor[HTML]{7EBF00}} \color[HTML]{000000} 0.708 \\
4k-6k & {\cellcolor[HTML]{50A800}} \color[HTML]{F1F1F1} 0.767 & {\cellcolor[HTML]{86C300}} \color[HTML]{000000} 0.699 & {\cellcolor[HTML]{86C300}} \color[HTML]{000000} 0.699 & {\cellcolor[HTML]{86C300}} \color[HTML]{000000} 0.699 & {\cellcolor[HTML]{FF5200}} \color[HTML]{F1F1F1} 0.329 & {\cellcolor[HTML]{9CCE00}} \color[HTML]{000000} 0.671 & {\cellcolor[HTML]{FFEA00}} \color[HTML]{000000} 0.521 & {\cellcolor[HTML]{7CBE00}} \color[HTML]{000000} 0.712 \\
6k-8k & {\cellcolor[HTML]{068300}} \color[HTML]{F1F1F1} 0.859 & {\cellcolor[HTML]{329900}} \color[HTML]{F1F1F1} 0.804 & {\cellcolor[HTML]{2A9500}} \color[HTML]{F1F1F1} 0.815 & {\cellcolor[HTML]{329900}} \color[HTML]{F1F1F1} 0.804 & {\cellcolor[HTML]{FF4600}} \color[HTML]{F1F1F1} 0.315 & {\cellcolor[HTML]{66B300}} \color[HTML]{F1F1F1} 0.739 & {\cellcolor[HTML]{D6EB00}} \color[HTML]{000000} 0.598 & {\cellcolor[HTML]{3A9D00}} \color[HTML]{F1F1F1} 0.793 \\
8k-10k & {\cellcolor[HTML]{0C8600}} \color[HTML]{F1F1F1} 0.852 & {\cellcolor[HTML]{5AAD00}} \color[HTML]{F1F1F1} 0.753 & {\cellcolor[HTML]{5AAD00}} \color[HTML]{F1F1F1} 0.753 & {\cellcolor[HTML]{52A900}} \color[HTML]{F1F1F1} 0.765 & {\cellcolor[HTML]{FF8600}} \color[HTML]{F1F1F1} 0.395 & {\cellcolor[HTML]{78BC00}} \color[HTML]{F1F1F1} 0.716 & {\cellcolor[HTML]{FFF200}} \color[HTML]{000000} 0.531 & {\cellcolor[HTML]{52A900}} \color[HTML]{F1F1F1} 0.765 \\
10k-12k & {\cellcolor[HTML]{008000}} \color[HTML]{F1F1F1} 0.868 & {\cellcolor[HTML]{3C9E00}} \color[HTML]{F1F1F1} 0.791 & {\cellcolor[HTML]{46A300}} \color[HTML]{F1F1F1} 0.780 & {\cellcolor[HTML]{3C9E00}} \color[HTML]{F1F1F1} 0.791 & {\cellcolor[HTML]{FF5A00}} \color[HTML]{F1F1F1} 0.341 & {\cellcolor[HTML]{68B400}} \color[HTML]{F1F1F1} 0.736 & {\cellcolor[HTML]{FFDE00}} \color[HTML]{000000} 0.505 & {\cellcolor[HTML]{56AB00}} \color[HTML]{F1F1F1} 0.758 \\
12k-14k & {\cellcolor[HTML]{40A000}} \color[HTML]{F1F1F1} 0.787 & {\cellcolor[HTML]{62B100}} \color[HTML]{F1F1F1} 0.745 & {\cellcolor[HTML]{50A800}} \color[HTML]{F1F1F1} 0.766 & {\cellcolor[HTML]{50A800}} \color[HTML]{F1F1F1} 0.766 & {\cellcolor[HTML]{FF2800}} \color[HTML]{F1F1F1} 0.277 & {\cellcolor[HTML]{2E9700}} \color[HTML]{F1F1F1} 0.809 & {\cellcolor[HTML]{FFB800}} \color[HTML]{000000} 0.457 & {\cellcolor[HTML]{5AAD00}} \color[HTML]{F1F1F1} 0.755 \\
14k-16k & {\cellcolor[HTML]{52A900}} \color[HTML]{F1F1F1} 0.764 & {\cellcolor[HTML]{90C800}} \color[HTML]{000000} 0.685 & {\cellcolor[HTML]{9ACD00}} \color[HTML]{000000} 0.674 & {\cellcolor[HTML]{9ACD00}} \color[HTML]{000000} 0.674 & {\cellcolor[HTML]{FF5800}} \color[HTML]{F1F1F1} 0.337 & {\cellcolor[HTML]{7EBF00}} \color[HTML]{000000} 0.708 & {\cellcolor[HTML]{FFC400}} \color[HTML]{000000} 0.472 & {\cellcolor[HTML]{A2D100}} \color[HTML]{000000} 0.663 \\
16k-24k & {\cellcolor[HTML]{2C9600}} \color[HTML]{F1F1F1} 0.812 & {\cellcolor[HTML]{58AC00}} \color[HTML]{F1F1F1} 0.757 & {\cellcolor[HTML]{5EAF00}} \color[HTML]{F1F1F1} 0.750 & {\cellcolor[HTML]{52A900}} \color[HTML]{F1F1F1} 0.764 & {\cellcolor[HTML]{FF3E00}} \color[HTML]{F1F1F1} 0.306 & {\cellcolor[HTML]{74BA00}} \color[HTML]{F1F1F1} 0.722 & {\cellcolor[HTML]{FFB800}} \color[HTML]{000000} 0.458 & {\cellcolor[HTML]{6EB700}} \color[HTML]{F1F1F1} 0.729 \\
24k-32k & {\cellcolor[HTML]{7ABD00}} \color[HTML]{F1F1F1} 0.713 & {\cellcolor[HTML]{7ABD00}} \color[HTML]{F1F1F1} 0.713 & {\cellcolor[HTML]{82C100}} \color[HTML]{000000} 0.703 & {\cellcolor[HTML]{82C100}} \color[HTML]{000000} 0.703 & {\cellcolor[HTML]{FF3000}} \color[HTML]{F1F1F1} 0.287 & {\cellcolor[HTML]{92C900}} \color[HTML]{000000} 0.683 & {\cellcolor[HTML]{FFA600}} \color[HTML]{000000} 0.436 & {\cellcolor[HTML]{8AC500}} \color[HTML]{000000} 0.693 \\
32k-49k & {\cellcolor[HTML]{DCEE00}} \color[HTML]{000000} 0.591 & {\cellcolor[HTML]{52A900}} \color[HTML]{F1F1F1} 0.765 & {\cellcolor[HTML]{58AC00}} \color[HTML]{F1F1F1} 0.757 & {\cellcolor[HTML]{52A900}} \color[HTML]{F1F1F1} 0.765 & {\cellcolor[HTML]{FF0000}} \color[HTML]{F1F1F1} 0.226 & {\cellcolor[HTML]{66B300}} \color[HTML]{F1F1F1} 0.739 & {\cellcolor[HTML]{FFDC00}} \color[HTML]{000000} 0.504 & {\cellcolor[HTML]{52A900}} \color[HTML]{F1F1F1} 0.765 \\
49k-65k & {\cellcolor[HTML]{FF4800}} \color[HTML]{F1F1F1} 0.317 & {\cellcolor[HTML]{7ABD00}} \color[HTML]{F1F1F1} 0.713 & {\cellcolor[HTML]{72B900}} \color[HTML]{F1F1F1} 0.723 & {\cellcolor[HTML]{72B900}} \color[HTML]{F1F1F1} 0.723 & {\cellcolor[HTML]{FF6600}} \color[HTML]{F1F1F1} 0.356 & {\cellcolor[HTML]{92C900}} \color[HTML]{000000} 0.683 & {\cellcolor[HTML]{FFDE00}} \color[HTML]{000000} 0.505 & {\cellcolor[HTML]{7ABD00}} \color[HTML]{F1F1F1} 0.713 \\
\hline
\begin{tabular}[c]{@{}l@{}}{In-Domain}\\ {(0k-8k)}\end{tabular} & {\cellcolor[HTML]{389C00}} \color[HTML]{F1F1F1} 0.797 & {\cellcolor[HTML]{5EAF00}} \color[HTML]{F1F1F1} 0.749 & {\cellcolor[HTML]{5EAF00}} \color[HTML]{F1F1F1} 0.749 & {\cellcolor[HTML]{60B000}} \color[HTML]{F1F1F1} 0.747 & {\cellcolor[HTML]{FFCC00}} \color[HTML]{000000} 0.482 & {\cellcolor[HTML]{7CBE00}} \color[HTML]{000000} 0.711 & {\cellcolor[HTML]{CCE600}} \color[HTML]{000000} 0.610 & {\cellcolor[HTML]{66B300}} \color[HTML]{F1F1F1} 0.738 \\
\begin{tabular}[c]{@{}l@{}}{OOD}\\ {(8k-65k)}\end{tabular} & {\cellcolor[HTML]{7EBF00}} \color[HTML]{000000} 0.709 & {\cellcolor[HTML]{64B200}} \color[HTML]{F1F1F1} 0.741 & {\cellcolor[HTML]{66B300}} \color[HTML]{F1F1F1} 0.739 & {\cellcolor[HTML]{62B100}} \color[HTML]{F1F1F1} 0.745 & {\cellcolor[HTML]{FF4400}} \color[HTML]{F1F1F1} 0.311 & {\cellcolor[HTML]{72B900}} \color[HTML]{F1F1F1} 0.724 & {\cellcolor[HTML]{FFCA00}} \color[HTML]{000000} 0.481 & {\cellcolor[HTML]{6CB600}} \color[HTML]{F1F1F1} 0.730 \\
\begin{tabular}[c]{@{}l@{}}{Long-OOD}\\ {(32k-65k)}\end{tabular} & {\cellcolor[HTML]{FFBC00}} \color[HTML]{000000} 0.463 & {\cellcolor[HTML]{64B200}} \color[HTML]{F1F1F1} 0.741 & {\cellcolor[HTML]{64B200}} \color[HTML]{F1F1F1} 0.741 & {\cellcolor[HTML]{60B000}} \color[HTML]{F1F1F1} 0.745 & {\cellcolor[HTML]{FF3000}} \color[HTML]{F1F1F1} 0.287 & {\cellcolor[HTML]{7ABD00}} \color[HTML]{F1F1F1} 0.713 & {\cellcolor[HTML]{FFDE00}} \color[HTML]{000000} 0.504 & {\cellcolor[HTML]{64B200}} \color[HTML]{F1F1F1} 0.741 \\
\begin{tabular}[c]{@{}l@{}}{Full}\\ {(0k-65k)}\end{tabular} & {\cellcolor[HTML]{64B200}} \color[HTML]{F1F1F1} 0.741 & {\cellcolor[HTML]{62B100}} \color[HTML]{F1F1F1} 0.744 & {\cellcolor[HTML]{64B200}} \color[HTML]{F1F1F1} 0.743 & {\cellcolor[HTML]{60B000}} \color[HTML]{F1F1F1} 0.746 & {\cellcolor[HTML]{FF7400}} \color[HTML]{F1F1F1} 0.372 & {\cellcolor[HTML]{76BB00}} \color[HTML]{F1F1F1} 0.720 & {\cellcolor[HTML]{FFF000}} \color[HTML]{000000} 0.527 & {\cellcolor[HTML]{6AB500}} \color[HTML]{F1F1F1} 0.733 \\
\bottomrule
\end{tabular}}
\caption{Associative layers ablation on the MT dataset for Gemma-3-1B-IT model, metric - EM. Middle and upper layers representations are the most important for associative memory; ARMT with only top-4 associative blocks keeps almost the same performance as the full ARMT.}
\label{tab:mt_best_assoc_ablation}
\end{center}
\end{table*}

\begin{table*}[t]

\begin{center}
\resizebox{0.9\textwidth}{!}{
\begin{tabular}{lcccccccc}
\toprule
\begin{tabular}[c]{@{}l@{}}\textbf{Model\//}\\ \textbf{Lengths}\end{tabular} & \begin{tabular}[c]{@{}l@{}}\textbf{Base,}\\ \textbf{GR-100+,}\\ \textbf{8k}\end{tabular} & \begin{tabular}[c]{@{}l@{}}\textbf{ARMT,}\\ \textbf{GR-100+,}\\ \textbf{8k}\end{tabular} & \begin{tabular}[c]{@{}l@{}}\textbf{ARMT,}\\ \textbf{ GR-100+, 8k,}\\ \textbf{w/o layers 0-6}\end{tabular} & \begin{tabular}[c]{@{}l@{}}\textbf{ARMT,}\\ \textbf{ GR-100+, 8k,}\\ \textbf{w/o layers 7-12}\end{tabular} & \begin{tabular}[c]{@{}l@{}}\textbf{ARMT,}\\ \textbf{ GR-100+, 8k,}\\ \textbf{w/o layers 13-18}\end{tabular} & \begin{tabular}[c]{@{}l@{}}\textbf{ARMT,}\\ \textbf{ GR-100+, 8k,}\\ \textbf{w/o layers 19-25}\end{tabular} & \begin{tabular}[c]{@{}l@{}}\textbf{ARMT,}\\ \textbf{ GR-100+, 8k,}\\ \textbf{only top-1 layer}\end{tabular} & \begin{tabular}[c]{@{}l@{}}\textbf{ARMT,}\\ \textbf{ GR-100+, 8k,}\\ \textbf{only top-4 layers}\end{tabular} \\
\midrule
0k-1k & {\cellcolor[HTML]{088400}} \color[HTML]{F1F1F1} 0.380 & {\cellcolor[HTML]{349A00}} \color[HTML]{F1F1F1} 0.358 & {\cellcolor[HTML]{349A00}} \color[HTML]{F1F1F1} 0.358 & {\cellcolor[HTML]{349A00}} \color[HTML]{F1F1F1} 0.358 & {\cellcolor[HTML]{349A00}} \color[HTML]{F1F1F1} 0.358 & {\cellcolor[HTML]{349A00}} \color[HTML]{F1F1F1} 0.358 & {\cellcolor[HTML]{349A00}} \color[HTML]{F1F1F1} 0.358 & {\cellcolor[HTML]{349A00}} \color[HTML]{F1F1F1} 0.358 \\
1k-2k & {\cellcolor[HTML]{008000}} \color[HTML]{F1F1F1} 0.385 & {\cellcolor[HTML]{9ECF00}} \color[HTML]{000000} 0.303 & {\cellcolor[HTML]{98CC00}} \color[HTML]{000000} 0.306 & {\cellcolor[HTML]{A2D100}} \color[HTML]{000000} 0.301 & {\cellcolor[HTML]{FFD600}} \color[HTML]{000000} 0.232 & {\cellcolor[HTML]{FFDA00}} \color[HTML]{000000} 0.234 & {\cellcolor[HTML]{FFA600}} \color[HTML]{000000} 0.207 & {\cellcolor[HTML]{A6D300}} \color[HTML]{000000} 0.299 \\
2k-4k & {\cellcolor[HTML]{40A000}} \color[HTML]{F1F1F1} 0.352 & {\cellcolor[HTML]{CCE600}} \color[HTML]{000000} 0.279 & {\cellcolor[HTML]{D0E800}} \color[HTML]{000000} 0.277 & {\cellcolor[HTML]{CCE600}} \color[HTML]{000000} 0.279 & {\cellcolor[HTML]{FF7E00}} \color[HTML]{F1F1F1} 0.186 & {\cellcolor[HTML]{FFB000}} \color[HTML]{000000} 0.212 & {\cellcolor[HTML]{FF7200}} \color[HTML]{F1F1F1} 0.180 & {\cellcolor[HTML]{C0E000}} \color[HTML]{000000} 0.285 \\
4k-6k & {\cellcolor[HTML]{4EA700}} \color[HTML]{F1F1F1} 0.344 & {\cellcolor[HTML]{9ECF00}} \color[HTML]{000000} 0.303 & {\cellcolor[HTML]{AED700}} \color[HTML]{000000} 0.295 & {\cellcolor[HTML]{D8EC00}} \color[HTML]{000000} 0.273 & {\cellcolor[HTML]{FF5400}} \color[HTML]{F1F1F1} 0.165 & {\cellcolor[HTML]{FF9400}} \color[HTML]{000000} 0.198 & {\cellcolor[HTML]{FF4A00}} \color[HTML]{F1F1F1} 0.160 & {\cellcolor[HTML]{CEE700}} \color[HTML]{000000} 0.278 \\
6k-8k & {\cellcolor[HTML]{ACD600}} \color[HTML]{000000} 0.296 & {\cellcolor[HTML]{FEFF00}} \color[HTML]{000000} 0.254 & {\cellcolor[HTML]{FFF200}} \color[HTML]{000000} 0.246 & {\cellcolor[HTML]{FFEC00}} \color[HTML]{000000} 0.243 & {\cellcolor[HTML]{FF5400}} \color[HTML]{F1F1F1} 0.165 & {\cellcolor[HTML]{FF8000}} \color[HTML]{F1F1F1} 0.188 & {\cellcolor[HTML]{FF2C00}} \color[HTML]{F1F1F1} 0.144 & {\cellcolor[HTML]{FFEE00}} \color[HTML]{000000} 0.244 \\
8k-10k & {\cellcolor[HTML]{98CC00}} \color[HTML]{000000} 0.306 & {\cellcolor[HTML]{FFDC00}} \color[HTML]{000000} 0.235 & {\cellcolor[HTML]{FFE200}} \color[HTML]{000000} 0.238 & {\cellcolor[HTML]{FFBE00}} \color[HTML]{000000} 0.219 & {\cellcolor[HTML]{FF4000}} \color[HTML]{F1F1F1} 0.155 & {\cellcolor[HTML]{FF8400}} \color[HTML]{F1F1F1} 0.190 & {\cellcolor[HTML]{FF1600}} \color[HTML]{F1F1F1} 0.133 & {\cellcolor[HTML]{FFD800}} \color[HTML]{000000} 0.233 \\
10k-12k & {\cellcolor[HTML]{E0F000}} \color[HTML]{000000} 0.269 & {\cellcolor[HTML]{FFE800}} \color[HTML]{000000} 0.241 & {\cellcolor[HTML]{FFE000}} \color[HTML]{000000} 0.237 & {\cellcolor[HTML]{FFCA00}} \color[HTML]{000000} 0.226 & {\cellcolor[HTML]{FF3C00}} \color[HTML]{F1F1F1} 0.152 & {\cellcolor[HTML]{FF8000}} \color[HTML]{F1F1F1} 0.188 & {\cellcolor[HTML]{FF0E00}} \color[HTML]{F1F1F1} 0.129 & {\cellcolor[HTML]{FFE800}} \color[HTML]{000000} 0.241 \\
12k-14k & {\cellcolor[HTML]{CAE500}} \color[HTML]{000000} 0.280 & {\cellcolor[HTML]{FFE000}} \color[HTML]{000000} 0.237 & {\cellcolor[HTML]{FFD800}} \color[HTML]{000000} 0.233 & {\cellcolor[HTML]{FFC400}} \color[HTML]{000000} 0.223 & {\cellcolor[HTML]{FF4200}} \color[HTML]{F1F1F1} 0.156 & {\cellcolor[HTML]{FF6A00}} \color[HTML]{F1F1F1} 0.176 & {\cellcolor[HTML]{FF3000}} \color[HTML]{F1F1F1} 0.146 & {\cellcolor[HTML]{FFD200}} \color[HTML]{000000} 0.230 \\
14k-16k & {\cellcolor[HTML]{FFD400}} \color[HTML]{000000} 0.231 & {\cellcolor[HTML]{FFE600}} \color[HTML]{000000} 0.240 & {\cellcolor[HTML]{FFDA00}} \color[HTML]{000000} 0.234 & {\cellcolor[HTML]{FFC200}} \color[HTML]{000000} 0.222 & {\cellcolor[HTML]{FF4000}} \color[HTML]{F1F1F1} 0.154 & {\cellcolor[HTML]{FF5C00}} \color[HTML]{F1F1F1} 0.169 & {\cellcolor[HTML]{FF2400}} \color[HTML]{F1F1F1} 0.140 & {\cellcolor[HTML]{FFE000}} \color[HTML]{000000} 0.237 \\
16k-24k & {\cellcolor[HTML]{E0F000}} \color[HTML]{000000} 0.269 & {\cellcolor[HTML]{FFFE00}} \color[HTML]{000000} 0.252 & {\cellcolor[HTML]{FFF600}} \color[HTML]{000000} 0.248 & {\cellcolor[HTML]{FFC000}} \color[HTML]{000000} 0.220 & {\cellcolor[HTML]{FF4000}} \color[HTML]{F1F1F1} 0.155 & {\cellcolor[HTML]{FF7600}} \color[HTML]{F1F1F1} 0.182 & {\cellcolor[HTML]{FF0000}} \color[HTML]{F1F1F1} 0.121 & {\cellcolor[HTML]{FFF400}} \color[HTML]{000000} 0.247 \\
24k-32k & {\cellcolor[HTML]{FFF400}} \color[HTML]{000000} 0.247 & {\cellcolor[HTML]{FFA600}} \color[HTML]{000000} 0.207 & {\cellcolor[HTML]{FFB000}} \color[HTML]{000000} 0.212 & {\cellcolor[HTML]{FFA400}} \color[HTML]{000000} 0.206 & {\cellcolor[HTML]{FF2400}} \color[HTML]{F1F1F1} 0.140 & {\cellcolor[HTML]{FF8000}} \color[HTML]{F1F1F1} 0.187 & {\cellcolor[HTML]{FF0E00}} \color[HTML]{F1F1F1} 0.129 & {\cellcolor[HTML]{FFBA00}} \color[HTML]{000000} 0.217 \\
32k-49k & {\cellcolor[HTML]{FF6000}} \color[HTML]{F1F1F1} 0.171 & {\cellcolor[HTML]{FF9800}} \color[HTML]{000000} 0.200 & {\cellcolor[HTML]{FFB200}} \color[HTML]{000000} 0.213 & {\cellcolor[HTML]{FFC000}} \color[HTML]{000000} 0.220 & {\cellcolor[HTML]{FF3C00}} \color[HTML]{F1F1F1} 0.152 & {\cellcolor[HTML]{FF4A00}} \color[HTML]{F1F1F1} 0.160 & {\cellcolor[HTML]{FF3200}} \color[HTML]{F1F1F1} 0.147 & {\cellcolor[HTML]{FFD000}} \color[HTML]{000000} 0.229 \\
49k-65k & {\cellcolor[HTML]{FF8200}} \color[HTML]{F1F1F1} 0.189 & {\cellcolor[HTML]{EAF500}} \color[HTML]{000000} 0.264 & {\cellcolor[HTML]{70B800}} \color[HTML]{F1F1F1} 0.327 & {\cellcolor[HTML]{C0E000}} \color[HTML]{000000} 0.286 & {\cellcolor[HTML]{FFD800}} \color[HTML]{000000} 0.233 & {\cellcolor[HTML]{FFA200}} \color[HTML]{000000} 0.205 & {\cellcolor[HTML]{FFA600}} \color[HTML]{000000} 0.207 & {\cellcolor[HTML]{68B400}} \color[HTML]{F1F1F1} 0.331 \\
\hline
In-Domain (0k-8k) & {\cellcolor[HTML]{40A000}} \color[HTML]{F1F1F1} 0.351 & {\cellcolor[HTML]{A6D300}} \color[HTML]{000000} 0.299 & {\cellcolor[HTML]{ACD600}} \color[HTML]{000000} 0.296 & {\cellcolor[HTML]{B6DB00}} \color[HTML]{000000} 0.291 & {\cellcolor[HTML]{FFC000}} \color[HTML]{000000} 0.220 & {\cellcolor[HTML]{FFE000}} \color[HTML]{000000} 0.237 & {\cellcolor[HTML]{FFAA00}} \color[HTML]{000000} 0.209 & {\cellcolor[HTML]{B2D900}} \color[HTML]{000000} 0.293 \\
OOD (8k-65k) & {\cellcolor[HTML]{E6F300}} \color[HTML]{000000} 0.266 & {\cellcolor[HTML]{FFE200}} \color[HTML]{000000} 0.238 & {\cellcolor[HTML]{FFE000}} \color[HTML]{000000} 0.237 & {\cellcolor[HTML]{FFC200}} \color[HTML]{000000} 0.222 & {\cellcolor[HTML]{FF4000}} \color[HTML]{F1F1F1} 0.154 & {\cellcolor[HTML]{FF7200}} \color[HTML]{F1F1F1} 0.181 & {\cellcolor[HTML]{FF1A00}} \color[HTML]{F1F1F1} 0.135 & {\cellcolor[HTML]{FFE000}} \color[HTML]{000000} 0.237 \\
Long-OOD (32k-65k) & {\cellcolor[HTML]{FF6800}} \color[HTML]{F1F1F1} 0.175 & {\cellcolor[HTML]{FFB400}} \color[HTML]{000000} 0.215 & {\cellcolor[HTML]{FFE400}} \color[HTML]{000000} 0.239 & {\cellcolor[HTML]{FFDC00}} \color[HTML]{000000} 0.235 & {\cellcolor[HTML]{FF6000}} \color[HTML]{F1F1F1} 0.171 & {\cellcolor[HTML]{FF5E00}} \color[HTML]{F1F1F1} 0.170 & {\cellcolor[HTML]{FF4C00}} \color[HTML]{F1F1F1} 0.161 & {\cellcolor[HTML]{FFFE00}} \color[HTML]{000000} 0.253 \\
Full (0k-65k) & {\cellcolor[HTML]{98CC00}} \color[HTML]{000000} 0.306 & {\cellcolor[HTML]{E4F200}} \color[HTML]{000000} 0.266 & {\cellcolor[HTML]{E8F400}} \color[HTML]{000000} 0.264 & {\cellcolor[HTML]{FEFF00}} \color[HTML]{000000} 0.254 & {\cellcolor[HTML]{FF7C00}} \color[HTML]{F1F1F1} 0.185 & {\cellcolor[HTML]{FFA600}} \color[HTML]{000000} 0.207 & {\cellcolor[HTML]{FF5E00}} \color[HTML]{F1F1F1} 0.170 & {\cellcolor[HTML]{ECF600}} \color[HTML]{000000} 0.263 \\
\bottomrule
\end{tabular}}
\caption{Associative layers ablation on the GR-100+ dataset for Gemma-3-1B-IT model, metric - ROUGE-L. Middle and upper layers representations are the most important for associative memory; ARMT with only top-4 associative blocks keeps almost the same performance as the full ARMT.}
\label{tab:gov_report_best_assoc_ablation}
\end{center}
\end{table*}
\begin{table*}[t]

\begin{center}
\resizebox{0.7\linewidth}{!}{
\begin{tabular}{lccccc}
\toprule
\begin{tabular}[c]{@{}l@{}}\textbf{Model\//}\\ \textbf{Lengths}\end{tabular} & \begin{tabular}[c]{@{}l@{}}\textbf{Base,}\\ \textbf{GR-100+,}\\ \textbf{8k}\end{tabular} & \begin{tabular}[c]{@{}l@{}}\textbf{ARMT,}\\ \textbf{GR-100+,}\\ \textbf{8k}\end{tabular} &  \begin{tabular}[c]{@{}l@{}}\textbf{ARMT,}\\ \textbf{ GR-100+, 8k,}\\ \textbf{only top-4 layers}\end{tabular} &  \begin{tabular}[c]{@{}l@{}}\textbf{ARMT,}\\ \textbf{ GR-100+, 8k,}\\ \textbf{only top-4 layers,} \\\textbf{trained}\end{tabular} &  \begin{tabular}[c]{@{}l@{}}\textbf{ARMT,}\\ \textbf{ GR-100+, 8k,}\\ \textbf{pre-selected layers,} \\\textbf{trained}\end{tabular} \\
\midrule
0k-1k & {\cellcolor[HTML]{0A8500}} \color[HTML]{F1F1F1} 0.380 & {\cellcolor[HTML]{40A000}} \color[HTML]{F1F1F1} 0.358 & {\cellcolor[HTML]{40A000}} \color[HTML]{F1F1F1} 0.358 & {\cellcolor[HTML]{5AAD00}} \color[HTML]{F1F1F1} 0.347 & {\cellcolor[HTML]{349A00}} \color[HTML]{F1F1F1} 0.363 \\
1k-2k & {\cellcolor[HTML]{008000}} \color[HTML]{F1F1F1} 0.385 & {\cellcolor[HTML]{C4E200}} \color[HTML]{000000} 0.303 & {\cellcolor[HTML]{CCE600}} \color[HTML]{000000} 0.299 & {\cellcolor[HTML]{E0F000}} \color[HTML]{000000} 0.291 & {\cellcolor[HTML]{A2D100}} \color[HTML]{000000} 0.317 \\
2k-4k & {\cellcolor[HTML]{4EA700}} \color[HTML]{F1F1F1} 0.352 & {\cellcolor[HTML]{FCFE00}} \color[HTML]{000000} 0.279 & {\cellcolor[HTML]{EEF700}} \color[HTML]{000000} 0.285 & {\cellcolor[HTML]{FFE000}} \color[HTML]{000000} 0.265 & {\cellcolor[HTML]{C0E000}} \color[HTML]{000000} 0.304 \\
4k-6k & {\cellcolor[HTML]{62B100}} \color[HTML]{F1F1F1} 0.344 & {\cellcolor[HTML]{C4E200}} \color[HTML]{000000} 0.303 & {\cellcolor[HTML]{FEFF00}} \color[HTML]{000000} 0.278 & {\cellcolor[HTML]{FFD800}} \color[HTML]{000000} 0.262 & {\cellcolor[HTML]{EAF500}} \color[HTML]{000000} 0.287 \\
6k-8k & {\cellcolor[HTML]{D4EA00}} \color[HTML]{000000} 0.296 & {\cellcolor[HTML]{FFC600}} \color[HTML]{000000} 0.254 & {\cellcolor[HTML]{FFAE00}} \color[HTML]{000000} 0.244 & {\cellcolor[HTML]{FFA600}} \color[HTML]{000000} 0.241 & {\cellcolor[HTML]{FFBC00}} \color[HTML]{000000} 0.250 \\
8k-10k & {\cellcolor[HTML]{BCDE00}} \color[HTML]{000000} 0.306 & {\cellcolor[HTML]{FF9800}} \color[HTML]{000000} 0.235 & {\cellcolor[HTML]{FF9400}} \color[HTML]{000000} 0.233 & {\cellcolor[HTML]{FF9000}} \color[HTML]{000000} 0.232 & {\cellcolor[HTML]{FF8E00}} \color[HTML]{F1F1F1} 0.231 \\
10k-12k & {\cellcolor[HTML]{FFEA00}} \color[HTML]{000000} 0.269 & {\cellcolor[HTML]{FFA600}} \color[HTML]{000000} 0.241 & {\cellcolor[HTML]{FFA600}} \color[HTML]{000000} 0.241 & {\cellcolor[HTML]{FF9000}} \color[HTML]{000000} 0.232 & {\cellcolor[HTML]{FF8E00}} \color[HTML]{F1F1F1} 0.231 \\
12k-14k & {\cellcolor[HTML]{FAFD00}} \color[HTML]{000000} 0.280 & {\cellcolor[HTML]{FF9C00}} \color[HTML]{000000} 0.237 & {\cellcolor[HTML]{FF8C00}} \color[HTML]{F1F1F1} 0.230 & {\cellcolor[HTML]{FF7A00}} \color[HTML]{F1F1F1} 0.222 & {\cellcolor[HTML]{FF7600}} \color[HTML]{F1F1F1} 0.221 \\
14k-16k & {\cellcolor[HTML]{FF8E00}} \color[HTML]{F1F1F1} 0.231 & {\cellcolor[HTML]{FFA400}} \color[HTML]{000000} 0.240 & {\cellcolor[HTML]{FF9C00}} \color[HTML]{000000} 0.237 & {\cellcolor[HTML]{FF9400}} \color[HTML]{000000} 0.233 & {\cellcolor[HTML]{FFAC00}} \color[HTML]{000000} 0.243 \\
16k-24k & {\cellcolor[HTML]{FFEA00}} \color[HTML]{000000} 0.269 & {\cellcolor[HTML]{FFC000}} \color[HTML]{000000} 0.252 & {\cellcolor[HTML]{FFB400}} \color[HTML]{000000} 0.247 & {\cellcolor[HTML]{FFA800}} \color[HTML]{000000} 0.242 & {\cellcolor[HTML]{FFD200}} \color[HTML]{000000} 0.259 \\
24k-32k & {\cellcolor[HTML]{FFB400}} \color[HTML]{000000} 0.247 & {\cellcolor[HTML]{FF5600}} \color[HTML]{F1F1F1} 0.207 & {\cellcolor[HTML]{FF6E00}} \color[HTML]{F1F1F1} 0.217 & {\cellcolor[HTML]{FF2600}} \color[HTML]{F1F1F1} 0.187 & {\cellcolor[HTML]{FF4E00}} \color[HTML]{F1F1F1} 0.204 \\
32k-49k & {\cellcolor[HTML]{FF0000}} \color[HTML]{F1F1F1} 0.171 & {\cellcolor[HTML]{FF4400}} \color[HTML]{F1F1F1} 0.200 & {\cellcolor[HTML]{FF8A00}} \color[HTML]{F1F1F1} 0.229 & {\cellcolor[HTML]{FF5200}} \color[HTML]{F1F1F1} 0.206 & {\cellcolor[HTML]{FF5000}} \color[HTML]{F1F1F1} 0.205 \\
49k-65k & {\cellcolor[HTML]{FF2A00}} \color[HTML]{F1F1F1} 0.189 & {\cellcolor[HTML]{FFDE00}} \color[HTML]{000000} 0.264 & {\cellcolor[HTML]{80C000}} \color[HTML]{000000} 0.331 & {\cellcolor[HTML]{70B800}} \color[HTML]{F1F1F1} 0.338 & {\cellcolor[HTML]{FFB200}} \color[HTML]{000000} 0.246 \\
\hline
\begin{tabular}[c]{@{}l@{}}{In-Domain}\\ {(0k-8k)}\end{tabular} & {\cellcolor[HTML]{50A800}} \color[HTML]{F1F1F1} 0.351 & {\cellcolor[HTML]{CCE600}} \color[HTML]{000000} 0.299 & {\cellcolor[HTML]{DCEE00}} \color[HTML]{000000} 0.293 & {\cellcolor[HTML]{F8FC00}} \color[HTML]{000000} 0.281 & {\cellcolor[HTML]{C0E000}} \color[HTML]{000000} 0.304 \\
\begin{tabular}[c]{@{}l@{}}{OOD}\\ {(8k-65k)}\end{tabular} & {\cellcolor[HTML]{FFE200}} \color[HTML]{000000} 0.266 & {\cellcolor[HTML]{FF9E00}} \color[HTML]{000000} 0.238 & {\cellcolor[HTML]{FF9C00}} \color[HTML]{000000} 0.237 & {\cellcolor[HTML]{FF8A00}} \color[HTML]{F1F1F1} 0.229 & {\cellcolor[HTML]{FF9600}} \color[HTML]{000000} 0.234 \\
\begin{tabular}[c]{@{}l@{}}{Long-OOD}\\ {(32k-65k)}\end{tabular} & {\cellcolor[HTML]{FF0800}} \color[HTML]{F1F1F1} 0.175 & {\cellcolor[HTML]{FF6800}} \color[HTML]{F1F1F1} 0.215 & {\cellcolor[HTML]{FFC200}} \color[HTML]{000000} 0.253 & {\cellcolor[HTML]{FF9C00}} \color[HTML]{000000} 0.236 & {\cellcolor[HTML]{FF6600}} \color[HTML]{F1F1F1} 0.214 \\
\begin{tabular}[c]{@{}l@{}}{Full}\\ {(0k-65k)}\end{tabular} & {\cellcolor[HTML]{BCDE00}} \color[HTML]{000000} 0.306 & {\cellcolor[HTML]{FFE400}} \color[HTML]{000000} 0.266 & {\cellcolor[HTML]{FFDA00}} \color[HTML]{000000} 0.263 & {\cellcolor[HTML]{FFC400}} \color[HTML]{000000} 0.254 & {\cellcolor[HTML]{FFE400}} \color[HTML]{000000} 0.267 \\
\bottomrule

\end{tabular}}
\caption{Associative layers ablation on the GR-100+ for Gemma-3-1B-IT model, metric - ROUGE-L. ARMT with only top-4 associative blocks keeps almost the same performance as the full ARMT even without training. Model with 5 associative blocks (approximately 20\%) achieves the same performance as the full ARMT.}
\label{tab:gov_report_best_assoc_ablation_with_train}
\end{center}
\end{table*}

We conducted a layer-wise ablation for associative layers in a trained ARMT model for the MT and GR-100+ datasets. The results are presented in~\Cref{tab:gov_report_best_all_layers_assoc_ablation_0_12,tab:gov_report_best_all_layers_assoc_ablation_13_26,tab:mt_best_all_layers_assoc_ablation_0_12,tab:mt_best_all_layers_assoc_ablation_13_26}. For the GR-100+ dataset, the most important representations are from the 14th, 25th, 9th, and 13th layers, while other layers contribute substantially less. For the MT dataset, the most important representations are from the 14th, 18th, 25th, and 19th layers. Layers are numbered from the starting embeddings layer (i. e., the 0th layer goes after the embeddings layer, while the 25th layer goes right before the last layer). We suppose that the middle layers are the most important for the associative memory, as these layers contain the more abstract representation than the lower or higher layers. On the lower layers, these representations are still not formed, while on the higher ones, they are close to the target tokens. 

\begin{table*}[h]

\begin{center}
\resizebox{1.0\textwidth}{!}{
\begin{tabular}{lccccccccccccccc}
\toprule
\begin{tabular}[c]{@{}l@{}}\textbf{Model\//}\\ \textbf{Lengths}\end{tabular} & \begin{tabular}[c]{@{}l@{}}\textbf{Base,}\\ \textbf{GR-100+,}\\ \textbf{8k}\end{tabular} & \begin{tabular}[c]{@{}l@{}}\textbf{ARMT,}\\ \textbf{GR-100+,}\\ \textbf{8k}\end{tabular} & \begin{tabular}[c]{@{}l@{}}\textbf{W/o}\\ \textbf{layer}\\ \textbf{0}\end{tabular} & \begin{tabular}[c]{@{}l@{}}\textbf{W/o}\\ \textbf{layer}\\ \textbf{1}\end{tabular} & \begin{tabular}[c]{@{}l@{}}\textbf{W/o}\\ \textbf{layer}\\ \textbf{2}\end{tabular} & \begin{tabular}[c]{@{}l@{}}\textbf{W/o}\\ \textbf{layer}\\ \textbf{3}\end{tabular} & \begin{tabular}[c]{@{}l@{}}\textbf{W/o}\\ \textbf{layer}\\ \textbf{4}\end{tabular} & \begin{tabular}[c]{@{}l@{}}\textbf{W/o}\\ \textbf{layer}\\ \textbf{5}\end{tabular} & \begin{tabular}[c]{@{}l@{}}\textbf{W/o}\\ \textbf{layer}\\ \textbf{6}\end{tabular} & \begin{tabular}[c]{@{}l@{}}\textbf{W/o}\\ \textbf{layer}\\ \textbf{7}\end{tabular} & \begin{tabular}[c]{@{}l@{}}\textbf{W/o}\\ \textbf{layer}\\ \textbf{8}\end{tabular}  & \begin{tabular}[c]{@{}l@{}}\textbf{W/o}\\ \textbf{layer}\\ \textbf{9}\end{tabular} & \begin{tabular}[c]{@{}l@{}}\textbf{W/o}\\ \textbf{layer}\\ \textbf{10}\end{tabular} & \begin{tabular}[c]{@{}l@{}}\textbf{W/o}\\ \textbf{layer}\\ \textbf{11}\end{tabular} & \begin{tabular}[c]{@{}l@{}}\textbf{W/o}\\ \textbf{layer}\\ \textbf{12}\end{tabular} \\
\midrule
0k-1k & {\cellcolor[HTML]{0A8500}} \color[HTML]{F1F1F1} 0.380 & {\cellcolor[HTML]{40A000}} \color[HTML]{F1F1F1} 0.358 & {\cellcolor[HTML]{40A000}} \color[HTML]{F1F1F1} 0.358 & {\cellcolor[HTML]{40A000}} \color[HTML]{F1F1F1} 0.358 & {\cellcolor[HTML]{40A000}} \color[HTML]{F1F1F1} 0.358 & {\cellcolor[HTML]{40A000}} \color[HTML]{F1F1F1} 0.358 & {\cellcolor[HTML]{40A000}} \color[HTML]{F1F1F1} 0.358 & {\cellcolor[HTML]{40A000}} \color[HTML]{F1F1F1} 0.358 & {\cellcolor[HTML]{40A000}} \color[HTML]{F1F1F1} 0.358 & {\cellcolor[HTML]{40A000}} \color[HTML]{F1F1F1} 0.358 & {\cellcolor[HTML]{40A000}} \color[HTML]{F1F1F1} 0.358 & {\cellcolor[HTML]{40A000}} \color[HTML]{F1F1F1} 0.358 & {\cellcolor[HTML]{40A000}} \color[HTML]{F1F1F1} 0.358 & {\cellcolor[HTML]{40A000}} \color[HTML]{F1F1F1} 0.358 & {\cellcolor[HTML]{40A000}} \color[HTML]{F1F1F1} 0.358 \\
1k-2k & {\cellcolor[HTML]{008000}} \color[HTML]{F1F1F1} 0.385 & {\cellcolor[HTML]{C2E100}} \color[HTML]{000000} 0.303 & {\cellcolor[HTML]{B8DC00}} \color[HTML]{000000} 0.307 & {\cellcolor[HTML]{CAE500}} \color[HTML]{000000} 0.300 & {\cellcolor[HTML]{D6EB00}} \color[HTML]{000000} 0.295 & {\cellcolor[HTML]{C4E200}} \color[HTML]{000000} 0.302 & {\cellcolor[HTML]{D2E900}} \color[HTML]{000000} 0.296 & {\cellcolor[HTML]{CCE600}} \color[HTML]{000000} 0.299 & {\cellcolor[HTML]{C2E100}} \color[HTML]{000000} 0.303 & {\cellcolor[HTML]{D8EC00}} \color[HTML]{000000} 0.294 & {\cellcolor[HTML]{C2E100}} \color[HTML]{000000} 0.303 & {\cellcolor[HTML]{E2F100}} \color[HTML]{000000} 0.290 & {\cellcolor[HTML]{C4E200}} \color[HTML]{000000} 0.302 & {\cellcolor[HTML]{C2E100}} \color[HTML]{000000} 0.303 & {\cellcolor[HTML]{C4E200}} \color[HTML]{000000} 0.302 \\
2k-4k & {\cellcolor[HTML]{4EA700}} \color[HTML]{F1F1F1} 0.352 & {\cellcolor[HTML]{FCFE00}} \color[HTML]{000000} 0.279 & {\cellcolor[HTML]{FEFF00}} \color[HTML]{000000} 0.278 & {\cellcolor[HTML]{FFF600}} \color[HTML]{000000} 0.274 & {\cellcolor[HTML]{FCFE00}} \color[HTML]{000000} 0.279 & {\cellcolor[HTML]{FAFD00}} \color[HTML]{000000} 0.280 & {\cellcolor[HTML]{F2F900}} \color[HTML]{000000} 0.283 & {\cellcolor[HTML]{F2F900}} \color[HTML]{000000} 0.283 & {\cellcolor[HTML]{FCFE00}} \color[HTML]{000000} 0.279 & {\cellcolor[HTML]{FEFF00}} \color[HTML]{000000} 0.278 & {\cellcolor[HTML]{FFFE00}} \color[HTML]{000000} 0.277 & {\cellcolor[HTML]{FFEA00}} \color[HTML]{000000} 0.269 & {\cellcolor[HTML]{F4FA00}} \color[HTML]{000000} 0.282 & {\cellcolor[HTML]{FAFD00}} \color[HTML]{000000} 0.280 & {\cellcolor[HTML]{D2E900}} \color[HTML]{000000} 0.296 \\
4k-6k & {\cellcolor[HTML]{60B000}} \color[HTML]{F1F1F1} 0.344 & {\cellcolor[HTML]{C2E100}} \color[HTML]{000000} 0.303 & {\cellcolor[HTML]{CEE700}} \color[HTML]{000000} 0.298 & {\cellcolor[HTML]{CCE600}} \color[HTML]{000000} 0.299 & {\cellcolor[HTML]{CEE700}} \color[HTML]{000000} 0.298 & {\cellcolor[HTML]{D0E800}} \color[HTML]{000000} 0.297 & {\cellcolor[HTML]{D0E800}} \color[HTML]{000000} 0.297 & {\cellcolor[HTML]{CEE700}} \color[HTML]{000000} 0.298 & {\cellcolor[HTML]{C8E400}} \color[HTML]{000000} 0.301 & {\cellcolor[HTML]{DCEE00}} \color[HTML]{000000} 0.292 & {\cellcolor[HTML]{C4E200}} \color[HTML]{000000} 0.302 & {\cellcolor[HTML]{FAFD00}} \color[HTML]{000000} 0.280 & {\cellcolor[HTML]{CCE600}} \color[HTML]{000000} 0.299 & {\cellcolor[HTML]{C8E400}} \color[HTML]{000000} 0.301 & {\cellcolor[HTML]{EEF700}} \color[HTML]{000000} 0.285 \\
6k-8k & {\cellcolor[HTML]{D2E900}} \color[HTML]{000000} 0.296 & {\cellcolor[HTML]{FFC800}} \color[HTML]{000000} 0.254 & {\cellcolor[HTML]{FFB600}} \color[HTML]{000000} 0.247 & {\cellcolor[HTML]{FFC400}} \color[HTML]{000000} 0.253 & {\cellcolor[HTML]{FFCA00}} \color[HTML]{000000} 0.255 & {\cellcolor[HTML]{FFB400}} \color[HTML]{000000} 0.246 & {\cellcolor[HTML]{FFCA00}} \color[HTML]{000000} 0.255 & {\cellcolor[HTML]{FFCC00}} \color[HTML]{000000} 0.256 & {\cellcolor[HTML]{FFC800}} \color[HTML]{000000} 0.254 & {\cellcolor[HTML]{FFB400}} \color[HTML]{000000} 0.246 & {\cellcolor[HTML]{FFBC00}} \color[HTML]{000000} 0.249 & {\cellcolor[HTML]{FFA600}} \color[HTML]{000000} 0.240 & {\cellcolor[HTML]{FFC200}} \color[HTML]{000000} 0.252 & {\cellcolor[HTML]{FFC800}} \color[HTML]{000000} 0.254 & {\cellcolor[HTML]{FFC800}} \color[HTML]{000000} 0.254 \\
8k-10k & {\cellcolor[HTML]{BCDE00}} \color[HTML]{000000} 0.306 & {\cellcolor[HTML]{FF9A00}} \color[HTML]{000000} 0.235 & {\cellcolor[HTML]{FF8A00}} \color[HTML]{F1F1F1} 0.228 & {\cellcolor[HTML]{FF9200}} \color[HTML]{000000} 0.232 & {\cellcolor[HTML]{FF9800}} \color[HTML]{000000} 0.234 & {\cellcolor[HTML]{FFA400}} \color[HTML]{000000} 0.239 & {\cellcolor[HTML]{FFAC00}} \color[HTML]{000000} 0.243 & {\cellcolor[HTML]{FF9600}} \color[HTML]{000000} 0.233 & {\cellcolor[HTML]{FF9E00}} \color[HTML]{000000} 0.237 & {\cellcolor[HTML]{FFAC00}} \color[HTML]{000000} 0.243 & {\cellcolor[HTML]{FF9E00}} \color[HTML]{000000} 0.237 & {\cellcolor[HTML]{FF6000}} \color[HTML]{F1F1F1} 0.211 & {\cellcolor[HTML]{FF9C00}} \color[HTML]{000000} 0.236 & {\cellcolor[HTML]{FF9800}} \color[HTML]{000000} 0.234 & {\cellcolor[HTML]{FF9000}} \color[HTML]{000000} 0.231 \\
10k-12k & {\cellcolor[HTML]{FFEA00}} \color[HTML]{000000} 0.269 & {\cellcolor[HTML]{FFA800}} \color[HTML]{000000} 0.241 & {\cellcolor[HTML]{FF9000}} \color[HTML]{000000} 0.231 & {\cellcolor[HTML]{FF9C00}} \color[HTML]{000000} 0.236 & {\cellcolor[HTML]{FFA800}} \color[HTML]{000000} 0.241 & {\cellcolor[HTML]{FFB000}} \color[HTML]{000000} 0.244 & {\cellcolor[HTML]{FFA600}} \color[HTML]{000000} 0.240 & {\cellcolor[HTML]{FFA400}} \color[HTML]{000000} 0.239 & {\cellcolor[HTML]{FFB000}} \color[HTML]{000000} 0.244 & {\cellcolor[HTML]{FFA600}} \color[HTML]{000000} 0.240 & {\cellcolor[HTML]{FFA800}} \color[HTML]{000000} 0.241 & {\cellcolor[HTML]{FF8600}} \color[HTML]{F1F1F1} 0.227 & {\cellcolor[HTML]{FFA400}} \color[HTML]{000000} 0.239 & {\cellcolor[HTML]{FFAA00}} \color[HTML]{000000} 0.242 & {\cellcolor[HTML]{FFAA00}} \color[HTML]{000000} 0.242 \\
12k-14k & {\cellcolor[HTML]{FAFD00}} \color[HTML]{000000} 0.280 & {\cellcolor[HTML]{FF9E00}} \color[HTML]{000000} 0.237 & {\cellcolor[HTML]{FFA000}} \color[HTML]{000000} 0.238 & {\cellcolor[HTML]{FF9E00}} \color[HTML]{000000} 0.237 & {\cellcolor[HTML]{FF9800}} \color[HTML]{000000} 0.234 & {\cellcolor[HTML]{FF8E00}} \color[HTML]{F1F1F1} 0.230 & {\cellcolor[HTML]{FF9600}} \color[HTML]{000000} 0.233 & {\cellcolor[HTML]{FF9200}} \color[HTML]{000000} 0.232 & {\cellcolor[HTML]{FF9A00}} \color[HTML]{000000} 0.235 & {\cellcolor[HTML]{FF9E00}} \color[HTML]{000000} 0.237 & {\cellcolor[HTML]{FF9000}} \color[HTML]{000000} 0.231 & {\cellcolor[HTML]{FF6C00}} \color[HTML]{F1F1F1} 0.216 & {\cellcolor[HTML]{FF9800}} \color[HTML]{000000} 0.234 & {\cellcolor[HTML]{FF9A00}} \color[HTML]{000000} 0.235 & {\cellcolor[HTML]{FF8600}} \color[HTML]{F1F1F1} 0.227 \\
14k-16k & {\cellcolor[HTML]{FF9000}} \color[HTML]{000000} 0.231 & {\cellcolor[HTML]{FFA600}} \color[HTML]{000000} 0.240 & {\cellcolor[HTML]{FFA000}} \color[HTML]{000000} 0.238 & {\cellcolor[HTML]{FFAA00}} \color[HTML]{000000} 0.242 & {\cellcolor[HTML]{FF9C00}} \color[HTML]{000000} 0.236 & {\cellcolor[HTML]{FFA000}} \color[HTML]{000000} 0.238 & {\cellcolor[HTML]{FFA400}} \color[HTML]{000000} 0.239 & {\cellcolor[HTML]{FFA600}} \color[HTML]{000000} 0.240 & {\cellcolor[HTML]{FFA600}} \color[HTML]{000000} 0.240 & {\cellcolor[HTML]{FFA600}} \color[HTML]{000000} 0.240 & {\cellcolor[HTML]{FFAC00}} \color[HTML]{000000} 0.243 & {\cellcolor[HTML]{FF7200}} \color[HTML]{F1F1F1} 0.218 & {\cellcolor[HTML]{FFA600}} \color[HTML]{000000} 0.240 & {\cellcolor[HTML]{FFA600}} \color[HTML]{000000} 0.240 & {\cellcolor[HTML]{FFA400}} \color[HTML]{000000} 0.239 \\
16k-24k & {\cellcolor[HTML]{FFEA00}} \color[HTML]{000000} 0.269 & {\cellcolor[HTML]{FFC200}} \color[HTML]{000000} 0.252 & {\cellcolor[HTML]{FFBE00}} \color[HTML]{000000} 0.250 & {\cellcolor[HTML]{FFB800}} \color[HTML]{000000} 0.248 & {\cellcolor[HTML]{FFC800}} \color[HTML]{000000} 0.254 & {\cellcolor[HTML]{FFCE00}} \color[HTML]{000000} 0.257 & {\cellcolor[HTML]{FFC400}} \color[HTML]{000000} 0.253 & {\cellcolor[HTML]{FFD000}} \color[HTML]{000000} 0.258 & {\cellcolor[HTML]{FFBC00}} \color[HTML]{000000} 0.249 & {\cellcolor[HTML]{FFBC00}} \color[HTML]{000000} 0.249 & {\cellcolor[HTML]{FFCE00}} \color[HTML]{000000} 0.257 & {\cellcolor[HTML]{FF8A00}} \color[HTML]{F1F1F1} 0.228 & {\cellcolor[HTML]{FFC800}} \color[HTML]{000000} 0.254 & {\cellcolor[HTML]{FFC000}} \color[HTML]{000000} 0.251 & {\cellcolor[HTML]{FFB400}} \color[HTML]{000000} 0.246 \\
24k-32k & {\cellcolor[HTML]{FFB600}} \color[HTML]{000000} 0.247 & {\cellcolor[HTML]{FF5800}} \color[HTML]{F1F1F1} 0.207 & {\cellcolor[HTML]{FF6400}} \color[HTML]{F1F1F1} 0.212 & {\cellcolor[HTML]{FF4E00}} \color[HTML]{F1F1F1} 0.203 & {\cellcolor[HTML]{FF5200}} \color[HTML]{F1F1F1} 0.205 & {\cellcolor[HTML]{FF6800}} \color[HTML]{F1F1F1} 0.214 & {\cellcolor[HTML]{FF5000}} \color[HTML]{F1F1F1} 0.204 & {\cellcolor[HTML]{FF5E00}} \color[HTML]{F1F1F1} 0.210 & {\cellcolor[HTML]{FF5000}} \color[HTML]{F1F1F1} 0.204 & {\cellcolor[HTML]{FF5200}} \color[HTML]{F1F1F1} 0.205 & {\cellcolor[HTML]{FF6C00}} \color[HTML]{F1F1F1} 0.216 & {\cellcolor[HTML]{FF2E00}} \color[HTML]{F1F1F1} 0.190 & {\cellcolor[HTML]{FF5400}} \color[HTML]{F1F1F1} 0.206 & {\cellcolor[HTML]{FF5000}} \color[HTML]{F1F1F1} 0.204 & {\cellcolor[HTML]{FF6C00}} \color[HTML]{F1F1F1} 0.216 \\
32k-49k & {\cellcolor[HTML]{FF0200}} \color[HTML]{F1F1F1} 0.171 & {\cellcolor[HTML]{FF4600}} \color[HTML]{F1F1F1} 0.200 & {\cellcolor[HTML]{FF4400}} \color[HTML]{F1F1F1} 0.199 & {\cellcolor[HTML]{FF4E00}} \color[HTML]{F1F1F1} 0.203 & {\cellcolor[HTML]{FF5400}} \color[HTML]{F1F1F1} 0.206 & {\cellcolor[HTML]{FF6800}} \color[HTML]{F1F1F1} 0.214 & {\cellcolor[HTML]{FF5200}} \color[HTML]{F1F1F1} 0.205 & {\cellcolor[HTML]{FF6000}} \color[HTML]{F1F1F1} 0.211 & {\cellcolor[HTML]{FF5000}} \color[HTML]{F1F1F1} 0.204 & {\cellcolor[HTML]{FF6C00}} \color[HTML]{F1F1F1} 0.216 & {\cellcolor[HTML]{FF7E00}} \color[HTML]{F1F1F1} 0.223 & {\cellcolor[HTML]{FF0000}} \color[HTML]{F1F1F1} 0.170 & {\cellcolor[HTML]{FF4200}} \color[HTML]{F1F1F1} 0.198 & {\cellcolor[HTML]{FF5000}} \color[HTML]{F1F1F1} 0.204 & {\cellcolor[HTML]{FF8400}} \color[HTML]{F1F1F1} 0.226 \\
49k-65k & {\cellcolor[HTML]{FF2C00}} \color[HTML]{F1F1F1} 0.189 & {\cellcolor[HTML]{FFDE00}} \color[HTML]{000000} 0.264 & {\cellcolor[HTML]{FFF000}} \color[HTML]{000000} 0.271 & {\cellcolor[HTML]{60B000}} \color[HTML]{F1F1F1} 0.344 & {\cellcolor[HTML]{FEFF00}} \color[HTML]{000000} 0.278 & {\cellcolor[HTML]{F0F800}} \color[HTML]{000000} 0.284 & {\cellcolor[HTML]{FFF200}} \color[HTML]{000000} 0.272 & {\cellcolor[HTML]{B2D900}} \color[HTML]{000000} 0.310 & {\cellcolor[HTML]{FFDE00}} \color[HTML]{000000} 0.264 & {\cellcolor[HTML]{F4FA00}} \color[HTML]{000000} 0.282 & {\cellcolor[HTML]{108800}} \color[HTML]{F1F1F1} 0.378 & {\cellcolor[HTML]{FF0000}} \color[HTML]{F1F1F1} 0.170 & {\cellcolor[HTML]{FFDE00}} \color[HTML]{000000} 0.264 & {\cellcolor[HTML]{FFDE00}} \color[HTML]{000000} 0.264 & {\cellcolor[HTML]{F0F800}} \color[HTML]{000000} 0.284 \\
\hline
In-Domain (0k-8k) & {\cellcolor[HTML]{50A800}} \color[HTML]{F1F1F1} 0.351 & {\cellcolor[HTML]{CCE600}} \color[HTML]{000000} 0.299 & {\cellcolor[HTML]{D0E800}} \color[HTML]{000000} 0.297 & {\cellcolor[HTML]{D2E900}} \color[HTML]{000000} 0.296 & {\cellcolor[HTML]{D2E900}} \color[HTML]{000000} 0.297 & {\cellcolor[HTML]{D2E900}} \color[HTML]{000000} 0.296 & {\cellcolor[HTML]{D0E800}} \color[HTML]{000000} 0.297 & {\cellcolor[HTML]{CEE700}} \color[HTML]{000000} 0.298 & {\cellcolor[HTML]{CCE600}} \color[HTML]{000000} 0.299 & {\cellcolor[HTML]{DAED00}} \color[HTML]{000000} 0.293 & {\cellcolor[HTML]{D0E800}} \color[HTML]{000000} 0.297 & {\cellcolor[HTML]{E8F400}} \color[HTML]{000000} 0.287 & {\cellcolor[HTML]{CEE700}} \color[HTML]{000000} 0.298 & {\cellcolor[HTML]{CCE600}} \color[HTML]{000000} 0.299 & {\cellcolor[HTML]{CCE600}} \color[HTML]{000000} 0.299 \\
OOD (8k-65k) & {\cellcolor[HTML]{FFE200}} \color[HTML]{000000} 0.266 & {\cellcolor[HTML]{FFA000}} \color[HTML]{000000} 0.238 & {\cellcolor[HTML]{FF9A00}} \color[HTML]{000000} 0.235 & {\cellcolor[HTML]{FF9E00}} \color[HTML]{000000} 0.237 & {\cellcolor[HTML]{FF9E00}} \color[HTML]{000000} 0.237 & {\cellcolor[HTML]{FFA400}} \color[HTML]{000000} 0.239 & {\cellcolor[HTML]{FFA200}} \color[HTML]{000000} 0.238 & {\cellcolor[HTML]{FFA200}} \color[HTML]{000000} 0.238 & {\cellcolor[HTML]{FFA000}} \color[HTML]{000000} 0.238 & {\cellcolor[HTML]{FFA400}} \color[HTML]{000000} 0.239 & {\cellcolor[HTML]{FFA800}} \color[HTML]{000000} 0.241 & {\cellcolor[HTML]{FF6C00}} \color[HTML]{F1F1F1} 0.216 & {\cellcolor[HTML]{FFA000}} \color[HTML]{000000} 0.237 & {\cellcolor[HTML]{FF9E00}} \color[HTML]{000000} 0.237 & {\cellcolor[HTML]{FF9C00}} \color[HTML]{000000} 0.236 \\
Long-OOD (32k-65k) & {\cellcolor[HTML]{FF0C00}} \color[HTML]{F1F1F1} 0.175 & {\cellcolor[HTML]{FF6A00}} \color[HTML]{F1F1F1} 0.215 & {\cellcolor[HTML]{FF6C00}} \color[HTML]{F1F1F1} 0.216 & {\cellcolor[HTML]{FF9C00}} \color[HTML]{000000} 0.236 & {\cellcolor[HTML]{FF7C00}} \color[HTML]{F1F1F1} 0.223 & {\cellcolor[HTML]{FF8E00}} \color[HTML]{F1F1F1} 0.230 & {\cellcolor[HTML]{FF7800}} \color[HTML]{F1F1F1} 0.220 & {\cellcolor[HTML]{FF9800}} \color[HTML]{000000} 0.234 & {\cellcolor[HTML]{FF7000}} \color[HTML]{F1F1F1} 0.218 & {\cellcolor[HTML]{FF9000}} \color[HTML]{000000} 0.231 & {\cellcolor[HTML]{FFD200}} \color[HTML]{000000} 0.259 & {\cellcolor[HTML]{FF0000}} \color[HTML]{F1F1F1} 0.170 & {\cellcolor[HTML]{FF6600}} \color[HTML]{F1F1F1} 0.213 & {\cellcolor[HTML]{FF7000}} \color[HTML]{F1F1F1} 0.218 & {\cellcolor[HTML]{FFA400}} \color[HTML]{000000} 0.239 \\
Full (0k-65k) & {\cellcolor[HTML]{BCDE00}} \color[HTML]{000000} 0.306 & {\cellcolor[HTML]{FFE400}} \color[HTML]{000000} 0.266 & {\cellcolor[HTML]{FFDE00}} \color[HTML]{000000} 0.264 & {\cellcolor[HTML]{FFE000}} \color[HTML]{000000} 0.265 & {\cellcolor[HTML]{FFE000}} \color[HTML]{000000} 0.265 & {\cellcolor[HTML]{FFE400}} \color[HTML]{000000} 0.266 & {\cellcolor[HTML]{FFE400}} \color[HTML]{000000} 0.266 & {\cellcolor[HTML]{FFE400}} \color[HTML]{000000} 0.266 & {\cellcolor[HTML]{FFE400}} \color[HTML]{000000} 0.266 & {\cellcolor[HTML]{FFE000}} \color[HTML]{000000} 0.264 & {\cellcolor[HTML]{FFE600}} \color[HTML]{000000} 0.267 & {\cellcolor[HTML]{FFBC00}} \color[HTML]{000000} 0.249 & {\cellcolor[HTML]{FFE400}} \color[HTML]{000000} 0.266 & {\cellcolor[HTML]{FFE400}} \color[HTML]{000000} 0.266 & {\cellcolor[HTML]{FFE200}} \color[HTML]{000000} 0.265 \\
\bottomrule
\end{tabular}}
\caption{Ablation for all associative layers on the GovReport-100+ dataset for ARMT with Gemma-3-1B-IT model, metric - ROUGE-L. Ablated associative layers from 0 to 12.}
\label{tab:gov_report_best_all_layers_assoc_ablation_0_12}
\end{center}
\end{table*}
\begin{table*}[h]

\begin{center}
\resizebox{1.0\textwidth}{!}{
\begin{tabular}{lccccccccccccccc}
\toprule
\begin{tabular}[c]{@{}l@{}}\textbf{Model\//}\\ \textbf{Lengths}\end{tabular} & \begin{tabular}[c]{@{}l@{}}\textbf{Base,}\\ \textbf{GR-100+,}\\ \textbf{8k}\end{tabular} & \begin{tabular}[c]{@{}l@{}}\textbf{ARMT,}\\ \textbf{GR-100+,}\\ \textbf{8k}\end{tabular} & \begin{tabular}[c]{@{}l@{}}\textbf{W/o}\\ \textbf{layer}\\ \textbf{13}\end{tabular} & \begin{tabular}[c]{@{}l@{}}\textbf{W/o}\\ \textbf{layer}\\ \textbf{14}\end{tabular} & \begin{tabular}[c]{@{}l@{}}\textbf{W/o}\\ \textbf{layer}\\ \textbf{15}\end{tabular} & \begin{tabular}[c]{@{}l@{}}\textbf{W/o}\\ \textbf{layer}\\ \textbf{16}\end{tabular} & \begin{tabular}[c]{@{}l@{}}\textbf{W/o}\\ \textbf{layer}\\ \textbf{17}\end{tabular} & \begin{tabular}[c]{@{}l@{}}\textbf{W/o}\\ \textbf{layer}\\ \textbf{18}\end{tabular} & \begin{tabular}[c]{@{}l@{}}\textbf{W/o}\\ \textbf{layer}\\ \textbf{19}\end{tabular} & \begin{tabular}[c]{@{}l@{}}\textbf{W/o}\\ \textbf{layer}\\ \textbf{20}\end{tabular} & \begin{tabular}[c]{@{}l@{}}\textbf{W/o}\\ \textbf{layer}\\ \textbf{21}\end{tabular} & \begin{tabular}[c]{@{}l@{}}\textbf{W/o}\\ \textbf{layer}\\ \textbf{22}\end{tabular} & \begin{tabular}[c]{@{}l@{}}\textbf{W/o}\\ \textbf{layer}\\ \textbf{23}\end{tabular} & \begin{tabular}[c]{@{}l@{}}\textbf{W/o}\\ \textbf{layer}\\ \textbf{24}\end{tabular} & \begin{tabular}[c]{@{}l@{}}\textbf{W/o}\\ \textbf{layer}\\ \textbf{25}\end{tabular}\\
\midrule
0k-1k & {\cellcolor[HTML]{0A8500}} \color[HTML]{F1F1F1} 0.380 & {\cellcolor[HTML]{3A9D00}} \color[HTML]{F1F1F1} 0.358 & {\cellcolor[HTML]{3A9D00}} \color[HTML]{F1F1F1} 0.358 & {\cellcolor[HTML]{3A9D00}} \color[HTML]{F1F1F1} 0.358 & {\cellcolor[HTML]{3A9D00}} \color[HTML]{F1F1F1} 0.358 & {\cellcolor[HTML]{3A9D00}} \color[HTML]{F1F1F1} 0.358 & {\cellcolor[HTML]{3A9D00}} \color[HTML]{F1F1F1} 0.358 & {\cellcolor[HTML]{3A9D00}} \color[HTML]{F1F1F1} 0.358 & {\cellcolor[HTML]{3A9D00}} \color[HTML]{F1F1F1} 0.358 & {\cellcolor[HTML]{3A9D00}} \color[HTML]{F1F1F1} 0.358 & {\cellcolor[HTML]{3A9D00}} \color[HTML]{F1F1F1} 0.358 & {\cellcolor[HTML]{3A9D00}} \color[HTML]{F1F1F1} 0.358 & {\cellcolor[HTML]{3A9D00}} \color[HTML]{F1F1F1} 0.358 & {\cellcolor[HTML]{3A9D00}} \color[HTML]{F1F1F1} 0.358 & {\cellcolor[HTML]{3A9D00}} \color[HTML]{F1F1F1} 0.358 \\
1k-2k & {\cellcolor[HTML]{008000}} \color[HTML]{F1F1F1} 0.385 & {\cellcolor[HTML]{B0D800}} \color[HTML]{000000} 0.303 & {\cellcolor[HTML]{B6DB00}} \color[HTML]{000000} 0.301 & {\cellcolor[HTML]{FFC200}} \color[HTML]{000000} 0.239 & {\cellcolor[HTML]{AED700}} \color[HTML]{000000} 0.304 & {\cellcolor[HTML]{B0D800}} \color[HTML]{000000} 0.303 & {\cellcolor[HTML]{B0D800}} \color[HTML]{000000} 0.303 & {\cellcolor[HTML]{D0E800}} \color[HTML]{000000} 0.289 & {\cellcolor[HTML]{CAE500}} \color[HTML]{000000} 0.291 & {\cellcolor[HTML]{B0D800}} \color[HTML]{000000} 0.303 & {\cellcolor[HTML]{B0D800}} \color[HTML]{000000} 0.303 & {\cellcolor[HTML]{B0D800}} \color[HTML]{000000} 0.303 & {\cellcolor[HTML]{B0D800}} \color[HTML]{000000} 0.303 & {\cellcolor[HTML]{C0E000}} \color[HTML]{000000} 0.296 & {\cellcolor[HTML]{FFA600}} \color[HTML]{000000} 0.226 \\
2k-4k & {\cellcolor[HTML]{46A300}} \color[HTML]{F1F1F1} 0.352 & {\cellcolor[HTML]{E4F200}} \color[HTML]{000000} 0.279 & {\cellcolor[HTML]{EAF500}} \color[HTML]{000000} 0.277 & {\cellcolor[HTML]{FF9600}} \color[HTML]{000000} 0.219 & {\cellcolor[HTML]{E4F200}} \color[HTML]{000000} 0.279 & {\cellcolor[HTML]{E4F200}} \color[HTML]{000000} 0.279 & {\cellcolor[HTML]{E4F200}} \color[HTML]{000000} 0.279 & {\cellcolor[HTML]{F0F800}} \color[HTML]{000000} 0.274 & {\cellcolor[HTML]{E8F400}} \color[HTML]{000000} 0.278 & {\cellcolor[HTML]{E4F200}} \color[HTML]{000000} 0.279 & {\cellcolor[HTML]{E4F200}} \color[HTML]{000000} 0.279 & {\cellcolor[HTML]{E4F200}} \color[HTML]{000000} 0.279 & {\cellcolor[HTML]{E4F200}} \color[HTML]{000000} 0.279 & {\cellcolor[HTML]{E4F200}} \color[HTML]{000000} 0.279 & {\cellcolor[HTML]{FF7E00}} \color[HTML]{F1F1F1} 0.208 \\
4k-6k & {\cellcolor[HTML]{58AC00}} \color[HTML]{F1F1F1} 0.344 & {\cellcolor[HTML]{B0D800}} \color[HTML]{000000} 0.303 & {\cellcolor[HTML]{F8FC00}} \color[HTML]{000000} 0.270 & {\cellcolor[HTML]{FF6800}} \color[HTML]{F1F1F1} 0.197 & {\cellcolor[HTML]{B4DA00}} \color[HTML]{000000} 0.302 & {\cellcolor[HTML]{B0D800}} \color[HTML]{000000} 0.303 & {\cellcolor[HTML]{B0D800}} \color[HTML]{000000} 0.303 & {\cellcolor[HTML]{D4EA00}} \color[HTML]{000000} 0.287 & {\cellcolor[HTML]{BEDF00}} \color[HTML]{000000} 0.297 & {\cellcolor[HTML]{B0D800}} \color[HTML]{000000} 0.303 & {\cellcolor[HTML]{B0D800}} \color[HTML]{000000} 0.303 & {\cellcolor[HTML]{B0D800}} \color[HTML]{000000} 0.303 & {\cellcolor[HTML]{B0D800}} \color[HTML]{000000} 0.303 & {\cellcolor[HTML]{B6DB00}} \color[HTML]{000000} 0.301 & {\cellcolor[HTML]{FF5800}} \color[HTML]{F1F1F1} 0.190 \\
6k-8k & {\cellcolor[HTML]{C0E000}} \color[HTML]{000000} 0.296 & {\cellcolor[HTML]{FFE200}} \color[HTML]{000000} 0.254 & {\cellcolor[HTML]{FFC200}} \color[HTML]{000000} 0.239 & {\cellcolor[HTML]{FF4A00}} \color[HTML]{F1F1F1} 0.184 & {\cellcolor[HTML]{FFE200}} \color[HTML]{000000} 0.254 & {\cellcolor[HTML]{FFE000}} \color[HTML]{000000} 0.253 & {\cellcolor[HTML]{FFE200}} \color[HTML]{000000} 0.254 & {\cellcolor[HTML]{FFF600}} \color[HTML]{000000} 0.263 & {\cellcolor[HTML]{FFE000}} \color[HTML]{000000} 0.253 & {\cellcolor[HTML]{FFE200}} \color[HTML]{000000} 0.254 & {\cellcolor[HTML]{FFE200}} \color[HTML]{000000} 0.254 & {\cellcolor[HTML]{FFE200}} \color[HTML]{000000} 0.254 & {\cellcolor[HTML]{FFE200}} \color[HTML]{000000} 0.254 & {\cellcolor[HTML]{FFE000}} \color[HTML]{000000} 0.253 & {\cellcolor[HTML]{FF6400}} \color[HTML]{F1F1F1} 0.196 \\
8k-10k & {\cellcolor[HTML]{AAD500}} \color[HTML]{000000} 0.306 & {\cellcolor[HTML]{FFBA00}} \color[HTML]{000000} 0.235 & {\cellcolor[HTML]{FF9E00}} \color[HTML]{000000} 0.222 & {\cellcolor[HTML]{FF3A00}} \color[HTML]{F1F1F1} 0.176 & {\cellcolor[HTML]{FFBA00}} \color[HTML]{000000} 0.235 & {\cellcolor[HTML]{FFC600}} \color[HTML]{000000} 0.241 & {\cellcolor[HTML]{FFBA00}} \color[HTML]{000000} 0.235 & {\cellcolor[HTML]{FFB400}} \color[HTML]{000000} 0.232 & {\cellcolor[HTML]{FFC600}} \color[HTML]{000000} 0.241 & {\cellcolor[HTML]{FFBA00}} \color[HTML]{000000} 0.235 & {\cellcolor[HTML]{FFBA00}} \color[HTML]{000000} 0.235 & {\cellcolor[HTML]{FFBA00}} \color[HTML]{000000} 0.235 & {\cellcolor[HTML]{FFBA00}} \color[HTML]{000000} 0.235 & {\cellcolor[HTML]{FFB800}} \color[HTML]{000000} 0.234 & {\cellcolor[HTML]{FF5400}} \color[HTML]{F1F1F1} 0.188 \\
10k-12k & {\cellcolor[HTML]{FAFD00}} \color[HTML]{000000} 0.269 & {\cellcolor[HTML]{FFC600}} \color[HTML]{000000} 0.241 & {\cellcolor[HTML]{FFA200}} \color[HTML]{000000} 0.224 & {\cellcolor[HTML]{FF4000}} \color[HTML]{F1F1F1} 0.179 & {\cellcolor[HTML]{FFC800}} \color[HTML]{000000} 0.242 & {\cellcolor[HTML]{FFC200}} \color[HTML]{000000} 0.239 & {\cellcolor[HTML]{FFC600}} \color[HTML]{000000} 0.241 & {\cellcolor[HTML]{FFB400}} \color[HTML]{000000} 0.232 & {\cellcolor[HTML]{FFBE00}} \color[HTML]{000000} 0.237 & {\cellcolor[HTML]{FFC600}} \color[HTML]{000000} 0.241 & {\cellcolor[HTML]{FFC600}} \color[HTML]{000000} 0.241 & {\cellcolor[HTML]{FFC600}} \color[HTML]{000000} 0.241 & {\cellcolor[HTML]{FFC600}} \color[HTML]{000000} 0.241 & {\cellcolor[HTML]{FFC600}} \color[HTML]{000000} 0.241 & {\cellcolor[HTML]{FF5000}} \color[HTML]{F1F1F1} 0.186 \\
12k-14k & {\cellcolor[HTML]{E2F100}} \color[HTML]{000000} 0.280 & {\cellcolor[HTML]{FFBE00}} \color[HTML]{000000} 0.237 & {\cellcolor[HTML]{FF8E00}} \color[HTML]{F1F1F1} 0.215 & {\cellcolor[HTML]{FF3A00}} \color[HTML]{F1F1F1} 0.176 & {\cellcolor[HTML]{FFBE00}} \color[HTML]{000000} 0.237 & {\cellcolor[HTML]{FFBE00}} \color[HTML]{000000} 0.237 & {\cellcolor[HTML]{FFBE00}} \color[HTML]{000000} 0.237 & {\cellcolor[HTML]{FFB800}} \color[HTML]{000000} 0.234 & {\cellcolor[HTML]{FFBC00}} \color[HTML]{000000} 0.236 & {\cellcolor[HTML]{FFBE00}} \color[HTML]{000000} 0.237 & {\cellcolor[HTML]{FFC000}} \color[HTML]{000000} 0.238 & {\cellcolor[HTML]{FFBE00}} \color[HTML]{000000} 0.237 & {\cellcolor[HTML]{FFBE00}} \color[HTML]{000000} 0.237 & {\cellcolor[HTML]{FFBA00}} \color[HTML]{000000} 0.235 & {\cellcolor[HTML]{FF3600}} \color[HTML]{F1F1F1} 0.174 \\
14k-16k & {\cellcolor[HTML]{FFB000}} \color[HTML]{000000} 0.231 & {\cellcolor[HTML]{FFC400}} \color[HTML]{000000} 0.240 & {\cellcolor[HTML]{FF8A00}} \color[HTML]{F1F1F1} 0.213 & {\cellcolor[HTML]{FF3C00}} \color[HTML]{F1F1F1} 0.177 & {\cellcolor[HTML]{FFC600}} \color[HTML]{000000} 0.241 & {\cellcolor[HTML]{FFC600}} \color[HTML]{000000} 0.241 & {\cellcolor[HTML]{FFC400}} \color[HTML]{000000} 0.240 & {\cellcolor[HTML]{FFB600}} \color[HTML]{000000} 0.233 & {\cellcolor[HTML]{FFC800}} \color[HTML]{000000} 0.242 & {\cellcolor[HTML]{FFC400}} \color[HTML]{000000} 0.240 & {\cellcolor[HTML]{FFC200}} \color[HTML]{000000} 0.239 & {\cellcolor[HTML]{FFC400}} \color[HTML]{000000} 0.240 & {\cellcolor[HTML]{FFC400}} \color[HTML]{000000} 0.240 & {\cellcolor[HTML]{FFCA00}} \color[HTML]{000000} 0.243 & {\cellcolor[HTML]{FF2E00}} \color[HTML]{F1F1F1} 0.171 \\
16k-24k & {\cellcolor[HTML]{FAFD00}} \color[HTML]{000000} 0.269 & {\cellcolor[HTML]{FFDE00}} \color[HTML]{000000} 0.252 & {\cellcolor[HTML]{FF8600}} \color[HTML]{F1F1F1} 0.211 & {\cellcolor[HTML]{FF3C00}} \color[HTML]{F1F1F1} 0.177 & {\cellcolor[HTML]{FFDE00}} \color[HTML]{000000} 0.252 & {\cellcolor[HTML]{FFDC00}} \color[HTML]{000000} 0.251 & {\cellcolor[HTML]{FFDE00}} \color[HTML]{000000} 0.252 & {\cellcolor[HTML]{FFD800}} \color[HTML]{000000} 0.249 & {\cellcolor[HTML]{FFDE00}} \color[HTML]{000000} 0.252 & {\cellcolor[HTML]{FFDE00}} \color[HTML]{000000} 0.252 & {\cellcolor[HTML]{FFDE00}} \color[HTML]{000000} 0.252 & {\cellcolor[HTML]{FFDE00}} \color[HTML]{000000} 0.252 & {\cellcolor[HTML]{FFDE00}} \color[HTML]{000000} 0.252 & {\cellcolor[HTML]{FFDC00}} \color[HTML]{000000} 0.251 & {\cellcolor[HTML]{FF4800}} \color[HTML]{F1F1F1} 0.183 \\
24k-32k & {\cellcolor[HTML]{FFD400}} \color[HTML]{000000} 0.247 & {\cellcolor[HTML]{FF7C00}} \color[HTML]{F1F1F1} 0.207 & {\cellcolor[HTML]{FF8200}} \color[HTML]{F1F1F1} 0.209 & {\cellcolor[HTML]{FF1E00}} \color[HTML]{F1F1F1} 0.163 & {\cellcolor[HTML]{FF7C00}} \color[HTML]{F1F1F1} 0.207 & {\cellcolor[HTML]{FF7C00}} \color[HTML]{F1F1F1} 0.207 & {\cellcolor[HTML]{FF7C00}} \color[HTML]{F1F1F1} 0.207 & {\cellcolor[HTML]{FF8200}} \color[HTML]{F1F1F1} 0.209 & {\cellcolor[HTML]{FF7800}} \color[HTML]{F1F1F1} 0.205 & {\cellcolor[HTML]{FF7C00}} \color[HTML]{F1F1F1} 0.207 & {\cellcolor[HTML]{FF7C00}} \color[HTML]{F1F1F1} 0.207 & {\cellcolor[HTML]{FF7C00}} \color[HTML]{F1F1F1} 0.207 & {\cellcolor[HTML]{FF7C00}} \color[HTML]{F1F1F1} 0.207 & {\cellcolor[HTML]{FF7E00}} \color[HTML]{F1F1F1} 0.208 & {\cellcolor[HTML]{FF5800}} \color[HTML]{F1F1F1} 0.190 \\
32k-49k & {\cellcolor[HTML]{FF2E00}} \color[HTML]{F1F1F1} 0.171 & {\cellcolor[HTML]{FF6E00}} \color[HTML]{F1F1F1} 0.200 & {\cellcolor[HTML]{FF5600}} \color[HTML]{F1F1F1} 0.189 & {\cellcolor[HTML]{FF0000}} \color[HTML]{F1F1F1} 0.149 & {\cellcolor[HTML]{FF6E00}} \color[HTML]{F1F1F1} 0.200 & {\cellcolor[HTML]{FF6E00}} \color[HTML]{F1F1F1} 0.200 & {\cellcolor[HTML]{FF6E00}} \color[HTML]{F1F1F1} 0.200 & {\cellcolor[HTML]{FF9E00}} \color[HTML]{000000} 0.222 & {\cellcolor[HTML]{FF7200}} \color[HTML]{F1F1F1} 0.202 & {\cellcolor[HTML]{FF6E00}} \color[HTML]{F1F1F1} 0.200 & {\cellcolor[HTML]{FF6E00}} \color[HTML]{F1F1F1} 0.200 & {\cellcolor[HTML]{FF6E00}} \color[HTML]{F1F1F1} 0.200 & {\cellcolor[HTML]{FF7400}} \color[HTML]{F1F1F1} 0.203 & {\cellcolor[HTML]{FF7400}} \color[HTML]{F1F1F1} 0.203 & {\cellcolor[HTML]{FF1E00}} \color[HTML]{F1F1F1} 0.163 \\
49k-65k & {\cellcolor[HTML]{FF5600}} \color[HTML]{F1F1F1} 0.189 & {\cellcolor[HTML]{FFF800}} \color[HTML]{000000} 0.264 & {\cellcolor[HTML]{FFAE00}} \color[HTML]{000000} 0.230 & {\cellcolor[HTML]{FFC200}} \color[HTML]{000000} 0.239 & {\cellcolor[HTML]{FFF800}} \color[HTML]{000000} 0.264 & {\cellcolor[HTML]{FFF800}} \color[HTML]{000000} 0.264 & {\cellcolor[HTML]{FFF800}} \color[HTML]{000000} 0.264 & {\cellcolor[HTML]{F2F900}} \color[HTML]{000000} 0.273 & {\cellcolor[HTML]{FFF800}} \color[HTML]{000000} 0.264 & {\cellcolor[HTML]{FFF800}} \color[HTML]{000000} 0.264 & {\cellcolor[HTML]{FFF800}} \color[HTML]{000000} 0.264 & {\cellcolor[HTML]{FFF800}} \color[HTML]{000000} 0.264 & {\cellcolor[HTML]{FFF800}} \color[HTML]{000000} 0.264 & {\cellcolor[HTML]{FFF800}} \color[HTML]{000000} 0.264 & {\cellcolor[HTML]{FF7800}} \color[HTML]{F1F1F1} 0.205 \\
\hline
In-Domain (0k-8k) & {\cellcolor[HTML]{48A400}} \color[HTML]{F1F1F1} 0.351 & {\cellcolor[HTML]{BADD00}} \color[HTML]{000000} 0.299 & {\cellcolor[HTML]{D0E800}} \color[HTML]{000000} 0.289 & {\cellcolor[HTML]{FFC200}} \color[HTML]{000000} 0.239 & {\cellcolor[HTML]{BADD00}} \color[HTML]{000000} 0.299 & {\cellcolor[HTML]{BADD00}} \color[HTML]{000000} 0.299 & {\cellcolor[HTML]{BADD00}} \color[HTML]{000000} 0.299 & {\cellcolor[HTML]{C4E200}} \color[HTML]{000000} 0.294 & {\cellcolor[HTML]{C2E100}} \color[HTML]{000000} 0.295 & {\cellcolor[HTML]{BADD00}} \color[HTML]{000000} 0.299 & {\cellcolor[HTML]{BADD00}} \color[HTML]{000000} 0.299 & {\cellcolor[HTML]{BADD00}} \color[HTML]{000000} 0.299 & {\cellcolor[HTML]{BADD00}} \color[HTML]{000000} 0.299 & {\cellcolor[HTML]{BEDF00}} \color[HTML]{000000} 0.297 & {\cellcolor[HTML]{FFBA00}} \color[HTML]{000000} 0.235 \\
OOD (8k-65k) & {\cellcolor[HTML]{FFFC00}} \color[HTML]{000000} 0.266 & {\cellcolor[HTML]{FFC000}} \color[HTML]{000000} 0.238 & {\cellcolor[HTML]{FF9000}} \color[HTML]{000000} 0.215 & {\cellcolor[HTML]{FF3A00}} \color[HTML]{F1F1F1} 0.176 & {\cellcolor[HTML]{FFC000}} \color[HTML]{000000} 0.238 & {\cellcolor[HTML]{FFC200}} \color[HTML]{000000} 0.238 & {\cellcolor[HTML]{FFC000}} \color[HTML]{000000} 0.238 & {\cellcolor[HTML]{FFB800}} \color[HTML]{000000} 0.234 & {\cellcolor[HTML]{FFC000}} \color[HTML]{000000} 0.238 & {\cellcolor[HTML]{FFC000}} \color[HTML]{000000} 0.238 & {\cellcolor[HTML]{FFC000}} \color[HTML]{000000} 0.238 & {\cellcolor[HTML]{FFC000}} \color[HTML]{000000} 0.238 & {\cellcolor[HTML]{FFC000}} \color[HTML]{000000} 0.238 & {\cellcolor[HTML]{FFC000}} \color[HTML]{000000} 0.238 & {\cellcolor[HTML]{FF4400}} \color[HTML]{F1F1F1} 0.180 \\
Long-OOD (32k-65k) & {\cellcolor[HTML]{FF3800}} \color[HTML]{F1F1F1} 0.175 & {\cellcolor[HTML]{FF8E00}} \color[HTML]{F1F1F1} 0.215 & {\cellcolor[HTML]{FF6A00}} \color[HTML]{F1F1F1} 0.198 & {\cellcolor[HTML]{FF2C00}} \color[HTML]{F1F1F1} 0.170 & {\cellcolor[HTML]{FF8E00}} \color[HTML]{F1F1F1} 0.215 & {\cellcolor[HTML]{FF8E00}} \color[HTML]{F1F1F1} 0.215 & {\cellcolor[HTML]{FF8E00}} \color[HTML]{F1F1F1} 0.215 & {\cellcolor[HTML]{FFB600}} \color[HTML]{000000} 0.234 & {\cellcolor[HTML]{FF9200}} \color[HTML]{000000} 0.216 & {\cellcolor[HTML]{FF8E00}} \color[HTML]{F1F1F1} 0.215 & {\cellcolor[HTML]{FF8E00}} \color[HTML]{F1F1F1} 0.215 & {\cellcolor[HTML]{FF8E00}} \color[HTML]{F1F1F1} 0.215 & {\cellcolor[HTML]{FF9200}} \color[HTML]{000000} 0.217 & {\cellcolor[HTML]{FF9200}} \color[HTML]{000000} 0.217 & {\cellcolor[HTML]{FF3200}} \color[HTML]{F1F1F1} 0.173 \\
Full (0k-65k) & {\cellcolor[HTML]{AAD500}} \color[HTML]{000000} 0.306 & {\cellcolor[HTML]{FFFE00}} \color[HTML]{000000} 0.266 & {\cellcolor[HTML]{FFDA00}} \color[HTML]{000000} 0.250 & {\cellcolor[HTML]{FF7A00}} \color[HTML]{F1F1F1} 0.205 & {\cellcolor[HTML]{FFFE00}} \color[HTML]{000000} 0.267 & {\cellcolor[HTML]{FFFE00}} \color[HTML]{000000} 0.267 & {\cellcolor[HTML]{FFFE00}} \color[HTML]{000000} 0.266 & {\cellcolor[HTML]{FFF400}} \color[HTML]{000000} 0.262 & {\cellcolor[HTML]{FFFA00}} \color[HTML]{000000} 0.265 & {\cellcolor[HTML]{FFFE00}} \color[HTML]{000000} 0.266 & {\cellcolor[HTML]{FFFE00}} \color[HTML]{000000} 0.266 & {\cellcolor[HTML]{FFFE00}} \color[HTML]{000000} 0.266 & {\cellcolor[HTML]{FFFE00}} \color[HTML]{000000} 0.266 & {\cellcolor[HTML]{FFFC00}} \color[HTML]{000000} 0.265 & {\cellcolor[HTML]{FF7A00}} \color[HTML]{F1F1F1} 0.206 \\
\bottomrule
\end{tabular}}
\caption{Ablation for all associative layers on the GovReport-100+ dataset for ARMT with Gemma-3-1B-IT model, metric - ROUGE-L. Ablated associative layers from 13 to 25.}
\label{tab:gov_report_best_all_layers_assoc_ablation_13_26}
\end{center}
\end{table*}
\begin{table*}[h]

\begin{center}
\resizebox{1.0\textwidth}{!}{
\begin{tabular}{lccccccccccccccc}
\toprule
\begin{tabular}[c]{@{}l@{}}\textbf{Model\//}\\ \textbf{Lengths}\end{tabular} & \begin{tabular}[c]{@{}l@{}}\textbf{Base,}\\ \textbf{MT,}\\ \textbf{8k}\end{tabular} & \begin{tabular}[c]{@{}l@{}}\textbf{ARMT,}\\ \textbf{MT,}\\ \textbf{8k}\end{tabular} & \begin{tabular}[c]{@{}l@{}}\textbf{W/o}\\ \textbf{layer}\\ \textbf{0}\end{tabular} & \begin{tabular}[c]{@{}l@{}}\textbf{W/o}\\ \textbf{layer}\\ \textbf{1}\end{tabular} & \begin{tabular}[c]{@{}l@{}}\textbf{W/o}\\ \textbf{layer}\\ \textbf{2}\end{tabular} & \begin{tabular}[c]{@{}l@{}}\textbf{W/o}\\ \textbf{layer}\\ \textbf{3}\end{tabular} & \begin{tabular}[c]{@{}l@{}}\textbf{W/o}\\ \textbf{layer}\\ \textbf{4}\end{tabular} & \begin{tabular}[c]{@{}l@{}}\textbf{W/o}\\ \textbf{layer}\\ \textbf{5}\end{tabular} & \begin{tabular}[c]{@{}l@{}}\textbf{W/o}\\ \textbf{layer}\\ \textbf{6}\end{tabular} & \begin{tabular}[c]{@{}l@{}}\textbf{W/o}\\ \textbf{layer}\\ \textbf{7}\end{tabular} & \begin{tabular}[c]{@{}l@{}}\textbf{W/o}\\ \textbf{layer}\\ \textbf{8}\end{tabular}  & \begin{tabular}[c]{@{}l@{}}\textbf{W/o}\\ \textbf{layer}\\ \textbf{9}\end{tabular} & \begin{tabular}[c]{@{}l@{}}\textbf{W/o}\\ \textbf{layer}\\ \textbf{10}\end{tabular} & \begin{tabular}[c]{@{}l@{}}\textbf{W/o}\\ \textbf{layer}\\ \textbf{11}\end{tabular} & \begin{tabular}[c]{@{}l@{}}\textbf{W/o}\\ \textbf{layer}\\ \textbf{12}\end{tabular} \\
\midrule
0k-1k & {\cellcolor[HTML]{42A100}} \color[HTML]{F1F1F1} 0.795 & {\cellcolor[HTML]{68B400}} \color[HTML]{F1F1F1} 0.756 & {\cellcolor[HTML]{68B400}} \color[HTML]{F1F1F1} 0.756 & {\cellcolor[HTML]{68B400}} \color[HTML]{F1F1F1} 0.756 & {\cellcolor[HTML]{68B400}} \color[HTML]{F1F1F1} 0.756 & {\cellcolor[HTML]{68B400}} \color[HTML]{F1F1F1} 0.756 & {\cellcolor[HTML]{68B400}} \color[HTML]{F1F1F1} 0.756 & {\cellcolor[HTML]{68B400}} \color[HTML]{F1F1F1} 0.756 & {\cellcolor[HTML]{68B400}} \color[HTML]{F1F1F1} 0.756 & {\cellcolor[HTML]{68B400}} \color[HTML]{F1F1F1} 0.756 & {\cellcolor[HTML]{68B400}} \color[HTML]{F1F1F1} 0.756 & {\cellcolor[HTML]{68B400}} \color[HTML]{F1F1F1} 0.756 & {\cellcolor[HTML]{68B400}} \color[HTML]{F1F1F1} 0.756 & {\cellcolor[HTML]{68B400}} \color[HTML]{F1F1F1} 0.756 & {\cellcolor[HTML]{68B400}} \color[HTML]{F1F1F1} 0.756 \\
1k-2k & {\cellcolor[HTML]{48A400}} \color[HTML]{F1F1F1} 0.790 & {\cellcolor[HTML]{74BA00}} \color[HTML]{F1F1F1} 0.743 & {\cellcolor[HTML]{7CBE00}} \color[HTML]{000000} 0.733 & {\cellcolor[HTML]{7CBE00}} \color[HTML]{000000} 0.733 & {\cellcolor[HTML]{7CBE00}} \color[HTML]{000000} 0.733 & {\cellcolor[HTML]{7CBE00}} \color[HTML]{000000} 0.733 & {\cellcolor[HTML]{7CBE00}} \color[HTML]{000000} 0.733 & {\cellcolor[HTML]{7CBE00}} \color[HTML]{000000} 0.733 & {\cellcolor[HTML]{74BA00}} \color[HTML]{F1F1F1} 0.743 & {\cellcolor[HTML]{74BA00}} \color[HTML]{F1F1F1} 0.743 & {\cellcolor[HTML]{74BA00}} \color[HTML]{F1F1F1} 0.743 & {\cellcolor[HTML]{7CBE00}} \color[HTML]{000000} 0.733 & {\cellcolor[HTML]{74BA00}} \color[HTML]{F1F1F1} 0.743 & {\cellcolor[HTML]{74BA00}} \color[HTML]{F1F1F1} 0.743 & {\cellcolor[HTML]{7CBE00}} \color[HTML]{000000} 0.733 \\
2k-4k & {\cellcolor[HTML]{56AB00}} \color[HTML]{F1F1F1} 0.774 & {\cellcolor[HTML]{7ABD00}} \color[HTML]{F1F1F1} 0.736 & {\cellcolor[HTML]{7ABD00}} \color[HTML]{F1F1F1} 0.736 & {\cellcolor[HTML]{7ABD00}} \color[HTML]{F1F1F1} 0.736 & {\cellcolor[HTML]{7ABD00}} \color[HTML]{F1F1F1} 0.736 & {\cellcolor[HTML]{7ABD00}} \color[HTML]{F1F1F1} 0.736 & {\cellcolor[HTML]{7ABD00}} \color[HTML]{F1F1F1} 0.736 & {\cellcolor[HTML]{7ABD00}} \color[HTML]{F1F1F1} 0.736 & {\cellcolor[HTML]{7ABD00}} \color[HTML]{F1F1F1} 0.736 & {\cellcolor[HTML]{7ABD00}} \color[HTML]{F1F1F1} 0.736 & {\cellcolor[HTML]{82C100}} \color[HTML]{000000} 0.726 & {\cellcolor[HTML]{7ABD00}} \color[HTML]{F1F1F1} 0.736 & {\cellcolor[HTML]{7ABD00}} \color[HTML]{F1F1F1} 0.736 & {\cellcolor[HTML]{7ABD00}} \color[HTML]{F1F1F1} 0.736 & {\cellcolor[HTML]{7ABD00}} \color[HTML]{F1F1F1} 0.736 \\
4k-6k & {\cellcolor[HTML]{5CAE00}} \color[HTML]{F1F1F1} 0.767 & {\cellcolor[HTML]{9CCE00}} \color[HTML]{000000} 0.699 & {\cellcolor[HTML]{9CCE00}} \color[HTML]{000000} 0.699 & {\cellcolor[HTML]{9CCE00}} \color[HTML]{000000} 0.699 & {\cellcolor[HTML]{AAD500}} \color[HTML]{000000} 0.685 & {\cellcolor[HTML]{9CCE00}} \color[HTML]{000000} 0.699 & {\cellcolor[HTML]{9CCE00}} \color[HTML]{000000} 0.699 & {\cellcolor[HTML]{9CCE00}} \color[HTML]{000000} 0.699 & {\cellcolor[HTML]{9CCE00}} \color[HTML]{000000} 0.699 & {\cellcolor[HTML]{9CCE00}} \color[HTML]{000000} 0.699 & {\cellcolor[HTML]{9CCE00}} \color[HTML]{000000} 0.699 & {\cellcolor[HTML]{AAD500}} \color[HTML]{000000} 0.685 & {\cellcolor[HTML]{9CCE00}} \color[HTML]{000000} 0.699 & {\cellcolor[HTML]{9CCE00}} \color[HTML]{000000} 0.699 & {\cellcolor[HTML]{9CCE00}} \color[HTML]{000000} 0.699 \\
6k-8k & {\cellcolor[HTML]{088400}} \color[HTML]{F1F1F1} 0.859 & {\cellcolor[HTML]{3A9D00}} \color[HTML]{F1F1F1} 0.804 & {\cellcolor[HTML]{3A9D00}} \color[HTML]{F1F1F1} 0.804 & {\cellcolor[HTML]{3A9D00}} \color[HTML]{F1F1F1} 0.804 & {\cellcolor[HTML]{309800}} \color[HTML]{F1F1F1} 0.815 & {\cellcolor[HTML]{309800}} \color[HTML]{F1F1F1} 0.815 & {\cellcolor[HTML]{3A9D00}} \color[HTML]{F1F1F1} 0.804 & {\cellcolor[HTML]{309800}} \color[HTML]{F1F1F1} 0.815 & {\cellcolor[HTML]{3A9D00}} \color[HTML]{F1F1F1} 0.804 & {\cellcolor[HTML]{3A9D00}} \color[HTML]{F1F1F1} 0.804 & {\cellcolor[HTML]{309800}} \color[HTML]{F1F1F1} 0.815 & {\cellcolor[HTML]{3A9D00}} \color[HTML]{F1F1F1} 0.804 & {\cellcolor[HTML]{3A9D00}} \color[HTML]{F1F1F1} 0.804 & {\cellcolor[HTML]{3A9D00}} \color[HTML]{F1F1F1} 0.804 & {\cellcolor[HTML]{3A9D00}} \color[HTML]{F1F1F1} 0.804 \\
8k-10k & {\cellcolor[HTML]{0E8700}} \color[HTML]{F1F1F1} 0.852 & {\cellcolor[HTML]{6AB500}} \color[HTML]{F1F1F1} 0.753 & {\cellcolor[HTML]{6AB500}} \color[HTML]{F1F1F1} 0.753 & {\cellcolor[HTML]{6AB500}} \color[HTML]{F1F1F1} 0.753 & {\cellcolor[HTML]{76BB00}} \color[HTML]{F1F1F1} 0.741 & {\cellcolor[HTML]{5EAF00}} \color[HTML]{F1F1F1} 0.765 & {\cellcolor[HTML]{76BB00}} \color[HTML]{F1F1F1} 0.741 & {\cellcolor[HTML]{82C100}} \color[HTML]{000000} 0.728 & {\cellcolor[HTML]{6AB500}} \color[HTML]{F1F1F1} 0.753 & {\cellcolor[HTML]{6AB500}} \color[HTML]{F1F1F1} 0.753 & {\cellcolor[HTML]{52A900}} \color[HTML]{F1F1F1} 0.778 & {\cellcolor[HTML]{6AB500}} \color[HTML]{F1F1F1} 0.753 & {\cellcolor[HTML]{76BB00}} \color[HTML]{F1F1F1} 0.741 & {\cellcolor[HTML]{6AB500}} \color[HTML]{F1F1F1} 0.753 & {\cellcolor[HTML]{76BB00}} \color[HTML]{F1F1F1} 0.741 \\
10k-12k & {\cellcolor[HTML]{008000}} \color[HTML]{F1F1F1} 0.868 & {\cellcolor[HTML]{46A300}} \color[HTML]{F1F1F1} 0.791 & {\cellcolor[HTML]{46A300}} \color[HTML]{F1F1F1} 0.791 & {\cellcolor[HTML]{5AAD00}} \color[HTML]{F1F1F1} 0.769 & {\cellcolor[HTML]{46A300}} \color[HTML]{F1F1F1} 0.791 & {\cellcolor[HTML]{3C9E00}} \color[HTML]{F1F1F1} 0.802 & {\cellcolor[HTML]{46A300}} \color[HTML]{F1F1F1} 0.791 & {\cellcolor[HTML]{3C9E00}} \color[HTML]{F1F1F1} 0.802 & {\cellcolor[HTML]{46A300}} \color[HTML]{F1F1F1} 0.791 & {\cellcolor[HTML]{46A300}} \color[HTML]{F1F1F1} 0.791 & {\cellcolor[HTML]{329900}} \color[HTML]{F1F1F1} 0.813 & {\cellcolor[HTML]{46A300}} \color[HTML]{F1F1F1} 0.791 & {\cellcolor[HTML]{46A300}} \color[HTML]{F1F1F1} 0.791 & {\cellcolor[HTML]{46A300}} \color[HTML]{F1F1F1} 0.791 & {\cellcolor[HTML]{46A300}} \color[HTML]{F1F1F1} 0.791 \\
12k-14k & {\cellcolor[HTML]{4AA500}} \color[HTML]{F1F1F1} 0.787 & {\cellcolor[HTML]{72B900}} \color[HTML]{F1F1F1} 0.745 & {\cellcolor[HTML]{7CBE00}} \color[HTML]{000000} 0.734 & {\cellcolor[HTML]{68B400}} \color[HTML]{F1F1F1} 0.755 & {\cellcolor[HTML]{7CBE00}} \color[HTML]{000000} 0.734 & {\cellcolor[HTML]{86C300}} \color[HTML]{000000} 0.723 & {\cellcolor[HTML]{7CBE00}} \color[HTML]{000000} 0.734 & {\cellcolor[HTML]{7CBE00}} \color[HTML]{000000} 0.734 & {\cellcolor[HTML]{7CBE00}} \color[HTML]{000000} 0.734 & {\cellcolor[HTML]{72B900}} \color[HTML]{F1F1F1} 0.745 & {\cellcolor[HTML]{5EAF00}} \color[HTML]{F1F1F1} 0.766 & {\cellcolor[HTML]{72B900}} \color[HTML]{F1F1F1} 0.745 & {\cellcolor[HTML]{72B900}} \color[HTML]{F1F1F1} 0.745 & {\cellcolor[HTML]{72B900}} \color[HTML]{F1F1F1} 0.745 & {\cellcolor[HTML]{72B900}} \color[HTML]{F1F1F1} 0.745 \\
14k-16k & {\cellcolor[HTML]{60B000}} \color[HTML]{F1F1F1} 0.764 & {\cellcolor[HTML]{AAD500}} \color[HTML]{000000} 0.685 & {\cellcolor[HTML]{AAD500}} \color[HTML]{000000} 0.685 & {\cellcolor[HTML]{9ECF00}} \color[HTML]{000000} 0.697 & {\cellcolor[HTML]{AAD500}} \color[HTML]{000000} 0.685 & {\cellcolor[HTML]{B4DA00}} \color[HTML]{000000} 0.674 & {\cellcolor[HTML]{AAD500}} \color[HTML]{000000} 0.685 & {\cellcolor[HTML]{AAD500}} \color[HTML]{000000} 0.685 & {\cellcolor[HTML]{AAD500}} \color[HTML]{000000} 0.685 & {\cellcolor[HTML]{AAD500}} \color[HTML]{000000} 0.685 & {\cellcolor[HTML]{B4DA00}} \color[HTML]{000000} 0.674 & {\cellcolor[HTML]{AAD500}} \color[HTML]{000000} 0.685 & {\cellcolor[HTML]{AAD500}} \color[HTML]{000000} 0.685 & {\cellcolor[HTML]{AAD500}} \color[HTML]{000000} 0.685 & {\cellcolor[HTML]{AAD500}} \color[HTML]{000000} 0.685 \\
16k-24k & {\cellcolor[HTML]{349A00}} \color[HTML]{F1F1F1} 0.812 & {\cellcolor[HTML]{66B300}} \color[HTML]{F1F1F1} 0.757 & {\cellcolor[HTML]{60B000}} \color[HTML]{F1F1F1} 0.764 & {\cellcolor[HTML]{60B000}} \color[HTML]{F1F1F1} 0.764 & {\cellcolor[HTML]{60B000}} \color[HTML]{F1F1F1} 0.764 & {\cellcolor[HTML]{6CB600}} \color[HTML]{F1F1F1} 0.750 & {\cellcolor[HTML]{60B000}} \color[HTML]{F1F1F1} 0.764 & {\cellcolor[HTML]{66B300}} \color[HTML]{F1F1F1} 0.757 & {\cellcolor[HTML]{60B000}} \color[HTML]{F1F1F1} 0.764 & {\cellcolor[HTML]{66B300}} \color[HTML]{F1F1F1} 0.757 & {\cellcolor[HTML]{60B000}} \color[HTML]{F1F1F1} 0.764 & {\cellcolor[HTML]{5AAD00}} \color[HTML]{F1F1F1} 0.771 & {\cellcolor[HTML]{60B000}} \color[HTML]{F1F1F1} 0.764 & {\cellcolor[HTML]{66B300}} \color[HTML]{F1F1F1} 0.757 & {\cellcolor[HTML]{60B000}} \color[HTML]{F1F1F1} 0.764 \\
24k-32k & {\cellcolor[HTML]{90C800}} \color[HTML]{000000} 0.713 & {\cellcolor[HTML]{90C800}} \color[HTML]{000000} 0.713 & {\cellcolor[HTML]{90C800}} \color[HTML]{000000} 0.713 & {\cellcolor[HTML]{86C300}} \color[HTML]{000000} 0.723 & {\cellcolor[HTML]{98CC00}} \color[HTML]{000000} 0.703 & {\cellcolor[HTML]{90C800}} \color[HTML]{000000} 0.713 & {\cellcolor[HTML]{90C800}} \color[HTML]{000000} 0.713 & {\cellcolor[HTML]{98CC00}} \color[HTML]{000000} 0.703 & {\cellcolor[HTML]{98CC00}} \color[HTML]{000000} 0.703 & {\cellcolor[HTML]{98CC00}} \color[HTML]{000000} 0.703 & {\cellcolor[HTML]{A2D100}} \color[HTML]{000000} 0.693 & {\cellcolor[HTML]{90C800}} \color[HTML]{000000} 0.713 & {\cellcolor[HTML]{90C800}} \color[HTML]{000000} 0.713 & {\cellcolor[HTML]{90C800}} \color[HTML]{000000} 0.713 & {\cellcolor[HTML]{98CC00}} \color[HTML]{000000} 0.703 \\
32k-49k & {\cellcolor[HTML]{FFFE00}} \color[HTML]{000000} 0.591 & {\cellcolor[HTML]{5EAF00}} \color[HTML]{F1F1F1} 0.765 & {\cellcolor[HTML]{66B300}} \color[HTML]{F1F1F1} 0.757 & {\cellcolor[HTML]{5EAF00}} \color[HTML]{F1F1F1} 0.765 & {\cellcolor[HTML]{66B300}} \color[HTML]{F1F1F1} 0.757 & {\cellcolor[HTML]{66B300}} \color[HTML]{F1F1F1} 0.757 & {\cellcolor[HTML]{66B300}} \color[HTML]{F1F1F1} 0.757 & {\cellcolor[HTML]{66B300}} \color[HTML]{F1F1F1} 0.757 & {\cellcolor[HTML]{66B300}} \color[HTML]{F1F1F1} 0.757 & {\cellcolor[HTML]{66B300}} \color[HTML]{F1F1F1} 0.757 & {\cellcolor[HTML]{66B300}} \color[HTML]{F1F1F1} 0.757 & {\cellcolor[HTML]{5EAF00}} \color[HTML]{F1F1F1} 0.765 & {\cellcolor[HTML]{66B300}} \color[HTML]{F1F1F1} 0.757 & {\cellcolor[HTML]{5EAF00}} \color[HTML]{F1F1F1} 0.765 & {\cellcolor[HTML]{66B300}} \color[HTML]{F1F1F1} 0.757 \\
49k-65k & {\cellcolor[HTML]{FF0000}} \color[HTML]{F1F1F1} 0.317 & {\cellcolor[HTML]{90C800}} \color[HTML]{000000} 0.713 & {\cellcolor[HTML]{90C800}} \color[HTML]{000000} 0.713 & {\cellcolor[HTML]{86C300}} \color[HTML]{000000} 0.723 & {\cellcolor[HTML]{90C800}} \color[HTML]{000000} 0.713 & {\cellcolor[HTML]{7CBE00}} \color[HTML]{000000} 0.733 & {\cellcolor[HTML]{86C300}} \color[HTML]{000000} 0.723 & {\cellcolor[HTML]{A2D100}} \color[HTML]{000000} 0.693 & {\cellcolor[HTML]{90C800}} \color[HTML]{000000} 0.713 & {\cellcolor[HTML]{90C800}} \color[HTML]{000000} 0.713 & {\cellcolor[HTML]{74BA00}} \color[HTML]{F1F1F1} 0.743 & {\cellcolor[HTML]{90C800}} \color[HTML]{000000} 0.713 & {\cellcolor[HTML]{90C800}} \color[HTML]{000000} 0.713 & {\cellcolor[HTML]{90C800}} \color[HTML]{000000} 0.713 & {\cellcolor[HTML]{90C800}} \color[HTML]{000000} 0.713 \\
\hline
In-Domain (0k-8k) & {\cellcolor[HTML]{40A000}} \color[HTML]{F1F1F1} 0.797 & {\cellcolor[HTML]{6EB700}} \color[HTML]{F1F1F1} 0.749 & {\cellcolor[HTML]{70B800}} \color[HTML]{F1F1F1} 0.747 & {\cellcolor[HTML]{70B800}} \color[HTML]{F1F1F1} 0.747 & {\cellcolor[HTML]{70B800}} \color[HTML]{F1F1F1} 0.747 & {\cellcolor[HTML]{6EB700}} \color[HTML]{F1F1F1} 0.749 & {\cellcolor[HTML]{70B800}} \color[HTML]{F1F1F1} 0.747 & {\cellcolor[HTML]{6EB700}} \color[HTML]{F1F1F1} 0.749 & {\cellcolor[HTML]{6EB700}} \color[HTML]{F1F1F1} 0.749 & {\cellcolor[HTML]{6EB700}} \color[HTML]{F1F1F1} 0.749 & {\cellcolor[HTML]{6EB700}} \color[HTML]{F1F1F1} 0.749 & {\cellcolor[HTML]{72B900}} \color[HTML]{F1F1F1} 0.744 & {\cellcolor[HTML]{6EB700}} \color[HTML]{F1F1F1} 0.749 & {\cellcolor[HTML]{6EB700}} \color[HTML]{F1F1F1} 0.749 & {\cellcolor[HTML]{70B800}} \color[HTML]{F1F1F1} 0.747 \\
OOD (8k-65k) & {\cellcolor[HTML]{92C900}} \color[HTML]{000000} 0.709 & {\cellcolor[HTML]{74BA00}} \color[HTML]{F1F1F1} 0.741 & {\cellcolor[HTML]{76BB00}} \color[HTML]{F1F1F1} 0.740 & {\cellcolor[HTML]{72B900}} \color[HTML]{F1F1F1} 0.745 & {\cellcolor[HTML]{78BC00}} \color[HTML]{F1F1F1} 0.738 & {\cellcolor[HTML]{76BB00}} \color[HTML]{F1F1F1} 0.740 & {\cellcolor[HTML]{76BB00}} \color[HTML]{F1F1F1} 0.740 & {\cellcolor[HTML]{7CBE00}} \color[HTML]{000000} 0.734 & {\cellcolor[HTML]{76BB00}} \color[HTML]{F1F1F1} 0.739 & {\cellcolor[HTML]{76BB00}} \color[HTML]{F1F1F1} 0.739 & {\cellcolor[HTML]{6EB700}} \color[HTML]{F1F1F1} 0.749 & {\cellcolor[HTML]{72B900}} \color[HTML]{F1F1F1} 0.744 & {\cellcolor[HTML]{76BB00}} \color[HTML]{F1F1F1} 0.740 & {\cellcolor[HTML]{74BA00}} \color[HTML]{F1F1F1} 0.741 & {\cellcolor[HTML]{76BB00}} \color[HTML]{F1F1F1} 0.739 \\
Long-OOD (32k-65k) & {\cellcolor[HTML]{FF8600}} \color[HTML]{F1F1F1} 0.463 & {\cellcolor[HTML]{76BB00}} \color[HTML]{F1F1F1} 0.741 & {\cellcolor[HTML]{7ABD00}} \color[HTML]{F1F1F1} 0.736 & {\cellcolor[HTML]{70B800}} \color[HTML]{F1F1F1} 0.745 & {\cellcolor[HTML]{7ABD00}} \color[HTML]{F1F1F1} 0.736 & {\cellcolor[HTML]{70B800}} \color[HTML]{F1F1F1} 0.746 & {\cellcolor[HTML]{74BA00}} \color[HTML]{F1F1F1} 0.741 & {\cellcolor[HTML]{82C100}} \color[HTML]{000000} 0.727 & {\cellcolor[HTML]{7ABD00}} \color[HTML]{F1F1F1} 0.736 & {\cellcolor[HTML]{7ABD00}} \color[HTML]{F1F1F1} 0.736 & {\cellcolor[HTML]{6CB600}} \color[HTML]{F1F1F1} 0.750 & {\cellcolor[HTML]{76BB00}} \color[HTML]{F1F1F1} 0.741 & {\cellcolor[HTML]{7ABD00}} \color[HTML]{F1F1F1} 0.736 & {\cellcolor[HTML]{76BB00}} \color[HTML]{F1F1F1} 0.741 & {\cellcolor[HTML]{7ABD00}} \color[HTML]{F1F1F1} 0.736 \\
Full (0k-65k) & {\cellcolor[HTML]{76BB00}} \color[HTML]{F1F1F1} 0.741 & {\cellcolor[HTML]{72B900}} \color[HTML]{F1F1F1} 0.744 & {\cellcolor[HTML]{74BA00}} \color[HTML]{F1F1F1} 0.743 & {\cellcolor[HTML]{70B800}} \color[HTML]{F1F1F1} 0.746 & {\cellcolor[HTML]{76BB00}} \color[HTML]{F1F1F1} 0.741 & {\cellcolor[HTML]{72B900}} \color[HTML]{F1F1F1} 0.743 & {\cellcolor[HTML]{74BA00}} \color[HTML]{F1F1F1} 0.743 & {\cellcolor[HTML]{76BB00}} \color[HTML]{F1F1F1} 0.739 & {\cellcolor[HTML]{74BA00}} \color[HTML]{F1F1F1} 0.743 & {\cellcolor[HTML]{74BA00}} \color[HTML]{F1F1F1} 0.743 & {\cellcolor[HTML]{6EB700}} \color[HTML]{F1F1F1} 0.749 & {\cellcolor[HTML]{72B900}} \color[HTML]{F1F1F1} 0.744 & {\cellcolor[HTML]{72B900}} \color[HTML]{F1F1F1} 0.743 & {\cellcolor[HTML]{72B900}} \color[HTML]{F1F1F1} 0.744 & {\cellcolor[HTML]{74BA00}} \color[HTML]{F1F1F1} 0.742 \\
\bottomrule
\end{tabular}}
\caption{Ablation for all associative layers on the MT dataset for ARMT with Gemma-3-1B-IT model, metric - EM. Ablated associative layers from 0 to 12.}
\label{tab:mt_best_all_layers_assoc_ablation_0_12}
\end{center}
\end{table*}
\begin{table*}[h]

\begin{center}
\resizebox{1.0\textwidth}{!}{
\begin{tabular}{lccccccccccccccc}
\toprule
\begin{tabular}[c]{@{}l@{}}\textbf{Model\//}\\ \textbf{Lengths}\end{tabular} & \begin{tabular}[c]{@{}l@{}}\textbf{Base,}\\ \textbf{MT,}\\ \textbf{8k}\end{tabular} & \begin{tabular}[c]{@{}l@{}}\textbf{ARMT,}\\ \textbf{MT,}\\ \textbf{8k}\end{tabular} & \begin{tabular}[c]{@{}l@{}}\textbf{W/o}\\ \textbf{layer}\\ \textbf{13}\end{tabular} & \begin{tabular}[c]{@{}l@{}}\textbf{W/o}\\ \textbf{layer}\\ \textbf{14}\end{tabular} & \begin{tabular}[c]{@{}l@{}}\textbf{W/o}\\ \textbf{layer}\\ \textbf{15}\end{tabular} & \begin{tabular}[c]{@{}l@{}}\textbf{W/o}\\ \textbf{layer}\\ \textbf{16}\end{tabular} & \begin{tabular}[c]{@{}l@{}}\textbf{W/o}\\ \textbf{layer}\\ \textbf{17}\end{tabular} & \begin{tabular}[c]{@{}l@{}}\textbf{W/o}\\ \textbf{layer}\\ \textbf{18}\end{tabular} & \begin{tabular}[c]{@{}l@{}}\textbf{W/o}\\ \textbf{layer}\\ \textbf{19}\end{tabular} & \begin{tabular}[c]{@{}l@{}}\textbf{W/o}\\ \textbf{layer}\\ \textbf{20}\end{tabular} & \begin{tabular}[c]{@{}l@{}}\textbf{W/o}\\ \textbf{layer}\\ \textbf{21}\end{tabular} & \begin{tabular}[c]{@{}l@{}}\textbf{W/o}\\ \textbf{layer}\\ \textbf{22}\end{tabular} & \begin{tabular}[c]{@{}l@{}}\textbf{W/o}\\ \textbf{layer}\\ \textbf{23}\end{tabular} & \begin{tabular}[c]{@{}l@{}}\textbf{W/o}\\ \textbf{layer}\\ \textbf{24}\end{tabular} & \begin{tabular}[c]{@{}l@{}}\textbf{W/o}\\ \textbf{layer}\\ \textbf{25}\end{tabular}\\
\midrule
0k-1k & {\cellcolor[HTML]{3C9E00}} \color[HTML]{F1F1F1} 0.795 & {\cellcolor[HTML]{5CAE00}} \color[HTML]{F1F1F1} 0.756 & {\cellcolor[HTML]{5CAE00}} \color[HTML]{F1F1F1} 0.756 & {\cellcolor[HTML]{5CAE00}} \color[HTML]{F1F1F1} 0.756 & {\cellcolor[HTML]{5CAE00}} \color[HTML]{F1F1F1} 0.756 & {\cellcolor[HTML]{5CAE00}} \color[HTML]{F1F1F1} 0.756 & {\cellcolor[HTML]{5CAE00}} \color[HTML]{F1F1F1} 0.756 & {\cellcolor[HTML]{5CAE00}} \color[HTML]{F1F1F1} 0.756 & {\cellcolor[HTML]{5CAE00}} \color[HTML]{F1F1F1} 0.756 & {\cellcolor[HTML]{5CAE00}} \color[HTML]{F1F1F1} 0.756 & {\cellcolor[HTML]{5CAE00}} \color[HTML]{F1F1F1} 0.756 & {\cellcolor[HTML]{5CAE00}} \color[HTML]{F1F1F1} 0.756 & {\cellcolor[HTML]{5CAE00}} \color[HTML]{F1F1F1} 0.756 & {\cellcolor[HTML]{5CAE00}} \color[HTML]{F1F1F1} 0.756 & {\cellcolor[HTML]{5CAE00}} \color[HTML]{F1F1F1} 0.756 \\
1k-2k & {\cellcolor[HTML]{40A000}} \color[HTML]{F1F1F1} 0.790 & {\cellcolor[HTML]{68B400}} \color[HTML]{F1F1F1} 0.743 & {\cellcolor[HTML]{70B800}} \color[HTML]{F1F1F1} 0.733 & {\cellcolor[HTML]{D6EB00}} \color[HTML]{000000} 0.610 & {\cellcolor[HTML]{68B400}} \color[HTML]{F1F1F1} 0.743 & {\cellcolor[HTML]{68B400}} \color[HTML]{F1F1F1} 0.743 & {\cellcolor[HTML]{68B400}} \color[HTML]{F1F1F1} 0.743 & {\cellcolor[HTML]{90C800}} \color[HTML]{000000} 0.695 & {\cellcolor[HTML]{90C800}} \color[HTML]{000000} 0.695 & {\cellcolor[HTML]{68B400}} \color[HTML]{F1F1F1} 0.743 & {\cellcolor[HTML]{68B400}} \color[HTML]{F1F1F1} 0.743 & {\cellcolor[HTML]{68B400}} \color[HTML]{F1F1F1} 0.743 & {\cellcolor[HTML]{60B000}} \color[HTML]{F1F1F1} 0.752 & {\cellcolor[HTML]{68B400}} \color[HTML]{F1F1F1} 0.743 & {\cellcolor[HTML]{70B800}} \color[HTML]{F1F1F1} 0.733 \\
2k-4k & {\cellcolor[HTML]{4EA700}} \color[HTML]{F1F1F1} 0.774 & {\cellcolor[HTML]{6EB700}} \color[HTML]{F1F1F1} 0.736 & {\cellcolor[HTML]{6EB700}} \color[HTML]{F1F1F1} 0.736 & {\cellcolor[HTML]{FFC400}} \color[HTML]{000000} 0.491 & {\cellcolor[HTML]{6EB700}} \color[HTML]{F1F1F1} 0.736 & {\cellcolor[HTML]{6EB700}} \color[HTML]{F1F1F1} 0.736 & {\cellcolor[HTML]{6EB700}} \color[HTML]{F1F1F1} 0.736 & {\cellcolor[HTML]{76BB00}} \color[HTML]{F1F1F1} 0.726 & {\cellcolor[HTML]{6EB700}} \color[HTML]{F1F1F1} 0.736 & {\cellcolor[HTML]{6EB700}} \color[HTML]{F1F1F1} 0.736 & {\cellcolor[HTML]{6EB700}} \color[HTML]{F1F1F1} 0.736 & {\cellcolor[HTML]{6EB700}} \color[HTML]{F1F1F1} 0.736 & {\cellcolor[HTML]{76BB00}} \color[HTML]{F1F1F1} 0.726 & {\cellcolor[HTML]{6EB700}} \color[HTML]{F1F1F1} 0.736 & {\cellcolor[HTML]{76BB00}} \color[HTML]{F1F1F1} 0.726 \\
4k-6k & {\cellcolor[HTML]{54AA00}} \color[HTML]{F1F1F1} 0.767 & {\cellcolor[HTML]{8CC600}} \color[HTML]{000000} 0.699 & {\cellcolor[HTML]{8CC600}} \color[HTML]{000000} 0.699 & {\cellcolor[HTML]{FF6A00}} \color[HTML]{F1F1F1} 0.384 & {\cellcolor[HTML]{8CC600}} \color[HTML]{000000} 0.699 & {\cellcolor[HTML]{8CC600}} \color[HTML]{000000} 0.699 & {\cellcolor[HTML]{8CC600}} \color[HTML]{000000} 0.699 & {\cellcolor[HTML]{A4D200}} \color[HTML]{000000} 0.671 & {\cellcolor[HTML]{82C100}} \color[HTML]{000000} 0.712 & {\cellcolor[HTML]{8CC600}} \color[HTML]{000000} 0.699 & {\cellcolor[HTML]{8CC600}} \color[HTML]{000000} 0.699 & {\cellcolor[HTML]{8CC600}} \color[HTML]{000000} 0.699 & {\cellcolor[HTML]{98CC00}} \color[HTML]{000000} 0.685 & {\cellcolor[HTML]{8CC600}} \color[HTML]{000000} 0.699 & {\cellcolor[HTML]{A4D200}} \color[HTML]{000000} 0.671 \\
6k-8k & {\cellcolor[HTML]{068300}} \color[HTML]{F1F1F1} 0.859 & {\cellcolor[HTML]{349A00}} \color[HTML]{F1F1F1} 0.804 & {\cellcolor[HTML]{2C9600}} \color[HTML]{F1F1F1} 0.815 & {\cellcolor[HTML]{FF6000}} \color[HTML]{F1F1F1} 0.370 & {\cellcolor[HTML]{349A00}} \color[HTML]{F1F1F1} 0.804 & {\cellcolor[HTML]{349A00}} \color[HTML]{F1F1F1} 0.804 & {\cellcolor[HTML]{349A00}} \color[HTML]{F1F1F1} 0.804 & {\cellcolor[HTML]{50A800}} \color[HTML]{F1F1F1} 0.772 & {\cellcolor[HTML]{3E9F00}} \color[HTML]{F1F1F1} 0.793 & {\cellcolor[HTML]{349A00}} \color[HTML]{F1F1F1} 0.804 & {\cellcolor[HTML]{349A00}} \color[HTML]{F1F1F1} 0.804 & {\cellcolor[HTML]{349A00}} \color[HTML]{F1F1F1} 0.804 & {\cellcolor[HTML]{3E9F00}} \color[HTML]{F1F1F1} 0.793 & {\cellcolor[HTML]{349A00}} \color[HTML]{F1F1F1} 0.804 & {\cellcolor[HTML]{46A300}} \color[HTML]{F1F1F1} 0.783 \\
8k-10k & {\cellcolor[HTML]{0C8600}} \color[HTML]{F1F1F1} 0.852 & {\cellcolor[HTML]{60B000}} \color[HTML]{F1F1F1} 0.753 & {\cellcolor[HTML]{6AB500}} \color[HTML]{F1F1F1} 0.741 & {\cellcolor[HTML]{FF7E00}} \color[HTML]{F1F1F1} 0.407 & {\cellcolor[HTML]{60B000}} \color[HTML]{F1F1F1} 0.753 & {\cellcolor[HTML]{60B000}} \color[HTML]{F1F1F1} 0.753 & {\cellcolor[HTML]{60B000}} \color[HTML]{F1F1F1} 0.753 & {\cellcolor[HTML]{7EBF00}} \color[HTML]{000000} 0.716 & {\cellcolor[HTML]{56AB00}} \color[HTML]{F1F1F1} 0.765 & {\cellcolor[HTML]{60B000}} \color[HTML]{F1F1F1} 0.753 & {\cellcolor[HTML]{60B000}} \color[HTML]{F1F1F1} 0.753 & {\cellcolor[HTML]{60B000}} \color[HTML]{F1F1F1} 0.753 & {\cellcolor[HTML]{56AB00}} \color[HTML]{F1F1F1} 0.765 & {\cellcolor[HTML]{60B000}} \color[HTML]{F1F1F1} 0.753 & {\cellcolor[HTML]{6AB500}} \color[HTML]{F1F1F1} 0.741 \\
10k-12k & {\cellcolor[HTML]{008000}} \color[HTML]{F1F1F1} 0.868 & {\cellcolor[HTML]{40A000}} \color[HTML]{F1F1F1} 0.791 & {\cellcolor[HTML]{40A000}} \color[HTML]{F1F1F1} 0.791 & {\cellcolor[HTML]{FF3E00}} \color[HTML]{F1F1F1} 0.330 & {\cellcolor[HTML]{40A000}} \color[HTML]{F1F1F1} 0.791 & {\cellcolor[HTML]{40A000}} \color[HTML]{F1F1F1} 0.791 & {\cellcolor[HTML]{40A000}} \color[HTML]{F1F1F1} 0.791 & {\cellcolor[HTML]{6EB700}} \color[HTML]{F1F1F1} 0.736 & {\cellcolor[HTML]{48A400}} \color[HTML]{F1F1F1} 0.780 & {\cellcolor[HTML]{40A000}} \color[HTML]{F1F1F1} 0.791 & {\cellcolor[HTML]{40A000}} \color[HTML]{F1F1F1} 0.791 & {\cellcolor[HTML]{40A000}} \color[HTML]{F1F1F1} 0.791 & {\cellcolor[HTML]{52A900}} \color[HTML]{F1F1F1} 0.769 & {\cellcolor[HTML]{52A900}} \color[HTML]{F1F1F1} 0.769 & {\cellcolor[HTML]{52A900}} \color[HTML]{F1F1F1} 0.769 \\
12k-14k & {\cellcolor[HTML]{42A100}} \color[HTML]{F1F1F1} 0.787 & {\cellcolor[HTML]{66B300}} \color[HTML]{F1F1F1} 0.745 & {\cellcolor[HTML]{6EB700}} \color[HTML]{F1F1F1} 0.734 & {\cellcolor[HTML]{FF0000}} \color[HTML]{F1F1F1} 0.255 & {\cellcolor[HTML]{66B300}} \color[HTML]{F1F1F1} 0.745 & {\cellcolor[HTML]{66B300}} \color[HTML]{F1F1F1} 0.745 & {\cellcolor[HTML]{66B300}} \color[HTML]{F1F1F1} 0.745 & {\cellcolor[HTML]{ACD600}} \color[HTML]{000000} 0.660 & {\cellcolor[HTML]{5EAF00}} \color[HTML]{F1F1F1} 0.755 & {\cellcolor[HTML]{66B300}} \color[HTML]{F1F1F1} 0.745 & {\cellcolor[HTML]{66B300}} \color[HTML]{F1F1F1} 0.745 & {\cellcolor[HTML]{66B300}} \color[HTML]{F1F1F1} 0.745 & {\cellcolor[HTML]{5EAF00}} \color[HTML]{F1F1F1} 0.755 & {\cellcolor[HTML]{5EAF00}} \color[HTML]{F1F1F1} 0.755 & {\cellcolor[HTML]{6EB700}} \color[HTML]{F1F1F1} 0.734 \\
14k-16k & {\cellcolor[HTML]{56AB00}} \color[HTML]{F1F1F1} 0.764 & {\cellcolor[HTML]{98CC00}} \color[HTML]{000000} 0.685 & {\cellcolor[HTML]{98CC00}} \color[HTML]{000000} 0.685 & {\cellcolor[HTML]{FF5600}} \color[HTML]{F1F1F1} 0.360 & {\cellcolor[HTML]{98CC00}} \color[HTML]{000000} 0.685 & {\cellcolor[HTML]{98CC00}} \color[HTML]{000000} 0.685 & {\cellcolor[HTML]{98CC00}} \color[HTML]{000000} 0.685 & {\cellcolor[HTML]{D0E800}} \color[HTML]{000000} 0.618 & {\cellcolor[HTML]{8EC700}} \color[HTML]{000000} 0.697 & {\cellcolor[HTML]{98CC00}} \color[HTML]{000000} 0.685 & {\cellcolor[HTML]{98CC00}} \color[HTML]{000000} 0.685 & {\cellcolor[HTML]{98CC00}} \color[HTML]{000000} 0.685 & {\cellcolor[HTML]{7CBE00}} \color[HTML]{000000} 0.719 & {\cellcolor[HTML]{8EC700}} \color[HTML]{000000} 0.697 & {\cellcolor[HTML]{8EC700}} \color[HTML]{000000} 0.697 \\
16k-24k & {\cellcolor[HTML]{2E9700}} \color[HTML]{F1F1F1} 0.812 & {\cellcolor[HTML]{5CAE00}} \color[HTML]{F1F1F1} 0.757 & {\cellcolor[HTML]{5CAE00}} \color[HTML]{F1F1F1} 0.757 & {\cellcolor[HTML]{FF4000}} \color[HTML]{F1F1F1} 0.333 & {\cellcolor[HTML]{5CAE00}} \color[HTML]{F1F1F1} 0.757 & {\cellcolor[HTML]{5CAE00}} \color[HTML]{F1F1F1} 0.757 & {\cellcolor[HTML]{5CAE00}} \color[HTML]{F1F1F1} 0.757 & {\cellcolor[HTML]{7EBF00}} \color[HTML]{000000} 0.715 & {\cellcolor[HTML]{68B400}} \color[HTML]{F1F1F1} 0.743 & {\cellcolor[HTML]{5CAE00}} \color[HTML]{F1F1F1} 0.757 & {\cellcolor[HTML]{56AB00}} \color[HTML]{F1F1F1} 0.764 & {\cellcolor[HTML]{5CAE00}} \color[HTML]{F1F1F1} 0.757 & {\cellcolor[HTML]{56AB00}} \color[HTML]{F1F1F1} 0.764 & {\cellcolor[HTML]{56AB00}} \color[HTML]{F1F1F1} 0.764 & {\cellcolor[HTML]{62B100}} \color[HTML]{F1F1F1} 0.750 \\
24k-32k & {\cellcolor[HTML]{80C000}} \color[HTML]{000000} 0.713 & {\cellcolor[HTML]{80C000}} \color[HTML]{000000} 0.713 & {\cellcolor[HTML]{88C400}} \color[HTML]{000000} 0.703 & {\cellcolor[HTML]{FF1A00}} \color[HTML]{F1F1F1} 0.287 & {\cellcolor[HTML]{80C000}} \color[HTML]{000000} 0.713 & {\cellcolor[HTML]{80C000}} \color[HTML]{000000} 0.713 & {\cellcolor[HTML]{80C000}} \color[HTML]{000000} 0.713 & {\cellcolor[HTML]{C2E100}} \color[HTML]{000000} 0.634 & {\cellcolor[HTML]{88C400}} \color[HTML]{000000} 0.703 & {\cellcolor[HTML]{80C000}} \color[HTML]{000000} 0.713 & {\cellcolor[HTML]{88C400}} \color[HTML]{000000} 0.703 & {\cellcolor[HTML]{80C000}} \color[HTML]{000000} 0.713 & {\cellcolor[HTML]{88C400}} \color[HTML]{000000} 0.703 & {\cellcolor[HTML]{80C000}} \color[HTML]{000000} 0.713 & {\cellcolor[HTML]{78BC00}} \color[HTML]{F1F1F1} 0.723 \\
32k-49k & {\cellcolor[HTML]{E6F300}} \color[HTML]{000000} 0.591 & {\cellcolor[HTML]{56AB00}} \color[HTML]{F1F1F1} 0.765 & {\cellcolor[HTML]{5CAE00}} \color[HTML]{F1F1F1} 0.757 & {\cellcolor[HTML]{FF1200}} \color[HTML]{F1F1F1} 0.278 & {\cellcolor[HTML]{56AB00}} \color[HTML]{F1F1F1} 0.765 & {\cellcolor[HTML]{56AB00}} \color[HTML]{F1F1F1} 0.765 & {\cellcolor[HTML]{5CAE00}} \color[HTML]{F1F1F1} 0.757 & {\cellcolor[HTML]{80C000}} \color[HTML]{000000} 0.713 & {\cellcolor[HTML]{56AB00}} \color[HTML]{F1F1F1} 0.765 & {\cellcolor[HTML]{56AB00}} \color[HTML]{F1F1F1} 0.765 & {\cellcolor[HTML]{56AB00}} \color[HTML]{F1F1F1} 0.765 & {\cellcolor[HTML]{56AB00}} \color[HTML]{F1F1F1} 0.765 & {\cellcolor[HTML]{56AB00}} \color[HTML]{F1F1F1} 0.765 & {\cellcolor[HTML]{64B200}} \color[HTML]{F1F1F1} 0.748 & {\cellcolor[HTML]{56AB00}} \color[HTML]{F1F1F1} 0.765 \\
49k-65k & {\cellcolor[HTML]{FF3200}} \color[HTML]{F1F1F1} 0.317 & {\cellcolor[HTML]{80C000}} \color[HTML]{000000} 0.713 & {\cellcolor[HTML]{80C000}} \color[HTML]{000000} 0.713 & {\cellcolor[HTML]{FF5400}} \color[HTML]{F1F1F1} 0.356 & {\cellcolor[HTML]{80C000}} \color[HTML]{000000} 0.713 & {\cellcolor[HTML]{80C000}} \color[HTML]{000000} 0.713 & {\cellcolor[HTML]{80C000}} \color[HTML]{000000} 0.713 & {\cellcolor[HTML]{9ACD00}} \color[HTML]{000000} 0.683 & {\cellcolor[HTML]{80C000}} \color[HTML]{000000} 0.713 & {\cellcolor[HTML]{80C000}} \color[HTML]{000000} 0.713 & {\cellcolor[HTML]{80C000}} \color[HTML]{000000} 0.713 & {\cellcolor[HTML]{80C000}} \color[HTML]{000000} 0.713 & {\cellcolor[HTML]{78BC00}} \color[HTML]{F1F1F1} 0.723 & {\cellcolor[HTML]{80C000}} \color[HTML]{000000} 0.713 & {\cellcolor[HTML]{70B800}} \color[HTML]{F1F1F1} 0.733 \\
\hline
In-Domain (0k-8k) & {\cellcolor[HTML]{3A9D00}} \color[HTML]{F1F1F1} 0.797 & {\cellcolor[HTML]{62B100}} \color[HTML]{F1F1F1} 0.749 & {\cellcolor[HTML]{62B100}} \color[HTML]{F1F1F1} 0.749 & {\cellcolor[HTML]{FFDE00}} \color[HTML]{000000} 0.522 & {\cellcolor[HTML]{62B100}} \color[HTML]{F1F1F1} 0.749 & {\cellcolor[HTML]{62B100}} \color[HTML]{F1F1F1} 0.749 & {\cellcolor[HTML]{62B100}} \color[HTML]{F1F1F1} 0.749 & {\cellcolor[HTML]{76BB00}} \color[HTML]{F1F1F1} 0.724 & {\cellcolor[HTML]{6CB600}} \color[HTML]{F1F1F1} 0.738 & {\cellcolor[HTML]{62B100}} \color[HTML]{F1F1F1} 0.749 & {\cellcolor[HTML]{62B100}} \color[HTML]{F1F1F1} 0.749 & {\cellcolor[HTML]{62B100}} \color[HTML]{F1F1F1} 0.749 & {\cellcolor[HTML]{66B300}} \color[HTML]{F1F1F1} 0.744 & {\cellcolor[HTML]{62B100}} \color[HTML]{F1F1F1} 0.749 & {\cellcolor[HTML]{6EB700}} \color[HTML]{F1F1F1} 0.735 \\
OOD (8k-65k) & {\cellcolor[HTML]{84C200}} \color[HTML]{000000} 0.709 & {\cellcolor[HTML]{68B400}} \color[HTML]{F1F1F1} 0.741 & {\cellcolor[HTML]{6CB600}} \color[HTML]{F1F1F1} 0.737 & {\cellcolor[HTML]{FF3800}} \color[HTML]{F1F1F1} 0.323 & {\cellcolor[HTML]{68B400}} \color[HTML]{F1F1F1} 0.741 & {\cellcolor[HTML]{68B400}} \color[HTML]{F1F1F1} 0.741 & {\cellcolor[HTML]{6AB500}} \color[HTML]{F1F1F1} 0.740 & {\cellcolor[HTML]{96CB00}} \color[HTML]{000000} 0.686 & {\cellcolor[HTML]{6AB500}} \color[HTML]{F1F1F1} 0.740 & {\cellcolor[HTML]{68B400}} \color[HTML]{F1F1F1} 0.741 & {\cellcolor[HTML]{68B400}} \color[HTML]{F1F1F1} 0.741 & {\cellcolor[HTML]{68B400}} \color[HTML]{F1F1F1} 0.741 & {\cellcolor[HTML]{64B200}} \color[HTML]{F1F1F1} 0.746 & {\cellcolor[HTML]{6AB500}} \color[HTML]{F1F1F1} 0.740 & {\cellcolor[HTML]{6AB500}} \color[HTML]{F1F1F1} 0.740 \\
Long-OOD (32k-65k) & {\cellcolor[HTML]{FFAC00}} \color[HTML]{000000} 0.463 & {\cellcolor[HTML]{6AB500}} \color[HTML]{F1F1F1} 0.741 & {\cellcolor[HTML]{6CB600}} \color[HTML]{F1F1F1} 0.736 & {\cellcolor[HTML]{FF3000}} \color[HTML]{F1F1F1} 0.314 & {\cellcolor[HTML]{6AB500}} \color[HTML]{F1F1F1} 0.741 & {\cellcolor[HTML]{6AB500}} \color[HTML]{F1F1F1} 0.741 & {\cellcolor[HTML]{6CB600}} \color[HTML]{F1F1F1} 0.736 & {\cellcolor[HTML]{8CC600}} \color[HTML]{000000} 0.699 & {\cellcolor[HTML]{6AB500}} \color[HTML]{F1F1F1} 0.741 & {\cellcolor[HTML]{6AB500}} \color[HTML]{F1F1F1} 0.741 & {\cellcolor[HTML]{6AB500}} \color[HTML]{F1F1F1} 0.741 & {\cellcolor[HTML]{6AB500}} \color[HTML]{F1F1F1} 0.741 & {\cellcolor[HTML]{66B300}} \color[HTML]{F1F1F1} 0.745 & {\cellcolor[HTML]{70B800}} \color[HTML]{F1F1F1} 0.732 & {\cellcolor[HTML]{62B100}} \color[HTML]{F1F1F1} 0.750 \\
Full (0k-65k) & {\cellcolor[HTML]{6AB500}} \color[HTML]{F1F1F1} 0.741 & {\cellcolor[HTML]{66B300}} \color[HTML]{F1F1F1} 0.744 & {\cellcolor[HTML]{6AB500}} \color[HTML]{F1F1F1} 0.741 & {\cellcolor[HTML]{FF7400}} \color[HTML]{F1F1F1} 0.394 & {\cellcolor[HTML]{66B300}} \color[HTML]{F1F1F1} 0.744 & {\cellcolor[HTML]{66B300}} \color[HTML]{F1F1F1} 0.744 & {\cellcolor[HTML]{68B400}} \color[HTML]{F1F1F1} 0.743 & {\cellcolor[HTML]{8CC600}} \color[HTML]{000000} 0.700 & {\cellcolor[HTML]{6AB500}} \color[HTML]{F1F1F1} 0.739 & {\cellcolor[HTML]{66B300}} \color[HTML]{F1F1F1} 0.744 & {\cellcolor[HTML]{66B300}} \color[HTML]{F1F1F1} 0.744 & {\cellcolor[HTML]{66B300}} \color[HTML]{F1F1F1} 0.744 & {\cellcolor[HTML]{66B300}} \color[HTML]{F1F1F1} 0.745 & {\cellcolor[HTML]{68B400}} \color[HTML]{F1F1F1} 0.743 & {\cellcolor[HTML]{6CB600}} \color[HTML]{F1F1F1} 0.739 \\
\bottomrule
\end{tabular}}
\caption{Ablation for all associative layers on the MT dataset for ARMT with Gemma-3-1B-IT model, metric - EM. Ablated associative layers from 13 to 25.}
\label{tab:mt_best_all_layers_assoc_ablation_13_26}
\end{center}
\end{table*}

We also compared the training dynamics during curriculum learning with top-4 associative blocks and with pre-selected associative blocks, the results are presented in~\Cref{tab:gov_report_best_assoc_ablation_with_train_full,tab:mt_best_assoc_ablation_with_train_full}. The pre-selected blocks show better final performance while not require additional layers ablation from a fully trained model. We suggest that this universal recipe for associative block selection before training can reduce the number of trained parameters almost without loss of performance.

\begin{table*}[t]

\begin{center}
\resizebox{1.\textwidth}{!}{
\begin{tabular}{lccccccccccc}
\toprule
\begin{tabular}[c]{@{}l@{}}\textbf{Model\//}\\ \textbf{Lengths}\end{tabular} & \begin{tabular}[c]{@{}l@{}}\textbf{Base,}\\ \textbf{GR-100+,}\\ \textbf{8k}\end{tabular} & \begin{tabular}[c]{@{}l@{}}\textbf{ARMT,}\\ \textbf{GR-100+,}\\ \textbf{2k}\end{tabular} & \begin{tabular}[c]{@{}l@{}}\textbf{ARMT,}\\ \textbf{GR-100+,}\\ \textbf{4k}\end{tabular} & \begin{tabular}[c]{@{}l@{}}\textbf{ARMT,}\\ \textbf{GR-100+,}\\ \textbf{8k}\end{tabular} &  \begin{tabular}[c]{@{}l@{}}\textbf{ARMT,}\\ \textbf{ GR-100+, 8k,}\\ \textbf{only top-4 layers}\end{tabular} & \begin{tabular}[c]{@{}l@{}}\textbf{ARMT,}\\ \textbf{ GR-100+, 2k,}\\ \textbf{only top-4 layers,} \\\textbf{trained}\end{tabular} & \begin{tabular}[c]{@{}l@{}}\textbf{ARMT,}\\ \textbf{ GR-100+, 4k,}\\ \textbf{only top-4 layers,} \\\textbf{trained}\end{tabular} & \begin{tabular}[c]{@{}l@{}}\textbf{ARMT,}\\ \textbf{ GR-100+, 8k,}\\ \textbf{only top-4 layers,} \\\textbf{trained}\end{tabular} &  \begin{tabular}[c]{@{}l@{}}\textbf{ARMT,}\\ \textbf{ GR-100+, 2k,}\\ \textbf{pre-selected layers,} \\\textbf{trained}\end{tabular} &  \begin{tabular}[c]{@{}l@{}}\textbf{ARMT,}\\ \textbf{ GR-100+, 4k,}\\ \textbf{pre-selected layers,} \\\textbf{trained}\end{tabular} &  \begin{tabular}[c]{@{}l@{}}\textbf{ARMT,}\\ \textbf{ GR-100+, 8k,}\\ \textbf{pre-selected layers,} \\\textbf{trained}\end{tabular} \\
\midrule
0k-1k & {\cellcolor[HTML]{068300}} \color[HTML]{F1F1F1} 0.380 & {\cellcolor[HTML]{249200}} \color[HTML]{F1F1F1} 0.357 & {\cellcolor[HTML]{249200}} \color[HTML]{F1F1F1} 0.357 & {\cellcolor[HTML]{229100}} \color[HTML]{F1F1F1} 0.358 & {\cellcolor[HTML]{229100}} \color[HTML]{F1F1F1} 0.358 & {\cellcolor[HTML]{1E8F00}} \color[HTML]{F1F1F1} 0.361 & {\cellcolor[HTML]{188C00}} \color[HTML]{F1F1F1} 0.366 & {\cellcolor[HTML]{329900}} \color[HTML]{F1F1F1} 0.347 & {\cellcolor[HTML]{2C9600}} \color[HTML]{F1F1F1} 0.351 & {\cellcolor[HTML]{2A9500}} \color[HTML]{F1F1F1} 0.353 & {\cellcolor[HTML]{1C8E00}} \color[HTML]{F1F1F1} 0.363 \\
1k-2k & {\cellcolor[HTML]{008000}} \color[HTML]{F1F1F1} 0.385 & {\cellcolor[HTML]{76BB00}} \color[HTML]{F1F1F1} 0.296 & {\cellcolor[HTML]{78BC00}} \color[HTML]{F1F1F1} 0.294 & {\cellcolor[HTML]{6CB600}} \color[HTML]{F1F1F1} 0.303 & {\cellcolor[HTML]{72B900}} \color[HTML]{F1F1F1} 0.299 & {\cellcolor[HTML]{90C800}} \color[HTML]{000000} 0.276 & {\cellcolor[HTML]{84C200}} \color[HTML]{000000} 0.285 & {\cellcolor[HTML]{7CBE00}} \color[HTML]{000000} 0.291 & {\cellcolor[HTML]{7EBF00}} \color[HTML]{000000} 0.290 & {\cellcolor[HTML]{5AAD00}} \color[HTML]{F1F1F1} 0.316 & {\cellcolor[HTML]{5AAD00}} \color[HTML]{F1F1F1} 0.317 \\
2k-4k & {\cellcolor[HTML]{2A9500}} \color[HTML]{F1F1F1} 0.352 & {\cellcolor[HTML]{A0D000}} \color[HTML]{000000} 0.264 & {\cellcolor[HTML]{92C900}} \color[HTML]{000000} 0.275 & {\cellcolor[HTML]{8CC600}} \color[HTML]{000000} 0.279 & {\cellcolor[HTML]{84C200}} \color[HTML]{000000} 0.285 & {\cellcolor[HTML]{B0D800}} \color[HTML]{000000} 0.252 & {\cellcolor[HTML]{9ECF00}} \color[HTML]{000000} 0.265 & {\cellcolor[HTML]{9ECF00}} \color[HTML]{000000} 0.265 & {\cellcolor[HTML]{A2D100}} \color[HTML]{000000} 0.263 & {\cellcolor[HTML]{74BA00}} \color[HTML]{F1F1F1} 0.297 & {\cellcolor[HTML]{6AB500}} \color[HTML]{F1F1F1} 0.304 \\
4k-6k & {\cellcolor[HTML]{369B00}} \color[HTML]{F1F1F1} 0.344 & {\cellcolor[HTML]{C0E000}} \color[HTML]{000000} 0.240 & {\cellcolor[HTML]{82C100}} \color[HTML]{000000} 0.286 & {\cellcolor[HTML]{6CB600}} \color[HTML]{F1F1F1} 0.303 & {\cellcolor[HTML]{8EC700}} \color[HTML]{000000} 0.278 & {\cellcolor[HTML]{DAED00}} \color[HTML]{000000} 0.220 & {\cellcolor[HTML]{A6D300}} \color[HTML]{000000} 0.260 & {\cellcolor[HTML]{A2D100}} \color[HTML]{000000} 0.262 & {\cellcolor[HTML]{C4E200}} \color[HTML]{000000} 0.237 & {\cellcolor[HTML]{8AC500}} \color[HTML]{000000} 0.280 & {\cellcolor[HTML]{82C100}} \color[HTML]{000000} 0.287 \\
6k-8k & {\cellcolor[HTML]{76BB00}} \color[HTML]{F1F1F1} 0.296 & {\cellcolor[HTML]{DCEE00}} \color[HTML]{000000} 0.219 & {\cellcolor[HTML]{C2E100}} \color[HTML]{000000} 0.238 & {\cellcolor[HTML]{AED700}} \color[HTML]{000000} 0.254 & {\cellcolor[HTML]{BADD00}} \color[HTML]{000000} 0.244 & {\cellcolor[HTML]{E6F300}} \color[HTML]{000000} 0.212 & {\cellcolor[HTML]{C2E100}} \color[HTML]{000000} 0.239 & {\cellcolor[HTML]{BEDF00}} \color[HTML]{000000} 0.241 & {\cellcolor[HTML]{FEFF00}} \color[HTML]{000000} 0.194 & {\cellcolor[HTML]{BEDF00}} \color[HTML]{000000} 0.242 & {\cellcolor[HTML]{B2D900}} \color[HTML]{000000} 0.250 \\
8k-10k & {\cellcolor[HTML]{68B400}} \color[HTML]{F1F1F1} 0.306 & {\cellcolor[HTML]{EEF700}} \color[HTML]{000000} 0.205 & {\cellcolor[HTML]{CAE500}} \color[HTML]{000000} 0.233 & {\cellcolor[HTML]{C6E300}} \color[HTML]{000000} 0.235 & {\cellcolor[HTML]{CAE500}} \color[HTML]{000000} 0.233 & {\cellcolor[HTML]{FFF200}} \color[HTML]{000000} 0.182 & {\cellcolor[HTML]{CEE700}} \color[HTML]{000000} 0.229 & {\cellcolor[HTML]{CAE500}} \color[HTML]{000000} 0.232 & {\cellcolor[HTML]{FFA600}} \color[HTML]{000000} 0.125 & {\cellcolor[HTML]{C8E400}} \color[HTML]{000000} 0.234 & {\cellcolor[HTML]{CCE600}} \color[HTML]{000000} 0.231 \\
10k-12k & {\cellcolor[HTML]{9ACD00}} \color[HTML]{000000} 0.269 & {\cellcolor[HTML]{E4F200}} \color[HTML]{000000} 0.213 & {\cellcolor[HTML]{DAED00}} \color[HTML]{000000} 0.221 & {\cellcolor[HTML]{BEDF00}} \color[HTML]{000000} 0.241 & {\cellcolor[HTML]{BEDF00}} \color[HTML]{000000} 0.241 & {\cellcolor[HTML]{FFFA00}} \color[HTML]{000000} 0.188 & {\cellcolor[HTML]{E0F000}} \color[HTML]{000000} 0.216 & {\cellcolor[HTML]{CAE500}} \color[HTML]{000000} 0.232 & {\cellcolor[HTML]{FF8E00}} \color[HTML]{F1F1F1} 0.108 & {\cellcolor[HTML]{D4EA00}} \color[HTML]{000000} 0.225 & {\cellcolor[HTML]{CCE600}} \color[HTML]{000000} 0.231 \\
12k-14k & {\cellcolor[HTML]{8AC500}} \color[HTML]{000000} 0.280 & {\cellcolor[HTML]{FEFF00}} \color[HTML]{000000} 0.193 & {\cellcolor[HTML]{D4EA00}} \color[HTML]{000000} 0.225 & {\cellcolor[HTML]{C4E200}} \color[HTML]{000000} 0.237 & {\cellcolor[HTML]{CEE700}} \color[HTML]{000000} 0.230 & {\cellcolor[HTML]{FFEC00}} \color[HTML]{000000} 0.178 & {\cellcolor[HTML]{D2E900}} \color[HTML]{000000} 0.226 & {\cellcolor[HTML]{D8EC00}} \color[HTML]{000000} 0.222 & {\cellcolor[HTML]{FF5800}} \color[HTML]{F1F1F1} 0.067 & {\cellcolor[HTML]{EEF700}} \color[HTML]{000000} 0.205 & {\cellcolor[HTML]{DAED00}} \color[HTML]{000000} 0.221 \\
14k-16k & {\cellcolor[HTML]{CCE600}} \color[HTML]{000000} 0.231 & {\cellcolor[HTML]{FFEE00}} \color[HTML]{000000} 0.180 & {\cellcolor[HTML]{DEEF00}} \color[HTML]{000000} 0.217 & {\cellcolor[HTML]{C0E000}} \color[HTML]{000000} 0.240 & {\cellcolor[HTML]{C4E200}} \color[HTML]{000000} 0.237 & {\cellcolor[HTML]{FFE600}} \color[HTML]{000000} 0.174 & {\cellcolor[HTML]{DEEF00}} \color[HTML]{000000} 0.218 & {\cellcolor[HTML]{CAE500}} \color[HTML]{000000} 0.233 & {\cellcolor[HTML]{FF4E00}} \color[HTML]{F1F1F1} 0.060 & {\cellcolor[HTML]{EAF500}} \color[HTML]{000000} 0.208 & {\cellcolor[HTML]{BCDE00}} \color[HTML]{000000} 0.243 \\
16k-24k & {\cellcolor[HTML]{9ACD00}} \color[HTML]{000000} 0.269 & {\cellcolor[HTML]{FF9800}} \color[HTML]{000000} 0.115 & {\cellcolor[HTML]{F6FB00}} \color[HTML]{000000} 0.200 & {\cellcolor[HTML]{B0D800}} \color[HTML]{000000} 0.252 & {\cellcolor[HTML]{B6DB00}} \color[HTML]{000000} 0.247 & {\cellcolor[HTML]{FFEE00}} \color[HTML]{000000} 0.179 & {\cellcolor[HTML]{CAE500}} \color[HTML]{000000} 0.232 & {\cellcolor[HTML]{BEDF00}} \color[HTML]{000000} 0.242 & {\cellcolor[HTML]{FF5200}} \color[HTML]{F1F1F1} 0.062 & {\cellcolor[HTML]{FFE000}} \color[HTML]{000000} 0.169 & {\cellcolor[HTML]{A6D300}} \color[HTML]{000000} 0.259 \\
24k-32k & {\cellcolor[HTML]{B6DB00}} \color[HTML]{000000} 0.247 & {\cellcolor[HTML]{FF1400}} \color[HTML]{F1F1F1} 0.016 & {\cellcolor[HTML]{FFDE00}} \color[HTML]{000000} 0.168 & {\cellcolor[HTML]{ECF600}} \color[HTML]{000000} 0.207 & {\cellcolor[HTML]{DEEF00}} \color[HTML]{000000} 0.217 & {\cellcolor[HTML]{FFD400}} \color[HTML]{000000} 0.160 & {\cellcolor[HTML]{FFFE00}} \color[HTML]{000000} 0.192 & {\cellcolor[HTML]{FFF800}} \color[HTML]{000000} 0.187 & {\cellcolor[HTML]{FF4400}} \color[HTML]{F1F1F1} 0.052 & {\cellcolor[HTML]{FF8C00}} \color[HTML]{F1F1F1} 0.106 & {\cellcolor[HTML]{F0F800}} \color[HTML]{000000} 0.204 \\
32k-49k & {\cellcolor[HTML]{FFE200}} \color[HTML]{000000} 0.171 & {\cellcolor[HTML]{FF0200}} \color[HTML]{F1F1F1} 0.002 & {\cellcolor[HTML]{FFAA00}} \color[HTML]{000000} 0.128 & {\cellcolor[HTML]{F6FB00}} \color[HTML]{000000} 0.200 & {\cellcolor[HTML]{CEE700}} \color[HTML]{000000} 0.229 & {\cellcolor[HTML]{FFB600}} \color[HTML]{000000} 0.138 & {\cellcolor[HTML]{FFF200}} \color[HTML]{000000} 0.183 & {\cellcolor[HTML]{EEF700}} \color[HTML]{000000} 0.206 & {\cellcolor[HTML]{FF2000}} \color[HTML]{F1F1F1} 0.025 & {\cellcolor[HTML]{FF7400}} \color[HTML]{F1F1F1} 0.088 & {\cellcolor[HTML]{EEF700}} \color[HTML]{000000} 0.205 \\
49k-65k & {\cellcolor[HTML]{FFFA00}} \color[HTML]{000000} 0.189 & {\cellcolor[HTML]{FF0000}} \color[HTML]{F1F1F1} 0.000 & {\cellcolor[HTML]{FF3400}} \color[HTML]{F1F1F1} 0.040 & {\cellcolor[HTML]{A0D000}} \color[HTML]{000000} 0.264 & {\cellcolor[HTML]{46A300}} \color[HTML]{F1F1F1} 0.331 & {\cellcolor[HTML]{E8F400}} \color[HTML]{000000} 0.210 & {\cellcolor[HTML]{80C000}} \color[HTML]{000000} 0.288 & {\cellcolor[HTML]{3E9F00}} \color[HTML]{F1F1F1} 0.338 & {\cellcolor[HTML]{FF1C00}} \color[HTML]{F1F1F1} 0.022 & {\cellcolor[HTML]{FF4000}} \color[HTML]{F1F1F1} 0.049 & {\cellcolor[HTML]{B8DC00}} \color[HTML]{000000} 0.246 \\
\hline
\begin{tabular}[c]{@{}l@{}}{In-Domain}\\ {(0k-8k)}\end{tabular} & {\cellcolor[HTML]{2C9600}} \color[HTML]{F1F1F1} 0.351 & {\cellcolor[HTML]{92C900}} \color[HTML]{000000} 0.275 & {\cellcolor[HTML]{7EBF00}} \color[HTML]{000000} 0.290 & {\cellcolor[HTML]{72B900}} \color[HTML]{F1F1F1} 0.299 & {\cellcolor[HTML]{7ABD00}} \color[HTML]{F1F1F1} 0.293 & {\cellcolor[HTML]{A0D000}} \color[HTML]{000000} 0.264 & {\cellcolor[HTML]{88C400}} \color[HTML]{000000} 0.283 & {\cellcolor[HTML]{8AC500}} \color[HTML]{000000} 0.281 & {\cellcolor[HTML]{9CCE00}} \color[HTML]{000000} 0.267 & {\cellcolor[HTML]{74BA00}} \color[HTML]{F1F1F1} 0.297 & {\cellcolor[HTML]{6AB500}} \color[HTML]{F1F1F1} 0.304 \\
\begin{tabular}[c]{@{}l@{}}{OOD}\\ {(8k-65k)}\end{tabular} & {\cellcolor[HTML]{9ECF00}} \color[HTML]{000000} 0.266 & {\cellcolor[HTML]{FFD600}} \color[HTML]{000000} 0.162 & {\cellcolor[HTML]{E6F300}} \color[HTML]{000000} 0.211 & {\cellcolor[HTML]{C2E100}} \color[HTML]{000000} 0.238 & {\cellcolor[HTML]{C4E200}} \color[HTML]{000000} 0.237 & {\cellcolor[HTML]{FFEC00}} \color[HTML]{000000} 0.178 & {\cellcolor[HTML]{D8EC00}} \color[HTML]{000000} 0.222 & {\cellcolor[HTML]{CEE700}} \color[HTML]{000000} 0.229 & {\cellcolor[HTML]{FF6800}} \color[HTML]{F1F1F1} 0.079 & {\cellcolor[HTML]{FCFE00}} \color[HTML]{000000} 0.195 & {\cellcolor[HTML]{C8E400}} \color[HTML]{000000} 0.234 \\
\begin{tabular}[c]{@{}l@{}}{Long-OOD}\\ {(32k-65k)}\end{tabular} & {\cellcolor[HTML]{FFE800}} \color[HTML]{000000} 0.175 & {\cellcolor[HTML]{FF0200}} \color[HTML]{F1F1F1} 0.002 & {\cellcolor[HTML]{FF8E00}} \color[HTML]{F1F1F1} 0.108 & {\cellcolor[HTML]{E2F100}} \color[HTML]{000000} 0.215 & {\cellcolor[HTML]{B0D800}} \color[HTML]{000000} 0.253 & {\cellcolor[HTML]{FFCC00}} \color[HTML]{000000} 0.155 & {\cellcolor[HTML]{ECF600}} \color[HTML]{000000} 0.207 & {\cellcolor[HTML]{C4E200}} \color[HTML]{000000} 0.236 & {\cellcolor[HTML]{FF2000}} \color[HTML]{F1F1F1} 0.024 & {\cellcolor[HTML]{FF6800}} \color[HTML]{F1F1F1} 0.079 & {\cellcolor[HTML]{E2F100}} \color[HTML]{000000} 0.214 \\
\begin{tabular}[c]{@{}l@{}}{Full}\\ {(0k-65k)}\end{tabular} & {\cellcolor[HTML]{68B400}} \color[HTML]{F1F1F1} 0.306 & {\cellcolor[HTML]{E2F100}} \color[HTML]{000000} 0.215 & {\cellcolor[HTML]{B6DB00}} \color[HTML]{000000} 0.248 & {\cellcolor[HTML]{9CCE00}} \color[HTML]{000000} 0.266 & {\cellcolor[HTML]{A2D100}} \color[HTML]{000000} 0.263 & {\cellcolor[HTML]{DEEF00}} \color[HTML]{000000} 0.218 & {\cellcolor[HTML]{B2D900}} \color[HTML]{000000} 0.250 & {\cellcolor[HTML]{AED700}} \color[HTML]{000000} 0.254 & {\cellcolor[HTML]{FFDC00}} \color[HTML]{000000} 0.167 & {\cellcolor[HTML]{BCDE00}} \color[HTML]{000000} 0.243 & {\cellcolor[HTML]{9CCE00}} \color[HTML]{000000} 0.267 \\
\bottomrule

\end{tabular}}
\caption{Associative layers ablation on the GR-100+ dataset for Gemma-3-1B-IT model, metric - ROUGE-L. ARMT with only top-4 associative blocks keeps almost the same performance as the full ARMT even without training. Model with 5 associative blocks (approximately 20\%) achieves the same performance as the full ARMT model.}
\label{tab:gov_report_best_assoc_ablation_with_train_full}
\end{center}
\end{table*}
\begin{table*}[t]

\begin{center}
\resizebox{1.\textwidth}{!}{
\begin{tabular}{lccccccccccc}
\toprule
\begin{tabular}[c]{@{}l@{}}\textbf{Model\//}\\ \textbf{Lengths}\end{tabular} & \begin{tabular}[c]{@{}l@{}}\textbf{Base,}\\ \textbf{MT, 8k}\end{tabular} & \begin{tabular}[c]{@{}l@{}}\textbf{ARMT,}\\ \textbf{MT, 2k}\end{tabular} & \begin{tabular}[c]{@{}l@{}}\textbf{ARMT,}\\ \textbf{MT, 4k}\end{tabular} & \begin{tabular}[c]{@{}l@{}}\textbf{ARMT,}\\ \textbf{MT, 8k}\end{tabular} &  \begin{tabular}[c]{@{}l@{}}\textbf{ARMT,}\\ \textbf{ MT, 8k,}\\ \textbf{only top-4 layers}\end{tabular} & \begin{tabular}[c]{@{}l@{}}\textbf{ARMT, MT, 2k,}\\ \textbf{only top-4 layers,} \\\textbf{trained}\end{tabular} & \begin{tabular}[c]{@{}l@{}}\textbf{ARMT, MT, 4k,}\\ \textbf{only top-4 layers,} \\\textbf{trained}\end{tabular} & \begin{tabular}[c]{@{}l@{}}\textbf{ARMT, MT, 8k,}\\ \textbf{only top-4 layers,} \\\textbf{trained}\end{tabular} &  \begin{tabular}[c]{@{}l@{}}\textbf{ARMT, MT, 2k,}\\ \textbf{pre-selected layers,} \\\textbf{trained}\end{tabular} &  \begin{tabular}[c]{@{}l@{}}\textbf{ARMT, MT, 4k,}\\ \textbf{pre-selected layers,} \\\textbf{trained}\end{tabular} &  \begin{tabular}[c]{@{}l@{}}\textbf{ARMT, MT, 8k,}\\ \textbf{pre-selected layers,} \\\textbf{trained}\end{tabular} \\
\midrule
0k-1k & {\cellcolor[HTML]{2A9500}} \color[HTML]{F1F1F1} 0.795 & {\cellcolor[HTML]{329900}} \color[HTML]{F1F1F1} 0.782 & {\cellcolor[HTML]{329900}} \color[HTML]{F1F1F1} 0.782 & {\cellcolor[HTML]{42A100}} \color[HTML]{F1F1F1} 0.756 & {\cellcolor[HTML]{42A100}} \color[HTML]{F1F1F1} 0.756 & {\cellcolor[HTML]{329900}} \color[HTML]{F1F1F1} 0.782 & {\cellcolor[HTML]{329900}} \color[HTML]{F1F1F1} 0.782 & {\cellcolor[HTML]{42A100}} \color[HTML]{F1F1F1} 0.756 & {\cellcolor[HTML]{329900}} \color[HTML]{F1F1F1} 0.782 & {\cellcolor[HTML]{42A100}} \color[HTML]{F1F1F1} 0.756 & {\cellcolor[HTML]{42A100}} \color[HTML]{F1F1F1} 0.756 \\
1k-2k & {\cellcolor[HTML]{2E9700}} \color[HTML]{F1F1F1} 0.790 & {\cellcolor[HTML]{54AA00}} \color[HTML]{F1F1F1} 0.724 & {\cellcolor[HTML]{5AAD00}} \color[HTML]{F1F1F1} 0.714 & {\cellcolor[HTML]{48A400}} \color[HTML]{F1F1F1} 0.743 & {\cellcolor[HTML]{54AA00}} \color[HTML]{F1F1F1} 0.724 & {\cellcolor[HTML]{5AAD00}} \color[HTML]{F1F1F1} 0.714 & {\cellcolor[HTML]{6AB500}} \color[HTML]{F1F1F1} 0.686 & {\cellcolor[HTML]{5AAD00}} \color[HTML]{F1F1F1} 0.714 & {\cellcolor[HTML]{54AA00}} \color[HTML]{F1F1F1} 0.724 & {\cellcolor[HTML]{60B000}} \color[HTML]{F1F1F1} 0.705 & {\cellcolor[HTML]{60B000}} \color[HTML]{F1F1F1} 0.705 \\
2k-4k & {\cellcolor[HTML]{369B00}} \color[HTML]{F1F1F1} 0.774 & {\cellcolor[HTML]{68B400}} \color[HTML]{F1F1F1} 0.689 & {\cellcolor[HTML]{64B200}} \color[HTML]{F1F1F1} 0.698 & {\cellcolor[HTML]{4CA600}} \color[HTML]{F1F1F1} 0.736 & {\cellcolor[HTML]{5EAF00}} \color[HTML]{F1F1F1} 0.708 & {\cellcolor[HTML]{7ABD00}} \color[HTML]{F1F1F1} 0.660 & {\cellcolor[HTML]{58AC00}} \color[HTML]{F1F1F1} 0.717 & {\cellcolor[HTML]{4CA600}} \color[HTML]{F1F1F1} 0.736 & {\cellcolor[HTML]{6EB700}} \color[HTML]{F1F1F1} 0.679 & {\cellcolor[HTML]{58AC00}} \color[HTML]{F1F1F1} 0.717 & {\cellcolor[HTML]{58AC00}} \color[HTML]{F1F1F1} 0.717 \\
4k-6k & {\cellcolor[HTML]{3A9D00}} \color[HTML]{F1F1F1} 0.767 & {\cellcolor[HTML]{94CA00}} \color[HTML]{000000} 0.616 & {\cellcolor[HTML]{84C200}} \color[HTML]{000000} 0.644 & {\cellcolor[HTML]{62B100}} \color[HTML]{F1F1F1} 0.699 & {\cellcolor[HTML]{5CAE00}} \color[HTML]{F1F1F1} 0.712 & {\cellcolor[HTML]{F4FA00}} \color[HTML]{000000} 0.452 & {\cellcolor[HTML]{62B100}} \color[HTML]{F1F1F1} 0.699 & {\cellcolor[HTML]{62B100}} \color[HTML]{F1F1F1} 0.699 & {\cellcolor[HTML]{FFEA00}} \color[HTML]{000000} 0.397 & {\cellcolor[HTML]{62B100}} \color[HTML]{F1F1F1} 0.699 & {\cellcolor[HTML]{5CAE00}} \color[HTML]{F1F1F1} 0.712 \\
6k-8k & {\cellcolor[HTML]{048200}} \color[HTML]{F1F1F1} 0.859 & {\cellcolor[HTML]{98CC00}} \color[HTML]{000000} 0.609 & {\cellcolor[HTML]{58AC00}} \color[HTML]{F1F1F1} 0.717 & {\cellcolor[HTML]{249200}} \color[HTML]{F1F1F1} 0.804 & {\cellcolor[HTML]{2C9600}} \color[HTML]{F1F1F1} 0.793 & {\cellcolor[HTML]{FF6C00}} \color[HTML]{F1F1F1} 0.185 & {\cellcolor[HTML]{2C9600}} \color[HTML]{F1F1F1} 0.793 & {\cellcolor[HTML]{1E8F00}} \color[HTML]{F1F1F1} 0.815 & {\cellcolor[HTML]{FF4C00}} \color[HTML]{F1F1F1} 0.130 & {\cellcolor[HTML]{1E8F00}} \color[HTML]{F1F1F1} 0.815 & {\cellcolor[HTML]{128900}} \color[HTML]{F1F1F1} 0.837 \\
8k-10k & {\cellcolor[HTML]{088400}} \color[HTML]{F1F1F1} 0.852 & {\cellcolor[HTML]{D4EA00}} \color[HTML]{000000} 0.506 & {\cellcolor[HTML]{42A100}} \color[HTML]{F1F1F1} 0.753 & {\cellcolor[HTML]{42A100}} \color[HTML]{F1F1F1} 0.753 & {\cellcolor[HTML]{3C9E00}} \color[HTML]{F1F1F1} 0.765 & {\cellcolor[HTML]{FF3A00}} \color[HTML]{F1F1F1} 0.099 & {\cellcolor[HTML]{2E9700}} \color[HTML]{F1F1F1} 0.790 & {\cellcolor[HTML]{188C00}} \color[HTML]{F1F1F1} 0.827 & {\cellcolor[HTML]{FF0600}} \color[HTML]{F1F1F1} 0.012 & {\cellcolor[HTML]{68B400}} \color[HTML]{F1F1F1} 0.691 & {\cellcolor[HTML]{108800}} \color[HTML]{F1F1F1} 0.840 \\
10k-12k & {\cellcolor[HTML]{008000}} \color[HTML]{F1F1F1} 0.868 & {\cellcolor[HTML]{FFDC00}} \color[HTML]{000000} 0.374 & {\cellcolor[HTML]{74BA00}} \color[HTML]{F1F1F1} 0.670 & {\cellcolor[HTML]{2C9600}} \color[HTML]{F1F1F1} 0.791 & {\cellcolor[HTML]{40A000}} \color[HTML]{F1F1F1} 0.758 & {\cellcolor[HTML]{FF0C00}} \color[HTML]{F1F1F1} 0.022 & {\cellcolor[HTML]{46A300}} \color[HTML]{F1F1F1} 0.747 & {\cellcolor[HTML]{269300}} \color[HTML]{F1F1F1} 0.802 & {\cellcolor[HTML]{FF0000}} \color[HTML]{F1F1F1} 0.000 & {\cellcolor[HTML]{46A300}} \color[HTML]{F1F1F1} 0.747 & {\cellcolor[HTML]{068300}} \color[HTML]{F1F1F1} 0.857 \\
12k-14k & {\cellcolor[HTML]{2E9700}} \color[HTML]{F1F1F1} 0.787 & {\cellcolor[HTML]{FFEE00}} \color[HTML]{000000} 0.404 & {\cellcolor[HTML]{60B000}} \color[HTML]{F1F1F1} 0.702 & {\cellcolor[HTML]{48A400}} \color[HTML]{F1F1F1} 0.745 & {\cellcolor[HTML]{42A100}} \color[HTML]{F1F1F1} 0.755 & {\cellcolor[HTML]{FF1200}} \color[HTML]{F1F1F1} 0.032 & {\cellcolor[HTML]{68B400}} \color[HTML]{F1F1F1} 0.691 & {\cellcolor[HTML]{349A00}} \color[HTML]{F1F1F1} 0.777 & {\cellcolor[HTML]{FF0000}} \color[HTML]{F1F1F1} 0.000 & {\cellcolor[HTML]{3C9E00}} \color[HTML]{F1F1F1} 0.766 & {\cellcolor[HTML]{349A00}} \color[HTML]{F1F1F1} 0.777 \\
14k-16k & {\cellcolor[HTML]{3C9E00}} \color[HTML]{F1F1F1} 0.764 & {\cellcolor[HTML]{FFC600}} \color[HTML]{000000} 0.337 & {\cellcolor[HTML]{86C300}} \color[HTML]{000000} 0.640 & {\cellcolor[HTML]{6AB500}} \color[HTML]{F1F1F1} 0.685 & {\cellcolor[HTML]{78BC00}} \color[HTML]{F1F1F1} 0.663 & {\cellcolor[HTML]{FF0600}} \color[HTML]{F1F1F1} 0.011 & {\cellcolor[HTML]{A6D300}} \color[HTML]{000000} 0.584 & {\cellcolor[HTML]{5EAF00}} \color[HTML]{F1F1F1} 0.708 & {\cellcolor[HTML]{FF0000}} \color[HTML]{F1F1F1} 0.000 & {\cellcolor[HTML]{A6D300}} \color[HTML]{000000} 0.584 & {\cellcolor[HTML]{6AB500}} \color[HTML]{F1F1F1} 0.685 \\
16k-24k & {\cellcolor[HTML]{209000}} \color[HTML]{F1F1F1} 0.812 & {\cellcolor[HTML]{FF8600}} \color[HTML]{F1F1F1} 0.229 & {\cellcolor[HTML]{82C100}} \color[HTML]{000000} 0.646 & {\cellcolor[HTML]{40A000}} \color[HTML]{F1F1F1} 0.757 & {\cellcolor[HTML]{50A800}} \color[HTML]{F1F1F1} 0.729 & {\cellcolor[HTML]{FF0000}} \color[HTML]{F1F1F1} 0.000 & {\cellcolor[HTML]{C0E000}} \color[HTML]{000000} 0.542 & {\cellcolor[HTML]{40A000}} \color[HTML]{F1F1F1} 0.757 & {\cellcolor[HTML]{FF0000}} \color[HTML]{F1F1F1} 0.000 & {\cellcolor[HTML]{9ACD00}} \color[HTML]{000000} 0.604 & {\cellcolor[HTML]{209000}} \color[HTML]{F1F1F1} 0.812 \\
24k-32k & {\cellcolor[HTML]{5AAD00}} \color[HTML]{F1F1F1} 0.713 & {\cellcolor[HTML]{FF7400}} \color[HTML]{F1F1F1} 0.198 & {\cellcolor[HTML]{A6D300}} \color[HTML]{000000} 0.584 & {\cellcolor[HTML]{5AAD00}} \color[HTML]{F1F1F1} 0.713 & {\cellcolor[HTML]{66B300}} \color[HTML]{F1F1F1} 0.693 & {\cellcolor[HTML]{FF0000}} \color[HTML]{F1F1F1} 0.000 & {\cellcolor[HTML]{FFE800}} \color[HTML]{000000} 0.396 & {\cellcolor[HTML]{3E9F00}} \color[HTML]{F1F1F1} 0.762 & {\cellcolor[HTML]{FF0000}} \color[HTML]{F1F1F1} 0.000 & {\cellcolor[HTML]{FEFF00}} \color[HTML]{000000} 0.436 & {\cellcolor[HTML]{3E9F00}} \color[HTML]{F1F1F1} 0.762 \\
32k-49k & {\cellcolor[HTML]{A2D100}} \color[HTML]{000000} 0.591 & {\cellcolor[HTML]{FF5000}} \color[HTML]{F1F1F1} 0.139 & {\cellcolor[HTML]{98CC00}} \color[HTML]{000000} 0.609 & {\cellcolor[HTML]{3C9E00}} \color[HTML]{F1F1F1} 0.765 & {\cellcolor[HTML]{3C9E00}} \color[HTML]{F1F1F1} 0.765 & {\cellcolor[HTML]{FF0000}} \color[HTML]{F1F1F1} 0.000 & {\cellcolor[HTML]{FF9800}} \color[HTML]{000000} 0.261 & {\cellcolor[HTML]{3C9E00}} \color[HTML]{F1F1F1} 0.765 & {\cellcolor[HTML]{FF0000}} \color[HTML]{F1F1F1} 0.000 & {\cellcolor[HTML]{FEFF00}} \color[HTML]{000000} 0.435 & {\cellcolor[HTML]{289400}} \color[HTML]{F1F1F1} 0.800 \\
49k-65k & {\cellcolor[HTML]{FFBA00}} \color[HTML]{000000} 0.317 & {\cellcolor[HTML]{FF3400}} \color[HTML]{F1F1F1} 0.089 & {\cellcolor[HTML]{9ACD00}} \color[HTML]{000000} 0.604 & {\cellcolor[HTML]{5AAD00}} \color[HTML]{F1F1F1} 0.713 & {\cellcolor[HTML]{5AAD00}} \color[HTML]{F1F1F1} 0.713 & {\cellcolor[HTML]{FF0000}} \color[HTML]{F1F1F1} 0.000 & {\cellcolor[HTML]{FFA200}} \color[HTML]{000000} 0.277 & {\cellcolor[HTML]{44A200}} \color[HTML]{F1F1F1} 0.752 & {\cellcolor[HTML]{FF0000}} \color[HTML]{F1F1F1} 0.000 & {\cellcolor[HTML]{E0F000}} \color[HTML]{000000} 0.485 & {\cellcolor[HTML]{269300}} \color[HTML]{F1F1F1} 0.802 \\
\hline
\begin{tabular}[c]{@{}l@{}}{In-Domain}\\ {(0k-8k)}\end{tabular} & {\cellcolor[HTML]{289400}} \color[HTML]{F1F1F1} 0.797 & {\cellcolor[HTML]{6AB500}} \color[HTML]{F1F1F1} 0.685 & {\cellcolor[HTML]{5CAE00}} \color[HTML]{F1F1F1} 0.711 & {\cellcolor[HTML]{46A300}} \color[HTML]{F1F1F1} 0.749 & {\cellcolor[HTML]{4CA600}} \color[HTML]{F1F1F1} 0.738 & {\cellcolor[HTML]{B2D900}} \color[HTML]{000000} 0.564 & {\cellcolor[HTML]{4EA700}} \color[HTML]{F1F1F1} 0.734 & {\cellcolor[HTML]{48A400}} \color[HTML]{F1F1F1} 0.744 & {\cellcolor[HTML]{BADD00}} \color[HTML]{000000} 0.551 & {\cellcolor[HTML]{4CA600}} \color[HTML]{F1F1F1} 0.738 & {\cellcolor[HTML]{48A400}} \color[HTML]{F1F1F1} 0.744 \\
\begin{tabular}[c]{@{}l@{}}{OOD}\\ {(8k-65k)}\end{tabular} & {\cellcolor[HTML]{5CAE00}} \color[HTML]{F1F1F1} 0.709 & {\cellcolor[HTML]{FF9E00}} \color[HTML]{000000} 0.271 & {\cellcolor[HTML]{82C100}} \color[HTML]{000000} 0.647 & {\cellcolor[HTML]{4AA500}} \color[HTML]{F1F1F1} 0.741 & {\cellcolor[HTML]{50A800}} \color[HTML]{F1F1F1} 0.730 & {\cellcolor[HTML]{FF0A00}} \color[HTML]{F1F1F1} 0.017 & {\cellcolor[HTML]{CCE600}} \color[HTML]{000000} 0.521 & {\cellcolor[HTML]{3A9D00}} \color[HTML]{F1F1F1} 0.767 & {\cellcolor[HTML]{FF0000}} \color[HTML]{F1F1F1} 0.001 & {\cellcolor[HTML]{A6D300}} \color[HTML]{000000} 0.586 & {\cellcolor[HTML]{2C9600}} \color[HTML]{F1F1F1} 0.793 \\
\begin{tabular}[c]{@{}l@{}}{Long-OOD}\\ {(32k-65k)}\end{tabular} & {\cellcolor[HTML]{EEF700}} \color[HTML]{000000} 0.463 & {\cellcolor[HTML]{FF4400}} \color[HTML]{F1F1F1} 0.116 & {\cellcolor[HTML]{9ACD00}} \color[HTML]{000000} 0.607 & {\cellcolor[HTML]{4AA500}} \color[HTML]{F1F1F1} 0.741 & {\cellcolor[HTML]{4AA500}} \color[HTML]{F1F1F1} 0.741 & {\cellcolor[HTML]{FF0000}} \color[HTML]{F1F1F1} 0.000 & {\cellcolor[HTML]{FF9E00}} \color[HTML]{000000} 0.268 & {\cellcolor[HTML]{40A000}} \color[HTML]{F1F1F1} 0.759 & {\cellcolor[HTML]{FF0000}} \color[HTML]{F1F1F1} 0.000 & {\cellcolor[HTML]{F0F800}} \color[HTML]{000000} 0.458 & {\cellcolor[HTML]{269300}} \color[HTML]{F1F1F1} 0.801 \\
\begin{tabular}[c]{@{}l@{}}{Full}\\ {(0k-65k)}\end{tabular} & {\cellcolor[HTML]{4AA500}} \color[HTML]{F1F1F1} 0.741 & {\cellcolor[HTML]{FFF600}} \color[HTML]{000000} 0.419 & {\cellcolor[HTML]{74BA00}} \color[HTML]{F1F1F1} 0.670 & {\cellcolor[HTML]{48A400}} \color[HTML]{F1F1F1} 0.744 & {\cellcolor[HTML]{4EA700}} \color[HTML]{F1F1F1} 0.733 & {\cellcolor[HTML]{FF7C00}} \color[HTML]{F1F1F1} 0.213 & {\cellcolor[HTML]{9ECF00}} \color[HTML]{000000} 0.597 & {\cellcolor[HTML]{40A000}} \color[HTML]{F1F1F1} 0.759 & {\cellcolor[HTML]{FF7400}} \color[HTML]{F1F1F1} 0.198 & {\cellcolor[HTML]{86C300}} \color[HTML]{000000} 0.640 & {\cellcolor[HTML]{369B00}} \color[HTML]{F1F1F1} 0.776 \\

\bottomrule

\end{tabular}}
\caption{Associative layers ablation on the MT dataset for Gemma-3-1B-IT model, metric - EM. ARMT with only top-4 associative blocks keeps almost the same performance as the full ARMT even without training. Model with 5 associative blocks (approximately 20\%) achieves the same performance as the full ARMT model.}
\label{tab:mt_best_assoc_ablation_with_train_full}
\end{center}
\end{table*}

\subsection{Ablation Study for Synthetic Training Data Generation}\label{app:gr_ablation}

\begin{figure}[t]
    \centering
    \includegraphics[width=\columnwidth]{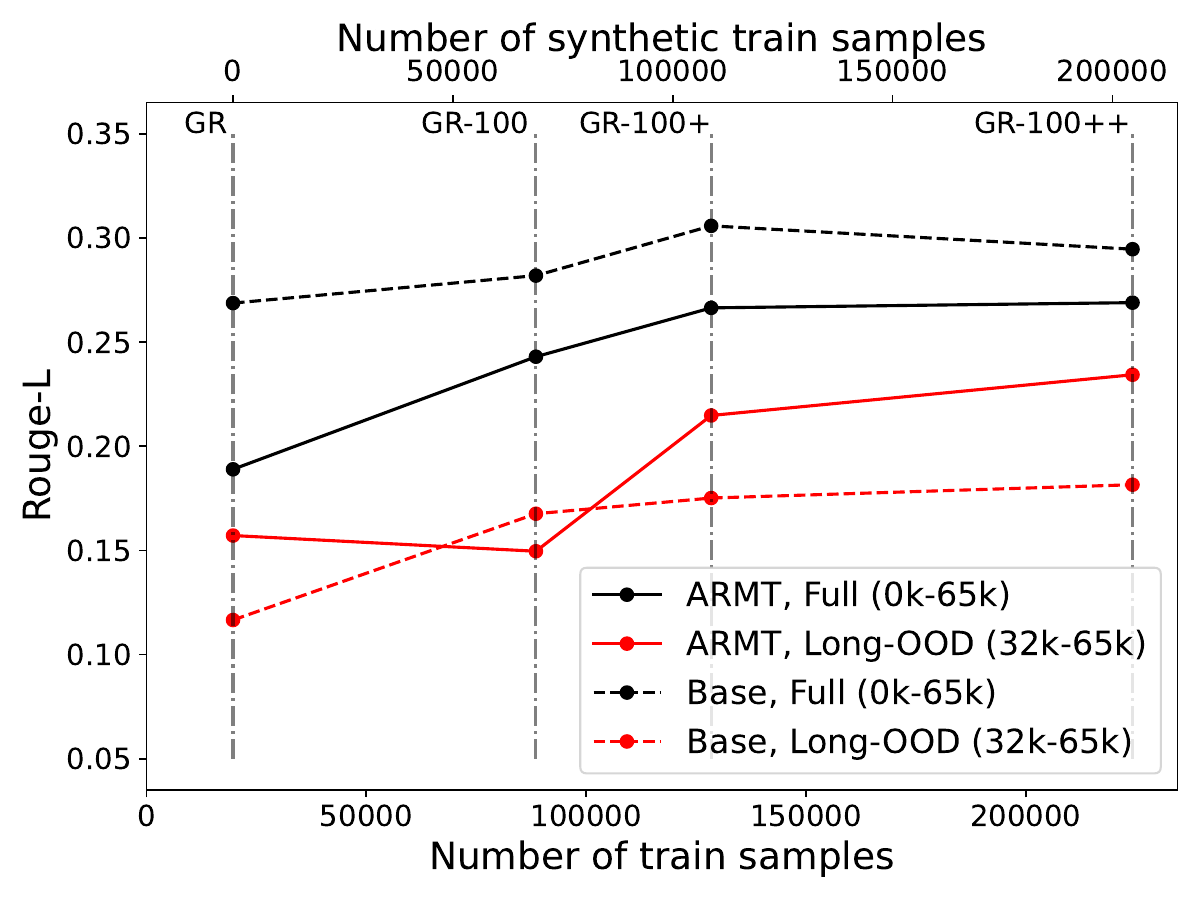}
    \caption{Scaling synthetic data for the GR dataset using ARMT-Gemma-1B-IT (ARMT) and Gemma-1B-IT (Base), trained up to context length 8192. The best performance is achieved with the GR-100+ dataset.}
    \label{fig:synth_scaling}
\end{figure}
As mentioned in the description of the GR dataset, the base version of the GR dataset is too small to finetune the ARMT model from scratch. In this section, we provide additional results with the base GR dataset without additional synthetic data; the results are provided in~\Cref{fig:synth_scaling} and in~\Cref{tab:gov_report_base}.
\begin{table*}[h]

\begin{center}
\footnotesize
\begin{tabular}{lccccc}
\toprule
\begin{tabular}[c]{@{}l@{}}\textbf{Model\//}\\ \textbf{Lengths}\end{tabular} & \begin{tabular}[c]{@{}l@{}}\textbf{Base,}\\ \textbf{No Fine-}\\ \textbf{Tuning}\end{tabular} & \begin{tabular}[c]{@{}l@{}}\textbf{Base,}\\ \textbf{GR, 8k}\end{tabular} & \begin{tabular}[c]{@{}l@{}}\textbf{ARMT,}\\ \textbf{GR, 2k}\end{tabular} & \begin{tabular}[c]{@{}l@{}}\textbf{ARMT,}\\ \textbf{GR, 4k}\end{tabular} & \begin{tabular}[c]{@{}l@{}}\textbf{ARMT,}\\ \textbf{GR, 8k}\end{tabular} \\
\midrule
0k-1k & {\cellcolor[HTML]{FFCE00}} \color[HTML]{000000} 0.188 & {\cellcolor[HTML]{128900}} \color[HTML]{F1F1F1} 0.347 & {\cellcolor[HTML]{209000}} \color[HTML]{F1F1F1} 0.340 & {\cellcolor[HTML]{389C00}} \color[HTML]{F1F1F1} 0.326 & {\cellcolor[HTML]{50A800}} \color[HTML]{F1F1F1} 0.313 \\
1k-2k & {\cellcolor[HTML]{FF9E00}} \color[HTML]{000000} 0.161 & {\cellcolor[HTML]{008000}} \color[HTML]{F1F1F1} 0.358 & {\cellcolor[HTML]{FFB800}} \color[HTML]{000000} 0.175 & {\cellcolor[HTML]{FFE000}} \color[HTML]{000000} 0.198 & {\cellcolor[HTML]{FFF800}} \color[HTML]{000000} 0.211 \\
2k-4k & {\cellcolor[HTML]{FF7800}} \color[HTML]{F1F1F1} 0.140 & {\cellcolor[HTML]{5EAF00}} \color[HTML]{F1F1F1} 0.305 & {\cellcolor[HTML]{FF8C00}} \color[HTML]{F1F1F1} 0.151 & {\cellcolor[HTML]{FFC800}} \color[HTML]{000000} 0.184 & {\cellcolor[HTML]{FFE400}} \color[HTML]{000000} 0.200 \\
4k-6k & {\cellcolor[HTML]{FF4A00}} \color[HTML]{F1F1F1} 0.114 & {\cellcolor[HTML]{7CBE00}} \color[HTML]{000000} 0.288 & {\cellcolor[HTML]{FF7000}} \color[HTML]{F1F1F1} 0.135 & {\cellcolor[HTML]{FFB800}} \color[HTML]{000000} 0.175 & {\cellcolor[HTML]{FFC000}} \color[HTML]{000000} 0.180 \\
6k-8k & {\cellcolor[HTML]{FF4800}} \color[HTML]{F1F1F1} 0.113 & {\cellcolor[HTML]{7ABD00}} \color[HTML]{F1F1F1} 0.289 & {\cellcolor[HTML]{FF8400}} \color[HTML]{F1F1F1} 0.146 & {\cellcolor[HTML]{FFA400}} \color[HTML]{000000} 0.164 & {\cellcolor[HTML]{FFBA00}} \color[HTML]{000000} 0.176 \\
8k-10k & {\cellcolor[HTML]{FF3C00}} \color[HTML]{F1F1F1} 0.106 & {\cellcolor[HTML]{9ACD00}} \color[HTML]{000000} 0.271 & {\cellcolor[HTML]{FF8400}} \color[HTML]{F1F1F1} 0.146 & {\cellcolor[HTML]{FF8E00}} \color[HTML]{F1F1F1} 0.152 & {\cellcolor[HTML]{FFA400}} \color[HTML]{000000} 0.164 \\
10k-12k & {\cellcolor[HTML]{FF3200}} \color[HTML]{F1F1F1} 0.101 & {\cellcolor[HTML]{E0F000}} \color[HTML]{000000} 0.232 & {\cellcolor[HTML]{FF6A00}} \color[HTML]{F1F1F1} 0.132 & {\cellcolor[HTML]{FFA800}} \color[HTML]{000000} 0.166 & {\cellcolor[HTML]{FFBA00}} \color[HTML]{000000} 0.177 \\
12k-14k & {\cellcolor[HTML]{FF2200}} \color[HTML]{F1F1F1} 0.091 & {\cellcolor[HTML]{D0E800}} \color[HTML]{000000} 0.241 & {\cellcolor[HTML]{FF5400}} \color[HTML]{F1F1F1} 0.120 & {\cellcolor[HTML]{FF8600}} \color[HTML]{F1F1F1} 0.147 & {\cellcolor[HTML]{FF9600}} \color[HTML]{000000} 0.156 \\
14k-16k & {\cellcolor[HTML]{FF1C00}} \color[HTML]{F1F1F1} 0.088 & {\cellcolor[HTML]{FAFD00}} \color[HTML]{000000} 0.218 & {\cellcolor[HTML]{FF6400}} \color[HTML]{F1F1F1} 0.128 & {\cellcolor[HTML]{FF8800}} \color[HTML]{F1F1F1} 0.149 & {\cellcolor[HTML]{FFAA00}} \color[HTML]{000000} 0.168 \\
16k-24k & {\cellcolor[HTML]{FF1200}} \color[HTML]{F1F1F1} 0.083 & {\cellcolor[HTML]{F2F900}} \color[HTML]{000000} 0.222 & {\cellcolor[HTML]{FF6400}} \color[HTML]{F1F1F1} 0.128 & {\cellcolor[HTML]{FF8200}} \color[HTML]{F1F1F1} 0.145 & {\cellcolor[HTML]{FFA400}} \color[HTML]{000000} 0.164 \\
24k-32k & {\cellcolor[HTML]{FF0A00}} \color[HTML]{F1F1F1} 0.078 & {\cellcolor[HTML]{FF8800}} \color[HTML]{F1F1F1} 0.149 & {\cellcolor[HTML]{FF4400}} \color[HTML]{F1F1F1} 0.111 & {\cellcolor[HTML]{FF7600}} \color[HTML]{F1F1F1} 0.138 & {\cellcolor[HTML]{FFB800}} \color[HTML]{000000} 0.175 \\
32k-49k & {\cellcolor[HTML]{FF0A00}} \color[HTML]{F1F1F1} 0.078 & {\cellcolor[HTML]{FF6600}} \color[HTML]{F1F1F1} 0.130 & {\cellcolor[HTML]{FF5A00}} \color[HTML]{F1F1F1} 0.123 & {\cellcolor[HTML]{FF7000}} \color[HTML]{F1F1F1} 0.135 & {\cellcolor[HTML]{FF8000}} \color[HTML]{F1F1F1} 0.144 \\
49k-65k & {\cellcolor[HTML]{FF3600}} \color[HTML]{F1F1F1} 0.103 & {\cellcolor[HTML]{FF0000}} \color[HTML]{F1F1F1} 0.072 & {\cellcolor[HTML]{FFBC00}} \color[HTML]{000000} 0.178 & {\cellcolor[HTML]{FFD200}} \color[HTML]{000000} 0.190 & {\cellcolor[HTML]{FFE600}} \color[HTML]{000000} 0.201 \\
\hline
In-Domain (0k-8k) & {\cellcolor[HTML]{FF7E00}} \color[HTML]{F1F1F1} 0.143 & {\cellcolor[HTML]{48A400}} \color[HTML]{F1F1F1} 0.317 & {\cellcolor[HTML]{FFD000}} \color[HTML]{000000} 0.189 & {\cellcolor[HTML]{FFF400}} \color[HTML]{000000} 0.209 & {\cellcolor[HTML]{FEFF00}} \color[HTML]{000000} 0.216 \\
OOD (8k-65k) & {\cellcolor[HTML]{FF2400}} \color[HTML]{F1F1F1} 0.092 & {\cellcolor[HTML]{ECF600}} \color[HTML]{000000} 0.226 & {\cellcolor[HTML]{FF6600}} \color[HTML]{F1F1F1} 0.130 & {\cellcolor[HTML]{FF8C00}} \color[HTML]{F1F1F1} 0.150 & {\cellcolor[HTML]{FFA600}} \color[HTML]{000000} 0.165 \\
Long-OOD (32k-65k) & {\cellcolor[HTML]{FF1400}} \color[HTML]{F1F1F1} 0.084 & {\cellcolor[HTML]{FF4E00}} \color[HTML]{F1F1F1} 0.117 & {\cellcolor[HTML]{FF7200}} \color[HTML]{F1F1F1} 0.136 & {\cellcolor[HTML]{FF8600}} \color[HTML]{F1F1F1} 0.148 & {\cellcolor[HTML]{FF9800}} \color[HTML]{000000} 0.157 \\
Full (0k-65k) & {\cellcolor[HTML]{FF4E00}} \color[HTML]{F1F1F1} 0.116 & {\cellcolor[HTML]{9ECF00}} \color[HTML]{000000} 0.269 & {\cellcolor[HTML]{FF9800}} \color[HTML]{000000} 0.157 & {\cellcolor[HTML]{FFBC00}} \color[HTML]{000000} 0.178 & {\cellcolor[HTML]{FFD000}} \color[HTML]{000000} 0.189 \\
\bottomrule
\end{tabular}
\caption{Best results on the GR dataset for Gemma-3-1B-IT model, metric - ROUGE-L.}
\label{tab:gov_report_base}
\end{center}
\end{table*}
One can notice that the base version of the GR dataset contains enough samples to finetune the Gemma-3-1B-IT model, but not enough to finetune the ARMT model.

\subsection{Ablation Study for SFT}\label{app:pretrain_ablation}
We also conducted additional SFT experiments for the ARMT model. We hypothesize that training with the cold-start weights for associative blocks and memory tokens could limit the performance of the model, which is trained on limited data for fine-tuning. To check this, we created an additional synthetic QA dataset with LLM's generated QA pairs over natural long-context samples. We continuously pre-trained the ARMT model with the Gemma-3-1B-IT backbone and finetuned it on the GR-100+ dataset with the standard curriculum learning setup. Due to limited resources, we pre-train the model only on the two segments of 1024 tokens. The results are presented in~\Cref{tab:gov_report_best_with_synth_pretrain}.
\begin{table*}[h]

\footnotesize
\begin{center}
\begin{tabular}{lccccc}
\toprule
\begin{tabular}[c]{@{}l@{}}\textbf{Model\//}\\ \textbf{Lengths}\end{tabular} & 
\begin{tabular}[c]{@{}l@{}}\textbf{Base,}\\ \textbf{No Fine-}\\ \textbf{Tuning}\end{tabular} & \begin{tabular}[c]{@{}l@{}}\textbf{Base,}\\ \textbf{GR-100+, 8k}\end{tabular} & \begin{tabular}[c]{@{}l@{}}\textbf{ARMT,}\\ \textbf{GR-100+, 8k}\end{tabular} & \begin{tabular}[c]{@{}l@{}}\textbf{ARMT,}\\ \textbf{GR-100+, 8k,}\\\textbf{Pretrained}\end{tabular} \\
\midrule
0k-1k & {\cellcolor[HTML]{FFEC00}} \color[HTML]{000000} 0.188 & {\cellcolor[HTML]{068300}} \color[HTML]{F1F1F1} 0.380 & {\cellcolor[HTML]{249200}} \color[HTML]{F1F1F1} 0.358 & {\cellcolor[HTML]{1E8F00}} \color[HTML]{F1F1F1} 0.363 \\
1k-2k & {\cellcolor[HTML]{FFC600}} \color[HTML]{000000} 0.161 & {\cellcolor[HTML]{008000}} \color[HTML]{F1F1F1} 0.385 & {\cellcolor[HTML]{72B900}} \color[HTML]{F1F1F1} 0.303 & {\cellcolor[HTML]{48A400}} \color[HTML]{F1F1F1} 0.332 \\
2k-4k & {\cellcolor[HTML]{FFAA00}} \color[HTML]{000000} 0.140 & {\cellcolor[HTML]{2E9700}} \color[HTML]{F1F1F1} 0.352 & {\cellcolor[HTML]{92C900}} \color[HTML]{000000} 0.279 & {\cellcolor[HTML]{8EC700}} \color[HTML]{000000} 0.282 \\
4k-6k & {\cellcolor[HTML]{FF8400}} \color[HTML]{F1F1F1} 0.114 & {\cellcolor[HTML]{389C00}} \color[HTML]{F1F1F1} 0.344 & {\cellcolor[HTML]{72B900}} \color[HTML]{F1F1F1} 0.303 & {\cellcolor[HTML]{8CC600}} \color[HTML]{000000} 0.284 \\
6k-8k & {\cellcolor[HTML]{FF8400}} \color[HTML]{F1F1F1} 0.113 & {\cellcolor[HTML]{7CBE00}} \color[HTML]{000000} 0.296 & {\cellcolor[HTML]{B6DB00}} \color[HTML]{000000} 0.254 & {\cellcolor[HTML]{94CA00}} \color[HTML]{000000} 0.278 \\
8k-10k & {\cellcolor[HTML]{FF7A00}} \color[HTML]{F1F1F1} 0.106 & {\cellcolor[HTML]{6EB700}} \color[HTML]{F1F1F1} 0.306 & {\cellcolor[HTML]{D0E800}} \color[HTML]{000000} 0.235 & {\cellcolor[HTML]{C4E200}} \color[HTML]{000000} 0.244 \\
10k-12k & {\cellcolor[HTML]{FF7200}} \color[HTML]{F1F1F1} 0.101 & {\cellcolor[HTML]{A0D000}} \color[HTML]{000000} 0.269 & {\cellcolor[HTML]{C8E400}} \color[HTML]{000000} 0.241 & {\cellcolor[HTML]{BADD00}} \color[HTML]{000000} 0.251 \\
12k-14k & {\cellcolor[HTML]{FF6400}} \color[HTML]{F1F1F1} 0.091 & {\cellcolor[HTML]{92C900}} \color[HTML]{000000} 0.280 & {\cellcolor[HTML]{CEE700}} \color[HTML]{000000} 0.237 & {\cellcolor[HTML]{C6E300}} \color[HTML]{000000} 0.242 \\
14k-16k & {\cellcolor[HTML]{FF6000}} \color[HTML]{F1F1F1} 0.088 & {\cellcolor[HTML]{D6EB00}} \color[HTML]{000000} 0.231 & {\cellcolor[HTML]{CAE500}} \color[HTML]{000000} 0.240 & {\cellcolor[HTML]{B8DC00}} \color[HTML]{000000} 0.252 \\
16k-24k & {\cellcolor[HTML]{FF5A00}} \color[HTML]{F1F1F1} 0.083 & {\cellcolor[HTML]{A0D000}} \color[HTML]{000000} 0.269 & {\cellcolor[HTML]{B8DC00}} \color[HTML]{000000} 0.252 & {\cellcolor[HTML]{BCDE00}} \color[HTML]{000000} 0.249 \\
24k-32k & {\cellcolor[HTML]{FF5200}} \color[HTML]{F1F1F1} 0.078 & {\cellcolor[HTML]{C0E000}} \color[HTML]{000000} 0.247 & {\cellcolor[HTML]{F8FC00}} \color[HTML]{000000} 0.207 & {\cellcolor[HTML]{EEF700}} \color[HTML]{000000} 0.214 \\
32k-49k & {\cellcolor[HTML]{FF5200}} \color[HTML]{F1F1F1} 0.078 & {\cellcolor[HTML]{FFD400}} \color[HTML]{000000} 0.171 & {\cellcolor[HTML]{FFFC00}} \color[HTML]{000000} 0.200 & {\cellcolor[HTML]{FFE600}} \color[HTML]{000000} 0.183 \\
49k-65k & {\cellcolor[HTML]{FF7600}} \color[HTML]{F1F1F1} 0.103 & {\cellcolor[HTML]{FFEE00}} \color[HTML]{000000} 0.189 & {\cellcolor[HTML]{A8D400}} \color[HTML]{000000} 0.264 & {\cellcolor[HTML]{FF0000}} \color[HTML]{F1F1F1} 0.018 \\
\hline
In-Domain (0k-8k) & {\cellcolor[HTML]{FFAE00}} \color[HTML]{000000} 0.143 & {\cellcolor[HTML]{2E9700}} \color[HTML]{F1F1F1} 0.351 & {\cellcolor[HTML]{76BB00}} \color[HTML]{F1F1F1} 0.299 & {\cellcolor[HTML]{6CB600}} \color[HTML]{F1F1F1} 0.307 \\
OOD (8k-65k) & {\cellcolor[HTML]{FF6600}} \color[HTML]{F1F1F1} 0.092 & {\cellcolor[HTML]{A6D300}} \color[HTML]{000000} 0.266 & {\cellcolor[HTML]{CCE600}} \color[HTML]{000000} 0.238 & {\cellcolor[HTML]{C8E400}} \color[HTML]{000000} 0.241 \\
Long-OOD (32k-65k) & {\cellcolor[HTML]{FF5A00}} \color[HTML]{F1F1F1} 0.084 & {\cellcolor[HTML]{FFDA00}} \color[HTML]{000000} 0.175 & {\cellcolor[HTML]{ECF600}} \color[HTML]{000000} 0.215 & {\cellcolor[HTML]{FFB000}} \color[HTML]{000000} 0.145 \\
Full (0k-65k) & {\cellcolor[HTML]{FF8800}} \color[HTML]{F1F1F1} 0.116 & {\cellcolor[HTML]{6EB700}} \color[HTML]{F1F1F1} 0.306 & {\cellcolor[HTML]{A4D200}} \color[HTML]{000000} 0.266 & {\cellcolor[HTML]{9CCE00}} \color[HTML]{000000} 0.272 \\
\bottomrule
\end{tabular}
\caption{Best results on the GovReport-100+ dataset for Gemma-3-1B-IT model, comparison with the pretraining on synthetic QA task, metric - ROUGE-L.}
\label{tab:gov_report_best_with_synth_pretrain}

\end{center}
\end{table*}

While this training setup allows us to achieve higher performance on in-domain lengths compared to the non-pretrained ARMT, the performance on long-context samples is worse. We hypothesize that this performance drop could arise from the limited length of texts in pre-training. 
However, this training setup shows promising results and is left for future research.

\begin{figure*}[ht!]
\centering
\begin{subfigure}{1.\columnwidth}
  \centering
  \includegraphics[width=1\columnwidth]{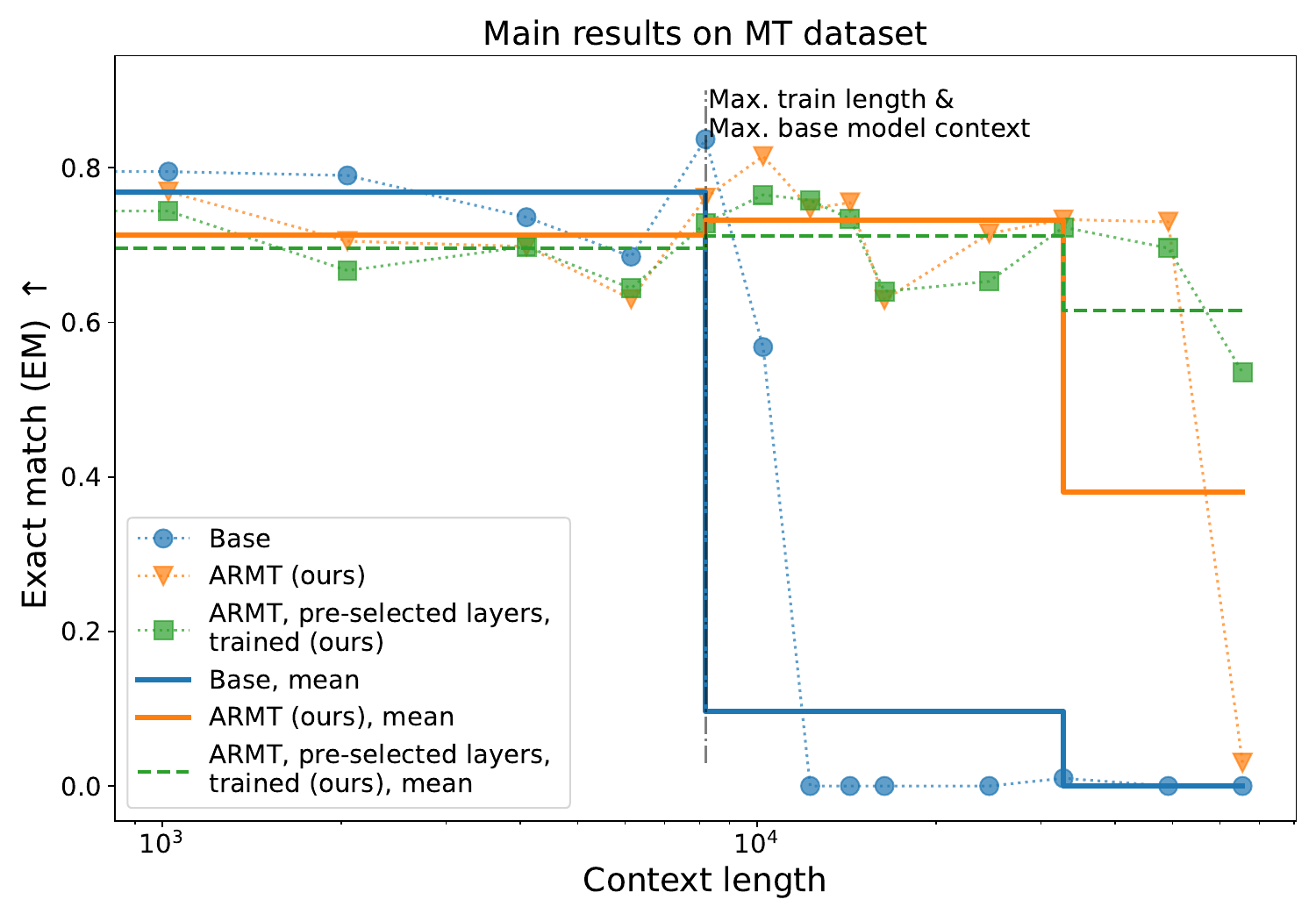}
  \caption{MT results.}
  \label{fig:main_res_mt_smollm}
\end{subfigure}%
\begin{subfigure}{1.\columnwidth}
  \centering
  \includegraphics[width=1.\columnwidth]{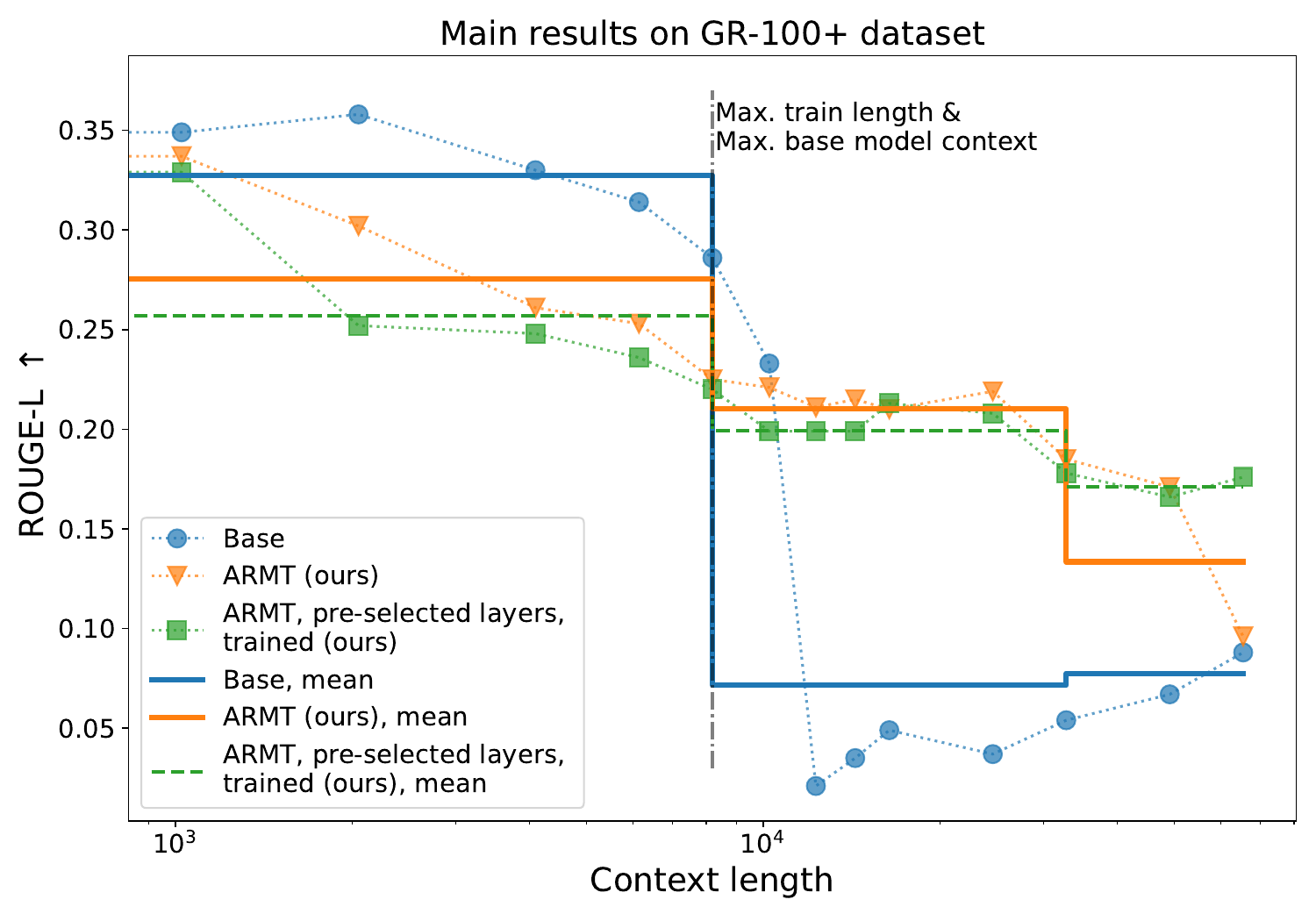}
  \caption{GR-100+ results.}
  \label{fig:main_res_gr100+_smollm}
\end{subfigure}
\caption{Main results with SmolLM-2-360M-IT model. \textbf{(1)} SmolLM-2-360M-IT (Base) with full attention trained up to context length 8192 drops in performance after maximum model length of 8k; \textbf{(2)} ARMT model with SmolLM-2-360M-IT backbone (ARMT) outperforms base model on OOD context length (after 8k); \textbf{(3)} ARMT fine-tuned with only six pre-selected associative layers (approximately 20\%) shows comparable or better performance than ARMT model with all associative layers.}
\label{fig:main_res_smollm}
\end{figure*}

\subsection{Ablation Study for ARMT Backbone}\label{app:smollm_ablation}

\begin{table*}[t]

\begin{center}
\resizebox{0.7\linewidth}{!}{
\begin{tabular}{lccccc}
\toprule
\begin{tabular}[c]{@{}l@{}}\textbf{Model\//}\\ \textbf{Lengths}\end{tabular} & 
\begin{tabular}[c]{@{}l@{}}\textbf{Base,}\\ \textbf{No Fine-}\\ \textbf{Tuning}\end{tabular} & \begin{tabular}[c]{@{}l@{}}\textbf{Base,}\\ \textbf{GR-100+, 8k}\end{tabular} & \begin{tabular}[c]{@{}l@{}}\textbf{ARMT,}\\ \textbf{GR-100+, 2k}\end{tabular} & \begin{tabular}[c]{@{}l@{}}\textbf{ARMT,}\\ \textbf{GR-100+, 4k}\end{tabular} & \begin{tabular}[c]{@{}l@{}}\textbf{ARMT,}\\ \textbf{GR-100+, 8k}\end{tabular} \\
\midrule
0k-1k & {\cellcolor[HTML]{FFC400}} \color[HTML]{000000} 0.145 & {\cellcolor[HTML]{0C8600}} \color[HTML]{F1F1F1} 0.349 & {\cellcolor[HTML]{209000}} \color[HTML]{F1F1F1} 0.335 & {\cellcolor[HTML]{269300}} \color[HTML]{F1F1F1} 0.332 & {\cellcolor[HTML]{1E8F00}} \color[HTML]{F1F1F1} 0.337 \\
1k-2k & {\cellcolor[HTML]{FFA200}} \color[HTML]{000000} 0.122 & {\cellcolor[HTML]{008000}} \color[HTML]{F1F1F1} 0.358 & {\cellcolor[HTML]{64B200}} \color[HTML]{F1F1F1} 0.289 & {\cellcolor[HTML]{5AAD00}} \color[HTML]{F1F1F1} 0.297 & {\cellcolor[HTML]{52A900}} \color[HTML]{F1F1F1} 0.302 \\
2k-4k & {\cellcolor[HTML]{FF9600}} \color[HTML]{000000} 0.113 & {\cellcolor[HTML]{289400}} \color[HTML]{F1F1F1} 0.330 & {\cellcolor[HTML]{B0D800}} \color[HTML]{000000} 0.238 & {\cellcolor[HTML]{94CA00}} \color[HTML]{000000} 0.257 & {\cellcolor[HTML]{8EC700}} \color[HTML]{000000} 0.261 \\
4k-6k & {\cellcolor[HTML]{FF7600}} \color[HTML]{F1F1F1} 0.091 & {\cellcolor[HTML]{40A000}} \color[HTML]{F1F1F1} 0.314 & {\cellcolor[HTML]{CAE500}} \color[HTML]{000000} 0.220 & {\cellcolor[HTML]{A6D300}} \color[HTML]{000000} 0.245 & {\cellcolor[HTML]{9ACD00}} \color[HTML]{000000} 0.253 \\
6k-8k & {\cellcolor[HTML]{FF7A00}} \color[HTML]{F1F1F1} 0.094 & {\cellcolor[HTML]{6AB500}} \color[HTML]{F1F1F1} 0.286 & {\cellcolor[HTML]{D6EB00}} \color[HTML]{000000} 0.212 & {\cellcolor[HTML]{D2E900}} \color[HTML]{000000} 0.215 & {\cellcolor[HTML]{C4E200}} \color[HTML]{000000} 0.225 \\
8k-10k & {\cellcolor[HTML]{FF7200}} \color[HTML]{F1F1F1} 0.089 & {\cellcolor[HTML]{B8DC00}} \color[HTML]{000000} 0.233 & {\cellcolor[HTML]{FFF400}} \color[HTML]{000000} 0.177 & {\cellcolor[HTML]{E6F300}} \color[HTML]{000000} 0.202 & {\cellcolor[HTML]{CAE500}} \color[HTML]{000000} 0.221 \\
10k-12k & {\cellcolor[HTML]{FF2800}} \color[HTML]{F1F1F1} 0.039 & {\cellcolor[HTML]{FF0E00}} \color[HTML]{F1F1F1} 0.021 & {\cellcolor[HTML]{FFD400}} \color[HTML]{000000} 0.156 & {\cellcolor[HTML]{EAF500}} \color[HTML]{000000} 0.199 & {\cellcolor[HTML]{D8EC00}} \color[HTML]{000000} 0.211 \\
12k-14k & {\cellcolor[HTML]{FF2800}} \color[HTML]{F1F1F1} 0.039 & {\cellcolor[HTML]{FF2200}} \color[HTML]{F1F1F1} 0.035 & {\cellcolor[HTML]{FFB800}} \color[HTML]{000000} 0.137 & {\cellcolor[HTML]{F0F800}} \color[HTML]{000000} 0.195 & {\cellcolor[HTML]{D2E900}} \color[HTML]{000000} 0.215 \\
14k-16k & {\cellcolor[HTML]{FF2800}} \color[HTML]{F1F1F1} 0.039 & {\cellcolor[HTML]{FF3800}} \color[HTML]{F1F1F1} 0.049 & {\cellcolor[HTML]{FF5C00}} \color[HTML]{F1F1F1} 0.074 & {\cellcolor[HTML]{F6FB00}} \color[HTML]{000000} 0.190 & {\cellcolor[HTML]{DAED00}} \color[HTML]{000000} 0.210 \\
16k-24k & {\cellcolor[HTML]{FF3600}} \color[HTML]{F1F1F1} 0.048 & {\cellcolor[HTML]{FF2600}} \color[HTML]{F1F1F1} 0.037 & {\cellcolor[HTML]{FF3200}} \color[HTML]{F1F1F1} 0.046 & {\cellcolor[HTML]{FFBE00}} \color[HTML]{000000} 0.140 & {\cellcolor[HTML]{CCE600}} \color[HTML]{000000} 0.219 \\
24k-32k & {\cellcolor[HTML]{FF2C00}} \color[HTML]{F1F1F1} 0.042 & {\cellcolor[HTML]{FF3E00}} \color[HTML]{F1F1F1} 0.054 & {\cellcolor[HTML]{FF0000}} \color[HTML]{F1F1F1} 0.011 & {\cellcolor[HTML]{FF4600}} \color[HTML]{F1F1F1} 0.059 & {\cellcolor[HTML]{FEFF00}} \color[HTML]{000000} 0.185 \\
32k-49k & {\cellcolor[HTML]{FF3000}} \color[HTML]{F1F1F1} 0.044 & {\cellcolor[HTML]{FF5200}} \color[HTML]{F1F1F1} 0.067 & {\cellcolor[HTML]{FF1400}} \color[HTML]{F1F1F1} 0.025 & {\cellcolor[HTML]{FF1E00}} \color[HTML]{F1F1F1} 0.032 & {\cellcolor[HTML]{FFEC00}} \color[HTML]{000000} 0.171 \\
49k-65k & {\cellcolor[HTML]{FF7000}} \color[HTML]{F1F1F1} 0.088 & {\cellcolor[HTML]{FF7000}} \color[HTML]{F1F1F1} 0.088 & {\cellcolor[HTML]{FF0400}} \color[HTML]{F1F1F1} 0.015 & {\cellcolor[HTML]{FF0000}} \color[HTML]{F1F1F1} 0.011 & {\cellcolor[HTML]{FF7C00}} \color[HTML]{F1F1F1} 0.096 \\
\hline
\begin{tabular}[c]{@{}l@{}}{In-Domain}\\ {(0k-8k)}\end{tabular} & {\cellcolor[HTML]{FF9600}} \color[HTML]{000000} 0.113 & {\cellcolor[HTML]{2C9600}} \color[HTML]{F1F1F1} 0.327 & {\cellcolor[HTML]{92C900}} \color[HTML]{000000} 0.258 & {\cellcolor[HTML]{82C100}} \color[HTML]{000000} 0.269 & {\cellcolor[HTML]{7ABD00}} \color[HTML]{F1F1F1} 0.275 \\
\begin{tabular}[c]{@{}l@{}}{OOD}\\ {(8k-65k)}\end{tabular} & {\cellcolor[HTML]{FF3A00}} \color[HTML]{F1F1F1} 0.050 & {\cellcolor[HTML]{FF5C00}} \color[HTML]{F1F1F1} 0.074 & {\cellcolor[HTML]{FF8C00}} \color[HTML]{F1F1F1} 0.106 & {\cellcolor[HTML]{FFEA00}} \color[HTML]{000000} 0.170 & {\cellcolor[HTML]{D8EC00}} \color[HTML]{000000} 0.211 \\
\begin{tabular}[c]{@{}l@{}}{Long-OOD}\\ {(32k-65k)}\end{tabular} & {\cellcolor[HTML]{FF3E00}} \color[HTML]{F1F1F1} 0.054 & {\cellcolor[HTML]{FF5800}} \color[HTML]{F1F1F1} 0.072 & {\cellcolor[HTML]{FF1000}} \color[HTML]{F1F1F1} 0.023 & {\cellcolor[HTML]{FF1600}} \color[HTML]{F1F1F1} 0.027 & {\cellcolor[HTML]{FFD200}} \color[HTML]{000000} 0.154 \\
\begin{tabular}[c]{@{}l@{}}{Full}\\ {(0k-65k)}\end{tabular} & {\cellcolor[HTML]{FF6400}} \color[HTML]{F1F1F1} 0.080 & {\cellcolor[HTML]{F4FA00}} \color[HTML]{000000} 0.192 & {\cellcolor[HTML]{FFF400}} \color[HTML]{000000} 0.178 & {\cellcolor[HTML]{D0E800}} \color[HTML]{000000} 0.216 & {\cellcolor[HTML]{ACD600}} \color[HTML]{000000} 0.241 \\
\bottomrule
\end{tabular}}

\end{center}
\caption{Best results on the GR-100+ dataset for SmolLM-2-360M-IT model, ROUGE-L. ARMT shows higher overall performance than the base model, and outperforms it on OOD and Long-OOD.}
\label{tab:gov_report_best_smollm}
\end{table*}
\begin{table*}[t]

\begin{center}
\resizebox{0.7\linewidth}{!}{
\begin{tabular}{lccccc}
\toprule
\begin{tabular}[c]{@{}l@{}}\textbf{Model\//}\\ \textbf{Lengths}\end{tabular} & \begin{tabular}[c]{@{}l@{}}\textbf{Base,}\\ \textbf{No Fine-}\\ \textbf{Tuning}\end{tabular} & \begin{tabular}[c]{@{}l@{}}\textbf{Base,}\\ \textbf{MT, 8k}\end{tabular} & \begin{tabular}[c]{@{}l@{}}\textbf{ARMT,}\\ \textbf{MT, 2k}\end{tabular} & \begin{tabular}[c]{@{}l@{}}\textbf{ARMT,}\\ \textbf{MT, 4k}\end{tabular} & \begin{tabular}[c]{@{}l@{}}\textbf{ARMT,}\\ \textbf{MT, 8k}\end{tabular} \\
\midrule
0k-1k & {\cellcolor[HTML]{FF0000}} \color[HTML]{F1F1F1} 0.000 & {\cellcolor[HTML]{188C00}} \color[HTML]{F1F1F1} 0.795 & {\cellcolor[HTML]{48A400}} \color[HTML]{F1F1F1} 0.718 & {\cellcolor[HTML]{309800}} \color[HTML]{F1F1F1} 0.756 & {\cellcolor[HTML]{289400}} \color[HTML]{F1F1F1} 0.769 \\
1k-2k & {\cellcolor[HTML]{FF0000}} \color[HTML]{F1F1F1} 0.000 & {\cellcolor[HTML]{1C8E00}} \color[HTML]{F1F1F1} 0.790 & {\cellcolor[HTML]{78BC00}} \color[HTML]{F1F1F1} 0.638 & {\cellcolor[HTML]{5CAE00}} \color[HTML]{F1F1F1} 0.686 & {\cellcolor[HTML]{50A800}} \color[HTML]{F1F1F1} 0.705 \\
2k-4k & {\cellcolor[HTML]{FF0000}} \color[HTML]{F1F1F1} 0.000 & {\cellcolor[HTML]{3C9E00}} \color[HTML]{F1F1F1} 0.736 & {\cellcolor[HTML]{76BB00}} \color[HTML]{F1F1F1} 0.642 & {\cellcolor[HTML]{48A400}} \color[HTML]{F1F1F1} 0.717 & {\cellcolor[HTML]{54AA00}} \color[HTML]{F1F1F1} 0.698 \\
4k-6k & {\cellcolor[HTML]{FF0000}} \color[HTML]{F1F1F1} 0.000 & {\cellcolor[HTML]{5CAE00}} \color[HTML]{F1F1F1} 0.685 & {\cellcolor[HTML]{8EC700}} \color[HTML]{000000} 0.603 & {\cellcolor[HTML]{7EBF00}} \color[HTML]{000000} 0.630 & {\cellcolor[HTML]{7EBF00}} \color[HTML]{000000} 0.630 \\
6k-8k & {\cellcolor[HTML]{FF0000}} \color[HTML]{F1F1F1} 0.000 & {\cellcolor[HTML]{008000}} \color[HTML]{F1F1F1} 0.837 & {\cellcolor[HTML]{92C900}} \color[HTML]{000000} 0.598 & {\cellcolor[HTML]{3A9D00}} \color[HTML]{F1F1F1} 0.739 & {\cellcolor[HTML]{2E9700}} \color[HTML]{F1F1F1} 0.761 \\
8k-10k & {\cellcolor[HTML]{FF0000}} \color[HTML]{F1F1F1} 0.000 & {\cellcolor[HTML]{A4D200}} \color[HTML]{000000} 0.568 & {\cellcolor[HTML]{E0F000}} \color[HTML]{000000} 0.469 & {\cellcolor[HTML]{148A00}} \color[HTML]{F1F1F1} 0.802 & {\cellcolor[HTML]{0C8600}} \color[HTML]{F1F1F1} 0.815 \\
10k-12k & {\cellcolor[HTML]{FF0600}} \color[HTML]{F1F1F1} 0.011 & {\cellcolor[HTML]{FF0000}} \color[HTML]{F1F1F1} 0.000 & {\cellcolor[HTML]{FFFE00}} \color[HTML]{000000} 0.418 & {\cellcolor[HTML]{58AC00}} \color[HTML]{F1F1F1} 0.692 & {\cellcolor[HTML]{369B00}} \color[HTML]{F1F1F1} 0.747 \\
12k-14k & {\cellcolor[HTML]{FF0000}} \color[HTML]{F1F1F1} 0.000 & {\cellcolor[HTML]{FF0000}} \color[HTML]{F1F1F1} 0.000 & {\cellcolor[HTML]{FF5400}} \color[HTML]{F1F1F1} 0.138 & {\cellcolor[HTML]{329900}} \color[HTML]{F1F1F1} 0.755 & {\cellcolor[HTML]{329900}} \color[HTML]{F1F1F1} 0.755 \\
14k-16k & {\cellcolor[HTML]{FF0000}} \color[HTML]{F1F1F1} 0.000 & {\cellcolor[HTML]{FF0000}} \color[HTML]{F1F1F1} 0.000 & {\cellcolor[HTML]{FF0000}} \color[HTML]{F1F1F1} 0.000 & {\cellcolor[HTML]{6AB500}} \color[HTML]{F1F1F1} 0.663 & {\cellcolor[HTML]{7EBF00}} \color[HTML]{000000} 0.629 \\
16k-24k & {\cellcolor[HTML]{FF0000}} \color[HTML]{F1F1F1} 0.000 & {\cellcolor[HTML]{FF0000}} \color[HTML]{F1F1F1} 0.000 & {\cellcolor[HTML]{FF0800}} \color[HTML]{F1F1F1} 0.014 & {\cellcolor[HTML]{4EA700}} \color[HTML]{F1F1F1} 0.708 & {\cellcolor[HTML]{4AA500}} \color[HTML]{F1F1F1} 0.715 \\
24k-32k & {\cellcolor[HTML]{FF0C00}} \color[HTML]{F1F1F1} 0.020 & {\cellcolor[HTML]{FF0600}} \color[HTML]{F1F1F1} 0.010 & {\cellcolor[HTML]{FF0000}} \color[HTML]{F1F1F1} 0.000 & {\cellcolor[HTML]{FF9C00}} \color[HTML]{000000} 0.257 & {\cellcolor[HTML]{3E9F00}} \color[HTML]{F1F1F1} 0.733 \\
32k-49k & {\cellcolor[HTML]{FF0400}} \color[HTML]{F1F1F1} 0.009 & {\cellcolor[HTML]{FF0000}} \color[HTML]{F1F1F1} 0.000 & {\cellcolor[HTML]{FF0000}} \color[HTML]{F1F1F1} 0.000 & {\cellcolor[HTML]{FF0000}} \color[HTML]{F1F1F1} 0.000 & {\cellcolor[HTML]{40A000}} \color[HTML]{F1F1F1} 0.730 \\
49k-65k & {\cellcolor[HTML]{FF0000}} \color[HTML]{F1F1F1} 0.000 & {\cellcolor[HTML]{FF0000}} \color[HTML]{F1F1F1} 0.000 & {\cellcolor[HTML]{FF0000}} \color[HTML]{F1F1F1} 0.000 & {\cellcolor[HTML]{FF0000}} \color[HTML]{F1F1F1} 0.000 & {\cellcolor[HTML]{FF1200}} \color[HTML]{F1F1F1} 0.030 \\
\hline
In-Domain (0k-8k) & {\cellcolor[HTML]{FF0000}} \color[HTML]{F1F1F1} 0.000 & {\cellcolor[HTML]{289400}} \color[HTML]{F1F1F1} 0.771 & {\cellcolor[HTML]{78BC00}} \color[HTML]{F1F1F1} 0.639 & {\cellcolor[HTML]{4EA700}} \color[HTML]{F1F1F1} 0.707 & {\cellcolor[HTML]{4AA500}} \color[HTML]{F1F1F1} 0.714 \\
OOD (8k-65k) & {\cellcolor[HTML]{FF0200}} \color[HTML]{F1F1F1} 0.005 & {\cellcolor[HTML]{FF2200}} \color[HTML]{F1F1F1} 0.058 & {\cellcolor[HTML]{FF4400}} \color[HTML]{F1F1F1} 0.112 & {\cellcolor[HTML]{DEEF00}} \color[HTML]{000000} 0.473 & {\cellcolor[HTML]{76BB00}} \color[HTML]{F1F1F1} 0.643 \\
Long-OOD (32k-65k) & {\cellcolor[HTML]{FF0200}} \color[HTML]{F1F1F1} 0.005 & {\cellcolor[HTML]{FF0000}} \color[HTML]{F1F1F1} 0.000 & {\cellcolor[HTML]{FF0000}} \color[HTML]{F1F1F1} 0.000 & {\cellcolor[HTML]{FF0000}} \color[HTML]{F1F1F1} 0.000 & {\cellcolor[HTML]{FFF600}} \color[HTML]{000000} 0.403 \\
Full (0k-65k) & {\cellcolor[HTML]{FF0000}} \color[HTML]{F1F1F1} 0.003 & {\cellcolor[HTML]{FFBE00}} \color[HTML]{000000} 0.313 & {\cellcolor[HTML]{FFB600}} \color[HTML]{000000} 0.300 & {\cellcolor[HTML]{AAD500}} \color[HTML]{000000} 0.557 & {\cellcolor[HTML]{66B300}} \color[HTML]{F1F1F1} 0.668 \\
\bottomrule
\end{tabular}}

\end{center}
\caption{Best results on the MT dataset for SmolLM-2-360M-IT model, metric - EM. ARMT shows better overall performance as the base model, and outperforms it on OOD and Long-OOD.}
\label{tab:mt_best_smollm}
\end{table*}
\begin{table*}[t]

\begin{center}
\resizebox{0.9\textwidth}{!}{
\begin{tabular}{lcccccccc}
\toprule
\begin{tabular}[c]{@{}l@{}}\textbf{Model\//}\\ \textbf{Lengths}\end{tabular} & \begin{tabular}[c]{@{}l@{}}\textbf{Base,}\\ \textbf{GR-100+,}\\ \textbf{8k}\end{tabular} & \begin{tabular}[c]{@{}l@{}}\textbf{ARMT,}\\ \textbf{GR-100+,}\\ \textbf{8k}\end{tabular} & \begin{tabular}[c]{@{}l@{}}\textbf{ARMT,}\\ \textbf{ GR-100+, 8k,}\\ \textbf{w/o layers 0-7}\end{tabular} & \begin{tabular}[c]{@{}l@{}}\textbf{ARMT,}\\ \textbf{ GR-100+, 8k,}\\ \textbf{w/o layers 8-15}\end{tabular} & \begin{tabular}[c]{@{}l@{}}\textbf{ARMT,}\\ \textbf{ GR-100+, 8k,}\\ \textbf{w/o layers 16-23}\end{tabular} & \begin{tabular}[c]{@{}l@{}}\textbf{ARMT,}\\ \textbf{ GR-100+, 8k,}\\ \textbf{w/o layers 24-31}\end{tabular} & \begin{tabular}[c]{@{}l@{}}\textbf{ARMT,}\\ \textbf{ GR-100+, 8k,}\\ \textbf{only top-1 layer}\end{tabular} & \begin{tabular}[c]{@{}l@{}}\textbf{ARMT,}\\ \textbf{ GR-100+, 8k,}\\ \textbf{only top-4 layers}\end{tabular}  \\
\midrule
0k-1k & {\cellcolor[HTML]{0C8600}} \color[HTML]{F1F1F1} 0.349 & {\cellcolor[HTML]{1E8F00}} \color[HTML]{F1F1F1} 0.337 & {\cellcolor[HTML]{1E8F00}} \color[HTML]{F1F1F1} 0.337 & {\cellcolor[HTML]{1E8F00}} \color[HTML]{F1F1F1} 0.338 & {\cellcolor[HTML]{249200}} \color[HTML]{F1F1F1} 0.334 & {\cellcolor[HTML]{289400}} \color[HTML]{F1F1F1} 0.331 & {\cellcolor[HTML]{2A9500}} \color[HTML]{F1F1F1} 0.330 & {\cellcolor[HTML]{269300}} \color[HTML]{F1F1F1} 0.332 \\
1k-2k & {\cellcolor[HTML]{008000}} \color[HTML]{F1F1F1} 0.358 & {\cellcolor[HTML]{54AA00}} \color[HTML]{F1F1F1} 0.302 & {\cellcolor[HTML]{56AB00}} \color[HTML]{F1F1F1} 0.301 & {\cellcolor[HTML]{58AC00}} \color[HTML]{F1F1F1} 0.299 & {\cellcolor[HTML]{FCFE00}} \color[HTML]{000000} 0.192 & {\cellcolor[HTML]{E2F100}} \color[HTML]{000000} 0.208 & {\cellcolor[HTML]{FFB600}} \color[HTML]{000000} 0.142 & {\cellcolor[HTML]{F4FA00}} \color[HTML]{000000} 0.197 \\
2k-4k & {\cellcolor[HTML]{2A9500}} \color[HTML]{F1F1F1} 0.330 & {\cellcolor[HTML]{92C900}} \color[HTML]{000000} 0.261 & {\cellcolor[HTML]{AAD500}} \color[HTML]{000000} 0.246 & {\cellcolor[HTML]{96CB00}} \color[HTML]{000000} 0.258 & {\cellcolor[HTML]{FFE800}} \color[HTML]{000000} 0.175 & {\cellcolor[HTML]{FFE800}} \color[HTML]{000000} 0.175 & {\cellcolor[HTML]{FFAA00}} \color[HTML]{000000} 0.134 & {\cellcolor[HTML]{FFF600}} \color[HTML]{000000} 0.183 \\
4k-6k & {\cellcolor[HTML]{42A100}} \color[HTML]{F1F1F1} 0.314 & {\cellcolor[HTML]{9ECF00}} \color[HTML]{000000} 0.253 & {\cellcolor[HTML]{CAE500}} \color[HTML]{000000} 0.224 & {\cellcolor[HTML]{BCDE00}} \color[HTML]{000000} 0.234 & {\cellcolor[HTML]{FFE800}} \color[HTML]{000000} 0.175 & {\cellcolor[HTML]{FFEE00}} \color[HTML]{000000} 0.178 & {\cellcolor[HTML]{FF8400}} \color[HTML]{F1F1F1} 0.109 & {\cellcolor[HTML]{FFCA00}} \color[HTML]{000000} 0.155 \\
6k-8k & {\cellcolor[HTML]{6CB600}} \color[HTML]{F1F1F1} 0.286 & {\cellcolor[HTML]{CAE500}} \color[HTML]{000000} 0.225 & {\cellcolor[HTML]{FFFC00}} \color[HTML]{000000} 0.187 & {\cellcolor[HTML]{D4EA00}} \color[HTML]{000000} 0.218 & {\cellcolor[HTML]{FFD000}} \color[HTML]{000000} 0.159 & {\cellcolor[HTML]{FFEC00}} \color[HTML]{000000} 0.177 & {\cellcolor[HTML]{FF9600}} \color[HTML]{000000} 0.120 & {\cellcolor[HTML]{FFB800}} \color[HTML]{000000} 0.143 \\
8k-10k & {\cellcolor[HTML]{BCDE00}} \color[HTML]{000000} 0.233 & {\cellcolor[HTML]{D0E800}} \color[HTML]{000000} 0.221 & {\cellcolor[HTML]{FFDC00}} \color[HTML]{000000} 0.166 & {\cellcolor[HTML]{E6F300}} \color[HTML]{000000} 0.206 & {\cellcolor[HTML]{FFC800}} \color[HTML]{000000} 0.153 & {\cellcolor[HTML]{FFD600}} \color[HTML]{000000} 0.163 & {\cellcolor[HTML]{FF9400}} \color[HTML]{000000} 0.119 & {\cellcolor[HTML]{FFBC00}} \color[HTML]{000000} 0.146 \\
10k-12k & {\cellcolor[HTML]{FF0000}} \color[HTML]{F1F1F1} 0.021 & {\cellcolor[HTML]{DEEF00}} \color[HTML]{000000} 0.211 & {\cellcolor[HTML]{FFC600}} \color[HTML]{000000} 0.152 & {\cellcolor[HTML]{DEEF00}} \color[HTML]{000000} 0.211 & {\cellcolor[HTML]{FFCC00}} \color[HTML]{000000} 0.156 & {\cellcolor[HTML]{FFE800}} \color[HTML]{000000} 0.174 & {\cellcolor[HTML]{FF8C00}} \color[HTML]{F1F1F1} 0.114 & {\cellcolor[HTML]{FFC800}} \color[HTML]{000000} 0.153 \\
12k-14k & {\cellcolor[HTML]{FF1400}} \color[HTML]{F1F1F1} 0.035 & {\cellcolor[HTML]{D8EC00}} \color[HTML]{000000} 0.215 & {\cellcolor[HTML]{FF8400}} \color[HTML]{F1F1F1} 0.109 & {\cellcolor[HTML]{EEF700}} \color[HTML]{000000} 0.201 & {\cellcolor[HTML]{FFC400}} \color[HTML]{000000} 0.151 & {\cellcolor[HTML]{FFD000}} \color[HTML]{000000} 0.159 & {\cellcolor[HTML]{FF8A00}} \color[HTML]{F1F1F1} 0.113 & {\cellcolor[HTML]{FFBC00}} \color[HTML]{000000} 0.146 \\
14k-16k & {\cellcolor[HTML]{FF2A00}} \color[HTML]{F1F1F1} 0.049 & {\cellcolor[HTML]{E0F000}} \color[HTML]{000000} 0.210 & {\cellcolor[HTML]{FF6A00}} \color[HTML]{F1F1F1} 0.092 & {\cellcolor[HTML]{F0F800}} \color[HTML]{000000} 0.199 & {\cellcolor[HTML]{FFBA00}} \color[HTML]{000000} 0.144 & {\cellcolor[HTML]{FFC800}} \color[HTML]{000000} 0.153 & {\cellcolor[HTML]{FF9600}} \color[HTML]{000000} 0.121 & {\cellcolor[HTML]{FFBC00}} \color[HTML]{000000} 0.146 \\
16k-24k & {\cellcolor[HTML]{FF1800}} \color[HTML]{F1F1F1} 0.037 & {\cellcolor[HTML]{D2E900}} \color[HTML]{000000} 0.219 & {\cellcolor[HTML]{FF6000}} \color[HTML]{F1F1F1} 0.085 & {\cellcolor[HTML]{FCFE00}} \color[HTML]{000000} 0.192 & {\cellcolor[HTML]{FFBC00}} \color[HTML]{000000} 0.146 & {\cellcolor[HTML]{FFCA00}} \color[HTML]{000000} 0.155 & {\cellcolor[HTML]{FF8A00}} \color[HTML]{F1F1F1} 0.113 & {\cellcolor[HTML]{FFB000}} \color[HTML]{000000} 0.138 \\
24k-32k & {\cellcolor[HTML]{FF3200}} \color[HTML]{F1F1F1} 0.054 & {\cellcolor[HTML]{FFF800}} \color[HTML]{000000} 0.185 & {\cellcolor[HTML]{FF4400}} \color[HTML]{F1F1F1} 0.066 & {\cellcolor[HTML]{FFDE00}} \color[HTML]{000000} 0.168 & {\cellcolor[HTML]{FFB400}} \color[HTML]{000000} 0.140 & {\cellcolor[HTML]{FFC400}} \color[HTML]{000000} 0.151 & {\cellcolor[HTML]{FF8C00}} \color[HTML]{F1F1F1} 0.114 & {\cellcolor[HTML]{FFB600}} \color[HTML]{000000} 0.142 \\
32k-49k & {\cellcolor[HTML]{FF4400}} \color[HTML]{F1F1F1} 0.067 & {\cellcolor[HTML]{FFE200}} \color[HTML]{000000} 0.171 & {\cellcolor[HTML]{FF4200}} \color[HTML]{F1F1F1} 0.065 & {\cellcolor[HTML]{FFC800}} \color[HTML]{000000} 0.153 & {\cellcolor[HTML]{FFBE00}} \color[HTML]{000000} 0.147 & {\cellcolor[HTML]{FFB600}} \color[HTML]{000000} 0.141 & {\cellcolor[HTML]{FF9A00}} \color[HTML]{000000} 0.123 & {\cellcolor[HTML]{FFAC00}} \color[HTML]{000000} 0.135 \\
49k-65k & {\cellcolor[HTML]{FF6400}} \color[HTML]{F1F1F1} 0.088 & {\cellcolor[HTML]{FF7000}} \color[HTML]{F1F1F1} 0.096 & {\cellcolor[HTML]{FF2A00}} \color[HTML]{F1F1F1} 0.049 & {\cellcolor[HTML]{FF9000}} \color[HTML]{000000} 0.116 & {\cellcolor[HTML]{FEFF00}} \color[HTML]{000000} 0.190 & {\cellcolor[HTML]{FF6800}} \color[HTML]{F1F1F1} 0.090 & {\cellcolor[HTML]{FF8A00}} \color[HTML]{F1F1F1} 0.113 & {\cellcolor[HTML]{FFD000}} \color[HTML]{000000} 0.158 \\
\hline
\begin{tabular}[c]{@{}l@{}}{In-Domain}\\ {(0k-8k)}\end{tabular} & {\cellcolor[HTML]{2E9700}} \color[HTML]{F1F1F1} 0.327 & {\cellcolor[HTML]{7CBE00}} \color[HTML]{000000} 0.275 & {\cellcolor[HTML]{96CB00}} \color[HTML]{000000} 0.259 & {\cellcolor[HTML]{86C300}} \color[HTML]{000000} 0.269 & {\cellcolor[HTML]{E6F300}} \color[HTML]{000000} 0.206 & {\cellcolor[HTML]{DCEE00}} \color[HTML]{000000} 0.213 & {\cellcolor[HTML]{FFDC00}} \color[HTML]{000000} 0.166 & {\cellcolor[HTML]{ECF600}} \color[HTML]{000000} 0.202 \\
\begin{tabular}[c]{@{}l@{}}{OOD}\\ {(8k-65k)}\end{tabular} & {\cellcolor[HTML]{FF4E00}} \color[HTML]{F1F1F1} 0.074 & {\cellcolor[HTML]{DEEF00}} \color[HTML]{000000} 0.211 & {\cellcolor[HTML]{FF8C00}} \color[HTML]{F1F1F1} 0.114 & {\cellcolor[HTML]{F4FA00}} \color[HTML]{000000} 0.197 & {\cellcolor[HTML]{FFC200}} \color[HTML]{000000} 0.150 & {\cellcolor[HTML]{FFD000}} \color[HTML]{000000} 0.158 & {\cellcolor[HTML]{FF9000}} \color[HTML]{000000} 0.116 & {\cellcolor[HTML]{FFBC00}} \color[HTML]{000000} 0.145 \\
\begin{tabular}[c]{@{}l@{}}{Long-OOD}\\ {(32k-65k)}\end{tabular} & {\cellcolor[HTML]{FF4C00}} \color[HTML]{F1F1F1} 0.072 & {\cellcolor[HTML]{FFC800}} \color[HTML]{000000} 0.154 & {\cellcolor[HTML]{FF3C00}} \color[HTML]{F1F1F1} 0.061 & {\cellcolor[HTML]{FFBA00}} \color[HTML]{000000} 0.144 & {\cellcolor[HTML]{FFCE00}} \color[HTML]{000000} 0.157 & {\cellcolor[HTML]{FFA400}} \color[HTML]{000000} 0.129 & {\cellcolor[HTML]{FF9600}} \color[HTML]{000000} 0.121 & {\cellcolor[HTML]{FFB400}} \color[HTML]{000000} 0.140 \\
\begin{tabular}[c]{@{}l@{}}{Full}\\ {(0k-65k)}\end{tabular} & {\cellcolor[HTML]{FAFD00}} \color[HTML]{000000} 0.192 & {\cellcolor[HTML]{B0D800}} \color[HTML]{000000} 0.241 & {\cellcolor[HTML]{FFF200}} \color[HTML]{000000} 0.182 & {\cellcolor[HTML]{C0E000}} \color[HTML]{000000} 0.231 & {\cellcolor[HTML]{FFEA00}} \color[HTML]{000000} 0.176 & {\cellcolor[HTML]{FFF600}} \color[HTML]{000000} 0.184 & {\cellcolor[HTML]{FFB400}} \color[HTML]{000000} 0.140 & {\cellcolor[HTML]{FFE400}} \color[HTML]{000000} 0.171 \\
\bottomrule
\end{tabular}}
\caption{Associative layers ablation on the GR-100+ dataset for SmolLM-2-360M-IT model, metric - ROUGE-L. Middle and upper layers representations are the most important for associative memory, but for SmolLM-2-360M-IT backbone the lower layers are also important.}
\label{tab:gov_report_best_assoc_ablation_smollm}
\end{center}
\end{table*}
\begin{table*}[t]

\begin{center}
\resizebox{0.9\textwidth}{!}{
\begin{tabular}{lcccccccc}
\toprule
\begin{tabular}[c]{@{}l@{}}\textbf{Model\//}\\ \textbf{Lengths}\end{tabular} & \begin{tabular}[c]{@{}l@{}}\textbf{Base,}\\ \textbf{MT,}\\ \textbf{8k}\end{tabular} & \begin{tabular}[c]{@{}l@{}}\textbf{ARMT,}\\ \textbf{MT,}\\ \textbf{8k}\end{tabular} & \begin{tabular}[c]{@{}l@{}}\textbf{ARMT,}\\ \textbf{ MT, 8k,}\\ \textbf{w/o layers 0-7}\end{tabular} & \begin{tabular}[c]{@{}l@{}}\textbf{ARMT,}\\ \textbf{ MT, 8k,}\\ \textbf{w/o layers 8-15}\end{tabular} & \begin{tabular}[c]{@{}l@{}}\textbf{ARMT,}\\ \textbf{ MT, 8k,}\\ \textbf{w/o layers 16-23}\end{tabular} & \begin{tabular}[c]{@{}l@{}}\textbf{ARMT,}\\ \textbf{ MT, 8k,}\\ \textbf{w/o layers 24-31}\end{tabular} & \begin{tabular}[c]{@{}l@{}}\textbf{ARMT,}\\ \textbf{ MT, 8k,}\\ \textbf{only top-1 layer}\end{tabular} & \begin{tabular}[c]{@{}l@{}}\textbf{ARMT,}\\ \textbf{ MT, 8k,}\\ \textbf{only top-4 layers}\end{tabular}  \\
\midrule
0k-1k & {\cellcolor[HTML]{188C00}} \color[HTML]{F1F1F1} 0.795 & {\cellcolor[HTML]{289400}} \color[HTML]{F1F1F1} 0.769 & {\cellcolor[HTML]{289400}} \color[HTML]{F1F1F1} 0.769 & {\cellcolor[HTML]{289400}} \color[HTML]{F1F1F1} 0.769 & {\cellcolor[HTML]{309800}} \color[HTML]{F1F1F1} 0.756 & {\cellcolor[HTML]{309800}} \color[HTML]{F1F1F1} 0.756 & {\cellcolor[HTML]{309800}} \color[HTML]{F1F1F1} 0.756 & {\cellcolor[HTML]{309800}} \color[HTML]{F1F1F1} 0.756 \\
1k-2k & {\cellcolor[HTML]{1C8E00}} \color[HTML]{F1F1F1} 0.790 & {\cellcolor[HTML]{50A800}} \color[HTML]{F1F1F1} 0.705 & {\cellcolor[HTML]{56AB00}} \color[HTML]{F1F1F1} 0.695 & {\cellcolor[HTML]{56AB00}} \color[HTML]{F1F1F1} 0.695 & {\cellcolor[HTML]{F4FA00}} \color[HTML]{000000} 0.438 & {\cellcolor[HTML]{84C200}} \color[HTML]{000000} 0.619 & {\cellcolor[HTML]{A2D100}} \color[HTML]{000000} 0.571 & {\cellcolor[HTML]{78BC00}} \color[HTML]{F1F1F1} 0.638 \\
2k-4k & {\cellcolor[HTML]{3C9E00}} \color[HTML]{F1F1F1} 0.736 & {\cellcolor[HTML]{54AA00}} \color[HTML]{F1F1F1} 0.698 & {\cellcolor[HTML]{4EA700}} \color[HTML]{F1F1F1} 0.708 & {\cellcolor[HTML]{54AA00}} \color[HTML]{F1F1F1} 0.698 & {\cellcolor[HTML]{FFEC00}} \color[HTML]{000000} 0.387 & {\cellcolor[HTML]{70B800}} \color[HTML]{F1F1F1} 0.651 & {\cellcolor[HTML]{88C400}} \color[HTML]{000000} 0.613 & {\cellcolor[HTML]{82C100}} \color[HTML]{000000} 0.623 \\
4k-6k & {\cellcolor[HTML]{5CAE00}} \color[HTML]{F1F1F1} 0.685 & {\cellcolor[HTML]{7EBF00}} \color[HTML]{000000} 0.630 & {\cellcolor[HTML]{64B200}} \color[HTML]{F1F1F1} 0.671 & {\cellcolor[HTML]{6CB600}} \color[HTML]{F1F1F1} 0.658 & {\cellcolor[HTML]{FFD000}} \color[HTML]{000000} 0.342 & {\cellcolor[HTML]{86C300}} \color[HTML]{000000} 0.616 & {\cellcolor[HTML]{B0D800}} \color[HTML]{000000} 0.548 & {\cellcolor[HTML]{6CB600}} \color[HTML]{F1F1F1} 0.658 \\
6k-8k & {\cellcolor[HTML]{008000}} \color[HTML]{F1F1F1} 0.837 & {\cellcolor[HTML]{2E9700}} \color[HTML]{F1F1F1} 0.761 & {\cellcolor[HTML]{48A400}} \color[HTML]{F1F1F1} 0.717 & {\cellcolor[HTML]{2E9700}} \color[HTML]{F1F1F1} 0.761 & {\cellcolor[HTML]{FFC000}} \color[HTML]{000000} 0.315 & {\cellcolor[HTML]{349A00}} \color[HTML]{F1F1F1} 0.750 & {\cellcolor[HTML]{92C900}} \color[HTML]{000000} 0.598 & {\cellcolor[HTML]{349A00}} \color[HTML]{F1F1F1} 0.750 \\
8k-10k & {\cellcolor[HTML]{A4D200}} \color[HTML]{000000} 0.568 & {\cellcolor[HTML]{0C8600}} \color[HTML]{F1F1F1} 0.815 & {\cellcolor[HTML]{1C8E00}} \color[HTML]{F1F1F1} 0.790 & {\cellcolor[HTML]{249200}} \color[HTML]{F1F1F1} 0.778 & {\cellcolor[HTML]{FFC400}} \color[HTML]{000000} 0.321 & {\cellcolor[HTML]{4AA500}} \color[HTML]{F1F1F1} 0.716 & {\cellcolor[HTML]{C2E100}} \color[HTML]{000000} 0.519 & {\cellcolor[HTML]{58AC00}} \color[HTML]{F1F1F1} 0.691 \\
10k-12k & {\cellcolor[HTML]{FF0000}} \color[HTML]{F1F1F1} 0.000 & {\cellcolor[HTML]{369B00}} \color[HTML]{F1F1F1} 0.747 & {\cellcolor[HTML]{3C9E00}} \color[HTML]{F1F1F1} 0.736 & {\cellcolor[HTML]{289400}} \color[HTML]{F1F1F1} 0.769 & {\cellcolor[HTML]{FFE400}} \color[HTML]{000000} 0.374 & {\cellcolor[HTML]{4AA500}} \color[HTML]{F1F1F1} 0.714 & {\cellcolor[HTML]{E4F200}} \color[HTML]{000000} 0.462 & {\cellcolor[HTML]{44A200}} \color[HTML]{F1F1F1} 0.725 \\
12k-14k & {\cellcolor[HTML]{FF0000}} \color[HTML]{F1F1F1} 0.000 & {\cellcolor[HTML]{329900}} \color[HTML]{F1F1F1} 0.755 & {\cellcolor[HTML]{329900}} \color[HTML]{F1F1F1} 0.755 & {\cellcolor[HTML]{4AA500}} \color[HTML]{F1F1F1} 0.713 & {\cellcolor[HTML]{FF9A00}} \color[HTML]{000000} 0.255 & {\cellcolor[HTML]{3E9F00}} \color[HTML]{F1F1F1} 0.734 & {\cellcolor[HTML]{DAED00}} \color[HTML]{000000} 0.479 & {\cellcolor[HTML]{44A200}} \color[HTML]{F1F1F1} 0.723 \\
14k-16k & {\cellcolor[HTML]{FF0000}} \color[HTML]{F1F1F1} 0.000 & {\cellcolor[HTML]{7EBF00}} \color[HTML]{000000} 0.629 & {\cellcolor[HTML]{84C200}} \color[HTML]{000000} 0.618 & {\cellcolor[HTML]{70B800}} \color[HTML]{F1F1F1} 0.652 & {\cellcolor[HTML]{FFB200}} \color[HTML]{000000} 0.292 & {\cellcolor[HTML]{A0D000}} \color[HTML]{000000} 0.573 & {\cellcolor[HTML]{FFDC00}} \color[HTML]{000000} 0.360 & {\cellcolor[HTML]{A8D400}} \color[HTML]{000000} 0.562 \\
16k-24k & {\cellcolor[HTML]{FF0000}} \color[HTML]{F1F1F1} 0.000 & {\cellcolor[HTML]{4AA500}} \color[HTML]{F1F1F1} 0.715 & {\cellcolor[HTML]{4AA500}} \color[HTML]{F1F1F1} 0.715 & {\cellcolor[HTML]{52A900}} \color[HTML]{F1F1F1} 0.701 & {\cellcolor[HTML]{FF9400}} \color[HTML]{000000} 0.243 & {\cellcolor[HTML]{70B800}} \color[HTML]{F1F1F1} 0.653 & {\cellcolor[HTML]{DAED00}} \color[HTML]{000000} 0.479 & {\cellcolor[HTML]{70B800}} \color[HTML]{F1F1F1} 0.653 \\
24k-32k & {\cellcolor[HTML]{FF0600}} \color[HTML]{F1F1F1} 0.010 & {\cellcolor[HTML]{3E9F00}} \color[HTML]{F1F1F1} 0.733 & {\cellcolor[HTML]{50A800}} \color[HTML]{F1F1F1} 0.703 & {\cellcolor[HTML]{5EAF00}} \color[HTML]{F1F1F1} 0.683 & {\cellcolor[HTML]{FFAE00}} \color[HTML]{000000} 0.287 & {\cellcolor[HTML]{76BB00}} \color[HTML]{F1F1F1} 0.644 & {\cellcolor[HTML]{FFE600}} \color[HTML]{000000} 0.376 & {\cellcolor[HTML]{8EC700}} \color[HTML]{000000} 0.604 \\
32k-49k & {\cellcolor[HTML]{FF0000}} \color[HTML]{F1F1F1} 0.000 & {\cellcolor[HTML]{40A000}} \color[HTML]{F1F1F1} 0.730 & {\cellcolor[HTML]{60B000}} \color[HTML]{F1F1F1} 0.678 & {\cellcolor[HTML]{40A000}} \color[HTML]{F1F1F1} 0.730 & {\cellcolor[HTML]{FF8E00}} \color[HTML]{F1F1F1} 0.235 & {\cellcolor[HTML]{7ABD00}} \color[HTML]{F1F1F1} 0.635 & {\cellcolor[HTML]{E6F300}} \color[HTML]{000000} 0.461 & {\cellcolor[HTML]{80C000}} \color[HTML]{000000} 0.626 \\
49k-65k & {\cellcolor[HTML]{FF0000}} \color[HTML]{F1F1F1} 0.000 & {\cellcolor[HTML]{FF1200}} \color[HTML]{F1F1F1} 0.030 & {\cellcolor[HTML]{BEDF00}} \color[HTML]{000000} 0.525 & {\cellcolor[HTML]{FF1200}} \color[HTML]{F1F1F1} 0.030 & {\cellcolor[HTML]{FF1200}} \color[HTML]{F1F1F1} 0.030 & {\cellcolor[HTML]{FF1200}} \color[HTML]{F1F1F1} 0.030 & {\cellcolor[HTML]{DCEE00}} \color[HTML]{000000} 0.475 & {\cellcolor[HTML]{76BB00}} \color[HTML]{F1F1F1} 0.644 \\
\hline
In-Domain (0k-8k) & {\cellcolor[HTML]{289400}} \color[HTML]{F1F1F1} 0.771 & {\cellcolor[HTML]{4AA500}} \color[HTML]{F1F1F1} 0.714 & {\cellcolor[HTML]{4CA600}} \color[HTML]{F1F1F1} 0.711 & {\cellcolor[HTML]{4AA500}} \color[HTML]{F1F1F1} 0.716 & {\cellcolor[HTML]{F2F900}} \color[HTML]{000000} 0.440 & {\cellcolor[HTML]{62B100}} \color[HTML]{F1F1F1} 0.676 & {\cellcolor[HTML]{88C400}} \color[HTML]{000000} 0.614 & {\cellcolor[HTML]{5EAF00}} \color[HTML]{F1F1F1} 0.681 \\
OOD (8k-65k) & {\cellcolor[HTML]{FF2200}} \color[HTML]{F1F1F1} 0.058 & {\cellcolor[HTML]{76BB00}} \color[HTML]{F1F1F1} 0.643 & {\cellcolor[HTML]{5AAD00}} \color[HTML]{F1F1F1} 0.689 & {\cellcolor[HTML]{7CBE00}} \color[HTML]{000000} 0.631 & {\cellcolor[HTML]{FF9800}} \color[HTML]{000000} 0.250 & {\cellcolor[HTML]{98CC00}} \color[HTML]{000000} 0.586 & {\cellcolor[HTML]{EAF500}} \color[HTML]{000000} 0.452 & {\cellcolor[HTML]{70B800}} \color[HTML]{F1F1F1} 0.652 \\
Long-OOD (32k-65k) & {\cellcolor[HTML]{FF0000}} \color[HTML]{F1F1F1} 0.000 & {\cellcolor[HTML]{FFF600}} \color[HTML]{000000} 0.403 & {\cellcolor[HTML]{8CC600}} \color[HTML]{000000} 0.606 & {\cellcolor[HTML]{FFF600}} \color[HTML]{000000} 0.403 & {\cellcolor[HTML]{FF5400}} \color[HTML]{F1F1F1} 0.139 & {\cellcolor[HTML]{FFD600}} \color[HTML]{000000} 0.352 & {\cellcolor[HTML]{E0F000}} \color[HTML]{000000} 0.468 & {\cellcolor[HTML]{7ABD00}} \color[HTML]{F1F1F1} 0.634 \\
Full (0k-65k) & {\cellcolor[HTML]{FFBE00}} \color[HTML]{000000} 0.313 & {\cellcolor[HTML]{66B300}} \color[HTML]{F1F1F1} 0.668 & {\cellcolor[HTML]{54AA00}} \color[HTML]{F1F1F1} 0.697 & {\cellcolor[HTML]{6AB500}} \color[HTML]{F1F1F1} 0.661 & {\cellcolor[HTML]{FFC200}} \color[HTML]{000000} 0.318 & {\cellcolor[HTML]{84C200}} \color[HTML]{000000} 0.618 & {\cellcolor[HTML]{C6E300}} \color[HTML]{000000} 0.510 & {\cellcolor[HTML]{6AB500}} \color[HTML]{F1F1F1} 0.662 \\
\bottomrule
\end{tabular}}
\caption{Associative layers ablation on the MT dataset for SmolLM-2-360M-IT model, metric - EM. Middle and upper layers representations are the most important for associative memory.}
\label{tab:mt_best_assoc_ablation_smollm}
\end{center}
\end{table*}

\Cref{tab:gov_report_best_smollm,tab:gov_report_best_assoc_ablation_smollm} show the main results on GR-100+ with ARMT model with SmolLM-2-360M-IT backbone and associative layer ablation for this model, while~\Cref{tab:mt_best_smollm,tab:mt_best_assoc_ablation_smollm} show the same results on the MT dataset.

We also expanded the associative memory ablation study on GR-100+ dataset on SmolLM-2-360M-IT model, the results are presented in~\Cref{tab:gov_report_best_all_layers_assoc_ablation_0_15_smollm,tab:gov_report_best_all_layers_assoc_ablation_16_32_smollm,tab:mt_best_all_layers_assoc_ablation_0_15_smollm,tab:mt_best_all_layers_assoc_ablation_16_32_smollm}. As one can see, the importance of associative layers also depends on the model, showing that for SmolLM-2-360M-IT on GR-100+ the 0-th layer is the most important one, while the 23-rd is the most important on MT. However, the top-4 associative layers by importance are 0-th, 23-rd, 30-th, and 20-th for GR-100+ and 23-rd, 24-th, 17-th, and 14-th for MT, so the importance of the middle layer representations is still high.

\begin{table*}[h]

\begin{center}
\resizebox{1.0\textwidth}{!}{
\begin{tabular}{lcccccccccccccccccc}
\toprule
\begin{tabular}[c]{@{}l@{}}\textbf{Model\//}\\ \textbf{Lengths}\end{tabular} & \begin{tabular}[c]{@{}l@{}}\textbf{Base,}\\ \textbf{GR-100+,}\\ \textbf{8k}\end{tabular} & \begin{tabular}[c]{@{}l@{}}\textbf{ARMT,}\\ \textbf{GR-100+,}\\ \textbf{8k}\end{tabular} & \begin{tabular}[c]{@{}l@{}}\textbf{W/o}\\ \textbf{layer}\\ \textbf{0}\end{tabular} & \begin{tabular}[c]{@{}l@{}}\textbf{W/o}\\ \textbf{layer}\\ \textbf{1}\end{tabular} & \begin{tabular}[c]{@{}l@{}}\textbf{W/o}\\ \textbf{layer}\\ \textbf{2}\end{tabular} & \begin{tabular}[c]{@{}l@{}}\textbf{W/o}\\ \textbf{layer}\\ \textbf{3}\end{tabular} & \begin{tabular}[c]{@{}l@{}}\textbf{W/o}\\ \textbf{layer}\\ \textbf{4}\end{tabular} & \begin{tabular}[c]{@{}l@{}}\textbf{W/o}\\ \textbf{layer}\\ \textbf{5}\end{tabular} & \begin{tabular}[c]{@{}l@{}}\textbf{W/o}\\ \textbf{layer}\\ \textbf{6}\end{tabular} & \begin{tabular}[c]{@{}l@{}}\textbf{W/o}\\ \textbf{layer}\\ \textbf{7}\end{tabular} & \begin{tabular}[c]{@{}l@{}}\textbf{W/o}\\ \textbf{layer}\\ \textbf{8}\end{tabular}  & \begin{tabular}[c]{@{}l@{}}\textbf{W/o}\\ \textbf{layer}\\ \textbf{9}\end{tabular} & \begin{tabular}[c]{@{}l@{}}\textbf{W/o}\\ \textbf{layer}\\ \textbf{10}\end{tabular} & \begin{tabular}[c]{@{}l@{}}\textbf{W/o}\\ \textbf{layer}\\ \textbf{11}\end{tabular} & \begin{tabular}[c]{@{}l@{}}\textbf{W/o}\\ \textbf{layer}\\ \textbf{12}\end{tabular} & \begin{tabular}[c]{@{}l@{}}\textbf{W/o}\\ \textbf{layer}\\ \textbf{13}\end{tabular} & \begin{tabular}[c]{@{}l@{}}\textbf{W/o}\\ \textbf{layer}\\ \textbf{14}\end{tabular} & \begin{tabular}[c]{@{}l@{}}\textbf{W/o}\\ \textbf{layer}\\ \textbf{15}\end{tabular} \\
\midrule
0k-1k & {\cellcolor[HTML]{0C8600}} \color[HTML]{F1F1F1} 0.349 & {\cellcolor[HTML]{1E8F00}} \color[HTML]{F1F1F1} 0.337 & {\cellcolor[HTML]{1E8F00}} \color[HTML]{F1F1F1} 0.337 & {\cellcolor[HTML]{1E8F00}} \color[HTML]{F1F1F1} 0.338 & {\cellcolor[HTML]{1E8F00}} \color[HTML]{F1F1F1} 0.337 & {\cellcolor[HTML]{1E8F00}} \color[HTML]{F1F1F1} 0.337 & {\cellcolor[HTML]{1E8F00}} \color[HTML]{F1F1F1} 0.337 & {\cellcolor[HTML]{1E8F00}} \color[HTML]{F1F1F1} 0.337 & {\cellcolor[HTML]{1E8F00}} \color[HTML]{F1F1F1} 0.337 & {\cellcolor[HTML]{1E8F00}} \color[HTML]{F1F1F1} 0.337 & {\cellcolor[HTML]{1E8F00}} \color[HTML]{F1F1F1} 0.338 & {\cellcolor[HTML]{1E8F00}} \color[HTML]{F1F1F1} 0.337 & {\cellcolor[HTML]{1E8F00}} \color[HTML]{F1F1F1} 0.337 & {\cellcolor[HTML]{1E8F00}} \color[HTML]{F1F1F1} 0.338 & {\cellcolor[HTML]{1E8F00}} \color[HTML]{F1F1F1} 0.337 & {\cellcolor[HTML]{1E8F00}} \color[HTML]{F1F1F1} 0.337 & {\cellcolor[HTML]{1E8F00}} \color[HTML]{F1F1F1} 0.337 & {\cellcolor[HTML]{1E8F00}} \color[HTML]{F1F1F1} 0.337 \\
1k-2k & {\cellcolor[HTML]{008000}} \color[HTML]{F1F1F1} 0.358 & {\cellcolor[HTML]{54AA00}} \color[HTML]{F1F1F1} 0.302 & {\cellcolor[HTML]{5AAD00}} \color[HTML]{F1F1F1} 0.298 & {\cellcolor[HTML]{56AB00}} \color[HTML]{F1F1F1} 0.301 & {\cellcolor[HTML]{52A900}} \color[HTML]{F1F1F1} 0.303 & {\cellcolor[HTML]{54AA00}} \color[HTML]{F1F1F1} 0.302 & {\cellcolor[HTML]{5AAD00}} \color[HTML]{F1F1F1} 0.298 & {\cellcolor[HTML]{58AC00}} \color[HTML]{F1F1F1} 0.299 & {\cellcolor[HTML]{52A900}} \color[HTML]{F1F1F1} 0.304 & {\cellcolor[HTML]{66B300}} \color[HTML]{F1F1F1} 0.290 & {\cellcolor[HTML]{66B300}} \color[HTML]{F1F1F1} 0.290 & {\cellcolor[HTML]{5AAD00}} \color[HTML]{F1F1F1} 0.298 & {\cellcolor[HTML]{5AAD00}} \color[HTML]{F1F1F1} 0.298 & {\cellcolor[HTML]{64B200}} \color[HTML]{F1F1F1} 0.292 & {\cellcolor[HTML]{5AAD00}} \color[HTML]{F1F1F1} 0.298 & {\cellcolor[HTML]{54AA00}} \color[HTML]{F1F1F1} 0.302 & {\cellcolor[HTML]{58AC00}} \color[HTML]{F1F1F1} 0.299 & {\cellcolor[HTML]{5EAF00}} \color[HTML]{F1F1F1} 0.295 \\
2k-4k & {\cellcolor[HTML]{2A9500}} \color[HTML]{F1F1F1} 0.330 & {\cellcolor[HTML]{92C900}} \color[HTML]{000000} 0.261 & {\cellcolor[HTML]{B0D800}} \color[HTML]{000000} 0.241 & {\cellcolor[HTML]{96CB00}} \color[HTML]{000000} 0.259 & {\cellcolor[HTML]{9ECF00}} \color[HTML]{000000} 0.253 & {\cellcolor[HTML]{9ECF00}} \color[HTML]{000000} 0.254 & {\cellcolor[HTML]{9ACD00}} \color[HTML]{000000} 0.256 & {\cellcolor[HTML]{A0D000}} \color[HTML]{000000} 0.252 & {\cellcolor[HTML]{96CB00}} \color[HTML]{000000} 0.258 & {\cellcolor[HTML]{94CA00}} \color[HTML]{000000} 0.260 & {\cellcolor[HTML]{9ACD00}} \color[HTML]{000000} 0.256 & {\cellcolor[HTML]{9ACD00}} \color[HTML]{000000} 0.256 & {\cellcolor[HTML]{96CB00}} \color[HTML]{000000} 0.259 & {\cellcolor[HTML]{9ACD00}} \color[HTML]{000000} 0.256 & {\cellcolor[HTML]{96CB00}} \color[HTML]{000000} 0.259 & {\cellcolor[HTML]{94CA00}} \color[HTML]{000000} 0.260 & {\cellcolor[HTML]{90C800}} \color[HTML]{000000} 0.263 & {\cellcolor[HTML]{92C900}} \color[HTML]{000000} 0.261 \\
4k-6k & {\cellcolor[HTML]{42A100}} \color[HTML]{F1F1F1} 0.314 & {\cellcolor[HTML]{9ECF00}} \color[HTML]{000000} 0.253 & {\cellcolor[HTML]{B6DB00}} \color[HTML]{000000} 0.237 & {\cellcolor[HTML]{98CC00}} \color[HTML]{000000} 0.257 & {\cellcolor[HTML]{9CCE00}} \color[HTML]{000000} 0.255 & {\cellcolor[HTML]{A0D000}} \color[HTML]{000000} 0.252 & {\cellcolor[HTML]{98CC00}} \color[HTML]{000000} 0.257 & {\cellcolor[HTML]{96CB00}} \color[HTML]{000000} 0.259 & {\cellcolor[HTML]{94CA00}} \color[HTML]{000000} 0.260 & {\cellcolor[HTML]{A0D000}} \color[HTML]{000000} 0.252 & {\cellcolor[HTML]{A4D200}} \color[HTML]{000000} 0.250 & {\cellcolor[HTML]{A6D300}} \color[HTML]{000000} 0.248 & {\cellcolor[HTML]{AED700}} \color[HTML]{000000} 0.243 & {\cellcolor[HTML]{9ECF00}} \color[HTML]{000000} 0.254 & {\cellcolor[HTML]{96CB00}} \color[HTML]{000000} 0.259 & {\cellcolor[HTML]{92C900}} \color[HTML]{000000} 0.261 & {\cellcolor[HTML]{9CCE00}} \color[HTML]{000000} 0.255 & {\cellcolor[HTML]{A4D200}} \color[HTML]{000000} 0.249 \\
6k-8k & {\cellcolor[HTML]{6CB600}} \color[HTML]{F1F1F1} 0.286 & {\cellcolor[HTML]{CAE500}} \color[HTML]{000000} 0.225 & {\cellcolor[HTML]{EEF700}} \color[HTML]{000000} 0.201 & {\cellcolor[HTML]{D0E800}} \color[HTML]{000000} 0.221 & {\cellcolor[HTML]{CCE600}} \color[HTML]{000000} 0.223 & {\cellcolor[HTML]{D6EB00}} \color[HTML]{000000} 0.217 & {\cellcolor[HTML]{CAE500}} \color[HTML]{000000} 0.225 & {\cellcolor[HTML]{D0E800}} \color[HTML]{000000} 0.220 & {\cellcolor[HTML]{C4E200}} \color[HTML]{000000} 0.228 & {\cellcolor[HTML]{D2E900}} \color[HTML]{000000} 0.219 & {\cellcolor[HTML]{CCE600}} \color[HTML]{000000} 0.223 & {\cellcolor[HTML]{D4EA00}} \color[HTML]{000000} 0.218 & {\cellcolor[HTML]{D2E900}} \color[HTML]{000000} 0.219 & {\cellcolor[HTML]{D0E800}} \color[HTML]{000000} 0.221 & {\cellcolor[HTML]{D0E800}} \color[HTML]{000000} 0.220 & {\cellcolor[HTML]{CAE500}} \color[HTML]{000000} 0.224 & {\cellcolor[HTML]{CCE600}} \color[HTML]{000000} 0.223 & {\cellcolor[HTML]{CAE500}} \color[HTML]{000000} 0.225 \\
8k-10k & {\cellcolor[HTML]{BCDE00}} \color[HTML]{000000} 0.233 & {\cellcolor[HTML]{D0E800}} \color[HTML]{000000} 0.221 & {\cellcolor[HTML]{FCFE00}} \color[HTML]{000000} 0.192 & {\cellcolor[HTML]{D2E900}} \color[HTML]{000000} 0.219 & {\cellcolor[HTML]{D6EB00}} \color[HTML]{000000} 0.216 & {\cellcolor[HTML]{CAE500}} \color[HTML]{000000} 0.224 & {\cellcolor[HTML]{D4EA00}} \color[HTML]{000000} 0.218 & {\cellcolor[HTML]{CEE700}} \color[HTML]{000000} 0.222 & {\cellcolor[HTML]{C2E100}} \color[HTML]{000000} 0.229 & {\cellcolor[HTML]{D4EA00}} \color[HTML]{000000} 0.218 & {\cellcolor[HTML]{CEE700}} \color[HTML]{000000} 0.222 & {\cellcolor[HTML]{DAED00}} \color[HTML]{000000} 0.214 & {\cellcolor[HTML]{D4EA00}} \color[HTML]{000000} 0.218 & {\cellcolor[HTML]{DCEE00}} \color[HTML]{000000} 0.213 & {\cellcolor[HTML]{D6EB00}} \color[HTML]{000000} 0.216 & {\cellcolor[HTML]{CEE700}} \color[HTML]{000000} 0.222 & {\cellcolor[HTML]{D2E900}} \color[HTML]{000000} 0.219 & {\cellcolor[HTML]{D6EB00}} \color[HTML]{000000} 0.217 \\
10k-12k & {\cellcolor[HTML]{FF0000}} \color[HTML]{F1F1F1} 0.021 & {\cellcolor[HTML]{DEEF00}} \color[HTML]{000000} 0.211 & {\cellcolor[HTML]{FFE000}} \color[HTML]{000000} 0.169 & {\cellcolor[HTML]{DAED00}} \color[HTML]{000000} 0.214 & {\cellcolor[HTML]{D8EC00}} \color[HTML]{000000} 0.215 & {\cellcolor[HTML]{E0F000}} \color[HTML]{000000} 0.210 & {\cellcolor[HTML]{D8EC00}} \color[HTML]{000000} 0.215 & {\cellcolor[HTML]{D6EB00}} \color[HTML]{000000} 0.217 & {\cellcolor[HTML]{D6EB00}} \color[HTML]{000000} 0.216 & {\cellcolor[HTML]{DAED00}} \color[HTML]{000000} 0.214 & {\cellcolor[HTML]{EAF500}} \color[HTML]{000000} 0.203 & {\cellcolor[HTML]{D0E800}} \color[HTML]{000000} 0.221 & {\cellcolor[HTML]{E2F100}} \color[HTML]{000000} 0.209 & {\cellcolor[HTML]{DEEF00}} \color[HTML]{000000} 0.211 & {\cellcolor[HTML]{D8EC00}} \color[HTML]{000000} 0.215 & {\cellcolor[HTML]{D6EB00}} \color[HTML]{000000} 0.217 & {\cellcolor[HTML]{D2E900}} \color[HTML]{000000} 0.219 & {\cellcolor[HTML]{DAED00}} \color[HTML]{000000} 0.214 \\
12k-14k & {\cellcolor[HTML]{FF1400}} \color[HTML]{F1F1F1} 0.035 & {\cellcolor[HTML]{D8EC00}} \color[HTML]{000000} 0.215 & {\cellcolor[HTML]{FFCA00}} \color[HTML]{000000} 0.155 & {\cellcolor[HTML]{D6EB00}} \color[HTML]{000000} 0.217 & {\cellcolor[HTML]{D0E800}} \color[HTML]{000000} 0.220 & {\cellcolor[HTML]{DAED00}} \color[HTML]{000000} 0.214 & {\cellcolor[HTML]{D8EC00}} \color[HTML]{000000} 0.215 & {\cellcolor[HTML]{E0F000}} \color[HTML]{000000} 0.210 & {\cellcolor[HTML]{CCE600}} \color[HTML]{000000} 0.223 & {\cellcolor[HTML]{DCEE00}} \color[HTML]{000000} 0.213 & {\cellcolor[HTML]{DCEE00}} \color[HTML]{000000} 0.213 & {\cellcolor[HTML]{D2E900}} \color[HTML]{000000} 0.219 & {\cellcolor[HTML]{E2F100}} \color[HTML]{000000} 0.209 & {\cellcolor[HTML]{DCEE00}} \color[HTML]{000000} 0.213 & {\cellcolor[HTML]{D2E900}} \color[HTML]{000000} 0.219 & {\cellcolor[HTML]{D6EB00}} \color[HTML]{000000} 0.217 & {\cellcolor[HTML]{DCEE00}} \color[HTML]{000000} 0.213 & {\cellcolor[HTML]{D2E900}} \color[HTML]{000000} 0.219 \\
14k-16k & {\cellcolor[HTML]{FF2A00}} \color[HTML]{F1F1F1} 0.049 & {\cellcolor[HTML]{E0F000}} \color[HTML]{000000} 0.210 & {\cellcolor[HTML]{FFA600}} \color[HTML]{000000} 0.131 & {\cellcolor[HTML]{DAED00}} \color[HTML]{000000} 0.214 & {\cellcolor[HTML]{D4EA00}} \color[HTML]{000000} 0.218 & {\cellcolor[HTML]{DEEF00}} \color[HTML]{000000} 0.211 & {\cellcolor[HTML]{D4EA00}} \color[HTML]{000000} 0.218 & {\cellcolor[HTML]{DCEE00}} \color[HTML]{000000} 0.212 & {\cellcolor[HTML]{DAED00}} \color[HTML]{000000} 0.214 & {\cellcolor[HTML]{DEEF00}} \color[HTML]{000000} 0.211 & {\cellcolor[HTML]{DCEE00}} \color[HTML]{000000} 0.212 & {\cellcolor[HTML]{D6EB00}} \color[HTML]{000000} 0.216 & {\cellcolor[HTML]{D6EB00}} \color[HTML]{000000} 0.217 & {\cellcolor[HTML]{DEEF00}} \color[HTML]{000000} 0.211 & {\cellcolor[HTML]{D4EA00}} \color[HTML]{000000} 0.218 & {\cellcolor[HTML]{E0F000}} \color[HTML]{000000} 0.210 & {\cellcolor[HTML]{D2E900}} \color[HTML]{000000} 0.219 & {\cellcolor[HTML]{DEEF00}} \color[HTML]{000000} 0.211 \\
16k-24k & {\cellcolor[HTML]{FF1800}} \color[HTML]{F1F1F1} 0.037 & {\cellcolor[HTML]{D2E900}} \color[HTML]{000000} 0.219 & {\cellcolor[HTML]{FF8200}} \color[HTML]{F1F1F1} 0.107 & {\cellcolor[HTML]{D4EA00}} \color[HTML]{000000} 0.218 & {\cellcolor[HTML]{DCEE00}} \color[HTML]{000000} 0.213 & {\cellcolor[HTML]{D8EC00}} \color[HTML]{000000} 0.215 & {\cellcolor[HTML]{DCEE00}} \color[HTML]{000000} 0.212 & {\cellcolor[HTML]{D8EC00}} \color[HTML]{000000} 0.215 & {\cellcolor[HTML]{CCE600}} \color[HTML]{000000} 0.223 & {\cellcolor[HTML]{D6EB00}} \color[HTML]{000000} 0.216 & {\cellcolor[HTML]{DCEE00}} \color[HTML]{000000} 0.213 & {\cellcolor[HTML]{D4EA00}} \color[HTML]{000000} 0.218 & {\cellcolor[HTML]{E8F400}} \color[HTML]{000000} 0.205 & {\cellcolor[HTML]{D4EA00}} \color[HTML]{000000} 0.218 & {\cellcolor[HTML]{D6EB00}} \color[HTML]{000000} 0.217 & {\cellcolor[HTML]{D0E800}} \color[HTML]{000000} 0.220 & {\cellcolor[HTML]{C6E300}} \color[HTML]{000000} 0.227 & {\cellcolor[HTML]{CEE700}} \color[HTML]{000000} 0.222 \\
24k-32k & {\cellcolor[HTML]{FF3200}} \color[HTML]{F1F1F1} 0.054 & {\cellcolor[HTML]{FFF800}} \color[HTML]{000000} 0.185 & {\cellcolor[HTML]{FF4600}} \color[HTML]{F1F1F1} 0.068 & {\cellcolor[HTML]{F0F800}} \color[HTML]{000000} 0.199 & {\cellcolor[HTML]{FCFE00}} \color[HTML]{000000} 0.192 & {\cellcolor[HTML]{F2F900}} \color[HTML]{000000} 0.198 & {\cellcolor[HTML]{F6FB00}} \color[HTML]{000000} 0.196 & {\cellcolor[HTML]{FFFC00}} \color[HTML]{000000} 0.188 & {\cellcolor[HTML]{EEF700}} \color[HTML]{000000} 0.201 & {\cellcolor[HTML]{EAF500}} \color[HTML]{000000} 0.203 & {\cellcolor[HTML]{E8F400}} \color[HTML]{000000} 0.205 & {\cellcolor[HTML]{FFF600}} \color[HTML]{000000} 0.183 & {\cellcolor[HTML]{FFDC00}} \color[HTML]{000000} 0.166 & {\cellcolor[HTML]{F6FB00}} \color[HTML]{000000} 0.196 & {\cellcolor[HTML]{FFFE00}} \color[HTML]{000000} 0.189 & {\cellcolor[HTML]{F6FB00}} \color[HTML]{000000} 0.196 & {\cellcolor[HTML]{F8FC00}} \color[HTML]{000000} 0.194 & {\cellcolor[HTML]{FFF600}} \color[HTML]{000000} 0.183 \\
32k-49k & {\cellcolor[HTML]{FF4400}} \color[HTML]{F1F1F1} 0.067 & {\cellcolor[HTML]{FFE200}} \color[HTML]{000000} 0.171 & {\cellcolor[HTML]{FF3E00}} \color[HTML]{F1F1F1} 0.063 & {\cellcolor[HTML]{FCFE00}} \color[HTML]{000000} 0.192 & {\cellcolor[HTML]{FFFE00}} \color[HTML]{000000} 0.189 & {\cellcolor[HTML]{FFC600}} \color[HTML]{000000} 0.152 & {\cellcolor[HTML]{FFF200}} \color[HTML]{000000} 0.181 & {\cellcolor[HTML]{FFE600}} \color[HTML]{000000} 0.173 & {\cellcolor[HTML]{FFE800}} \color[HTML]{000000} 0.175 & {\cellcolor[HTML]{FFE000}} \color[HTML]{000000} 0.169 & {\cellcolor[HTML]{FFF200}} \color[HTML]{000000} 0.181 & {\cellcolor[HTML]{FFFE00}} \color[HTML]{000000} 0.189 & {\cellcolor[HTML]{FFC000}} \color[HTML]{000000} 0.148 & {\cellcolor[HTML]{FFE200}} \color[HTML]{000000} 0.170 & {\cellcolor[HTML]{FFF400}} \color[HTML]{000000} 0.182 & {\cellcolor[HTML]{FFE800}} \color[HTML]{000000} 0.174 & {\cellcolor[HTML]{F2F900}} \color[HTML]{000000} 0.198 & {\cellcolor[HTML]{FFCA00}} \color[HTML]{000000} 0.154 \\
49k-65k & {\cellcolor[HTML]{FF6400}} \color[HTML]{F1F1F1} 0.088 & {\cellcolor[HTML]{FF7000}} \color[HTML]{F1F1F1} 0.096 & {\cellcolor[HTML]{FF3E00}} \color[HTML]{F1F1F1} 0.062 & {\cellcolor[HTML]{FFD000}} \color[HTML]{000000} 0.159 & {\cellcolor[HTML]{FF7600}} \color[HTML]{F1F1F1} 0.099 & {\cellcolor[HTML]{FF5200}} \color[HTML]{F1F1F1} 0.076 & {\cellcolor[HTML]{FF7400}} \color[HTML]{F1F1F1} 0.098 & {\cellcolor[HTML]{FF9A00}} \color[HTML]{000000} 0.123 & {\cellcolor[HTML]{FFAE00}} \color[HTML]{000000} 0.136 & {\cellcolor[HTML]{FF7000}} \color[HTML]{F1F1F1} 0.095 & {\cellcolor[HTML]{FF5000}} \color[HTML]{F1F1F1} 0.074 & {\cellcolor[HTML]{FF9200}} \color[HTML]{000000} 0.118 & {\cellcolor[HTML]{FF5800}} \color[HTML]{F1F1F1} 0.080 & {\cellcolor[HTML]{FF6000}} \color[HTML]{F1F1F1} 0.085 & {\cellcolor[HTML]{FF6C00}} \color[HTML]{F1F1F1} 0.093 & {\cellcolor[HTML]{FFA200}} \color[HTML]{000000} 0.128 & {\cellcolor[HTML]{FF5A00}} \color[HTML]{F1F1F1} 0.081 & {\cellcolor[HTML]{FF4E00}} \color[HTML]{F1F1F1} 0.073 \\
\hline
\begin{tabular}[c]{@{}l@{}}{In-Domain}\\ {(0k-8k)}\end{tabular} & {\cellcolor[HTML]{2E9700}} \color[HTML]{F1F1F1} 0.327 & {\cellcolor[HTML]{7CBE00}} \color[HTML]{000000} 0.275 & {\cellcolor[HTML]{90C800}} \color[HTML]{000000} 0.262 & {\cellcolor[HTML]{7EBF00}} \color[HTML]{000000} 0.275 & {\cellcolor[HTML]{7EBF00}} \color[HTML]{000000} 0.274 & {\cellcolor[HTML]{82C100}} \color[HTML]{000000} 0.272 & {\cellcolor[HTML]{7EBF00}} \color[HTML]{000000} 0.274 & {\cellcolor[HTML]{80C000}} \color[HTML]{000000} 0.273 & {\cellcolor[HTML]{7ABD00}} \color[HTML]{F1F1F1} 0.277 & {\cellcolor[HTML]{82C100}} \color[HTML]{000000} 0.271 & {\cellcolor[HTML]{84C200}} \color[HTML]{000000} 0.271 & {\cellcolor[HTML]{84C200}} \color[HTML]{000000} 0.271 & {\cellcolor[HTML]{84C200}} \color[HTML]{000000} 0.271 & {\cellcolor[HTML]{82C100}} \color[HTML]{000000} 0.272 & {\cellcolor[HTML]{7EBF00}} \color[HTML]{000000} 0.274 & {\cellcolor[HTML]{7ABD00}} \color[HTML]{F1F1F1} 0.276 & {\cellcolor[HTML]{7CBE00}} \color[HTML]{000000} 0.275 & {\cellcolor[HTML]{80C000}} \color[HTML]{000000} 0.273 \\
\begin{tabular}[c]{@{}l@{}}{OOD}\\ {(8k-65k)}\end{tabular} & {\cellcolor[HTML]{FF4E00}} \color[HTML]{F1F1F1} 0.074 & {\cellcolor[HTML]{DEEF00}} \color[HTML]{000000} 0.211 & {\cellcolor[HTML]{FFB600}} \color[HTML]{000000} 0.141 & {\cellcolor[HTML]{DAED00}} \color[HTML]{000000} 0.214 & {\cellcolor[HTML]{DCEE00}} \color[HTML]{000000} 0.213 & {\cellcolor[HTML]{E0F000}} \color[HTML]{000000} 0.210 & {\cellcolor[HTML]{DEEF00}} \color[HTML]{000000} 0.212 & {\cellcolor[HTML]{DEEF00}} \color[HTML]{000000} 0.211 & {\cellcolor[HTML]{D4EA00}} \color[HTML]{000000} 0.217 & {\cellcolor[HTML]{DEEF00}} \color[HTML]{000000} 0.211 & {\cellcolor[HTML]{E0F000}} \color[HTML]{000000} 0.210 & {\cellcolor[HTML]{DAED00}} \color[HTML]{000000} 0.213 & {\cellcolor[HTML]{E8F400}} \color[HTML]{000000} 0.205 & {\cellcolor[HTML]{E0F000}} \color[HTML]{000000} 0.209 & {\cellcolor[HTML]{DCEE00}} \color[HTML]{000000} 0.213 & {\cellcolor[HTML]{DAED00}} \color[HTML]{000000} 0.213 & {\cellcolor[HTML]{D8EC00}} \color[HTML]{000000} 0.216 & {\cellcolor[HTML]{DEEF00}} \color[HTML]{000000} 0.211 \\
\begin{tabular}[c]{@{}l@{}}{Long-OOD}\\ {(32k-65k)}\end{tabular} & {\cellcolor[HTML]{FF4C00}} \color[HTML]{F1F1F1} 0.072 & {\cellcolor[HTML]{FFC800}} \color[HTML]{000000} 0.154 & {\cellcolor[HTML]{FF3E00}} \color[HTML]{F1F1F1} 0.063 & {\cellcolor[HTML]{FFF800}} \color[HTML]{000000} 0.184 & {\cellcolor[HTML]{FFDE00}} \color[HTML]{000000} 0.168 & {\cellcolor[HTML]{FFAC00}} \color[HTML]{000000} 0.134 & {\cellcolor[HTML]{FFD400}} \color[HTML]{000000} 0.162 & {\cellcolor[HTML]{FFD400}} \color[HTML]{000000} 0.161 & {\cellcolor[HTML]{FFDC00}} \color[HTML]{000000} 0.166 & {\cellcolor[HTML]{FFC600}} \color[HTML]{000000} 0.152 & {\cellcolor[HTML]{FFCC00}} \color[HTML]{000000} 0.156 & {\cellcolor[HTML]{FFE600}} \color[HTML]{000000} 0.173 & {\cellcolor[HTML]{FFA800}} \color[HTML]{000000} 0.132 & {\cellcolor[HTML]{FFC400}} \color[HTML]{000000} 0.150 & {\cellcolor[HTML]{FFD400}} \color[HTML]{000000} 0.161 & {\cellcolor[HTML]{FFD800}} \color[HTML]{000000} 0.163 & {\cellcolor[HTML]{FFE200}} \color[HTML]{000000} 0.171 & {\cellcolor[HTML]{FFAC00}} \color[HTML]{000000} 0.135 \\
\begin{tabular}[c]{@{}l@{}}{Full}\\ {(0k-65k)}\end{tabular} & {\cellcolor[HTML]{FAFD00}} \color[HTML]{000000} 0.192 & {\cellcolor[HTML]{B0D800}} \color[HTML]{000000} 0.241 & {\cellcolor[HTML]{F2F900}} \color[HTML]{000000} 0.198 & {\cellcolor[HTML]{AED700}} \color[HTML]{000000} 0.242 & {\cellcolor[HTML]{B0D800}} \color[HTML]{000000} 0.241 & {\cellcolor[HTML]{B4DA00}} \color[HTML]{000000} 0.239 & {\cellcolor[HTML]{B0D800}} \color[HTML]{000000} 0.241 & {\cellcolor[HTML]{B2D900}} \color[HTML]{000000} 0.240 & {\cellcolor[HTML]{AAD500}} \color[HTML]{000000} 0.245 & {\cellcolor[HTML]{B4DA00}} \color[HTML]{000000} 0.239 & {\cellcolor[HTML]{B4DA00}} \color[HTML]{000000} 0.238 & {\cellcolor[HTML]{B2D900}} \color[HTML]{000000} 0.240 & {\cellcolor[HTML]{B8DC00}} \color[HTML]{000000} 0.236 & {\cellcolor[HTML]{B4DA00}} \color[HTML]{000000} 0.239 & {\cellcolor[HTML]{B0D800}} \color[HTML]{000000} 0.242 & {\cellcolor[HTML]{AED700}} \color[HTML]{000000} 0.243 & {\cellcolor[HTML]{AED700}} \color[HTML]{000000} 0.243 & {\cellcolor[HTML]{B2D900}} \color[HTML]{000000} 0.240 \\
\bottomrule
\end{tabular}}
\caption{Ablation for all associative layers on the GovReport-100+ dataset for ARMT with SmolLM-2-360M-IT model, metric - ROUGE-L. Ablated associative layers from 0 to 15.}
\label{tab:gov_report_best_all_layers_assoc_ablation_0_15_smollm}
\end{center}
\end{table*}
\begin{table*}[h]

\begin{center}
\resizebox{1.0\textwidth}{!}{
\begin{tabular}{lcccccccccccccccccc}
\toprule
\begin{tabular}[c]{@{}l@{}}\textbf{Model\//}\\ \textbf{Lengths}\end{tabular} & \begin{tabular}[c]{@{}l@{}}\textbf{Base,}\\ \textbf{GR-100+,}\\ \textbf{8k}\end{tabular} & \begin{tabular}[c]{@{}l@{}}\textbf{ARMT,}\\ \textbf{GR-100+,}\\ \textbf{8k}\end{tabular} & \begin{tabular}[c]{@{}l@{}}\textbf{W/o}\\ \textbf{layer}\\ \textbf{16}\end{tabular} & \begin{tabular}[c]{@{}l@{}}\textbf{W/o}\\ \textbf{layer}\\ \textbf{17}\end{tabular} & \begin{tabular}[c]{@{}l@{}}\textbf{W/o}\\ \textbf{layer}\\ \textbf{18}\end{tabular} & \begin{tabular}[c]{@{}l@{}}\textbf{W/o}\\ \textbf{layer}\\ \textbf{19}\end{tabular} & \begin{tabular}[c]{@{}l@{}}\textbf{W/o}\\ \textbf{layer}\\ \textbf{20}\end{tabular} & \begin{tabular}[c]{@{}l@{}}\textbf{W/o}\\ \textbf{layer}\\ \textbf{21}\end{tabular} & \begin{tabular}[c]{@{}l@{}}\textbf{W/o}\\ \textbf{layer}\\ \textbf{22}\end{tabular} & \begin{tabular}[c]{@{}l@{}}\textbf{W/o}\\ \textbf{layer}\\ \textbf{23}\end{tabular} & \begin{tabular}[c]{@{}l@{}}\textbf{W/o}\\ \textbf{layer}\\ \textbf{24}\end{tabular} & \begin{tabular}[c]{@{}l@{}}\textbf{W/o}\\ \textbf{layer}\\ \textbf{25}\end{tabular} & \begin{tabular}[c]{@{}l@{}}\textbf{W/o}\\ \textbf{layer}\\ \textbf{26}\end{tabular} & \begin{tabular}[c]{@{}l@{}}\textbf{W/o}\\ \textbf{layer}\\ \textbf{27}\end{tabular} & \begin{tabular}[c]{@{}l@{}}\textbf{W/o}\\ \textbf{layer}\\ \textbf{28}\end{tabular} & \begin{tabular}[c]{@{}l@{}}\textbf{W/o}\\ \textbf{layer}\\ \textbf{29}\end{tabular} & \begin{tabular}[c]{@{}l@{}}\textbf{W/o}\\ \textbf{layer}\\ \textbf{30}\end{tabular} & \begin{tabular}[c]{@{}l@{}}\textbf{W/o}\\ \textbf{layer}\\ \textbf{31}\end{tabular}\\
\midrule
0k-1k & {\cellcolor[HTML]{0C8600}} \color[HTML]{F1F1F1} 0.349 & {\cellcolor[HTML]{1E8F00}} \color[HTML]{F1F1F1} 0.337 & {\cellcolor[HTML]{1E8F00}} \color[HTML]{F1F1F1} 0.337 & {\cellcolor[HTML]{1E8F00}} \color[HTML]{F1F1F1} 0.337 & {\cellcolor[HTML]{1E8F00}} \color[HTML]{F1F1F1} 0.337 & {\cellcolor[HTML]{1E8F00}} \color[HTML]{F1F1F1} 0.338 & {\cellcolor[HTML]{1E8F00}} \color[HTML]{F1F1F1} 0.337 & {\cellcolor[HTML]{1E8F00}} \color[HTML]{F1F1F1} 0.338 & {\cellcolor[HTML]{229100}} \color[HTML]{F1F1F1} 0.335 & {\cellcolor[HTML]{1E8F00}} \color[HTML]{F1F1F1} 0.337 & {\cellcolor[HTML]{209000}} \color[HTML]{F1F1F1} 0.336 & {\cellcolor[HTML]{1E8F00}} \color[HTML]{F1F1F1} 0.337 & {\cellcolor[HTML]{1E8F00}} \color[HTML]{F1F1F1} 0.337 & {\cellcolor[HTML]{1E8F00}} \color[HTML]{F1F1F1} 0.337 & {\cellcolor[HTML]{1E8F00}} \color[HTML]{F1F1F1} 0.338 & {\cellcolor[HTML]{209000}} \color[HTML]{F1F1F1} 0.336 & {\cellcolor[HTML]{1E8F00}} \color[HTML]{F1F1F1} 0.337 & {\cellcolor[HTML]{1E8F00}} \color[HTML]{F1F1F1} 0.337 \\
1k-2k & {\cellcolor[HTML]{008000}} \color[HTML]{F1F1F1} 0.358 & {\cellcolor[HTML]{54AA00}} \color[HTML]{F1F1F1} 0.302 & {\cellcolor[HTML]{5EAF00}} \color[HTML]{F1F1F1} 0.296 & {\cellcolor[HTML]{5EAF00}} \color[HTML]{F1F1F1} 0.296 & {\cellcolor[HTML]{5EAF00}} \color[HTML]{F1F1F1} 0.295 & {\cellcolor[HTML]{6AB500}} \color[HTML]{F1F1F1} 0.287 & {\cellcolor[HTML]{6EB700}} \color[HTML]{F1F1F1} 0.285 & {\cellcolor[HTML]{6CB600}} \color[HTML]{F1F1F1} 0.286 & {\cellcolor[HTML]{80C000}} \color[HTML]{000000} 0.273 & {\cellcolor[HTML]{B4DA00}} \color[HTML]{000000} 0.239 & {\cellcolor[HTML]{6AB500}} \color[HTML]{F1F1F1} 0.288 & {\cellcolor[HTML]{5AAD00}} \color[HTML]{F1F1F1} 0.298 & {\cellcolor[HTML]{52A900}} \color[HTML]{F1F1F1} 0.303 & {\cellcolor[HTML]{7ABD00}} \color[HTML]{F1F1F1} 0.277 & {\cellcolor[HTML]{5AAD00}} \color[HTML]{F1F1F1} 0.298 & {\cellcolor[HTML]{74BA00}} \color[HTML]{F1F1F1} 0.281 & {\cellcolor[HTML]{8AC500}} \color[HTML]{000000} 0.267 & {\cellcolor[HTML]{64B200}} \color[HTML]{F1F1F1} 0.291 \\
2k-4k & {\cellcolor[HTML]{2A9500}} \color[HTML]{F1F1F1} 0.330 & {\cellcolor[HTML]{92C900}} \color[HTML]{000000} 0.261 & {\cellcolor[HTML]{A2D100}} \color[HTML]{000000} 0.251 & {\cellcolor[HTML]{9ACD00}} \color[HTML]{000000} 0.256 & {\cellcolor[HTML]{94CA00}} \color[HTML]{000000} 0.260 & {\cellcolor[HTML]{98CC00}} \color[HTML]{000000} 0.257 & {\cellcolor[HTML]{96CB00}} \color[HTML]{000000} 0.258 & {\cellcolor[HTML]{96CB00}} \color[HTML]{000000} 0.259 & {\cellcolor[HTML]{9CCE00}} \color[HTML]{000000} 0.255 & {\cellcolor[HTML]{DCEE00}} \color[HTML]{000000} 0.212 & {\cellcolor[HTML]{B0D800}} \color[HTML]{000000} 0.241 & {\cellcolor[HTML]{94CA00}} \color[HTML]{000000} 0.260 & {\cellcolor[HTML]{90C800}} \color[HTML]{000000} 0.262 & {\cellcolor[HTML]{AAD500}} \color[HTML]{000000} 0.246 & {\cellcolor[HTML]{9ACD00}} \color[HTML]{000000} 0.256 & {\cellcolor[HTML]{A6D300}} \color[HTML]{000000} 0.248 & {\cellcolor[HTML]{B8DC00}} \color[HTML]{000000} 0.236 & {\cellcolor[HTML]{8EC700}} \color[HTML]{000000} 0.264 \\
4k-6k & {\cellcolor[HTML]{42A100}} \color[HTML]{F1F1F1} 0.314 & {\cellcolor[HTML]{9ECF00}} \color[HTML]{000000} 0.253 & {\cellcolor[HTML]{9ECF00}} \color[HTML]{000000} 0.253 & {\cellcolor[HTML]{9ECF00}} \color[HTML]{000000} 0.253 & {\cellcolor[HTML]{9ECF00}} \color[HTML]{000000} 0.253 & {\cellcolor[HTML]{B4DA00}} \color[HTML]{000000} 0.239 & {\cellcolor[HTML]{B2D900}} \color[HTML]{000000} 0.240 & {\cellcolor[HTML]{9ECF00}} \color[HTML]{000000} 0.253 & {\cellcolor[HTML]{BADD00}} \color[HTML]{000000} 0.235 & {\cellcolor[HTML]{C6E300}} \color[HTML]{000000} 0.227 & {\cellcolor[HTML]{AAD500}} \color[HTML]{000000} 0.246 & {\cellcolor[HTML]{96CB00}} \color[HTML]{000000} 0.258 & {\cellcolor[HTML]{90C800}} \color[HTML]{000000} 0.263 & {\cellcolor[HTML]{B6DB00}} \color[HTML]{000000} 0.237 & {\cellcolor[HTML]{94CA00}} \color[HTML]{000000} 0.260 & {\cellcolor[HTML]{AAD500}} \color[HTML]{000000} 0.246 & {\cellcolor[HTML]{B2D900}} \color[HTML]{000000} 0.240 & {\cellcolor[HTML]{9ECF00}} \color[HTML]{000000} 0.253 \\
6k-8k & {\cellcolor[HTML]{6CB600}} \color[HTML]{F1F1F1} 0.286 & {\cellcolor[HTML]{CAE500}} \color[HTML]{000000} 0.225 & {\cellcolor[HTML]{CEE700}} \color[HTML]{000000} 0.222 & {\cellcolor[HTML]{C2E100}} \color[HTML]{000000} 0.229 & {\cellcolor[HTML]{D6EB00}} \color[HTML]{000000} 0.216 & {\cellcolor[HTML]{C8E400}} \color[HTML]{000000} 0.226 & {\cellcolor[HTML]{C0E000}} \color[HTML]{000000} 0.231 & {\cellcolor[HTML]{D6EB00}} \color[HTML]{000000} 0.216 & {\cellcolor[HTML]{C0E000}} \color[HTML]{000000} 0.231 & {\cellcolor[HTML]{DCEE00}} \color[HTML]{000000} 0.213 & {\cellcolor[HTML]{D6EB00}} \color[HTML]{000000} 0.217 & {\cellcolor[HTML]{C2E100}} \color[HTML]{000000} 0.230 & {\cellcolor[HTML]{C8E400}} \color[HTML]{000000} 0.226 & {\cellcolor[HTML]{D8EC00}} \color[HTML]{000000} 0.215 & {\cellcolor[HTML]{C8E400}} \color[HTML]{000000} 0.226 & {\cellcolor[HTML]{C8E400}} \color[HTML]{000000} 0.226 & {\cellcolor[HTML]{CAE500}} \color[HTML]{000000} 0.224 & {\cellcolor[HTML]{CCE600}} \color[HTML]{000000} 0.223 \\
8k-10k & {\cellcolor[HTML]{BCDE00}} \color[HTML]{000000} 0.233 & {\cellcolor[HTML]{D0E800}} \color[HTML]{000000} 0.221 & {\cellcolor[HTML]{CCE600}} \color[HTML]{000000} 0.223 & {\cellcolor[HTML]{CAE500}} \color[HTML]{000000} 0.224 & {\cellcolor[HTML]{D0E800}} \color[HTML]{000000} 0.221 & {\cellcolor[HTML]{E8F400}} \color[HTML]{000000} 0.204 & {\cellcolor[HTML]{E2F100}} \color[HTML]{000000} 0.208 & {\cellcolor[HTML]{D6EB00}} \color[HTML]{000000} 0.217 & {\cellcolor[HTML]{E0F000}} \color[HTML]{000000} 0.210 & {\cellcolor[HTML]{FAFD00}} \color[HTML]{000000} 0.193 & {\cellcolor[HTML]{E0F000}} \color[HTML]{000000} 0.210 & {\cellcolor[HTML]{D8EC00}} \color[HTML]{000000} 0.215 & {\cellcolor[HTML]{D6EB00}} \color[HTML]{000000} 0.216 & {\cellcolor[HTML]{E2F100}} \color[HTML]{000000} 0.208 & {\cellcolor[HTML]{D0E800}} \color[HTML]{000000} 0.220 & {\cellcolor[HTML]{E4F200}} \color[HTML]{000000} 0.207 & {\cellcolor[HTML]{ECF600}} \color[HTML]{000000} 0.202 & {\cellcolor[HTML]{D6EB00}} \color[HTML]{000000} 0.216 \\
10k-12k & {\cellcolor[HTML]{FF0000}} \color[HTML]{F1F1F1} 0.021 & {\cellcolor[HTML]{DEEF00}} \color[HTML]{000000} 0.211 & {\cellcolor[HTML]{DEEF00}} \color[HTML]{000000} 0.211 & {\cellcolor[HTML]{E0F000}} \color[HTML]{000000} 0.210 & {\cellcolor[HTML]{D2E900}} \color[HTML]{000000} 0.219 & {\cellcolor[HTML]{E2F100}} \color[HTML]{000000} 0.208 & {\cellcolor[HTML]{F2F900}} \color[HTML]{000000} 0.198 & {\cellcolor[HTML]{E6F300}} \color[HTML]{000000} 0.206 & {\cellcolor[HTML]{DAED00}} \color[HTML]{000000} 0.214 & {\cellcolor[HTML]{FFFC00}} \color[HTML]{000000} 0.188 & {\cellcolor[HTML]{EEF700}} \color[HTML]{000000} 0.201 & {\cellcolor[HTML]{E0F000}} \color[HTML]{000000} 0.210 & {\cellcolor[HTML]{D6EB00}} \color[HTML]{000000} 0.216 & {\cellcolor[HTML]{E2F100}} \color[HTML]{000000} 0.208 & {\cellcolor[HTML]{DEEF00}} \color[HTML]{000000} 0.211 & {\cellcolor[HTML]{DCEE00}} \color[HTML]{000000} 0.213 & {\cellcolor[HTML]{DCEE00}} \color[HTML]{000000} 0.212 & {\cellcolor[HTML]{DEEF00}} \color[HTML]{000000} 0.211 \\
12k-14k & {\cellcolor[HTML]{FF1400}} \color[HTML]{F1F1F1} 0.035 & {\cellcolor[HTML]{D8EC00}} \color[HTML]{000000} 0.215 & {\cellcolor[HTML]{CCE600}} \color[HTML]{000000} 0.223 & {\cellcolor[HTML]{E2F100}} \color[HTML]{000000} 0.208 & {\cellcolor[HTML]{D8EC00}} \color[HTML]{000000} 0.215 & {\cellcolor[HTML]{E6F300}} \color[HTML]{000000} 0.206 & {\cellcolor[HTML]{F4FA00}} \color[HTML]{000000} 0.197 & {\cellcolor[HTML]{E2F100}} \color[HTML]{000000} 0.208 & {\cellcolor[HTML]{DAED00}} \color[HTML]{000000} 0.214 & {\cellcolor[HTML]{FFF000}} \color[HTML]{000000} 0.180 & {\cellcolor[HTML]{DAED00}} \color[HTML]{000000} 0.214 & {\cellcolor[HTML]{D8EC00}} \color[HTML]{000000} 0.215 & {\cellcolor[HTML]{DCEE00}} \color[HTML]{000000} 0.212 & {\cellcolor[HTML]{DEEF00}} \color[HTML]{000000} 0.211 & {\cellcolor[HTML]{D2E900}} \color[HTML]{000000} 0.219 & {\cellcolor[HTML]{ECF600}} \color[HTML]{000000} 0.202 & {\cellcolor[HTML]{E4F200}} \color[HTML]{000000} 0.207 & {\cellcolor[HTML]{DAED00}} \color[HTML]{000000} 0.214 \\
14k-16k & {\cellcolor[HTML]{FF2A00}} \color[HTML]{F1F1F1} 0.049 & {\cellcolor[HTML]{E0F000}} \color[HTML]{000000} 0.210 & {\cellcolor[HTML]{DCEE00}} \color[HTML]{000000} 0.212 & {\cellcolor[HTML]{DCEE00}} \color[HTML]{000000} 0.213 & {\cellcolor[HTML]{DEEF00}} \color[HTML]{000000} 0.211 & {\cellcolor[HTML]{DCEE00}} \color[HTML]{000000} 0.213 & {\cellcolor[HTML]{ECF600}} \color[HTML]{000000} 0.202 & {\cellcolor[HTML]{E2F100}} \color[HTML]{000000} 0.209 & {\cellcolor[HTML]{DCEE00}} \color[HTML]{000000} 0.213 & {\cellcolor[HTML]{EAF500}} \color[HTML]{000000} 0.203 & {\cellcolor[HTML]{DEEF00}} \color[HTML]{000000} 0.211 & {\cellcolor[HTML]{DAED00}} \color[HTML]{000000} 0.214 & {\cellcolor[HTML]{E6F300}} \color[HTML]{000000} 0.206 & {\cellcolor[HTML]{DEEF00}} \color[HTML]{000000} 0.211 & {\cellcolor[HTML]{DCEE00}} \color[HTML]{000000} 0.212 & {\cellcolor[HTML]{DCEE00}} \color[HTML]{000000} 0.213 & {\cellcolor[HTML]{F0F800}} \color[HTML]{000000} 0.200 & {\cellcolor[HTML]{E0F000}} \color[HTML]{000000} 0.210 \\
16k-24k & {\cellcolor[HTML]{FF1800}} \color[HTML]{F1F1F1} 0.037 & {\cellcolor[HTML]{D2E900}} \color[HTML]{000000} 0.219 & {\cellcolor[HTML]{CCE600}} \color[HTML]{000000} 0.223 & {\cellcolor[HTML]{CEE700}} \color[HTML]{000000} 0.222 & {\cellcolor[HTML]{D0E800}} \color[HTML]{000000} 0.221 & {\cellcolor[HTML]{E2F100}} \color[HTML]{000000} 0.209 & {\cellcolor[HTML]{E2F100}} \color[HTML]{000000} 0.209 & {\cellcolor[HTML]{DCEE00}} \color[HTML]{000000} 0.213 & {\cellcolor[HTML]{DEEF00}} \color[HTML]{000000} 0.211 & {\cellcolor[HTML]{FCFE00}} \color[HTML]{000000} 0.191 & {\cellcolor[HTML]{E2F100}} \color[HTML]{000000} 0.208 & {\cellcolor[HTML]{D6EB00}} \color[HTML]{000000} 0.216 & {\cellcolor[HTML]{D6EB00}} \color[HTML]{000000} 0.216 & {\cellcolor[HTML]{D6EB00}} \color[HTML]{000000} 0.216 & {\cellcolor[HTML]{CEE700}} \color[HTML]{000000} 0.222 & {\cellcolor[HTML]{EAF500}} \color[HTML]{000000} 0.203 & {\cellcolor[HTML]{FCFE00}} \color[HTML]{000000} 0.191 & {\cellcolor[HTML]{D0E800}} \color[HTML]{000000} 0.220 \\
24k-32k & {\cellcolor[HTML]{FF3200}} \color[HTML]{F1F1F1} 0.054 & {\cellcolor[HTML]{FFF800}} \color[HTML]{000000} 0.185 & {\cellcolor[HTML]{EEF700}} \color[HTML]{000000} 0.201 & {\cellcolor[HTML]{FFFA00}} \color[HTML]{000000} 0.186 & {\cellcolor[HTML]{F6FB00}} \color[HTML]{000000} 0.196 & {\cellcolor[HTML]{EEF700}} \color[HTML]{000000} 0.201 & {\cellcolor[HTML]{F6FB00}} \color[HTML]{000000} 0.196 & {\cellcolor[HTML]{FCFE00}} \color[HTML]{000000} 0.192 & {\cellcolor[HTML]{FFF800}} \color[HTML]{000000} 0.185 & {\cellcolor[HTML]{FFD800}} \color[HTML]{000000} 0.164 & {\cellcolor[HTML]{F4FA00}} \color[HTML]{000000} 0.197 & {\cellcolor[HTML]{F6FB00}} \color[HTML]{000000} 0.195 & {\cellcolor[HTML]{F2F900}} \color[HTML]{000000} 0.198 & {\cellcolor[HTML]{FFF400}} \color[HTML]{000000} 0.182 & {\cellcolor[HTML]{FFFE00}} \color[HTML]{000000} 0.189 & {\cellcolor[HTML]{FFF800}} \color[HTML]{000000} 0.185 & {\cellcolor[HTML]{FFF800}} \color[HTML]{000000} 0.185 & {\cellcolor[HTML]{FCFE00}} \color[HTML]{000000} 0.192 \\
32k-49k & {\cellcolor[HTML]{FF4400}} \color[HTML]{F1F1F1} 0.067 & {\cellcolor[HTML]{FFE200}} \color[HTML]{000000} 0.171 & {\cellcolor[HTML]{FFFE00}} \color[HTML]{000000} 0.189 & {\cellcolor[HTML]{FFEA00}} \color[HTML]{000000} 0.176 & {\cellcolor[HTML]{FFDE00}} \color[HTML]{000000} 0.168 & {\cellcolor[HTML]{FFE400}} \color[HTML]{000000} 0.172 & {\cellcolor[HTML]{FFA600}} \color[HTML]{000000} 0.131 & {\cellcolor[HTML]{FFCA00}} \color[HTML]{000000} 0.155 & {\cellcolor[HTML]{FFF200}} \color[HTML]{000000} 0.181 & {\cellcolor[HTML]{FFC400}} \color[HTML]{000000} 0.151 & {\cellcolor[HTML]{FFE600}} \color[HTML]{000000} 0.173 & {\cellcolor[HTML]{FFD600}} \color[HTML]{000000} 0.163 & {\cellcolor[HTML]{FFEA00}} \color[HTML]{000000} 0.176 & {\cellcolor[HTML]{FFE200}} \color[HTML]{000000} 0.171 & {\cellcolor[HTML]{FFD400}} \color[HTML]{000000} 0.161 & {\cellcolor[HTML]{FFD000}} \color[HTML]{000000} 0.159 & {\cellcolor[HTML]{FFCA00}} \color[HTML]{000000} 0.155 & {\cellcolor[HTML]{FFD800}} \color[HTML]{000000} 0.164 \\
49k-65k & {\cellcolor[HTML]{FF6400}} \color[HTML]{F1F1F1} 0.088 & {\cellcolor[HTML]{FF7000}} \color[HTML]{F1F1F1} 0.096 & {\cellcolor[HTML]{FF8000}} \color[HTML]{F1F1F1} 0.106 & {\cellcolor[HTML]{FF5600}} \color[HTML]{F1F1F1} 0.078 & {\cellcolor[HTML]{FF8400}} \color[HTML]{F1F1F1} 0.109 & {\cellcolor[HTML]{FF5800}} \color[HTML]{F1F1F1} 0.080 & {\cellcolor[HTML]{FF3200}} \color[HTML]{F1F1F1} 0.054 & {\cellcolor[HTML]{FF6A00}} \color[HTML]{F1F1F1} 0.092 & {\cellcolor[HTML]{FF6400}} \color[HTML]{F1F1F1} 0.087 & {\cellcolor[HTML]{FF8400}} \color[HTML]{F1F1F1} 0.109 & {\cellcolor[HTML]{FFBA00}} \color[HTML]{000000} 0.144 & {\cellcolor[HTML]{FF8A00}} \color[HTML]{F1F1F1} 0.113 & {\cellcolor[HTML]{FF6800}} \color[HTML]{F1F1F1} 0.090 & {\cellcolor[HTML]{FF6E00}} \color[HTML]{F1F1F1} 0.094 & {\cellcolor[HTML]{FF6C00}} \color[HTML]{F1F1F1} 0.093 & {\cellcolor[HTML]{FF6200}} \color[HTML]{F1F1F1} 0.086 & {\cellcolor[HTML]{FF8C00}} \color[HTML]{F1F1F1} 0.114 & {\cellcolor[HTML]{FF7C00}} \color[HTML]{F1F1F1} 0.103 \\
\hline
\begin{tabular}[c]{@{}l@{}}{In-Domain}\\ {(0k-8k)}\end{tabular} & {\cellcolor[HTML]{2E9700}} \color[HTML]{F1F1F1} 0.327 & {\cellcolor[HTML]{7CBE00}} \color[HTML]{000000} 0.275 & {\cellcolor[HTML]{82C100}} \color[HTML]{000000} 0.271 & {\cellcolor[HTML]{7EBF00}} \color[HTML]{000000} 0.274 & {\cellcolor[HTML]{82C100}} \color[HTML]{000000} 0.272 & {\cellcolor[HTML]{86C300}} \color[HTML]{000000} 0.269 & {\cellcolor[HTML]{84C200}} \color[HTML]{000000} 0.270 & {\cellcolor[HTML]{84C200}} \color[HTML]{000000} 0.270 & {\cellcolor[HTML]{8CC600}} \color[HTML]{000000} 0.266 & {\cellcolor[HTML]{AAD500}} \color[HTML]{000000} 0.245 & {\cellcolor[HTML]{8CC600}} \color[HTML]{000000} 0.265 & {\cellcolor[HTML]{7CBE00}} \color[HTML]{000000} 0.276 & {\cellcolor[HTML]{78BC00}} \color[HTML]{F1F1F1} 0.278 & {\cellcolor[HTML]{90C800}} \color[HTML]{000000} 0.262 & {\cellcolor[HTML]{7CBE00}} \color[HTML]{000000} 0.275 & {\cellcolor[HTML]{8AC500}} \color[HTML]{000000} 0.267 & {\cellcolor[HTML]{94CA00}} \color[HTML]{000000} 0.260 & {\cellcolor[HTML]{80C000}} \color[HTML]{000000} 0.273 \\
\begin{tabular}[c]{@{}l@{}}{OOD}\\ {(8k-65k)}\end{tabular} & {\cellcolor[HTML]{FF4E00}} \color[HTML]{F1F1F1} 0.074 & {\cellcolor[HTML]{DEEF00}} \color[HTML]{000000} 0.211 & {\cellcolor[HTML]{D8EC00}} \color[HTML]{000000} 0.215 & {\cellcolor[HTML]{DEEF00}} \color[HTML]{000000} 0.211 & {\cellcolor[HTML]{DCEE00}} \color[HTML]{000000} 0.213 & {\cellcolor[HTML]{E8F400}} \color[HTML]{000000} 0.205 & {\cellcolor[HTML]{F2F900}} \color[HTML]{000000} 0.198 & {\cellcolor[HTML]{E6F300}} \color[HTML]{000000} 0.206 & {\cellcolor[HTML]{E2F100}} \color[HTML]{000000} 0.208 & {\cellcolor[HTML]{FFFA00}} \color[HTML]{000000} 0.187 & {\cellcolor[HTML]{E6F300}} \color[HTML]{000000} 0.206 & {\cellcolor[HTML]{E0F000}} \color[HTML]{000000} 0.210 & {\cellcolor[HTML]{E0F000}} \color[HTML]{000000} 0.210 & {\cellcolor[HTML]{E6F300}} \color[HTML]{000000} 0.206 & {\cellcolor[HTML]{DCEE00}} \color[HTML]{000000} 0.212 & {\cellcolor[HTML]{EAF500}} \color[HTML]{000000} 0.203 & {\cellcolor[HTML]{F2F900}} \color[HTML]{000000} 0.199 & {\cellcolor[HTML]{E0F000}} \color[HTML]{000000} 0.210 \\
\begin{tabular}[c]{@{}l@{}}{Long-OOD}\\ {(32k-65k)}\end{tabular} & {\cellcolor[HTML]{FF4C00}} \color[HTML]{F1F1F1} 0.072 & {\cellcolor[HTML]{FFC800}} \color[HTML]{000000} 0.154 & {\cellcolor[HTML]{FFE200}} \color[HTML]{000000} 0.170 & {\cellcolor[HTML]{FFC800}} \color[HTML]{000000} 0.153 & {\cellcolor[HTML]{FFCA00}} \color[HTML]{000000} 0.154 & {\cellcolor[HTML]{FFC400}} \color[HTML]{000000} 0.151 & {\cellcolor[HTML]{FF8C00}} \color[HTML]{F1F1F1} 0.113 & {\cellcolor[HTML]{FFB400}} \color[HTML]{000000} 0.140 & {\cellcolor[HTML]{FFD200}} \color[HTML]{000000} 0.159 & {\cellcolor[HTML]{FFB600}} \color[HTML]{000000} 0.141 & {\cellcolor[HTML]{FFDC00}} \color[HTML]{000000} 0.166 & {\cellcolor[HTML]{FFC600}} \color[HTML]{000000} 0.151 & {\cellcolor[HTML]{FFCC00}} \color[HTML]{000000} 0.156 & {\cellcolor[HTML]{FFC800}} \color[HTML]{000000} 0.153 & {\cellcolor[HTML]{FFBC00}} \color[HTML]{000000} 0.145 & {\cellcolor[HTML]{FFB800}} \color[HTML]{000000} 0.142 & {\cellcolor[HTML]{FFBC00}} \color[HTML]{000000} 0.146 & {\cellcolor[HTML]{FFC200}} \color[HTML]{000000} 0.150 \\
\begin{tabular}[c]{@{}l@{}}{Full}\\ {(0k-65k)}\end{tabular} & {\cellcolor[HTML]{FAFD00}} \color[HTML]{000000} 0.192 & {\cellcolor[HTML]{B0D800}} \color[HTML]{000000} 0.241 & {\cellcolor[HTML]{B0D800}} \color[HTML]{000000} 0.242 & {\cellcolor[HTML]{B2D900}} \color[HTML]{000000} 0.240 & {\cellcolor[HTML]{B2D900}} \color[HTML]{000000} 0.241 & {\cellcolor[HTML]{BADD00}} \color[HTML]{000000} 0.235 & {\cellcolor[HTML]{BEDF00}} \color[HTML]{000000} 0.232 & {\cellcolor[HTML]{B8DC00}} \color[HTML]{000000} 0.236 & {\cellcolor[HTML]{BADD00}} \color[HTML]{000000} 0.235 & {\cellcolor[HTML]{DAED00}} \color[HTML]{000000} 0.214 & {\cellcolor[HTML]{BCDE00}} \color[HTML]{000000} 0.234 & {\cellcolor[HTML]{B0D800}} \color[HTML]{000000} 0.241 & {\cellcolor[HTML]{B0D800}} \color[HTML]{000000} 0.242 & {\cellcolor[HTML]{BEDF00}} \color[HTML]{000000} 0.232 & {\cellcolor[HTML]{B0D800}} \color[HTML]{000000} 0.242 & {\cellcolor[HTML]{BEDF00}} \color[HTML]{000000} 0.233 & {\cellcolor[HTML]{C6E300}} \color[HTML]{000000} 0.227 & {\cellcolor[HTML]{B2D900}} \color[HTML]{000000} 0.240 \\
\bottomrule
\end{tabular}}
\caption{Ablation for all associative layers on the GovReport-100+ dataset for ARMT with SmolLM-2-360M-IT model, metric - ROUGE-L. Ablated associative layers from 16 to 31.}
\label{tab:gov_report_best_all_layers_assoc_ablation_16_32_smollm}
\end{center}
\end{table*}
\begin{table*}[h]

\begin{center}
\resizebox{1.0\textwidth}{!}{
\begin{tabular}{lcccccccccccccccccc}
\toprule
\begin{tabular}[c]{@{}l@{}}\textbf{Model\//}\\ \textbf{Lengths}\end{tabular} & \begin{tabular}[c]{@{}l@{}}\textbf{Base,}\\ \textbf{MT,}\\ \textbf{8k}\end{tabular} & \begin{tabular}[c]{@{}l@{}}\textbf{ARMT,}\\ \textbf{MT,}\\ \textbf{8k}\end{tabular} & \begin{tabular}[c]{@{}l@{}}\textbf{W/o}\\ \textbf{layer}\\ \textbf{0}\end{tabular} & \begin{tabular}[c]{@{}l@{}}\textbf{W/o}\\ \textbf{layer}\\ \textbf{1}\end{tabular} & \begin{tabular}[c]{@{}l@{}}\textbf{W/o}\\ \textbf{layer}\\ \textbf{2}\end{tabular} & \begin{tabular}[c]{@{}l@{}}\textbf{W/o}\\ \textbf{layer}\\ \textbf{3}\end{tabular} & \begin{tabular}[c]{@{}l@{}}\textbf{W/o}\\ \textbf{layer}\\ \textbf{4}\end{tabular} & \begin{tabular}[c]{@{}l@{}}\textbf{W/o}\\ \textbf{layer}\\ \textbf{5}\end{tabular} & \begin{tabular}[c]{@{}l@{}}\textbf{W/o}\\ \textbf{layer}\\ \textbf{6}\end{tabular} & \begin{tabular}[c]{@{}l@{}}\textbf{W/o}\\ \textbf{layer}\\ \textbf{7}\end{tabular} & \begin{tabular}[c]{@{}l@{}}\textbf{W/o}\\ \textbf{layer}\\ \textbf{8}\end{tabular}  & \begin{tabular}[c]{@{}l@{}}\textbf{W/o}\\ \textbf{layer}\\ \textbf{9}\end{tabular} & \begin{tabular}[c]{@{}l@{}}\textbf{W/o}\\ \textbf{layer}\\ \textbf{10}\end{tabular} & \begin{tabular}[c]{@{}l@{}}\textbf{W/o}\\ \textbf{layer}\\ \textbf{11}\end{tabular} & \begin{tabular}[c]{@{}l@{}}\textbf{W/o}\\ \textbf{layer}\\ \textbf{12}\end{tabular} & \begin{tabular}[c]{@{}l@{}}\textbf{W/o}\\ \textbf{layer}\\ \textbf{13}\end{tabular} & \begin{tabular}[c]{@{}l@{}}\textbf{W/o}\\ \textbf{layer}\\ \textbf{14}\end{tabular} & \begin{tabular}[c]{@{}l@{}}\textbf{W/o}\\ \textbf{layer}\\ \textbf{15}\end{tabular} \\
\midrule
0k-1k & {\cellcolor[HTML]{188C00}} \color[HTML]{F1F1F1} 0.795 & {\cellcolor[HTML]{289400}} \color[HTML]{F1F1F1} 0.769 & {\cellcolor[HTML]{289400}} \color[HTML]{F1F1F1} 0.769 & {\cellcolor[HTML]{289400}} \color[HTML]{F1F1F1} 0.769 & {\cellcolor[HTML]{289400}} \color[HTML]{F1F1F1} 0.769 & {\cellcolor[HTML]{289400}} \color[HTML]{F1F1F1} 0.769 & {\cellcolor[HTML]{289400}} \color[HTML]{F1F1F1} 0.769 & {\cellcolor[HTML]{289400}} \color[HTML]{F1F1F1} 0.769 & {\cellcolor[HTML]{289400}} \color[HTML]{F1F1F1} 0.769 & {\cellcolor[HTML]{289400}} \color[HTML]{F1F1F1} 0.769 & {\cellcolor[HTML]{289400}} \color[HTML]{F1F1F1} 0.769 & {\cellcolor[HTML]{289400}} \color[HTML]{F1F1F1} 0.769 & {\cellcolor[HTML]{289400}} \color[HTML]{F1F1F1} 0.769 & {\cellcolor[HTML]{289400}} \color[HTML]{F1F1F1} 0.769 & {\cellcolor[HTML]{289400}} \color[HTML]{F1F1F1} 0.769 & {\cellcolor[HTML]{289400}} \color[HTML]{F1F1F1} 0.769 & {\cellcolor[HTML]{289400}} \color[HTML]{F1F1F1} 0.769 & {\cellcolor[HTML]{289400}} \color[HTML]{F1F1F1} 0.769 \\
1k-2k & {\cellcolor[HTML]{1C8E00}} \color[HTML]{F1F1F1} 0.790 & {\cellcolor[HTML]{50A800}} \color[HTML]{F1F1F1} 0.705 & {\cellcolor[HTML]{56AB00}} \color[HTML]{F1F1F1} 0.695 & {\cellcolor[HTML]{50A800}} \color[HTML]{F1F1F1} 0.705 & {\cellcolor[HTML]{50A800}} \color[HTML]{F1F1F1} 0.705 & {\cellcolor[HTML]{50A800}} \color[HTML]{F1F1F1} 0.705 & {\cellcolor[HTML]{50A800}} \color[HTML]{F1F1F1} 0.705 & {\cellcolor[HTML]{50A800}} \color[HTML]{F1F1F1} 0.705 & {\cellcolor[HTML]{50A800}} \color[HTML]{F1F1F1} 0.705 & {\cellcolor[HTML]{56AB00}} \color[HTML]{F1F1F1} 0.695 & {\cellcolor[HTML]{50A800}} \color[HTML]{F1F1F1} 0.705 & {\cellcolor[HTML]{50A800}} \color[HTML]{F1F1F1} 0.705 & {\cellcolor[HTML]{50A800}} \color[HTML]{F1F1F1} 0.705 & {\cellcolor[HTML]{56AB00}} \color[HTML]{F1F1F1} 0.695 & {\cellcolor[HTML]{50A800}} \color[HTML]{F1F1F1} 0.705 & {\cellcolor[HTML]{50A800}} \color[HTML]{F1F1F1} 0.705 & {\cellcolor[HTML]{50A800}} \color[HTML]{F1F1F1} 0.705 & {\cellcolor[HTML]{50A800}} \color[HTML]{F1F1F1} 0.705 \\
2k-4k & {\cellcolor[HTML]{3C9E00}} \color[HTML]{F1F1F1} 0.736 & {\cellcolor[HTML]{54AA00}} \color[HTML]{F1F1F1} 0.698 & {\cellcolor[HTML]{5AAD00}} \color[HTML]{F1F1F1} 0.689 & {\cellcolor[HTML]{5AAD00}} \color[HTML]{F1F1F1} 0.689 & {\cellcolor[HTML]{54AA00}} \color[HTML]{F1F1F1} 0.698 & {\cellcolor[HTML]{60B000}} \color[HTML]{F1F1F1} 0.679 & {\cellcolor[HTML]{54AA00}} \color[HTML]{F1F1F1} 0.698 & {\cellcolor[HTML]{54AA00}} \color[HTML]{F1F1F1} 0.698 & {\cellcolor[HTML]{54AA00}} \color[HTML]{F1F1F1} 0.698 & {\cellcolor[HTML]{4EA700}} \color[HTML]{F1F1F1} 0.708 & {\cellcolor[HTML]{54AA00}} \color[HTML]{F1F1F1} 0.698 & {\cellcolor[HTML]{5AAD00}} \color[HTML]{F1F1F1} 0.689 & {\cellcolor[HTML]{5AAD00}} \color[HTML]{F1F1F1} 0.689 & {\cellcolor[HTML]{5AAD00}} \color[HTML]{F1F1F1} 0.689 & {\cellcolor[HTML]{54AA00}} \color[HTML]{F1F1F1} 0.698 & {\cellcolor[HTML]{54AA00}} \color[HTML]{F1F1F1} 0.698 & {\cellcolor[HTML]{54AA00}} \color[HTML]{F1F1F1} 0.698 & {\cellcolor[HTML]{54AA00}} \color[HTML]{F1F1F1} 0.698 \\
4k-6k & {\cellcolor[HTML]{5CAE00}} \color[HTML]{F1F1F1} 0.685 & {\cellcolor[HTML]{7EBF00}} \color[HTML]{000000} 0.630 & {\cellcolor[HTML]{76BB00}} \color[HTML]{F1F1F1} 0.644 & {\cellcolor[HTML]{6CB600}} \color[HTML]{F1F1F1} 0.658 & {\cellcolor[HTML]{76BB00}} \color[HTML]{F1F1F1} 0.644 & {\cellcolor[HTML]{76BB00}} \color[HTML]{F1F1F1} 0.644 & {\cellcolor[HTML]{76BB00}} \color[HTML]{F1F1F1} 0.644 & {\cellcolor[HTML]{7EBF00}} \color[HTML]{000000} 0.630 & {\cellcolor[HTML]{76BB00}} \color[HTML]{F1F1F1} 0.644 & {\cellcolor[HTML]{6CB600}} \color[HTML]{F1F1F1} 0.658 & {\cellcolor[HTML]{76BB00}} \color[HTML]{F1F1F1} 0.644 & {\cellcolor[HTML]{76BB00}} \color[HTML]{F1F1F1} 0.644 & {\cellcolor[HTML]{76BB00}} \color[HTML]{F1F1F1} 0.644 & {\cellcolor[HTML]{76BB00}} \color[HTML]{F1F1F1} 0.644 & {\cellcolor[HTML]{76BB00}} \color[HTML]{F1F1F1} 0.644 & {\cellcolor[HTML]{7EBF00}} \color[HTML]{000000} 0.630 & {\cellcolor[HTML]{6CB600}} \color[HTML]{F1F1F1} 0.658 & {\cellcolor[HTML]{6CB600}} \color[HTML]{F1F1F1} 0.658 \\
6k-8k & {\cellcolor[HTML]{008000}} \color[HTML]{F1F1F1} 0.837 & {\cellcolor[HTML]{2E9700}} \color[HTML]{F1F1F1} 0.761 & {\cellcolor[HTML]{2E9700}} \color[HTML]{F1F1F1} 0.761 & {\cellcolor[HTML]{2E9700}} \color[HTML]{F1F1F1} 0.761 & {\cellcolor[HTML]{2E9700}} \color[HTML]{F1F1F1} 0.761 & {\cellcolor[HTML]{269300}} \color[HTML]{F1F1F1} 0.772 & {\cellcolor[HTML]{349A00}} \color[HTML]{F1F1F1} 0.750 & {\cellcolor[HTML]{269300}} \color[HTML]{F1F1F1} 0.772 & {\cellcolor[HTML]{269300}} \color[HTML]{F1F1F1} 0.772 & {\cellcolor[HTML]{2E9700}} \color[HTML]{F1F1F1} 0.761 & {\cellcolor[HTML]{2E9700}} \color[HTML]{F1F1F1} 0.761 & {\cellcolor[HTML]{2E9700}} \color[HTML]{F1F1F1} 0.761 & {\cellcolor[HTML]{269300}} \color[HTML]{F1F1F1} 0.772 & {\cellcolor[HTML]{269300}} \color[HTML]{F1F1F1} 0.772 & {\cellcolor[HTML]{2E9700}} \color[HTML]{F1F1F1} 0.761 & {\cellcolor[HTML]{2E9700}} \color[HTML]{F1F1F1} 0.761 & {\cellcolor[HTML]{349A00}} \color[HTML]{F1F1F1} 0.750 & {\cellcolor[HTML]{2E9700}} \color[HTML]{F1F1F1} 0.761 \\
8k-10k & {\cellcolor[HTML]{A4D200}} \color[HTML]{000000} 0.568 & {\cellcolor[HTML]{0C8600}} \color[HTML]{F1F1F1} 0.815 & {\cellcolor[HTML]{0C8600}} \color[HTML]{F1F1F1} 0.815 & {\cellcolor[HTML]{0C8600}} \color[HTML]{F1F1F1} 0.815 & {\cellcolor[HTML]{148A00}} \color[HTML]{F1F1F1} 0.802 & {\cellcolor[HTML]{148A00}} \color[HTML]{F1F1F1} 0.802 & {\cellcolor[HTML]{148A00}} \color[HTML]{F1F1F1} 0.802 & {\cellcolor[HTML]{1C8E00}} \color[HTML]{F1F1F1} 0.790 & {\cellcolor[HTML]{0C8600}} \color[HTML]{F1F1F1} 0.815 & {\cellcolor[HTML]{1C8E00}} \color[HTML]{F1F1F1} 0.790 & {\cellcolor[HTML]{148A00}} \color[HTML]{F1F1F1} 0.802 & {\cellcolor[HTML]{148A00}} \color[HTML]{F1F1F1} 0.802 & {\cellcolor[HTML]{0C8600}} \color[HTML]{F1F1F1} 0.815 & {\cellcolor[HTML]{1C8E00}} \color[HTML]{F1F1F1} 0.790 & {\cellcolor[HTML]{148A00}} \color[HTML]{F1F1F1} 0.802 & {\cellcolor[HTML]{148A00}} \color[HTML]{F1F1F1} 0.802 & {\cellcolor[HTML]{2C9600}} \color[HTML]{F1F1F1} 0.765 & {\cellcolor[HTML]{148A00}} \color[HTML]{F1F1F1} 0.802 \\
10k-12k & {\cellcolor[HTML]{FF0000}} \color[HTML]{F1F1F1} 0.000 & {\cellcolor[HTML]{369B00}} \color[HTML]{F1F1F1} 0.747 & {\cellcolor[HTML]{3C9E00}} \color[HTML]{F1F1F1} 0.736 & {\cellcolor[HTML]{369B00}} \color[HTML]{F1F1F1} 0.747 & {\cellcolor[HTML]{369B00}} \color[HTML]{F1F1F1} 0.747 & {\cellcolor[HTML]{309800}} \color[HTML]{F1F1F1} 0.758 & {\cellcolor[HTML]{369B00}} \color[HTML]{F1F1F1} 0.747 & {\cellcolor[HTML]{369B00}} \color[HTML]{F1F1F1} 0.747 & {\cellcolor[HTML]{369B00}} \color[HTML]{F1F1F1} 0.747 & {\cellcolor[HTML]{309800}} \color[HTML]{F1F1F1} 0.758 & {\cellcolor[HTML]{3C9E00}} \color[HTML]{F1F1F1} 0.736 & {\cellcolor[HTML]{369B00}} \color[HTML]{F1F1F1} 0.747 & {\cellcolor[HTML]{3C9E00}} \color[HTML]{F1F1F1} 0.736 & {\cellcolor[HTML]{369B00}} \color[HTML]{F1F1F1} 0.747 & {\cellcolor[HTML]{369B00}} \color[HTML]{F1F1F1} 0.747 & {\cellcolor[HTML]{369B00}} \color[HTML]{F1F1F1} 0.747 & {\cellcolor[HTML]{3C9E00}} \color[HTML]{F1F1F1} 0.736 & {\cellcolor[HTML]{369B00}} \color[HTML]{F1F1F1} 0.747 \\
12k-14k & {\cellcolor[HTML]{FF0000}} \color[HTML]{F1F1F1} 0.000 & {\cellcolor[HTML]{329900}} \color[HTML]{F1F1F1} 0.755 & {\cellcolor[HTML]{2A9500}} \color[HTML]{F1F1F1} 0.766 & {\cellcolor[HTML]{329900}} \color[HTML]{F1F1F1} 0.755 & {\cellcolor[HTML]{329900}} \color[HTML]{F1F1F1} 0.755 & {\cellcolor[HTML]{2A9500}} \color[HTML]{F1F1F1} 0.766 & {\cellcolor[HTML]{329900}} \color[HTML]{F1F1F1} 0.755 & {\cellcolor[HTML]{329900}} \color[HTML]{F1F1F1} 0.755 & {\cellcolor[HTML]{329900}} \color[HTML]{F1F1F1} 0.755 & {\cellcolor[HTML]{329900}} \color[HTML]{F1F1F1} 0.755 & {\cellcolor[HTML]{329900}} \color[HTML]{F1F1F1} 0.755 & {\cellcolor[HTML]{389C00}} \color[HTML]{F1F1F1} 0.745 & {\cellcolor[HTML]{329900}} \color[HTML]{F1F1F1} 0.755 & {\cellcolor[HTML]{389C00}} \color[HTML]{F1F1F1} 0.745 & {\cellcolor[HTML]{329900}} \color[HTML]{F1F1F1} 0.755 & {\cellcolor[HTML]{329900}} \color[HTML]{F1F1F1} 0.755 & {\cellcolor[HTML]{4AA500}} \color[HTML]{F1F1F1} 0.713 & {\cellcolor[HTML]{329900}} \color[HTML]{F1F1F1} 0.755 \\
14k-16k & {\cellcolor[HTML]{FF0000}} \color[HTML]{F1F1F1} 0.000 & {\cellcolor[HTML]{7EBF00}} \color[HTML]{000000} 0.629 & {\cellcolor[HTML]{7EBF00}} \color[HTML]{000000} 0.629 & {\cellcolor[HTML]{78BC00}} \color[HTML]{F1F1F1} 0.640 & {\cellcolor[HTML]{7EBF00}} \color[HTML]{000000} 0.629 & {\cellcolor[HTML]{84C200}} \color[HTML]{000000} 0.618 & {\cellcolor[HTML]{7EBF00}} \color[HTML]{000000} 0.629 & {\cellcolor[HTML]{6AB500}} \color[HTML]{F1F1F1} 0.663 & {\cellcolor[HTML]{7EBF00}} \color[HTML]{000000} 0.629 & {\cellcolor[HTML]{78BC00}} \color[HTML]{F1F1F1} 0.640 & {\cellcolor[HTML]{7EBF00}} \color[HTML]{000000} 0.629 & {\cellcolor[HTML]{6AB500}} \color[HTML]{F1F1F1} 0.663 & {\cellcolor[HTML]{70B800}} \color[HTML]{F1F1F1} 0.652 & {\cellcolor[HTML]{78BC00}} \color[HTML]{F1F1F1} 0.640 & {\cellcolor[HTML]{7EBF00}} \color[HTML]{000000} 0.629 & {\cellcolor[HTML]{7EBF00}} \color[HTML]{000000} 0.629 & {\cellcolor[HTML]{70B800}} \color[HTML]{F1F1F1} 0.652 & {\cellcolor[HTML]{7EBF00}} \color[HTML]{000000} 0.629 \\
16k-24k & {\cellcolor[HTML]{FF0000}} \color[HTML]{F1F1F1} 0.000 & {\cellcolor[HTML]{4AA500}} \color[HTML]{F1F1F1} 0.715 & {\cellcolor[HTML]{52A900}} \color[HTML]{F1F1F1} 0.701 & {\cellcolor[HTML]{42A100}} \color[HTML]{F1F1F1} 0.729 & {\cellcolor[HTML]{4EA700}} \color[HTML]{F1F1F1} 0.708 & {\cellcolor[HTML]{46A300}} \color[HTML]{F1F1F1} 0.722 & {\cellcolor[HTML]{46A300}} \color[HTML]{F1F1F1} 0.722 & {\cellcolor[HTML]{46A300}} \color[HTML]{F1F1F1} 0.722 & {\cellcolor[HTML]{46A300}} \color[HTML]{F1F1F1} 0.722 & {\cellcolor[HTML]{46A300}} \color[HTML]{F1F1F1} 0.722 & {\cellcolor[HTML]{52A900}} \color[HTML]{F1F1F1} 0.701 & {\cellcolor[HTML]{46A300}} \color[HTML]{F1F1F1} 0.722 & {\cellcolor[HTML]{4AA500}} \color[HTML]{F1F1F1} 0.715 & {\cellcolor[HTML]{4AA500}} \color[HTML]{F1F1F1} 0.715 & {\cellcolor[HTML]{46A300}} \color[HTML]{F1F1F1} 0.722 & {\cellcolor[HTML]{4AA500}} \color[HTML]{F1F1F1} 0.715 & {\cellcolor[HTML]{4AA500}} \color[HTML]{F1F1F1} 0.715 & {\cellcolor[HTML]{46A300}} \color[HTML]{F1F1F1} 0.722 \\
24k-32k & {\cellcolor[HTML]{FF0600}} \color[HTML]{F1F1F1} 0.010 & {\cellcolor[HTML]{3E9F00}} \color[HTML]{F1F1F1} 0.733 & {\cellcolor[HTML]{50A800}} \color[HTML]{F1F1F1} 0.703 & {\cellcolor[HTML]{3E9F00}} \color[HTML]{F1F1F1} 0.733 & {\cellcolor[HTML]{44A200}} \color[HTML]{F1F1F1} 0.723 & {\cellcolor[HTML]{3E9F00}} \color[HTML]{F1F1F1} 0.733 & {\cellcolor[HTML]{329900}} \color[HTML]{F1F1F1} 0.752 & {\cellcolor[HTML]{3E9F00}} \color[HTML]{F1F1F1} 0.733 & {\cellcolor[HTML]{389C00}} \color[HTML]{F1F1F1} 0.743 & {\cellcolor[HTML]{389C00}} \color[HTML]{F1F1F1} 0.743 & {\cellcolor[HTML]{44A200}} \color[HTML]{F1F1F1} 0.723 & {\cellcolor[HTML]{3E9F00}} \color[HTML]{F1F1F1} 0.733 & {\cellcolor[HTML]{389C00}} \color[HTML]{F1F1F1} 0.743 & {\cellcolor[HTML]{3E9F00}} \color[HTML]{F1F1F1} 0.733 & {\cellcolor[HTML]{389C00}} \color[HTML]{F1F1F1} 0.743 & {\cellcolor[HTML]{44A200}} \color[HTML]{F1F1F1} 0.723 & {\cellcolor[HTML]{3E9F00}} \color[HTML]{F1F1F1} 0.733 & {\cellcolor[HTML]{389C00}} \color[HTML]{F1F1F1} 0.743 \\
32k-49k & {\cellcolor[HTML]{FF0000}} \color[HTML]{F1F1F1} 0.000 & {\cellcolor[HTML]{40A000}} \color[HTML]{F1F1F1} 0.730 & {\cellcolor[HTML]{46A300}} \color[HTML]{F1F1F1} 0.722 & {\cellcolor[HTML]{3A9D00}} \color[HTML]{F1F1F1} 0.739 & {\cellcolor[HTML]{46A300}} \color[HTML]{F1F1F1} 0.722 & {\cellcolor[HTML]{4AA500}} \color[HTML]{F1F1F1} 0.713 & {\cellcolor[HTML]{40A000}} \color[HTML]{F1F1F1} 0.730 & {\cellcolor[HTML]{40A000}} \color[HTML]{F1F1F1} 0.730 & {\cellcolor[HTML]{40A000}} \color[HTML]{F1F1F1} 0.730 & {\cellcolor[HTML]{46A300}} \color[HTML]{F1F1F1} 0.722 & {\cellcolor[HTML]{3A9D00}} \color[HTML]{F1F1F1} 0.739 & {\cellcolor[HTML]{4AA500}} \color[HTML]{F1F1F1} 0.713 & {\cellcolor[HTML]{40A000}} \color[HTML]{F1F1F1} 0.730 & {\cellcolor[HTML]{3A9D00}} \color[HTML]{F1F1F1} 0.739 & {\cellcolor[HTML]{46A300}} \color[HTML]{F1F1F1} 0.722 & {\cellcolor[HTML]{40A000}} \color[HTML]{F1F1F1} 0.730 & {\cellcolor[HTML]{40A000}} \color[HTML]{F1F1F1} 0.730 & {\cellcolor[HTML]{3A9D00}} \color[HTML]{F1F1F1} 0.739 \\
49k-65k & {\cellcolor[HTML]{FF0000}} \color[HTML]{F1F1F1} 0.000 & {\cellcolor[HTML]{FF1200}} \color[HTML]{F1F1F1} 0.030 & {\cellcolor[HTML]{7CBE00}} \color[HTML]{000000} 0.634 & {\cellcolor[HTML]{FF1800}} \color[HTML]{F1F1F1} 0.040 & {\cellcolor[HTML]{FF1E00}} \color[HTML]{F1F1F1} 0.050 & {\cellcolor[HTML]{FF2400}} \color[HTML]{F1F1F1} 0.059 & {\cellcolor[HTML]{FF1800}} \color[HTML]{F1F1F1} 0.040 & {\cellcolor[HTML]{FF1800}} \color[HTML]{F1F1F1} 0.040 & {\cellcolor[HTML]{FF1200}} \color[HTML]{F1F1F1} 0.030 & {\cellcolor[HTML]{FF1800}} \color[HTML]{F1F1F1} 0.040 & {\cellcolor[HTML]{FF2400}} \color[HTML]{F1F1F1} 0.059 & {\cellcolor[HTML]{FF1800}} \color[HTML]{F1F1F1} 0.040 & {\cellcolor[HTML]{FF1200}} \color[HTML]{F1F1F1} 0.030 & {\cellcolor[HTML]{FF1200}} \color[HTML]{F1F1F1} 0.030 & {\cellcolor[HTML]{FF1800}} \color[HTML]{F1F1F1} 0.040 & {\cellcolor[HTML]{FF1200}} \color[HTML]{F1F1F1} 0.030 & {\cellcolor[HTML]{FF1200}} \color[HTML]{F1F1F1} 0.030 & {\cellcolor[HTML]{FF1200}} \color[HTML]{F1F1F1} 0.030 \\
\hline
\begin{tabular}[c]{@{}l@{}}{In-Domain}\\ {(0k-8k)}\end{tabular} & {\cellcolor[HTML]{289400}} \color[HTML]{F1F1F1} 0.771 & {\cellcolor[HTML]{4AA500}} \color[HTML]{F1F1F1} 0.714 & {\cellcolor[HTML]{4CA600}} \color[HTML]{F1F1F1} 0.711 & {\cellcolor[HTML]{48A400}} \color[HTML]{F1F1F1} 0.716 & {\cellcolor[HTML]{4AA500}} \color[HTML]{F1F1F1} 0.716 & {\cellcolor[HTML]{4AA500}} \color[HTML]{F1F1F1} 0.714 & {\cellcolor[HTML]{4AA500}} \color[HTML]{F1F1F1} 0.714 & {\cellcolor[HTML]{4AA500}} \color[HTML]{F1F1F1} 0.716 & {\cellcolor[HTML]{48A400}} \color[HTML]{F1F1F1} 0.718 & {\cellcolor[HTML]{48A400}} \color[HTML]{F1F1F1} 0.718 & {\cellcolor[HTML]{4AA500}} \color[HTML]{F1F1F1} 0.716 & {\cellcolor[HTML]{4AA500}} \color[HTML]{F1F1F1} 0.714 & {\cellcolor[HTML]{48A400}} \color[HTML]{F1F1F1} 0.716 & {\cellcolor[HTML]{4AA500}} \color[HTML]{F1F1F1} 0.714 & {\cellcolor[HTML]{4AA500}} \color[HTML]{F1F1F1} 0.716 & {\cellcolor[HTML]{4AA500}} \color[HTML]{F1F1F1} 0.714 & {\cellcolor[HTML]{4AA500}} \color[HTML]{F1F1F1} 0.716 & {\cellcolor[HTML]{48A400}} \color[HTML]{F1F1F1} 0.718 \\
\begin{tabular}[c]{@{}l@{}}{OOD}\\ {(8k-65k)}\end{tabular} & {\cellcolor[HTML]{FF2200}} \color[HTML]{F1F1F1} 0.058 & {\cellcolor[HTML]{76BB00}} \color[HTML]{F1F1F1} 0.643 & {\cellcolor[HTML]{4CA600}} \color[HTML]{F1F1F1} 0.711 & {\cellcolor[HTML]{72B900}} \color[HTML]{F1F1F1} 0.649 & {\cellcolor[HTML]{76BB00}} \color[HTML]{F1F1F1} 0.641 & {\cellcolor[HTML]{74BA00}} \color[HTML]{F1F1F1} 0.646 & {\cellcolor[HTML]{74BA00}} \color[HTML]{F1F1F1} 0.647 & {\cellcolor[HTML]{74BA00}} \color[HTML]{F1F1F1} 0.647 & {\cellcolor[HTML]{74BA00}} \color[HTML]{F1F1F1} 0.646 & {\cellcolor[HTML]{74BA00}} \color[HTML]{F1F1F1} 0.646 & {\cellcolor[HTML]{76BB00}} \color[HTML]{F1F1F1} 0.642 & {\cellcolor[HTML]{74BA00}} \color[HTML]{F1F1F1} 0.645 & {\cellcolor[HTML]{74BA00}} \color[HTML]{F1F1F1} 0.646 & {\cellcolor[HTML]{76BB00}} \color[HTML]{F1F1F1} 0.642 & {\cellcolor[HTML]{74BA00}} \color[HTML]{F1F1F1} 0.645 & {\cellcolor[HTML]{78BC00}} \color[HTML]{F1F1F1} 0.641 & {\cellcolor[HTML]{7ABD00}} \color[HTML]{F1F1F1} 0.635 & {\cellcolor[HTML]{74BA00}} \color[HTML]{F1F1F1} 0.646 \\
\begin{tabular}[c]{@{}l@{}}{Long-OOD}\\ {(32k-65k)}\end{tabular} & {\cellcolor[HTML]{FF0000}} \color[HTML]{F1F1F1} 0.000 & {\cellcolor[HTML]{FFF600}} \color[HTML]{000000} 0.403 & {\cellcolor[HTML]{5EAF00}} \color[HTML]{F1F1F1} 0.681 & {\cellcolor[HTML]{FFFC00}} \color[HTML]{000000} 0.412 & {\cellcolor[HTML]{FFF800}} \color[HTML]{000000} 0.408 & {\cellcolor[HTML]{FFF800}} \color[HTML]{000000} 0.407 & {\cellcolor[HTML]{FFF800}} \color[HTML]{000000} 0.407 & {\cellcolor[HTML]{FFF800}} \color[HTML]{000000} 0.407 & {\cellcolor[HTML]{FFF600}} \color[HTML]{000000} 0.403 & {\cellcolor[HTML]{FFF600}} \color[HTML]{000000} 0.403 & {\cellcolor[HTML]{FEFF00}} \color[HTML]{000000} 0.421 & {\cellcolor[HTML]{FFF200}} \color[HTML]{000000} 0.398 & {\cellcolor[HTML]{FFF600}} \color[HTML]{000000} 0.403 & {\cellcolor[HTML]{FFF800}} \color[HTML]{000000} 0.407 & {\cellcolor[HTML]{FFF600}} \color[HTML]{000000} 0.403 & {\cellcolor[HTML]{FFF600}} \color[HTML]{000000} 0.403 & {\cellcolor[HTML]{FFF600}} \color[HTML]{000000} 0.403 & {\cellcolor[HTML]{FFF800}} \color[HTML]{000000} 0.407 \\
\begin{tabular}[c]{@{}l@{}}{Full}\\ {(0k-65k)}\end{tabular} & {\cellcolor[HTML]{FFBE00}} \color[HTML]{000000} 0.313 & {\cellcolor[HTML]{66B300}} \color[HTML]{F1F1F1} 0.668 & {\cellcolor[HTML]{4CA600}} \color[HTML]{F1F1F1} 0.711 & {\cellcolor[HTML]{64B200}} \color[HTML]{F1F1F1} 0.673 & {\cellcolor[HTML]{66B300}} \color[HTML]{F1F1F1} 0.668 & {\cellcolor[HTML]{66B300}} \color[HTML]{F1F1F1} 0.670 & {\cellcolor[HTML]{64B200}} \color[HTML]{F1F1F1} 0.671 & {\cellcolor[HTML]{64B200}} \color[HTML]{F1F1F1} 0.672 & {\cellcolor[HTML]{64B200}} \color[HTML]{F1F1F1} 0.672 & {\cellcolor[HTML]{64B200}} \color[HTML]{F1F1F1} 0.672 & {\cellcolor[HTML]{66B300}} \color[HTML]{F1F1F1} 0.668 & {\cellcolor[HTML]{66B300}} \color[HTML]{F1F1F1} 0.669 & {\cellcolor[HTML]{64B200}} \color[HTML]{F1F1F1} 0.671 & {\cellcolor[HTML]{66B300}} \color[HTML]{F1F1F1} 0.668 & {\cellcolor[HTML]{66B300}} \color[HTML]{F1F1F1} 0.670 & {\cellcolor[HTML]{68B400}} \color[HTML]{F1F1F1} 0.667 & {\cellcolor[HTML]{68B400}} \color[HTML]{F1F1F1} 0.664 & {\cellcolor[HTML]{64B200}} \color[HTML]{F1F1F1} 0.672 \\
\bottomrule
\end{tabular}}
\caption{Ablation for all associative layers on the MT dataset for ARMT with SmolLM-2-360M-IT model, metric - EM. Ablated associative layers from 0 to 15.}
\label{tab:mt_best_all_layers_assoc_ablation_0_15_smollm}
\end{center}
\end{table*}
\begin{table*}[h]

\begin{center}
\resizebox{1.0\textwidth}{!}{
\begin{tabular}{lcccccccccccccccccc}
\toprule
\begin{tabular}[c]{@{}l@{}}\textbf{Model\//}\\ \textbf{Lengths}\end{tabular} & \begin{tabular}[c]{@{}l@{}}\textbf{Base,}\\ \textbf{MT,}\\ \textbf{8k}\end{tabular} & \begin{tabular}[c]{@{}l@{}}\textbf{ARMT,}\\ \textbf{MT,}\\ \textbf{8k}\end{tabular} & \begin{tabular}[c]{@{}l@{}}\textbf{W/o}\\ \textbf{layer}\\ \textbf{16}\end{tabular} & \begin{tabular}[c]{@{}l@{}}\textbf{W/o}\\ \textbf{layer}\\ \textbf{17}\end{tabular} & \begin{tabular}[c]{@{}l@{}}\textbf{W/o}\\ \textbf{layer}\\ \textbf{18}\end{tabular} & \begin{tabular}[c]{@{}l@{}}\textbf{W/o}\\ \textbf{layer}\\ \textbf{19}\end{tabular} & \begin{tabular}[c]{@{}l@{}}\textbf{W/o}\\ \textbf{layer}\\ \textbf{20}\end{tabular} & \begin{tabular}[c]{@{}l@{}}\textbf{W/o}\\ \textbf{layer}\\ \textbf{21}\end{tabular} & \begin{tabular}[c]{@{}l@{}}\textbf{W/o}\\ \textbf{layer}\\ \textbf{22}\end{tabular} & \begin{tabular}[c]{@{}l@{}}\textbf{W/o}\\ \textbf{layer}\\ \textbf{23}\end{tabular} & \begin{tabular}[c]{@{}l@{}}\textbf{W/o}\\ \textbf{layer}\\ \textbf{24}\end{tabular} & \begin{tabular}[c]{@{}l@{}}\textbf{W/o}\\ \textbf{layer}\\ \textbf{25}\end{tabular} & \begin{tabular}[c]{@{}l@{}}\textbf{W/o}\\ \textbf{layer}\\ \textbf{26}\end{tabular} & \begin{tabular}[c]{@{}l@{}}\textbf{W/o}\\ \textbf{layer}\\ \textbf{27}\end{tabular} & \begin{tabular}[c]{@{}l@{}}\textbf{W/o}\\ \textbf{layer}\\ \textbf{28}\end{tabular} & \begin{tabular}[c]{@{}l@{}}\textbf{W/o}\\ \textbf{layer}\\ \textbf{29}\end{tabular} & \begin{tabular}[c]{@{}l@{}}\textbf{W/o}\\ \textbf{layer}\\ \textbf{30}\end{tabular} & \begin{tabular}[c]{@{}l@{}}\textbf{W/o}\\ \textbf{layer}\\ \textbf{31}\end{tabular}\\
\midrule
0k-1k & {\cellcolor[HTML]{188C00}} \color[HTML]{F1F1F1} 0.795 & {\cellcolor[HTML]{289400}} \color[HTML]{F1F1F1} 0.769 & {\cellcolor[HTML]{289400}} \color[HTML]{F1F1F1} 0.769 & {\cellcolor[HTML]{289400}} \color[HTML]{F1F1F1} 0.769 & {\cellcolor[HTML]{289400}} \color[HTML]{F1F1F1} 0.769 & {\cellcolor[HTML]{289400}} \color[HTML]{F1F1F1} 0.769 & {\cellcolor[HTML]{289400}} \color[HTML]{F1F1F1} 0.769 & {\cellcolor[HTML]{289400}} \color[HTML]{F1F1F1} 0.769 & {\cellcolor[HTML]{289400}} \color[HTML]{F1F1F1} 0.769 & {\cellcolor[HTML]{309800}} \color[HTML]{F1F1F1} 0.756 & {\cellcolor[HTML]{289400}} \color[HTML]{F1F1F1} 0.769 & {\cellcolor[HTML]{289400}} \color[HTML]{F1F1F1} 0.769 & {\cellcolor[HTML]{289400}} \color[HTML]{F1F1F1} 0.769 & {\cellcolor[HTML]{289400}} \color[HTML]{F1F1F1} 0.769 & {\cellcolor[HTML]{309800}} \color[HTML]{F1F1F1} 0.756 & {\cellcolor[HTML]{289400}} \color[HTML]{F1F1F1} 0.769 & {\cellcolor[HTML]{309800}} \color[HTML]{F1F1F1} 0.756 & {\cellcolor[HTML]{289400}} \color[HTML]{F1F1F1} 0.769 \\
1k-2k & {\cellcolor[HTML]{1C8E00}} \color[HTML]{F1F1F1} 0.790 & {\cellcolor[HTML]{50A800}} \color[HTML]{F1F1F1} 0.705 & {\cellcolor[HTML]{50A800}} \color[HTML]{F1F1F1} 0.705 & {\cellcolor[HTML]{50A800}} \color[HTML]{F1F1F1} 0.705 & {\cellcolor[HTML]{50A800}} \color[HTML]{F1F1F1} 0.705 & {\cellcolor[HTML]{50A800}} \color[HTML]{F1F1F1} 0.705 & {\cellcolor[HTML]{50A800}} \color[HTML]{F1F1F1} 0.705 & {\cellcolor[HTML]{50A800}} \color[HTML]{F1F1F1} 0.705 & {\cellcolor[HTML]{50A800}} \color[HTML]{F1F1F1} 0.705 & {\cellcolor[HTML]{ECF600}} \color[HTML]{000000} 0.448 & {\cellcolor[HTML]{66B300}} \color[HTML]{F1F1F1} 0.667 & {\cellcolor[HTML]{4AA500}} \color[HTML]{F1F1F1} 0.714 & {\cellcolor[HTML]{50A800}} \color[HTML]{F1F1F1} 0.705 & {\cellcolor[HTML]{66B300}} \color[HTML]{F1F1F1} 0.667 & {\cellcolor[HTML]{5CAE00}} \color[HTML]{F1F1F1} 0.686 & {\cellcolor[HTML]{6EB700}} \color[HTML]{F1F1F1} 0.657 & {\cellcolor[HTML]{5CAE00}} \color[HTML]{F1F1F1} 0.686 & {\cellcolor[HTML]{56AB00}} \color[HTML]{F1F1F1} 0.695 \\
2k-4k & {\cellcolor[HTML]{3C9E00}} \color[HTML]{F1F1F1} 0.736 & {\cellcolor[HTML]{54AA00}} \color[HTML]{F1F1F1} 0.698 & {\cellcolor[HTML]{54AA00}} \color[HTML]{F1F1F1} 0.698 & {\cellcolor[HTML]{5AAD00}} \color[HTML]{F1F1F1} 0.689 & {\cellcolor[HTML]{54AA00}} \color[HTML]{F1F1F1} 0.698 & {\cellcolor[HTML]{54AA00}} \color[HTML]{F1F1F1} 0.698 & {\cellcolor[HTML]{54AA00}} \color[HTML]{F1F1F1} 0.698 & {\cellcolor[HTML]{54AA00}} \color[HTML]{F1F1F1} 0.698 & {\cellcolor[HTML]{5AAD00}} \color[HTML]{F1F1F1} 0.689 & {\cellcolor[HTML]{FFFC00}} \color[HTML]{000000} 0.415 & {\cellcolor[HTML]{54AA00}} \color[HTML]{F1F1F1} 0.698 & {\cellcolor[HTML]{54AA00}} \color[HTML]{F1F1F1} 0.698 & {\cellcolor[HTML]{54AA00}} \color[HTML]{F1F1F1} 0.698 & {\cellcolor[HTML]{54AA00}} \color[HTML]{F1F1F1} 0.698 & {\cellcolor[HTML]{66B300}} \color[HTML]{F1F1F1} 0.670 & {\cellcolor[HTML]{4EA700}} \color[HTML]{F1F1F1} 0.708 & {\cellcolor[HTML]{54AA00}} \color[HTML]{F1F1F1} 0.698 & {\cellcolor[HTML]{54AA00}} \color[HTML]{F1F1F1} 0.698 \\
4k-6k & {\cellcolor[HTML]{5CAE00}} \color[HTML]{F1F1F1} 0.685 & {\cellcolor[HTML]{7EBF00}} \color[HTML]{000000} 0.630 & {\cellcolor[HTML]{76BB00}} \color[HTML]{F1F1F1} 0.644 & {\cellcolor[HTML]{6CB600}} \color[HTML]{F1F1F1} 0.658 & {\cellcolor[HTML]{6CB600}} \color[HTML]{F1F1F1} 0.658 & {\cellcolor[HTML]{76BB00}} \color[HTML]{F1F1F1} 0.644 & {\cellcolor[HTML]{6CB600}} \color[HTML]{F1F1F1} 0.658 & {\cellcolor[HTML]{76BB00}} \color[HTML]{F1F1F1} 0.644 & {\cellcolor[HTML]{7EBF00}} \color[HTML]{000000} 0.630 & {\cellcolor[HTML]{FFE200}} \color[HTML]{000000} 0.370 & {\cellcolor[HTML]{86C300}} \color[HTML]{000000} 0.616 & {\cellcolor[HTML]{76BB00}} \color[HTML]{F1F1F1} 0.644 & {\cellcolor[HTML]{7EBF00}} \color[HTML]{000000} 0.630 & {\cellcolor[HTML]{6CB600}} \color[HTML]{F1F1F1} 0.658 & {\cellcolor[HTML]{6CB600}} \color[HTML]{F1F1F1} 0.658 & {\cellcolor[HTML]{6CB600}} \color[HTML]{F1F1F1} 0.658 & {\cellcolor[HTML]{7EBF00}} \color[HTML]{000000} 0.630 & {\cellcolor[HTML]{7EBF00}} \color[HTML]{000000} 0.630 \\
6k-8k & {\cellcolor[HTML]{008000}} \color[HTML]{F1F1F1} 0.837 & {\cellcolor[HTML]{2E9700}} \color[HTML]{F1F1F1} 0.761 & {\cellcolor[HTML]{2E9700}} \color[HTML]{F1F1F1} 0.761 & {\cellcolor[HTML]{48A400}} \color[HTML]{F1F1F1} 0.717 & {\cellcolor[HTML]{2E9700}} \color[HTML]{F1F1F1} 0.761 & {\cellcolor[HTML]{2E9700}} \color[HTML]{F1F1F1} 0.761 & {\cellcolor[HTML]{2E9700}} \color[HTML]{F1F1F1} 0.761 & {\cellcolor[HTML]{148A00}} \color[HTML]{F1F1F1} 0.804 & {\cellcolor[HTML]{2E9700}} \color[HTML]{F1F1F1} 0.761 & {\cellcolor[HTML]{FFD400}} \color[HTML]{000000} 0.348 & {\cellcolor[HTML]{56AB00}} \color[HTML]{F1F1F1} 0.696 & {\cellcolor[HTML]{2E9700}} \color[HTML]{F1F1F1} 0.761 & {\cellcolor[HTML]{2E9700}} \color[HTML]{F1F1F1} 0.761 & {\cellcolor[HTML]{349A00}} \color[HTML]{F1F1F1} 0.750 & {\cellcolor[HTML]{0C8600}} \color[HTML]{F1F1F1} 0.815 & {\cellcolor[HTML]{209000}} \color[HTML]{F1F1F1} 0.783 & {\cellcolor[HTML]{2E9700}} \color[HTML]{F1F1F1} 0.761 & {\cellcolor[HTML]{2E9700}} \color[HTML]{F1F1F1} 0.761 \\
8k-10k & {\cellcolor[HTML]{A4D200}} \color[HTML]{000000} 0.568 & {\cellcolor[HTML]{0C8600}} \color[HTML]{F1F1F1} 0.815 & {\cellcolor[HTML]{0C8600}} \color[HTML]{F1F1F1} 0.815 & {\cellcolor[HTML]{3A9D00}} \color[HTML]{F1F1F1} 0.741 & {\cellcolor[HTML]{148A00}} \color[HTML]{F1F1F1} 0.802 & {\cellcolor[HTML]{148A00}} \color[HTML]{F1F1F1} 0.802 & {\cellcolor[HTML]{0C8600}} \color[HTML]{F1F1F1} 0.815 & {\cellcolor[HTML]{148A00}} \color[HTML]{F1F1F1} 0.802 & {\cellcolor[HTML]{148A00}} \color[HTML]{F1F1F1} 0.802 & {\cellcolor[HTML]{FFB400}} \color[HTML]{000000} 0.296 & {\cellcolor[HTML]{329900}} \color[HTML]{F1F1F1} 0.753 & {\cellcolor[HTML]{0C8600}} \color[HTML]{F1F1F1} 0.815 & {\cellcolor[HTML]{0C8600}} \color[HTML]{F1F1F1} 0.815 & {\cellcolor[HTML]{148A00}} \color[HTML]{F1F1F1} 0.802 & {\cellcolor[HTML]{0C8600}} \color[HTML]{F1F1F1} 0.815 & {\cellcolor[HTML]{148A00}} \color[HTML]{F1F1F1} 0.802 & {\cellcolor[HTML]{0C8600}} \color[HTML]{F1F1F1} 0.815 & {\cellcolor[HTML]{148A00}} \color[HTML]{F1F1F1} 0.802 \\
10k-12k & {\cellcolor[HTML]{FF0000}} \color[HTML]{F1F1F1} 0.000 & {\cellcolor[HTML]{369B00}} \color[HTML]{F1F1F1} 0.747 & {\cellcolor[HTML]{3C9E00}} \color[HTML]{F1F1F1} 0.736 & {\cellcolor[HTML]{50A800}} \color[HTML]{F1F1F1} 0.703 & {\cellcolor[HTML]{369B00}} \color[HTML]{F1F1F1} 0.747 & {\cellcolor[HTML]{3C9E00}} \color[HTML]{F1F1F1} 0.736 & {\cellcolor[HTML]{369B00}} \color[HTML]{F1F1F1} 0.747 & {\cellcolor[HTML]{3C9E00}} \color[HTML]{F1F1F1} 0.736 & {\cellcolor[HTML]{369B00}} \color[HTML]{F1F1F1} 0.747 & {\cellcolor[HTML]{FFF800}} \color[HTML]{000000} 0.407 & {\cellcolor[HTML]{44A200}} \color[HTML]{F1F1F1} 0.725 & {\cellcolor[HTML]{44A200}} \color[HTML]{F1F1F1} 0.725 & {\cellcolor[HTML]{44A200}} \color[HTML]{F1F1F1} 0.725 & {\cellcolor[HTML]{289400}} \color[HTML]{F1F1F1} 0.769 & {\cellcolor[HTML]{44A200}} \color[HTML]{F1F1F1} 0.725 & {\cellcolor[HTML]{309800}} \color[HTML]{F1F1F1} 0.758 & {\cellcolor[HTML]{3C9E00}} \color[HTML]{F1F1F1} 0.736 & {\cellcolor[HTML]{309800}} \color[HTML]{F1F1F1} 0.758 \\
12k-14k & {\cellcolor[HTML]{FF0000}} \color[HTML]{F1F1F1} 0.000 & {\cellcolor[HTML]{329900}} \color[HTML]{F1F1F1} 0.755 & {\cellcolor[HTML]{329900}} \color[HTML]{F1F1F1} 0.755 & {\cellcolor[HTML]{329900}} \color[HTML]{F1F1F1} 0.755 & {\cellcolor[HTML]{329900}} \color[HTML]{F1F1F1} 0.755 & {\cellcolor[HTML]{329900}} \color[HTML]{F1F1F1} 0.755 & {\cellcolor[HTML]{329900}} \color[HTML]{F1F1F1} 0.755 & {\cellcolor[HTML]{329900}} \color[HTML]{F1F1F1} 0.755 & {\cellcolor[HTML]{329900}} \color[HTML]{F1F1F1} 0.755 & {\cellcolor[HTML]{FFB600}} \color[HTML]{000000} 0.298 & {\cellcolor[HTML]{389C00}} \color[HTML]{F1F1F1} 0.745 & {\cellcolor[HTML]{389C00}} \color[HTML]{F1F1F1} 0.745 & {\cellcolor[HTML]{2A9500}} \color[HTML]{F1F1F1} 0.766 & {\cellcolor[HTML]{329900}} \color[HTML]{F1F1F1} 0.755 & {\cellcolor[HTML]{3E9F00}} \color[HTML]{F1F1F1} 0.734 & {\cellcolor[HTML]{2A9500}} \color[HTML]{F1F1F1} 0.766 & {\cellcolor[HTML]{2A9500}} \color[HTML]{F1F1F1} 0.766 & {\cellcolor[HTML]{389C00}} \color[HTML]{F1F1F1} 0.745 \\
14k-16k & {\cellcolor[HTML]{FF0000}} \color[HTML]{F1F1F1} 0.000 & {\cellcolor[HTML]{7EBF00}} \color[HTML]{000000} 0.629 & {\cellcolor[HTML]{7EBF00}} \color[HTML]{000000} 0.629 & {\cellcolor[HTML]{70B800}} \color[HTML]{F1F1F1} 0.652 & {\cellcolor[HTML]{7EBF00}} \color[HTML]{000000} 0.629 & {\cellcolor[HTML]{84C200}} \color[HTML]{000000} 0.618 & {\cellcolor[HTML]{7EBF00}} \color[HTML]{000000} 0.629 & {\cellcolor[HTML]{7EBF00}} \color[HTML]{000000} 0.629 & {\cellcolor[HTML]{7EBF00}} \color[HTML]{000000} 0.629 & {\cellcolor[HTML]{FFC600}} \color[HTML]{000000} 0.326 & {\cellcolor[HTML]{78BC00}} \color[HTML]{F1F1F1} 0.640 & {\cellcolor[HTML]{7EBF00}} \color[HTML]{000000} 0.629 & {\cellcolor[HTML]{7EBF00}} \color[HTML]{000000} 0.629 & {\cellcolor[HTML]{84C200}} \color[HTML]{000000} 0.618 & {\cellcolor[HTML]{6AB500}} \color[HTML]{F1F1F1} 0.663 & {\cellcolor[HTML]{7EBF00}} \color[HTML]{000000} 0.629 & {\cellcolor[HTML]{7EBF00}} \color[HTML]{000000} 0.629 & {\cellcolor[HTML]{7EBF00}} \color[HTML]{000000} 0.629 \\
16k-24k & {\cellcolor[HTML]{FF0000}} \color[HTML]{F1F1F1} 0.000 & {\cellcolor[HTML]{4AA500}} \color[HTML]{F1F1F1} 0.715 & {\cellcolor[HTML]{4AA500}} \color[HTML]{F1F1F1} 0.715 & {\cellcolor[HTML]{5EAF00}} \color[HTML]{F1F1F1} 0.681 & {\cellcolor[HTML]{46A300}} \color[HTML]{F1F1F1} 0.722 & {\cellcolor[HTML]{4EA700}} \color[HTML]{F1F1F1} 0.708 & {\cellcolor[HTML]{4EA700}} \color[HTML]{F1F1F1} 0.708 & {\cellcolor[HTML]{4EA700}} \color[HTML]{F1F1F1} 0.708 & {\cellcolor[HTML]{4AA500}} \color[HTML]{F1F1F1} 0.715 & {\cellcolor[HTML]{FF9000}} \color[HTML]{000000} 0.236 & {\cellcolor[HTML]{5EAF00}} \color[HTML]{F1F1F1} 0.681 & {\cellcolor[HTML]{4AA500}} \color[HTML]{F1F1F1} 0.715 & {\cellcolor[HTML]{46A300}} \color[HTML]{F1F1F1} 0.722 & {\cellcolor[HTML]{46A300}} \color[HTML]{F1F1F1} 0.722 & {\cellcolor[HTML]{4AA500}} \color[HTML]{F1F1F1} 0.715 & {\cellcolor[HTML]{4EA700}} \color[HTML]{F1F1F1} 0.708 & {\cellcolor[HTML]{42A100}} \color[HTML]{F1F1F1} 0.729 & {\cellcolor[HTML]{4AA500}} \color[HTML]{F1F1F1} 0.715 \\
24k-32k & {\cellcolor[HTML]{FF0600}} \color[HTML]{F1F1F1} 0.010 & {\cellcolor[HTML]{3E9F00}} \color[HTML]{F1F1F1} 0.733 & {\cellcolor[HTML]{3E9F00}} \color[HTML]{F1F1F1} 0.733 & {\cellcolor[HTML]{58AC00}} \color[HTML]{F1F1F1} 0.693 & {\cellcolor[HTML]{3E9F00}} \color[HTML]{F1F1F1} 0.733 & {\cellcolor[HTML]{389C00}} \color[HTML]{F1F1F1} 0.743 & {\cellcolor[HTML]{389C00}} \color[HTML]{F1F1F1} 0.743 & {\cellcolor[HTML]{3E9F00}} \color[HTML]{F1F1F1} 0.733 & {\cellcolor[HTML]{44A200}} \color[HTML]{F1F1F1} 0.723 & {\cellcolor[HTML]{FFA800}} \color[HTML]{000000} 0.277 & {\cellcolor[HTML]{58AC00}} \color[HTML]{F1F1F1} 0.693 & {\cellcolor[HTML]{44A200}} \color[HTML]{F1F1F1} 0.723 & {\cellcolor[HTML]{44A200}} \color[HTML]{F1F1F1} 0.723 & {\cellcolor[HTML]{3E9F00}} \color[HTML]{F1F1F1} 0.733 & {\cellcolor[HTML]{4AA500}} \color[HTML]{F1F1F1} 0.713 & {\cellcolor[HTML]{3E9F00}} \color[HTML]{F1F1F1} 0.733 & {\cellcolor[HTML]{3E9F00}} \color[HTML]{F1F1F1} 0.733 & {\cellcolor[HTML]{3E9F00}} \color[HTML]{F1F1F1} 0.733 \\
32k-49k & {\cellcolor[HTML]{FF0000}} \color[HTML]{F1F1F1} 0.000 & {\cellcolor[HTML]{40A000}} \color[HTML]{F1F1F1} 0.730 & {\cellcolor[HTML]{46A300}} \color[HTML]{F1F1F1} 0.722 & {\cellcolor[HTML]{66B300}} \color[HTML]{F1F1F1} 0.670 & {\cellcolor[HTML]{46A300}} \color[HTML]{F1F1F1} 0.722 & {\cellcolor[HTML]{4AA500}} \color[HTML]{F1F1F1} 0.713 & {\cellcolor[HTML]{40A000}} \color[HTML]{F1F1F1} 0.730 & {\cellcolor[HTML]{40A000}} \color[HTML]{F1F1F1} 0.730 & {\cellcolor[HTML]{40A000}} \color[HTML]{F1F1F1} 0.730 & {\cellcolor[HTML]{FF9400}} \color[HTML]{000000} 0.243 & {\cellcolor[HTML]{50A800}} \color[HTML]{F1F1F1} 0.704 & {\cellcolor[HTML]{40A000}} \color[HTML]{F1F1F1} 0.730 & {\cellcolor[HTML]{40A000}} \color[HTML]{F1F1F1} 0.730 & {\cellcolor[HTML]{40A000}} \color[HTML]{F1F1F1} 0.730 & {\cellcolor[HTML]{46A300}} \color[HTML]{F1F1F1} 0.722 & {\cellcolor[HTML]{40A000}} \color[HTML]{F1F1F1} 0.730 & {\cellcolor[HTML]{40A000}} \color[HTML]{F1F1F1} 0.730 & {\cellcolor[HTML]{3A9D00}} \color[HTML]{F1F1F1} 0.739 \\
49k-65k & {\cellcolor[HTML]{FF0000}} \color[HTML]{F1F1F1} 0.000 & {\cellcolor[HTML]{FF1200}} \color[HTML]{F1F1F1} 0.030 & {\cellcolor[HTML]{FF1200}} \color[HTML]{F1F1F1} 0.030 & {\cellcolor[HTML]{FF1200}} \color[HTML]{F1F1F1} 0.030 & {\cellcolor[HTML]{FF1200}} \color[HTML]{F1F1F1} 0.030 & {\cellcolor[HTML]{FF1200}} \color[HTML]{F1F1F1} 0.030 & {\cellcolor[HTML]{FF2400}} \color[HTML]{F1F1F1} 0.059 & {\cellcolor[HTML]{FF1200}} \color[HTML]{F1F1F1} 0.030 & {\cellcolor[HTML]{FF1E00}} \color[HTML]{F1F1F1} 0.050 & {\cellcolor[HTML]{FF1200}} \color[HTML]{F1F1F1} 0.030 & {\cellcolor[HTML]{FF1200}} \color[HTML]{F1F1F1} 0.030 & {\cellcolor[HTML]{FF1200}} \color[HTML]{F1F1F1} 0.030 & {\cellcolor[HTML]{FF1200}} \color[HTML]{F1F1F1} 0.030 & {\cellcolor[HTML]{FF1200}} \color[HTML]{F1F1F1} 0.030 & {\cellcolor[HTML]{FF1200}} \color[HTML]{F1F1F1} 0.030 & {\cellcolor[HTML]{FF1800}} \color[HTML]{F1F1F1} 0.040 & {\cellcolor[HTML]{FF1200}} \color[HTML]{F1F1F1} 0.030 & {\cellcolor[HTML]{FF1200}} \color[HTML]{F1F1F1} 0.030 \\
\hline
\begin{tabular}[c]{@{}l@{}}{In-Domain}\\ {(0k-8k)}\end{tabular} & {\cellcolor[HTML]{289400}} \color[HTML]{F1F1F1} 0.771 & {\cellcolor[HTML]{4AA500}} \color[HTML]{F1F1F1} 0.714 & {\cellcolor[HTML]{4AA500}} \color[HTML]{F1F1F1} 0.716 & {\cellcolor[HTML]{4EA700}} \color[HTML]{F1F1F1} 0.707 & {\cellcolor[HTML]{48A400}} \color[HTML]{F1F1F1} 0.718 & {\cellcolor[HTML]{4AA500}} \color[HTML]{F1F1F1} 0.716 & {\cellcolor[HTML]{48A400}} \color[HTML]{F1F1F1} 0.718 & {\cellcolor[HTML]{44A200}} \color[HTML]{F1F1F1} 0.725 & {\cellcolor[HTML]{4CA600}} \color[HTML]{F1F1F1} 0.712 & {\cellcolor[HTML]{E6F300}} \color[HTML]{000000} 0.460 & {\cellcolor[HTML]{5AAD00}} \color[HTML]{F1F1F1} 0.689 & {\cellcolor[HTML]{48A400}} \color[HTML]{F1F1F1} 0.718 & {\cellcolor[HTML]{4AA500}} \color[HTML]{F1F1F1} 0.714 & {\cellcolor[HTML]{4EA700}} \color[HTML]{F1F1F1} 0.707 & {\cellcolor[HTML]{4AA500}} \color[HTML]{F1F1F1} 0.716 & {\cellcolor[HTML]{4AA500}} \color[HTML]{F1F1F1} 0.714 & {\cellcolor[HTML]{4EA700}} \color[HTML]{F1F1F1} 0.707 & {\cellcolor[HTML]{4CA600}} \color[HTML]{F1F1F1} 0.711 \\
\begin{tabular}[c]{@{}l@{}}{OOD}\\ {(8k-65k)}\end{tabular} & {\cellcolor[HTML]{FF2200}} \color[HTML]{F1F1F1} 0.058 & {\cellcolor[HTML]{76BB00}} \color[HTML]{F1F1F1} 0.643 & {\cellcolor[HTML]{76BB00}} \color[HTML]{F1F1F1} 0.641 & {\cellcolor[HTML]{88C400}} \color[HTML]{000000} 0.614 & {\cellcolor[HTML]{76BB00}} \color[HTML]{F1F1F1} 0.642 & {\cellcolor[HTML]{7ABD00}} \color[HTML]{F1F1F1} 0.637 & {\cellcolor[HTML]{74BA00}} \color[HTML]{F1F1F1} 0.647 & {\cellcolor[HTML]{78BC00}} \color[HTML]{F1F1F1} 0.640 & {\cellcolor[HTML]{76BB00}} \color[HTML]{F1F1F1} 0.643 & {\cellcolor[HTML]{FF9E00}} \color[HTML]{000000} 0.259 & {\cellcolor[HTML]{84C200}} \color[HTML]{000000} 0.620 & {\cellcolor[HTML]{78BC00}} \color[HTML]{F1F1F1} 0.638 & {\cellcolor[HTML]{76BB00}} \color[HTML]{F1F1F1} 0.642 & {\cellcolor[HTML]{74BA00}} \color[HTML]{F1F1F1} 0.644 & {\cellcolor[HTML]{78BC00}} \color[HTML]{F1F1F1} 0.639 & {\cellcolor[HTML]{74BA00}} \color[HTML]{F1F1F1} 0.644 & {\cellcolor[HTML]{74BA00}} \color[HTML]{F1F1F1} 0.646 & {\cellcolor[HTML]{76BB00}} \color[HTML]{F1F1F1} 0.643 \\
\begin{tabular}[c]{@{}l@{}}{Long-OOD}\\ {(32k-65k)}\end{tabular} & {\cellcolor[HTML]{FF0000}} \color[HTML]{F1F1F1} 0.000 & {\cellcolor[HTML]{FFF600}} \color[HTML]{000000} 0.403 & {\cellcolor[HTML]{FFF200}} \color[HTML]{000000} 0.398 & {\cellcolor[HTML]{FFE200}} \color[HTML]{000000} 0.371 & {\cellcolor[HTML]{FFF200}} \color[HTML]{000000} 0.398 & {\cellcolor[HTML]{FFF000}} \color[HTML]{000000} 0.394 & {\cellcolor[HTML]{FFFE00}} \color[HTML]{000000} 0.416 & {\cellcolor[HTML]{FFF600}} \color[HTML]{000000} 0.403 & {\cellcolor[HTML]{FFFC00}} \color[HTML]{000000} 0.412 & {\cellcolor[HTML]{FF5600}} \color[HTML]{F1F1F1} 0.143 & {\cellcolor[HTML]{FFEC00}} \color[HTML]{000000} 0.389 & {\cellcolor[HTML]{FFF600}} \color[HTML]{000000} 0.403 & {\cellcolor[HTML]{FFF600}} \color[HTML]{000000} 0.403 & {\cellcolor[HTML]{FFF600}} \color[HTML]{000000} 0.403 & {\cellcolor[HTML]{FFF200}} \color[HTML]{000000} 0.398 & {\cellcolor[HTML]{FFF800}} \color[HTML]{000000} 0.407 & {\cellcolor[HTML]{FFF600}} \color[HTML]{000000} 0.403 & {\cellcolor[HTML]{FFF800}} \color[HTML]{000000} 0.407 \\
\begin{tabular}[c]{@{}l@{}}{Full}\\ {(0k-65k)}\end{tabular} & {\cellcolor[HTML]{FFBE00}} \color[HTML]{000000} 0.313 & {\cellcolor[HTML]{66B300}} \color[HTML]{F1F1F1} 0.668 & {\cellcolor[HTML]{66B300}} \color[HTML]{F1F1F1} 0.668 & {\cellcolor[HTML]{72B900}} \color[HTML]{F1F1F1} 0.647 & {\cellcolor[HTML]{66B300}} \color[HTML]{F1F1F1} 0.669 & {\cellcolor[HTML]{68B400}} \color[HTML]{F1F1F1} 0.665 & {\cellcolor[HTML]{64B200}} \color[HTML]{F1F1F1} 0.672 & {\cellcolor[HTML]{66B300}} \color[HTML]{F1F1F1} 0.670 & {\cellcolor[HTML]{66B300}} \color[HTML]{F1F1F1} 0.668 & {\cellcolor[HTML]{FFCA00}} \color[HTML]{000000} 0.331 & {\cellcolor[HTML]{74BA00}} \color[HTML]{F1F1F1} 0.645 & {\cellcolor[HTML]{68B400}} \color[HTML]{F1F1F1} 0.667 & {\cellcolor[HTML]{66B300}} \color[HTML]{F1F1F1} 0.668 & {\cellcolor[HTML]{68B400}} \color[HTML]{F1F1F1} 0.667 & {\cellcolor[HTML]{68B400}} \color[HTML]{F1F1F1} 0.666 & {\cellcolor[HTML]{66B300}} \color[HTML]{F1F1F1} 0.669 & {\cellcolor[HTML]{66B300}} \color[HTML]{F1F1F1} 0.668 & {\cellcolor[HTML]{66B300}} \color[HTML]{F1F1F1} 0.668 \\
\bottomrule
\end{tabular}}
\caption{Ablation for all associative layers on the MT dataset for ARMT with SmolLM-2-360M-IT model, metric - EM. Ablated associative layers from 16 to 31.}
\label{tab:mt_best_all_layers_assoc_ablation_16_32_smollm}
\end{center}
\end{table*}

As was done for Gemma-3-1B-IT, we also show the training dynamics during the curriculum learning for ARMT with the SmolLM-2-360M-IT backbone with \emph{pre-selected associative blocks}, the results are presented in~\Cref{tab:mt_best_assoc_ablation_with_train_full_smollm,tab:gov_report_best_assoc_ablation_with_train_full_smollm}. The pre-selected blocks use the same universal pattern as described for Gemma-3-1B-IT, adding associative layers on last layer, two middle layers, two layers between third and fourth quarter, and one layer between first and second quarter. Here we added one more layer between the third and fourth quarters, as the SmolLM-2-360M-IT model has 32 layers instead of 26 for Gemma-3-1B-IT, resulting in associative layers on layers numbered 8, 16, 17, 23, 24, 31. Although this pattern does not add an associative layer on top-1 layer by importance for GR-100+, it still achieves a performance comparable to the full model, using only 20\% of associative layers. Moreover, on MT this approach improves performance on long-context samples compared to the full ARMT model.

\begin{table*}[t]

\begin{center}
\resizebox{1.\textwidth}{!}{
\begin{tabular}{lccccccc}
\toprule
\begin{tabular}[c]{@{}l@{}}\textbf{Model\//}\\ \textbf{Lengths}\end{tabular} & \begin{tabular}[c]{@{}l@{}}\textbf{Base,}\\ \textbf{MT, 8k}\end{tabular} & \begin{tabular}[c]{@{}l@{}}\textbf{ARMT,}\\ \textbf{MT, 2k}\end{tabular} & \begin{tabular}[c]{@{}l@{}}\textbf{ARMT,}\\ \textbf{MT, 4k}\end{tabular} & \begin{tabular}[c]{@{}l@{}}\textbf{ARMT,}\\ \textbf{MT, 8k}\end{tabular}  &  \begin{tabular}[c]{@{}l@{}}\textbf{ARMT, MT, 2k,}\\ \textbf{pre-selected layers,} \\\textbf{trained}\end{tabular} &  \begin{tabular}[c]{@{}l@{}}\textbf{ARMT, MT, 4k,}\\ \textbf{pre-selected layers,} \\\textbf{trained}\end{tabular} &  \begin{tabular}[c]{@{}l@{}}\textbf{ARMT, MT, 8k,}\\ \textbf{pre-selected layers,} \\\textbf{trained}\end{tabular} \\
\midrule
0k-1k & {\cellcolor[HTML]{188C00}} \color[HTML]{F1F1F1} 0.795 & {\cellcolor[HTML]{48A400}} \color[HTML]{F1F1F1} 0.718 & {\cellcolor[HTML]{309800}} \color[HTML]{F1F1F1} 0.756 & {\cellcolor[HTML]{289400}} \color[HTML]{F1F1F1} 0.769 & {\cellcolor[HTML]{40A000}} \color[HTML]{F1F1F1} 0.731 & {\cellcolor[HTML]{389C00}} \color[HTML]{F1F1F1} 0.744 & {\cellcolor[HTML]{389C00}} \color[HTML]{F1F1F1} 0.744 \\
1k-2k & {\cellcolor[HTML]{1C8E00}} \color[HTML]{F1F1F1} 0.790 & {\cellcolor[HTML]{78BC00}} \color[HTML]{F1F1F1} 0.638 & {\cellcolor[HTML]{5CAE00}} \color[HTML]{F1F1F1} 0.686 & {\cellcolor[HTML]{50A800}} \color[HTML]{F1F1F1} 0.705 & {\cellcolor[HTML]{72B900}} \color[HTML]{F1F1F1} 0.648 & {\cellcolor[HTML]{66B300}} \color[HTML]{F1F1F1} 0.667 & {\cellcolor[HTML]{66B300}} \color[HTML]{F1F1F1} 0.667 \\
2k-4k & {\cellcolor[HTML]{3C9E00}} \color[HTML]{F1F1F1} 0.736 & {\cellcolor[HTML]{76BB00}} \color[HTML]{F1F1F1} 0.642 & {\cellcolor[HTML]{48A400}} \color[HTML]{F1F1F1} 0.717 & {\cellcolor[HTML]{54AA00}} \color[HTML]{F1F1F1} 0.698 & {\cellcolor[HTML]{76BB00}} \color[HTML]{F1F1F1} 0.642 & {\cellcolor[HTML]{54AA00}} \color[HTML]{F1F1F1} 0.698 & {\cellcolor[HTML]{54AA00}} \color[HTML]{F1F1F1} 0.698 \\
4k-6k & {\cellcolor[HTML]{5CAE00}} \color[HTML]{F1F1F1} 0.685 & {\cellcolor[HTML]{8EC700}} \color[HTML]{000000} 0.603 & {\cellcolor[HTML]{7EBF00}} \color[HTML]{000000} 0.630 & {\cellcolor[HTML]{7EBF00}} \color[HTML]{000000} 0.630 & {\cellcolor[HTML]{A8D400}} \color[HTML]{000000} 0.562 & {\cellcolor[HTML]{6CB600}} \color[HTML]{F1F1F1} 0.658 & {\cellcolor[HTML]{76BB00}} \color[HTML]{F1F1F1} 0.644 \\
6k-8k & {\cellcolor[HTML]{008000}} \color[HTML]{F1F1F1} 0.837 & {\cellcolor[HTML]{92C900}} \color[HTML]{000000} 0.598 & {\cellcolor[HTML]{3A9D00}} \color[HTML]{F1F1F1} 0.739 & {\cellcolor[HTML]{2E9700}} \color[HTML]{F1F1F1} 0.761 & {\cellcolor[HTML]{B2D900}} \color[HTML]{000000} 0.543 & {\cellcolor[HTML]{42A100}} \color[HTML]{F1F1F1} 0.728 & {\cellcolor[HTML]{42A100}} \color[HTML]{F1F1F1} 0.728 \\
8k-10k & {\cellcolor[HTML]{A4D200}} \color[HTML]{000000} 0.568 & {\cellcolor[HTML]{E0F000}} \color[HTML]{000000} 0.469 & {\cellcolor[HTML]{148A00}} \color[HTML]{F1F1F1} 0.802 & {\cellcolor[HTML]{0C8600}} \color[HTML]{F1F1F1} 0.815 & {\cellcolor[HTML]{E0F000}} \color[HTML]{000000} 0.469 & {\cellcolor[HTML]{50A800}} \color[HTML]{F1F1F1} 0.704 & {\cellcolor[HTML]{2C9600}} \color[HTML]{F1F1F1} 0.765 \\
10k-12k & {\cellcolor[HTML]{FF0000}} \color[HTML]{F1F1F1} 0.000 & {\cellcolor[HTML]{FFFE00}} \color[HTML]{000000} 0.418 & {\cellcolor[HTML]{58AC00}} \color[HTML]{F1F1F1} 0.692 & {\cellcolor[HTML]{369B00}} \color[HTML]{F1F1F1} 0.747 & {\cellcolor[HTML]{F8FC00}} \color[HTML]{000000} 0.429 & {\cellcolor[HTML]{58AC00}} \color[HTML]{F1F1F1} 0.692 & {\cellcolor[HTML]{309800}} \color[HTML]{F1F1F1} 0.758 \\
12k-14k & {\cellcolor[HTML]{FF0000}} \color[HTML]{F1F1F1} 0.000 & {\cellcolor[HTML]{FF5400}} \color[HTML]{F1F1F1} 0.138 & {\cellcolor[HTML]{329900}} \color[HTML]{F1F1F1} 0.755 & {\cellcolor[HTML]{329900}} \color[HTML]{F1F1F1} 0.755 & {\cellcolor[HTML]{FFE200}} \color[HTML]{000000} 0.372 & {\cellcolor[HTML]{52A900}} \color[HTML]{F1F1F1} 0.702 & {\cellcolor[HTML]{3E9F00}} \color[HTML]{F1F1F1} 0.734 \\
14k-16k & {\cellcolor[HTML]{FF0000}} \color[HTML]{F1F1F1} 0.000 & {\cellcolor[HTML]{FF0000}} \color[HTML]{F1F1F1} 0.000 & {\cellcolor[HTML]{6AB500}} \color[HTML]{F1F1F1} 0.663 & {\cellcolor[HTML]{7EBF00}} \color[HTML]{000000} 0.629 & {\cellcolor[HTML]{FF7A00}} \color[HTML]{F1F1F1} 0.202 & {\cellcolor[HTML]{7EBF00}} \color[HTML]{000000} 0.629 & {\cellcolor[HTML]{78BC00}} \color[HTML]{F1F1F1} 0.640 \\
16k-24k & {\cellcolor[HTML]{FF0000}} \color[HTML]{F1F1F1} 0.000 & {\cellcolor[HTML]{FF0800}} \color[HTML]{F1F1F1} 0.014 & {\cellcolor[HTML]{4EA700}} \color[HTML]{F1F1F1} 0.708 & {\cellcolor[HTML]{4AA500}} \color[HTML]{F1F1F1} 0.715 & {\cellcolor[HTML]{FF5800}} \color[HTML]{F1F1F1} 0.146 & {\cellcolor[HTML]{80C000}} \color[HTML]{000000} 0.625 & {\cellcolor[HTML]{70B800}} \color[HTML]{F1F1F1} 0.653 \\
24k-32k & {\cellcolor[HTML]{FF0600}} \color[HTML]{F1F1F1} 0.010 & {\cellcolor[HTML]{FF0000}} \color[HTML]{F1F1F1} 0.000 & {\cellcolor[HTML]{FF9C00}} \color[HTML]{000000} 0.257 & {\cellcolor[HTML]{3E9F00}} \color[HTML]{F1F1F1} 0.733 & {\cellcolor[HTML]{FF4E00}} \color[HTML]{F1F1F1} 0.129 & {\cellcolor[HTML]{82C100}} \color[HTML]{000000} 0.624 & {\cellcolor[HTML]{44A200}} \color[HTML]{F1F1F1} 0.723 \\
32k-49k & {\cellcolor[HTML]{FF0000}} \color[HTML]{F1F1F1} 0.000 & {\cellcolor[HTML]{FF0000}} \color[HTML]{F1F1F1} 0.000 & {\cellcolor[HTML]{FF0000}} \color[HTML]{F1F1F1} 0.000 & {\cellcolor[HTML]{40A000}} \color[HTML]{F1F1F1} 0.730 & {\cellcolor[HTML]{FF2E00}} \color[HTML]{F1F1F1} 0.078 & {\cellcolor[HTML]{86C300}} \color[HTML]{000000} 0.617 & {\cellcolor[HTML]{56AB00}} \color[HTML]{F1F1F1} 0.696 \\
49k-65k & {\cellcolor[HTML]{FF0000}} \color[HTML]{F1F1F1} 0.000 & {\cellcolor[HTML]{FF0000}} \color[HTML]{F1F1F1} 0.000 & {\cellcolor[HTML]{FF0000}} \color[HTML]{F1F1F1} 0.000 & {\cellcolor[HTML]{FF1200}} \color[HTML]{F1F1F1} 0.030 & {\cellcolor[HTML]{FF4800}} \color[HTML]{F1F1F1} 0.119 & {\cellcolor[HTML]{D6EB00}} \color[HTML]{000000} 0.485 & {\cellcolor[HTML]{B8DC00}} \color[HTML]{000000} 0.535 \\
\hline
In-Domain (0k-8k) & {\cellcolor[HTML]{289400}} \color[HTML]{F1F1F1} 0.771 & {\cellcolor[HTML]{78BC00}} \color[HTML]{F1F1F1} 0.639 & {\cellcolor[HTML]{4EA700}} \color[HTML]{F1F1F1} 0.707 & {\cellcolor[HTML]{4AA500}} \color[HTML]{F1F1F1} 0.714 & {\cellcolor[HTML]{80C000}} \color[HTML]{000000} 0.626 & {\cellcolor[HTML]{54AA00}} \color[HTML]{F1F1F1} 0.698 & {\cellcolor[HTML]{56AB00}} \color[HTML]{F1F1F1} 0.696 \\
OOD (8k-65k) & {\cellcolor[HTML]{FF2200}} \color[HTML]{F1F1F1} 0.058 & {\cellcolor[HTML]{FF4400}} \color[HTML]{F1F1F1} 0.112 & {\cellcolor[HTML]{DEEF00}} \color[HTML]{000000} 0.473 & {\cellcolor[HTML]{76BB00}} \color[HTML]{F1F1F1} 0.643 & {\cellcolor[HTML]{FF8A00}} \color[HTML]{F1F1F1} 0.227 & {\cellcolor[HTML]{7CBE00}} \color[HTML]{000000} 0.631 & {\cellcolor[HTML]{5CAE00}} \color[HTML]{F1F1F1} 0.684 \\
Long-OOD (32k-65k) & {\cellcolor[HTML]{FF0000}} \color[HTML]{F1F1F1} 0.000 & {\cellcolor[HTML]{FF0000}} \color[HTML]{F1F1F1} 0.000 & {\cellcolor[HTML]{FF0000}} \color[HTML]{F1F1F1} 0.000 & {\cellcolor[HTML]{FFF600}} \color[HTML]{000000} 0.403 & {\cellcolor[HTML]{FF3A00}} \color[HTML]{F1F1F1} 0.097 & {\cellcolor[HTML]{ACD600}} \color[HTML]{000000} 0.555 & {\cellcolor[HTML]{84C200}} \color[HTML]{000000} 0.621 \\
Full (0k-65k) & {\cellcolor[HTML]{FFBE00}} \color[HTML]{000000} 0.313 & {\cellcolor[HTML]{FFB600}} \color[HTML]{000000} 0.300 & {\cellcolor[HTML]{AAD500}} \color[HTML]{000000} 0.557 & {\cellcolor[HTML]{66B300}} \color[HTML]{F1F1F1} 0.668 & {\cellcolor[HTML]{FFE000}} \color[HTML]{000000} 0.369 & {\cellcolor[HTML]{6EB700}} \color[HTML]{F1F1F1} 0.655 & {\cellcolor[HTML]{5AAD00}} \color[HTML]{F1F1F1} 0.688 \\
\bottomrule

\end{tabular}}
\caption{Associative layers ablation on the MT dataset for SmolLM-2-360M-IT model, metric - EM. Model with 6 associative blocks (approximately 20\%) achieves slightly lower performance as the full ARMT model.}
\label{tab:mt_best_assoc_ablation_with_train_full_smollm}
\end{center}
\end{table*}
\begin{table*}[t]

\begin{center}
\resizebox{1.\textwidth}{!}{
\begin{tabular}{lccccccc}
\toprule
\begin{tabular}[c]{@{}l@{}}\textbf{Model\//}\\ \textbf{Lengths}\end{tabular} & \begin{tabular}[c]{@{}l@{}}\textbf{Base,}\\ \textbf{GR-100+,}\\ \textbf{8k}\end{tabular} & \begin{tabular}[c]{@{}l@{}}\textbf{ARMT,}\\ \textbf{GR-100+,}\\ \textbf{2k}\end{tabular} & \begin{tabular}[c]{@{}l@{}}\textbf{ARMT,}\\ \textbf{GR-100+,}\\ \textbf{4k}\end{tabular} & \begin{tabular}[c]{@{}l@{}}\textbf{ARMT,}\\ \textbf{GR-100+,}\\ \textbf{8k}\end{tabular}  &  \begin{tabular}[c]{@{}l@{}}\textbf{ARMT,}\\ \textbf{GR-100+, 2k,}\\ \textbf{pre-selected layers,} \\\textbf{trained}\end{tabular} &  \begin{tabular}[c]{@{}l@{}}\textbf{ARMT,}\\ \textbf{GR-100+, 4k,}\\ \textbf{pre-selected layers,} \\\textbf{trained}\end{tabular} &  \begin{tabular}[c]{@{}l@{}}\textbf{ARMT,}\\ \textbf{GR-100+, 8k,}\\ \textbf{pre-selected layers,} \\\textbf{trained}\end{tabular} \\
\midrule
0k-1k & {\cellcolor[HTML]{0C8600}} \color[HTML]{F1F1F1} 0.349 & {\cellcolor[HTML]{209000}} \color[HTML]{F1F1F1} 0.335 & {\cellcolor[HTML]{269300}} \color[HTML]{F1F1F1} 0.332 & {\cellcolor[HTML]{1E8F00}} \color[HTML]{F1F1F1} 0.337 & {\cellcolor[HTML]{128900}} \color[HTML]{F1F1F1} 0.345 & {\cellcolor[HTML]{108800}} \color[HTML]{F1F1F1} 0.346 & {\cellcolor[HTML]{2A9500}} \color[HTML]{F1F1F1} 0.329 \\
1k-2k & {\cellcolor[HTML]{008000}} \color[HTML]{F1F1F1} 0.358 & {\cellcolor[HTML]{64B200}} \color[HTML]{F1F1F1} 0.289 & {\cellcolor[HTML]{5AAD00}} \color[HTML]{F1F1F1} 0.297 & {\cellcolor[HTML]{52A900}} \color[HTML]{F1F1F1} 0.302 & {\cellcolor[HTML]{8EC700}} \color[HTML]{000000} 0.261 & {\cellcolor[HTML]{94CA00}} \color[HTML]{000000} 0.257 & {\cellcolor[HTML]{9CCE00}} \color[HTML]{000000} 0.252 \\
2k-4k & {\cellcolor[HTML]{289400}} \color[HTML]{F1F1F1} 0.330 & {\cellcolor[HTML]{B0D800}} \color[HTML]{000000} 0.238 & {\cellcolor[HTML]{94CA00}} \color[HTML]{000000} 0.257 & {\cellcolor[HTML]{8EC700}} \color[HTML]{000000} 0.261 & {\cellcolor[HTML]{BCDE00}} \color[HTML]{000000} 0.230 & {\cellcolor[HTML]{AAD500}} \color[HTML]{000000} 0.242 & {\cellcolor[HTML]{A2D100}} \color[HTML]{000000} 0.248 \\
4k-6k & {\cellcolor[HTML]{40A000}} \color[HTML]{F1F1F1} 0.314 & {\cellcolor[HTML]{CAE500}} \color[HTML]{000000} 0.220 & {\cellcolor[HTML]{A6D300}} \color[HTML]{000000} 0.245 & {\cellcolor[HTML]{9ACD00}} \color[HTML]{000000} 0.253 & {\cellcolor[HTML]{E2F100}} \color[HTML]{000000} 0.204 & {\cellcolor[HTML]{C2E100}} \color[HTML]{000000} 0.226 & {\cellcolor[HTML]{B4DA00}} \color[HTML]{000000} 0.236 \\
6k-8k & {\cellcolor[HTML]{6AB500}} \color[HTML]{F1F1F1} 0.286 & {\cellcolor[HTML]{D6EB00}} \color[HTML]{000000} 0.212 & {\cellcolor[HTML]{D2E900}} \color[HTML]{000000} 0.215 & {\cellcolor[HTML]{C4E200}} \color[HTML]{000000} 0.225 & {\cellcolor[HTML]{FFF800}} \color[HTML]{000000} 0.180 & {\cellcolor[HTML]{CCE600}} \color[HTML]{000000} 0.219 & {\cellcolor[HTML]{CAE500}} \color[HTML]{000000} 0.220 \\
8k-10k & {\cellcolor[HTML]{B8DC00}} \color[HTML]{000000} 0.233 & {\cellcolor[HTML]{FFF400}} \color[HTML]{000000} 0.177 & {\cellcolor[HTML]{E6F300}} \color[HTML]{000000} 0.202 & {\cellcolor[HTML]{CAE500}} \color[HTML]{000000} 0.221 & {\cellcolor[HTML]{FFD600}} \color[HTML]{000000} 0.157 & {\cellcolor[HTML]{FCFE00}} \color[HTML]{000000} 0.186 & {\cellcolor[HTML]{EAF500}} \color[HTML]{000000} 0.199 \\
10k-12k & {\cellcolor[HTML]{FF0E00}} \color[HTML]{F1F1F1} 0.021 & {\cellcolor[HTML]{FFD400}} \color[HTML]{000000} 0.156 & {\cellcolor[HTML]{EAF500}} \color[HTML]{000000} 0.199 & {\cellcolor[HTML]{D8EC00}} \color[HTML]{000000} 0.211 & {\cellcolor[HTML]{FFDA00}} \color[HTML]{000000} 0.159 & {\cellcolor[HTML]{E6F300}} \color[HTML]{000000} 0.202 & {\cellcolor[HTML]{EAF500}} \color[HTML]{000000} 0.199 \\
12k-14k & {\cellcolor[HTML]{FF2200}} \color[HTML]{F1F1F1} 0.035 & {\cellcolor[HTML]{FFB800}} \color[HTML]{000000} 0.137 & {\cellcolor[HTML]{F0F800}} \color[HTML]{000000} 0.195 & {\cellcolor[HTML]{D2E900}} \color[HTML]{000000} 0.215 & {\cellcolor[HTML]{FFB400}} \color[HTML]{000000} 0.134 & {\cellcolor[HTML]{FCFE00}} \color[HTML]{000000} 0.186 & {\cellcolor[HTML]{EAF500}} \color[HTML]{000000} 0.199 \\
14k-16k & {\cellcolor[HTML]{FF3800}} \color[HTML]{F1F1F1} 0.049 & {\cellcolor[HTML]{FF5C00}} \color[HTML]{F1F1F1} 0.074 & {\cellcolor[HTML]{F6FB00}} \color[HTML]{000000} 0.190 & {\cellcolor[HTML]{DAED00}} \color[HTML]{000000} 0.210 & {\cellcolor[HTML]{FFD600}} \color[HTML]{000000} 0.157 & {\cellcolor[HTML]{E6F300}} \color[HTML]{000000} 0.202 & {\cellcolor[HTML]{D4EA00}} \color[HTML]{000000} 0.213 \\
16k-24k & {\cellcolor[HTML]{FF2600}} \color[HTML]{F1F1F1} 0.037 & {\cellcolor[HTML]{FF3200}} \color[HTML]{F1F1F1} 0.046 & {\cellcolor[HTML]{FFBE00}} \color[HTML]{000000} 0.140 & {\cellcolor[HTML]{CCE600}} \color[HTML]{000000} 0.219 & {\cellcolor[HTML]{FFD400}} \color[HTML]{000000} 0.156 & {\cellcolor[HTML]{E4F200}} \color[HTML]{000000} 0.203 & {\cellcolor[HTML]{DCEE00}} \color[HTML]{000000} 0.208 \\
24k-32k & {\cellcolor[HTML]{FF3E00}} \color[HTML]{F1F1F1} 0.054 & {\cellcolor[HTML]{FF0000}} \color[HTML]{F1F1F1} 0.011 & {\cellcolor[HTML]{FF4600}} \color[HTML]{F1F1F1} 0.059 & {\cellcolor[HTML]{FEFF00}} \color[HTML]{000000} 0.185 & {\cellcolor[HTML]{FFE200}} \color[HTML]{000000} 0.165 & {\cellcolor[HTML]{FFE800}} \color[HTML]{000000} 0.169 & {\cellcolor[HTML]{FFF600}} \color[HTML]{000000} 0.178 \\
32k-49k & {\cellcolor[HTML]{FF5200}} \color[HTML]{F1F1F1} 0.067 & {\cellcolor[HTML]{FF1400}} \color[HTML]{F1F1F1} 0.025 & {\cellcolor[HTML]{FF1E00}} \color[HTML]{F1F1F1} 0.032 & {\cellcolor[HTML]{FFEC00}} \color[HTML]{000000} 0.171 & {\cellcolor[HTML]{FFEA00}} \color[HTML]{000000} 0.170 & {\cellcolor[HTML]{FFE000}} \color[HTML]{000000} 0.164 & {\cellcolor[HTML]{FFE400}} \color[HTML]{000000} 0.166 \\
49k-65k & {\cellcolor[HTML]{FF7000}} \color[HTML]{F1F1F1} 0.088 & {\cellcolor[HTML]{FF0400}} \color[HTML]{F1F1F1} 0.015 & {\cellcolor[HTML]{FF0000}} \color[HTML]{F1F1F1} 0.011 & {\cellcolor[HTML]{FF7C00}} \color[HTML]{F1F1F1} 0.096 & {\cellcolor[HTML]{FFE200}} \color[HTML]{000000} 0.165 & {\cellcolor[HTML]{D4EA00}} \color[HTML]{000000} 0.214 & {\cellcolor[HTML]{FFF200}} \color[HTML]{000000} 0.176 \\
\hline
\begin{tabular}[c]{@{}l@{}}{In-Domain}\\ {(0k-8k)}\end{tabular} & {\cellcolor[HTML]{2C9600}} \color[HTML]{F1F1F1} 0.327 & {\cellcolor[HTML]{92C900}} \color[HTML]{000000} 0.258 & {\cellcolor[HTML]{82C100}} \color[HTML]{000000} 0.269 & {\cellcolor[HTML]{7ABD00}} \color[HTML]{F1F1F1} 0.275 & {\cellcolor[HTML]{A8D400}} \color[HTML]{000000} 0.244 & {\cellcolor[HTML]{94CA00}} \color[HTML]{000000} 0.258 & {\cellcolor[HTML]{94CA00}} \color[HTML]{000000} 0.257 \\
\begin{tabular}[c]{@{}l@{}}{OOD}\\ {(8k-65k)}\end{tabular} & {\cellcolor[HTML]{FF5C00}} \color[HTML]{F1F1F1} 0.074 & {\cellcolor[HTML]{FF8C00}} \color[HTML]{F1F1F1} 0.106 & {\cellcolor[HTML]{FFEA00}} \color[HTML]{000000} 0.170 & {\cellcolor[HTML]{D8EC00}} \color[HTML]{000000} 0.211 & {\cellcolor[HTML]{FFD200}} \color[HTML]{000000} 0.154 & {\cellcolor[HTML]{F2F900}} \color[HTML]{000000} 0.193 & {\cellcolor[HTML]{E8F400}} \color[HTML]{000000} 0.200 \\
\begin{tabular}[c]{@{}l@{}}{Long-OOD}\\ {(32k-65k)}\end{tabular} & {\cellcolor[HTML]{FF5800}} \color[HTML]{F1F1F1} 0.072 & {\cellcolor[HTML]{FF1000}} \color[HTML]{F1F1F1} 0.023 & {\cellcolor[HTML]{FF1600}} \color[HTML]{F1F1F1} 0.027 & {\cellcolor[HTML]{FFD200}} \color[HTML]{000000} 0.154 & {\cellcolor[HTML]{FFE800}} \color[HTML]{000000} 0.169 & {\cellcolor[HTML]{FFF200}} \color[HTML]{000000} 0.176 & {\cellcolor[HTML]{FFE800}} \color[HTML]{000000} 0.168 \\
\begin{tabular}[c]{@{}l@{}}{Full}\\ {(0k-65k)}\end{tabular} & {\cellcolor[HTML]{F4FA00}} \color[HTML]{000000} 0.192 & {\cellcolor[HTML]{FFF400}} \color[HTML]{000000} 0.178 & {\cellcolor[HTML]{D0E800}} \color[HTML]{000000} 0.216 & {\cellcolor[HTML]{ACD600}} \color[HTML]{000000} 0.241 & {\cellcolor[HTML]{EEF700}} \color[HTML]{000000} 0.196 & {\cellcolor[HTML]{C6E300}} \color[HTML]{000000} 0.223 & {\cellcolor[HTML]{C0E000}} \color[HTML]{000000} 0.227 \\
\bottomrule

\end{tabular}}
\caption{Associative layers ablation on the GR-100+ dataset for SmolLM-2-360M-IT model, metric - ROUGE-L. Model with 6 associative blocks (approximately 20\%) achieves slightly lower performance as the full ARMT model.}
\label{tab:gov_report_best_assoc_ablation_with_train_full_smollm}
\end{center}
\end{table*}

\section{Training Hyperparameters}\label{app:hyperpars}
The hyperparameters used during training are described in~\Cref{tab:hyperpars}. All models were trained using two NVIDIA H100-80GB GPUs. The full ARMT fine-tuning process takes approximately 48 hours per run.
\begin{table*}[h]

\begin{center}
\begin{tabular}{lccccc}
\toprule
\textbf{Dataset} & \textbf{Model} & \textbf{Length} & \begin{tabular}[c]{@{}l@{}}\textbf{Learning}\\ \textbf{Rate}\end{tabular} & \begin{tabular}[c]{@{}l@{}}\textbf{Training}\\ \textbf{Steps}\end{tabular} & \begin{tabular}[c]{@{}l@{}}\textbf{Total Batch}\\ \textbf{Size}\end{tabular} \\
\midrule
MT & Gemma-3-1B-IT & 8k & 1e-4 &  10.000 & 64 \\
MT & ARMT & 2k & 1e-4 &  10.000 & 64 \\
MT & ARMT & 4k & 1e-4 &  10.000 & 64 \\
MT & ARMT & 8k & 3e-5 &  10.000 & 64 \\
\hline
GR & Gemma-3-1B-IT & 8k & 1e-4 &  8.000 & 64 \\
GR & ARMT & 2k & 1e-4 &  4.000 & 64 \\
GR & ARMT & 4k & 3e-5 &  4.000 & 64 \\
GR & ARMT & 8k & 1e-5 &  4.000 & 64 \\
\hline
GR-100+ & Gemma-3-1B-IT & 8k & 1e-4 &  10.000 & 64 \\
GR-100+ & ARMT & 2k & 1e-4 &  5.000 & 64 \\
GR-100+ & ARMT & 4k & 3e-5 &  5.000 & 64 \\
GR-100+ & ARMT & 8k & 1e-5 &  5.000 & 64 \\

\hline
ContractNLI & Gemma-3-1B-IT & 8k & 1e-4 &  500 & 64 \\
ContractNLI & ARMT & 2k & 1e-4 &  250 & 64 \\
ContractNLI & ARMT & 4k & 3e-5 &  250 & 64 \\
ContractNLI & ARMT & 8k & 1e-5 &  250 & 64 \\

\bottomrule
\end{tabular}
\caption{Training hyperparameters. All models were trained with LoRA~\citep{lora} on all linear layers with rank=64, $\alpha$=128 and dropout=0.1. We used gradient clipping to the maximum value of 1.0 during training, and weight decay of 0.01. For ARMT model, we used 16 memory tokens and associative memory size of 64.}
\label{tab:hyperpars}
\end{center}
\end{table*}

\section{Memory Size Ablations}\label{app:mem_size_ablations}
In our experiments, the ARMT model without continued pretraining used 16 memory tokens, whereas the pretrained variant used 32 memory tokens. We hypothesize that a larger memory size is beneficial during pretraining, while 16 memory tokens are sufficient for direct fine-tuning. The comparison between ARMT models with 16 and 32 memory tokens, both trained without pretraining, is presented in~\Cref{tab:gov_report_best_mem_size_ablation,tab:mt_best_mem_size_ablation}. Overall, the model with 16 memory tokens achieves better performance across all data splits and demonstrates stronger generalization during curriculum learning.
\begin{table*}[t]

\begin{center}
\resizebox{0.8\linewidth}{!}{
\begin{tabular}{lccccccc}
\toprule
\begin{tabular}[c]{@{}l@{}}\textbf{Model\//}\\ \textbf{Lengths}\end{tabular} & \begin{tabular}[c]{@{}l@{}}\textbf{Base,}\\ \textbf{GR-100+,}\\ \textbf{8k}\end{tabular} & \begin{tabular}[c]{@{}l@{}}\textbf{ARMT,}\\ \textbf{GR-100+,}\\ \textbf{2k}\end{tabular} & \begin{tabular}[c]{@{}l@{}}\textbf{ARMT,}\\ \textbf{GR-100+,}\\ \textbf{4k}\end{tabular} & \begin{tabular}[c]{@{}l@{}}\textbf{ARMT,}\\ \textbf{GR-100+,}\\ \textbf{8k}\end{tabular} & \begin{tabular}[c]{@{}l@{}}\textbf{ARMT,}\\ \textbf{GR-100+,}\\ \textbf{2k, 32 mem}\end{tabular} & \begin{tabular}[c]{@{}l@{}}\textbf{ARMT,}\\ \textbf{GR-100+,}\\ \textbf{4k, 32 mem}\end{tabular} & \begin{tabular}[c]{@{}l@{}}\textbf{ARMT,}\\ \textbf{GR-100+,}\\ \textbf{8k, 32 mem}\end{tabular} \\
\midrule
0k-1k & {\cellcolor[HTML]{068300}} \color[HTML]{F1F1F1} 0.380 & {\cellcolor[HTML]{249200}} \color[HTML]{F1F1F1} 0.357 & {\cellcolor[HTML]{249200}} \color[HTML]{F1F1F1} 0.357 & {\cellcolor[HTML]{229100}} \color[HTML]{F1F1F1} 0.358 & {\cellcolor[HTML]{3C9E00}} \color[HTML]{F1F1F1} 0.339 & {\cellcolor[HTML]{2E9700}} \color[HTML]{F1F1F1} 0.349 & {\cellcolor[HTML]{329900}} \color[HTML]{F1F1F1} 0.346 \\
1k-2k & {\cellcolor[HTML]{008000}} \color[HTML]{F1F1F1} 0.385 & {\cellcolor[HTML]{76BB00}} \color[HTML]{F1F1F1} 0.296 & {\cellcolor[HTML]{78BC00}} \color[HTML]{F1F1F1} 0.294 & {\cellcolor[HTML]{6CB600}} \color[HTML]{F1F1F1} 0.303 & {\cellcolor[HTML]{76BB00}} \color[HTML]{F1F1F1} 0.296 & {\cellcolor[HTML]{72B900}} \color[HTML]{F1F1F1} 0.299 & {\cellcolor[HTML]{6AB500}} \color[HTML]{F1F1F1} 0.305 \\
2k-4k & {\cellcolor[HTML]{2A9500}} \color[HTML]{F1F1F1} 0.352 & {\cellcolor[HTML]{A0D000}} \color[HTML]{000000} 0.264 & {\cellcolor[HTML]{92C900}} \color[HTML]{000000} 0.275 & {\cellcolor[HTML]{8CC600}} \color[HTML]{000000} 0.279 & {\cellcolor[HTML]{B4DA00}} \color[HTML]{000000} 0.249 & {\cellcolor[HTML]{7EBF00}} \color[HTML]{000000} 0.290 & {\cellcolor[HTML]{8EC700}} \color[HTML]{000000} 0.277 \\
4k-6k & {\cellcolor[HTML]{369B00}} \color[HTML]{F1F1F1} 0.344 & {\cellcolor[HTML]{C0E000}} \color[HTML]{000000} 0.240 & {\cellcolor[HTML]{82C100}} \color[HTML]{000000} 0.286 & {\cellcolor[HTML]{6CB600}} \color[HTML]{F1F1F1} 0.303 & {\cellcolor[HTML]{C6E300}} \color[HTML]{000000} 0.235 & {\cellcolor[HTML]{A2D100}} \color[HTML]{000000} 0.263 & {\cellcolor[HTML]{7EBF00}} \color[HTML]{000000} 0.290 \\
6k-8k & {\cellcolor[HTML]{76BB00}} \color[HTML]{F1F1F1} 0.296 & {\cellcolor[HTML]{DCEE00}} \color[HTML]{000000} 0.219 & {\cellcolor[HTML]{C2E100}} \color[HTML]{000000} 0.238 & {\cellcolor[HTML]{AED700}} \color[HTML]{000000} 0.254 & {\cellcolor[HTML]{D2E900}} \color[HTML]{000000} 0.227 & {\cellcolor[HTML]{C2E100}} \color[HTML]{000000} 0.239 & {\cellcolor[HTML]{B6DB00}} \color[HTML]{000000} 0.248 \\
8k-10k & {\cellcolor[HTML]{68B400}} \color[HTML]{F1F1F1} 0.306 & {\cellcolor[HTML]{EEF700}} \color[HTML]{000000} 0.205 & {\cellcolor[HTML]{CAE500}} \color[HTML]{000000} 0.233 & {\cellcolor[HTML]{C6E300}} \color[HTML]{000000} 0.235 & {\cellcolor[HTML]{FFFC00}} \color[HTML]{000000} 0.190 & {\cellcolor[HTML]{C2E100}} \color[HTML]{000000} 0.238 & {\cellcolor[HTML]{BADD00}} \color[HTML]{000000} 0.244 \\
10k-12k & {\cellcolor[HTML]{9ACD00}} \color[HTML]{000000} 0.269 & {\cellcolor[HTML]{E4F200}} \color[HTML]{000000} 0.213 & {\cellcolor[HTML]{DAED00}} \color[HTML]{000000} 0.221 & {\cellcolor[HTML]{BEDF00}} \color[HTML]{000000} 0.241 & {\cellcolor[HTML]{FFD200}} \color[HTML]{000000} 0.159 & {\cellcolor[HTML]{D2E900}} \color[HTML]{000000} 0.227 & {\cellcolor[HTML]{D4EA00}} \color[HTML]{000000} 0.225 \\
12k-14k & {\cellcolor[HTML]{8AC500}} \color[HTML]{000000} 0.280 & {\cellcolor[HTML]{FEFF00}} \color[HTML]{000000} 0.193 & {\cellcolor[HTML]{D4EA00}} \color[HTML]{000000} 0.225 & {\cellcolor[HTML]{C4E200}} \color[HTML]{000000} 0.237 & {\cellcolor[HTML]{FF8A00}} \color[HTML]{F1F1F1} 0.105 & {\cellcolor[HTML]{F4FA00}} \color[HTML]{000000} 0.201 & {\cellcolor[HTML]{CAE500}} \color[HTML]{000000} 0.232 \\
14k-16k & {\cellcolor[HTML]{CCE600}} \color[HTML]{000000} 0.231 & {\cellcolor[HTML]{FFEE00}} \color[HTML]{000000} 0.180 & {\cellcolor[HTML]{DEEF00}} \color[HTML]{000000} 0.217 & {\cellcolor[HTML]{C0E000}} \color[HTML]{000000} 0.240 & {\cellcolor[HTML]{FF3800}} \color[HTML]{F1F1F1} 0.043 & {\cellcolor[HTML]{FFD000}} \color[HTML]{000000} 0.157 & {\cellcolor[HTML]{C0E000}} \color[HTML]{000000} 0.240 \\
16k-24k & {\cellcolor[HTML]{9ACD00}} \color[HTML]{000000} 0.269 & {\cellcolor[HTML]{FF9800}} \color[HTML]{000000} 0.115 & {\cellcolor[HTML]{F6FB00}} \color[HTML]{000000} 0.200 & {\cellcolor[HTML]{B0D800}} \color[HTML]{000000} 0.252 & {\cellcolor[HTML]{FF1400}} \color[HTML]{F1F1F1} 0.016 & {\cellcolor[HTML]{FF3C00}} \color[HTML]{F1F1F1} 0.046 & {\cellcolor[HTML]{BEDF00}} \color[HTML]{000000} 0.241 \\
24k-32k & {\cellcolor[HTML]{B6DB00}} \color[HTML]{000000} 0.247 & {\cellcolor[HTML]{FF1400}} \color[HTML]{F1F1F1} 0.016 & {\cellcolor[HTML]{FFDE00}} \color[HTML]{000000} 0.168 & {\cellcolor[HTML]{ECF600}} \color[HTML]{000000} 0.207 & {\cellcolor[HTML]{FF0000}} \color[HTML]{F1F1F1} 0.000 & {\cellcolor[HTML]{FF0000}} \color[HTML]{F1F1F1} 0.001 & {\cellcolor[HTML]{FFF800}} \color[HTML]{000000} 0.187 \\
32k-49k & {\cellcolor[HTML]{FFE200}} \color[HTML]{000000} 0.171 & {\cellcolor[HTML]{FF0200}} \color[HTML]{F1F1F1} 0.002 & {\cellcolor[HTML]{FFAA00}} \color[HTML]{000000} 0.128 & {\cellcolor[HTML]{F6FB00}} \color[HTML]{000000} 0.200 & {\cellcolor[HTML]{FF0000}} \color[HTML]{F1F1F1} 0.000 & {\cellcolor[HTML]{FF0000}} \color[HTML]{F1F1F1} 0.000 & {\cellcolor[HTML]{F6FB00}} \color[HTML]{000000} 0.199 \\
49k-65k & {\cellcolor[HTML]{FFFA00}} \color[HTML]{000000} 0.189 & {\cellcolor[HTML]{FF0000}} \color[HTML]{F1F1F1} 0.000 & {\cellcolor[HTML]{FF3400}} \color[HTML]{F1F1F1} 0.040 & {\cellcolor[HTML]{A0D000}} \color[HTML]{000000} 0.264 & {\cellcolor[HTML]{FF0000}} \color[HTML]{F1F1F1} 0.000 & {\cellcolor[HTML]{FF0000}} \color[HTML]{F1F1F1} 0.000 & {\cellcolor[HTML]{FEFF00}} \color[HTML]{000000} 0.194 \\
\hline
\begin{tabular}[c]{@{}l@{}}{In-Domain}\\ {(0k-8k)}\end{tabular} & {\cellcolor[HTML]{2C9600}} \color[HTML]{F1F1F1} 0.351 & {\cellcolor[HTML]{92C900}} \color[HTML]{000000} 0.275 & {\cellcolor[HTML]{7EBF00}} \color[HTML]{000000} 0.290 & {\cellcolor[HTML]{72B900}} \color[HTML]{F1F1F1} 0.299 & {\cellcolor[HTML]{9ACD00}} \color[HTML]{000000} 0.269 & {\cellcolor[HTML]{80C000}} \color[HTML]{000000} 0.288 & {\cellcolor[HTML]{7ABD00}} \color[HTML]{F1F1F1} 0.293 \\
\begin{tabular}[c]{@{}l@{}}{OOD}\\ {(8k-65k)}\end{tabular} & {\cellcolor[HTML]{9ECF00}} \color[HTML]{000000} 0.266 & {\cellcolor[HTML]{FFD600}} \color[HTML]{000000} 0.162 & {\cellcolor[HTML]{E6F300}} \color[HTML]{000000} 0.211 & {\cellcolor[HTML]{C2E100}} \color[HTML]{000000} 0.238 & {\cellcolor[HTML]{FF7800}} \color[HTML]{F1F1F1} 0.090 & {\cellcolor[HTML]{FFCC00}} \color[HTML]{000000} 0.154 & {\cellcolor[HTML]{CAE500}} \color[HTML]{000000} 0.232 \\
\begin{tabular}[c]{@{}l@{}}{Long-OOD}\\ {(32k-65k)}\end{tabular} & {\cellcolor[HTML]{FFE800}} \color[HTML]{000000} 0.175 & {\cellcolor[HTML]{FF0200}} \color[HTML]{F1F1F1} 0.002 & {\cellcolor[HTML]{FF8E00}} \color[HTML]{F1F1F1} 0.108 & {\cellcolor[HTML]{E2F100}} \color[HTML]{000000} 0.215 & {\cellcolor[HTML]{FF0000}} \color[HTML]{F1F1F1} 0.000 & {\cellcolor[HTML]{FF0000}} \color[HTML]{F1F1F1} 0.000 & {\cellcolor[HTML]{F8FC00}} \color[HTML]{000000} 0.198 \\
\begin{tabular}[c]{@{}l@{}}{Full}\\ {(0k-65k)}\end{tabular} & {\cellcolor[HTML]{68B400}} \color[HTML]{F1F1F1} 0.306 & {\cellcolor[HTML]{E2F100}} \color[HTML]{000000} 0.215 & {\cellcolor[HTML]{B6DB00}} \color[HTML]{000000} 0.248 & {\cellcolor[HTML]{9CCE00}} \color[HTML]{000000} 0.266 & {\cellcolor[HTML]{FFE600}} \color[HTML]{000000} 0.174 & {\cellcolor[HTML]{E0F000}} \color[HTML]{000000} 0.217 & {\cellcolor[HTML]{A4D200}} \color[HTML]{000000} 0.260 \\
\bottomrule

\end{tabular}}
\caption{Memory size ablation on the GR-100+ for Gemma-3-1B-IT model, metric - ROUGE-L. ARMT in base setup (with 16 memory tokens) performs better than model with 32 memory tokens.}
\label{tab:gov_report_best_mem_size_ablation}
\end{center}
\end{table*}
\begin{table*}[t]

\begin{center}
\resizebox{0.8\linewidth}{!}{
\begin{tabular}{lccccccc}
\toprule
\begin{tabular}[c]{@{}l@{}}\textbf{Model\//}\\ \textbf{Lengths}\end{tabular} & \begin{tabular}[c]{@{}l@{}}\textbf{Base,}\\ \textbf{MT,}\\ \textbf{8k}\end{tabular} & \begin{tabular}[c]{@{}l@{}}\textbf{ARMT,}\\ \textbf{MT,}\\ \textbf{2k}\end{tabular} & \begin{tabular}[c]{@{}l@{}}\textbf{ARMT,}\\ \textbf{MT,}\\ \textbf{4k}\end{tabular} & \begin{tabular}[c]{@{}l@{}}\textbf{ARMT,}\\ \textbf{MT,}\\ \textbf{8k}\end{tabular} & \begin{tabular}[c]{@{}l@{}}\textbf{ARMT,}\\ \textbf{MT,}\\ \textbf{2k, 32 mem}\end{tabular} & \begin{tabular}[c]{@{}l@{}}\textbf{ARMT,}\\ \textbf{MT,}\\ \textbf{4k, 32 mem}\end{tabular} & \begin{tabular}[c]{@{}l@{}}\textbf{ARMT,}\\ \textbf{MT,}\\ \textbf{8k, 32 mem}\end{tabular} \\
\midrule
0k-1k & {\cellcolor[HTML]{2A9500}} \color[HTML]{F1F1F1} 0.795 & {\cellcolor[HTML]{329900}} \color[HTML]{F1F1F1} 0.782 & {\cellcolor[HTML]{329900}} \color[HTML]{F1F1F1} 0.782 & {\cellcolor[HTML]{42A100}} \color[HTML]{F1F1F1} 0.756 & {\cellcolor[HTML]{229100}} \color[HTML]{F1F1F1} 0.808 & {\cellcolor[HTML]{229100}} \color[HTML]{F1F1F1} 0.808 & {\cellcolor[HTML]{329900}} \color[HTML]{F1F1F1} 0.782 \\
1k-2k & {\cellcolor[HTML]{2E9700}} \color[HTML]{F1F1F1} 0.790 & {\cellcolor[HTML]{54AA00}} \color[HTML]{F1F1F1} 0.724 & {\cellcolor[HTML]{5AAD00}} \color[HTML]{F1F1F1} 0.714 & {\cellcolor[HTML]{4AA500}} \color[HTML]{F1F1F1} 0.743 & {\cellcolor[HTML]{66B300}} \color[HTML]{F1F1F1} 0.695 & {\cellcolor[HTML]{72B900}} \color[HTML]{F1F1F1} 0.676 & {\cellcolor[HTML]{66B300}} \color[HTML]{F1F1F1} 0.695 \\
2k-4k & {\cellcolor[HTML]{389C00}} \color[HTML]{F1F1F1} 0.774 & {\cellcolor[HTML]{6AB500}} \color[HTML]{F1F1F1} 0.689 & {\cellcolor[HTML]{64B200}} \color[HTML]{F1F1F1} 0.698 & {\cellcolor[HTML]{4EA700}} \color[HTML]{F1F1F1} 0.736 & {\cellcolor[HTML]{86C300}} \color[HTML]{000000} 0.642 & {\cellcolor[HTML]{76BB00}} \color[HTML]{F1F1F1} 0.670 & {\cellcolor[HTML]{5EAF00}} \color[HTML]{F1F1F1} 0.708 \\
4k-6k & {\cellcolor[HTML]{3C9E00}} \color[HTML]{F1F1F1} 0.767 & {\cellcolor[HTML]{96CB00}} \color[HTML]{000000} 0.616 & {\cellcolor[HTML]{84C200}} \color[HTML]{000000} 0.644 & {\cellcolor[HTML]{64B200}} \color[HTML]{F1F1F1} 0.699 & {\cellcolor[HTML]{9ECF00}} \color[HTML]{000000} 0.603 & {\cellcolor[HTML]{84C200}} \color[HTML]{000000} 0.644 & {\cellcolor[HTML]{64B200}} \color[HTML]{F1F1F1} 0.699 \\
6k-8k & {\cellcolor[HTML]{048200}} \color[HTML]{F1F1F1} 0.859 & {\cellcolor[HTML]{9ACD00}} \color[HTML]{000000} 0.609 & {\cellcolor[HTML]{5AAD00}} \color[HTML]{F1F1F1} 0.717 & {\cellcolor[HTML]{269300}} \color[HTML]{F1F1F1} 0.804 & {\cellcolor[HTML]{FFBC00}} \color[HTML]{000000} 0.326 & {\cellcolor[HTML]{52A900}} \color[HTML]{F1F1F1} 0.728 & {\cellcolor[HTML]{269300}} \color[HTML]{F1F1F1} 0.804 \\
8k-10k & {\cellcolor[HTML]{088400}} \color[HTML]{F1F1F1} 0.852 & {\cellcolor[HTML]{D8EC00}} \color[HTML]{000000} 0.506 & {\cellcolor[HTML]{44A200}} \color[HTML]{F1F1F1} 0.753 & {\cellcolor[HTML]{44A200}} \color[HTML]{F1F1F1} 0.753 & {\cellcolor[HTML]{FF4200}} \color[HTML]{F1F1F1} 0.123 & {\cellcolor[HTML]{7EBF00}} \color[HTML]{000000} 0.654 & {\cellcolor[HTML]{108800}} \color[HTML]{F1F1F1} 0.840 \\
10k-12k & {\cellcolor[HTML]{008000}} \color[HTML]{F1F1F1} 0.868 & {\cellcolor[HTML]{FFD800}} \color[HTML]{000000} 0.374 & {\cellcolor[HTML]{76BB00}} \color[HTML]{F1F1F1} 0.670 & {\cellcolor[HTML]{2C9600}} \color[HTML]{F1F1F1} 0.791 & {\cellcolor[HTML]{FF5C00}} \color[HTML]{F1F1F1} 0.165 & {\cellcolor[HTML]{54AA00}} \color[HTML]{F1F1F1} 0.725 & {\cellcolor[HTML]{1A8D00}} \color[HTML]{F1F1F1} 0.824 \\
12k-14k & {\cellcolor[HTML]{309800}} \color[HTML]{F1F1F1} 0.787 & {\cellcolor[HTML]{FFEA00}} \color[HTML]{000000} 0.404 & {\cellcolor[HTML]{62B100}} \color[HTML]{F1F1F1} 0.702 & {\cellcolor[HTML]{48A400}} \color[HTML]{F1F1F1} 0.745 & {\cellcolor[HTML]{FF3200}} \color[HTML]{F1F1F1} 0.096 & {\cellcolor[HTML]{82C100}} \color[HTML]{000000} 0.649 & {\cellcolor[HTML]{56AB00}} \color[HTML]{F1F1F1} 0.723 \\
14k-16k & {\cellcolor[HTML]{3E9F00}} \color[HTML]{F1F1F1} 0.764 & {\cellcolor[HTML]{FFC200}} \color[HTML]{000000} 0.337 & {\cellcolor[HTML]{88C400}} \color[HTML]{000000} 0.640 & {\cellcolor[HTML]{6CB600}} \color[HTML]{F1F1F1} 0.685 & {\cellcolor[HTML]{FF1400}} \color[HTML]{F1F1F1} 0.045 & {\cellcolor[HTML]{94CA00}} \color[HTML]{000000} 0.618 & {\cellcolor[HTML]{6CB600}} \color[HTML]{F1F1F1} 0.685 \\
16k-24k & {\cellcolor[HTML]{209000}} \color[HTML]{F1F1F1} 0.812 & {\cellcolor[HTML]{FF8200}} \color[HTML]{F1F1F1} 0.229 & {\cellcolor[HTML]{84C200}} \color[HTML]{000000} 0.646 & {\cellcolor[HTML]{42A100}} \color[HTML]{F1F1F1} 0.757 & {\cellcolor[HTML]{FF1E00}} \color[HTML]{F1F1F1} 0.062 & {\cellcolor[HTML]{AAD500}} \color[HTML]{000000} 0.583 & {\cellcolor[HTML]{56AB00}} \color[HTML]{F1F1F1} 0.722 \\
24k-32k & {\cellcolor[HTML]{5CAE00}} \color[HTML]{F1F1F1} 0.713 & {\cellcolor[HTML]{FF7000}} \color[HTML]{F1F1F1} 0.198 & {\cellcolor[HTML]{A8D400}} \color[HTML]{000000} 0.584 & {\cellcolor[HTML]{5CAE00}} \color[HTML]{F1F1F1} 0.713 & {\cellcolor[HTML]{FF1000}} \color[HTML]{F1F1F1} 0.040 & {\cellcolor[HTML]{D2E900}} \color[HTML]{000000} 0.515 & {\cellcolor[HTML]{68B400}} \color[HTML]{F1F1F1} 0.693 \\
32k-49k & {\cellcolor[HTML]{A4D200}} \color[HTML]{000000} 0.591 & {\cellcolor[HTML]{FF4C00}} \color[HTML]{F1F1F1} 0.139 & {\cellcolor[HTML]{9ACD00}} \color[HTML]{000000} 0.609 & {\cellcolor[HTML]{3C9E00}} \color[HTML]{F1F1F1} 0.765 & {\cellcolor[HTML]{FF1200}} \color[HTML]{F1F1F1} 0.043 & {\cellcolor[HTML]{E8F400}} \color[HTML]{000000} 0.478 & {\cellcolor[HTML]{46A300}} \color[HTML]{F1F1F1} 0.748 \\
49k-65k & {\cellcolor[HTML]{FFB600}} \color[HTML]{000000} 0.317 & {\cellcolor[HTML]{FF2E00}} \color[HTML]{F1F1F1} 0.089 & {\cellcolor[HTML]{9CCE00}} \color[HTML]{000000} 0.604 & {\cellcolor[HTML]{5CAE00}} \color[HTML]{F1F1F1} 0.713 & {\cellcolor[HTML]{FF0000}} \color[HTML]{F1F1F1} 0.010 & {\cellcolor[HTML]{D8EC00}} \color[HTML]{000000} 0.505 & {\cellcolor[HTML]{50A800}} \color[HTML]{F1F1F1} 0.733 \\
\hline
\begin{tabular}[c]{@{}l@{}}{In-Domain}\\ {(0k-8k)}\end{tabular} & {\cellcolor[HTML]{2A9500}} \color[HTML]{F1F1F1} 0.797 & {\cellcolor[HTML]{6CB600}} \color[HTML]{F1F1F1} 0.685 & {\cellcolor[HTML]{5CAE00}} \color[HTML]{F1F1F1} 0.711 & {\cellcolor[HTML]{46A300}} \color[HTML]{F1F1F1} 0.749 & {\cellcolor[HTML]{98CC00}} \color[HTML]{000000} 0.612 & {\cellcolor[HTML]{62B100}} \color[HTML]{F1F1F1} 0.703 & {\cellcolor[HTML]{4EA700}} \color[HTML]{F1F1F1} 0.736 \\
\begin{tabular}[c]{@{}l@{}}{OOD}\\ {(8k-65k)}\end{tabular} & {\cellcolor[HTML]{5EAF00}} \color[HTML]{F1F1F1} 0.709 & {\cellcolor[HTML]{FF9A00}} \color[HTML]{000000} 0.271 & {\cellcolor[HTML]{82C100}} \color[HTML]{000000} 0.647 & {\cellcolor[HTML]{4AA500}} \color[HTML]{F1F1F1} 0.741 & {\cellcolor[HTML]{FF2200}} \color[HTML]{F1F1F1} 0.070 & {\cellcolor[HTML]{A8D400}} \color[HTML]{000000} 0.584 & {\cellcolor[HTML]{4AA500}} \color[HTML]{F1F1F1} 0.743 \\
\begin{tabular}[c]{@{}l@{}}{Long-OOD}\\ {(32k-65k)}\end{tabular} & {\cellcolor[HTML]{F0F800}} \color[HTML]{000000} 0.463 & {\cellcolor[HTML]{FF3E00}} \color[HTML]{F1F1F1} 0.116 & {\cellcolor[HTML]{9ACD00}} \color[HTML]{000000} 0.607 & {\cellcolor[HTML]{4AA500}} \color[HTML]{F1F1F1} 0.741 & {\cellcolor[HTML]{FF0A00}} \color[HTML]{F1F1F1} 0.028 & {\cellcolor[HTML]{E0F000}} \color[HTML]{000000} 0.491 & {\cellcolor[HTML]{4AA500}} \color[HTML]{F1F1F1} 0.741 \\
\begin{tabular}[c]{@{}l@{}}{Full}\\ {(0k-65k)}\end{tabular} & {\cellcolor[HTML]{4AA500}} \color[HTML]{F1F1F1} 0.741 & {\cellcolor[HTML]{FFF200}} \color[HTML]{000000} 0.419 & {\cellcolor[HTML]{76BB00}} \color[HTML]{F1F1F1} 0.670 & {\cellcolor[HTML]{48A400}} \color[HTML]{F1F1F1} 0.744 & {\cellcolor[HTML]{FF9600}} \color[HTML]{000000} 0.264 & {\cellcolor[HTML]{8EC700}} \color[HTML]{000000} 0.627 & {\cellcolor[HTML]{4CA600}} \color[HTML]{F1F1F1} 0.740 \\
\bottomrule

\end{tabular}}
\caption{Memory size ablation on the MT for Gemma-3-1B-IT model, metric - EM. ARMT in base setup (with 16 memory tokens) performs better than model with 32 memory tokens.}
\label{tab:mt_best_mem_size_ablation}
\end{center}
\end{table*}

\section{RMT Comparison}\label{app:rmt_ablations}
To compare the ARMT architecture with RMT under identical conditions on a real-world task, we trained the RMT model on the GR-100+ dataset using the same setup as the ARMT model. The results, presented in~\Cref{tab:gov_report_rmt_ablation}, show that ARMT consistently outperforms RMT in all data splits, according to the prior results on synthetic data~\citep{rodkin2024associative}.
\begin{table*}[t]

\begin{center}
\resizebox{1.\textwidth}{!}{
\begin{tabular}{lccccccc}
\toprule
\begin{tabular}[c]{@{}l@{}}\textbf{Model\//}\\ \textbf{Lengths}\end{tabular} & \begin{tabular}[c]{@{}l@{}}\textbf{Base,}\\ \textbf{GR-100+,}\\ \textbf{8k}\end{tabular} & \begin{tabular}[c]{@{}l@{}}\textbf{ARMT,}\\ \textbf{GR-100+,}\\ \textbf{2k}\end{tabular} & \begin{tabular}[c]{@{}l@{}}\textbf{ARMT,}\\ \textbf{GR-100+,}\\ \textbf{4k}\end{tabular} & \begin{tabular}[c]{@{}l@{}}\textbf{ARMT,}\\ \textbf{GR-100+,}\\ \textbf{8k}\end{tabular}  &  \begin{tabular}[c]{@{}l@{}}\textbf{RMT,}\\ \textbf{GR-100+, 2k}\end{tabular} &  \begin{tabular}[c]{@{}l@{}}\textbf{RMT,}\\ \textbf{GR-100+, 4k}\end{tabular} &  \begin{tabular}[c]{@{}l@{}}\textbf{RMT,}\\ \textbf{GR-100+, 8k}\end{tabular} \\
\midrule
0k-1k & {\cellcolor[HTML]{068300}} \color[HTML]{F1F1F1} 0.380 & {\cellcolor[HTML]{249200}} \color[HTML]{F1F1F1} 0.357 & {\cellcolor[HTML]{249200}} \color[HTML]{F1F1F1} 0.357 & {\cellcolor[HTML]{229100}} \color[HTML]{F1F1F1} 0.358 & {\cellcolor[HTML]{DEEF00}} \color[HTML]{000000} 0.218 & {\cellcolor[HTML]{BEDF00}} \color[HTML]{000000} 0.242 & {\cellcolor[HTML]{FAFD00}} \color[HTML]{000000} 0.196 \\
1k-2k & {\cellcolor[HTML]{008000}} \color[HTML]{F1F1F1} 0.385 & {\cellcolor[HTML]{76BB00}} \color[HTML]{F1F1F1} 0.296 & {\cellcolor[HTML]{78BC00}} \color[HTML]{F1F1F1} 0.294 & {\cellcolor[HTML]{6CB600}} \color[HTML]{F1F1F1} 0.303 & {\cellcolor[HTML]{F8FC00}} \color[HTML]{000000} 0.198 & {\cellcolor[HTML]{D6EB00}} \color[HTML]{000000} 0.224 & {\cellcolor[HTML]{A2D100}} \color[HTML]{000000} 0.263 \\
2k-4k & {\cellcolor[HTML]{2A9500}} \color[HTML]{F1F1F1} 0.352 & {\cellcolor[HTML]{A0D000}} \color[HTML]{000000} 0.264 & {\cellcolor[HTML]{92C900}} \color[HTML]{000000} 0.275 & {\cellcolor[HTML]{8CC600}} \color[HTML]{000000} 0.279 & {\cellcolor[HTML]{FFAE00}} \color[HTML]{000000} 0.132 & {\cellcolor[HTML]{FFDE00}} \color[HTML]{000000} 0.168 & {\cellcolor[HTML]{DAED00}} \color[HTML]{000000} 0.220 \\
4k-6k & {\cellcolor[HTML]{369B00}} \color[HTML]{F1F1F1} 0.344 & {\cellcolor[HTML]{C0E000}} \color[HTML]{000000} 0.240 & {\cellcolor[HTML]{82C100}} \color[HTML]{000000} 0.286 & {\cellcolor[HTML]{6CB600}} \color[HTML]{F1F1F1} 0.303 & {\cellcolor[HTML]{FFB600}} \color[HTML]{000000} 0.138 & {\cellcolor[HTML]{FFE200}} \color[HTML]{000000} 0.170 & {\cellcolor[HTML]{E2F100}} \color[HTML]{000000} 0.214 \\
6k-8k & {\cellcolor[HTML]{76BB00}} \color[HTML]{F1F1F1} 0.296 & {\cellcolor[HTML]{DCEE00}} \color[HTML]{000000} 0.219 & {\cellcolor[HTML]{C2E100}} \color[HTML]{000000} 0.238 & {\cellcolor[HTML]{AED700}} \color[HTML]{000000} 0.254 & {\cellcolor[HTML]{FFA200}} \color[HTML]{000000} 0.122 & {\cellcolor[HTML]{FFDA00}} \color[HTML]{000000} 0.165 & {\cellcolor[HTML]{D2E900}} \color[HTML]{000000} 0.226 \\
8k-10k & {\cellcolor[HTML]{68B400}} \color[HTML]{F1F1F1} 0.306 & {\cellcolor[HTML]{EEF700}} \color[HTML]{000000} 0.205 & {\cellcolor[HTML]{CAE500}} \color[HTML]{000000} 0.233 & {\cellcolor[HTML]{C6E300}} \color[HTML]{000000} 0.235 & {\cellcolor[HTML]{FF8A00}} \color[HTML]{F1F1F1} 0.104 & {\cellcolor[HTML]{FFCA00}} \color[HTML]{000000} 0.152 & {\cellcolor[HTML]{FFDE00}} \color[HTML]{000000} 0.167 \\
10k-12k & {\cellcolor[HTML]{9ACD00}} \color[HTML]{000000} 0.269 & {\cellcolor[HTML]{E4F200}} \color[HTML]{000000} 0.213 & {\cellcolor[HTML]{DAED00}} \color[HTML]{000000} 0.221 & {\cellcolor[HTML]{BEDF00}} \color[HTML]{000000} 0.241 & {\cellcolor[HTML]{FF9E00}} \color[HTML]{000000} 0.119 & {\cellcolor[HTML]{FFD000}} \color[HTML]{000000} 0.157 & {\cellcolor[HTML]{FFFC00}} \color[HTML]{000000} 0.190 \\
12k-14k & {\cellcolor[HTML]{8AC500}} \color[HTML]{000000} 0.280 & {\cellcolor[HTML]{FEFF00}} \color[HTML]{000000} 0.193 & {\cellcolor[HTML]{D4EA00}} \color[HTML]{000000} 0.225 & {\cellcolor[HTML]{C4E200}} \color[HTML]{000000} 0.237 & {\cellcolor[HTML]{FF9E00}} \color[HTML]{000000} 0.119 & {\cellcolor[HTML]{FFA800}} \color[HTML]{000000} 0.127 & {\cellcolor[HTML]{FFE200}} \color[HTML]{000000} 0.171 \\
14k-16k & {\cellcolor[HTML]{CCE600}} \color[HTML]{000000} 0.231 & {\cellcolor[HTML]{FFEE00}} \color[HTML]{000000} 0.180 & {\cellcolor[HTML]{DEEF00}} \color[HTML]{000000} 0.217 & {\cellcolor[HTML]{C0E000}} \color[HTML]{000000} 0.240 & {\cellcolor[HTML]{FF9600}} \color[HTML]{000000} 0.114 & {\cellcolor[HTML]{FF9A00}} \color[HTML]{000000} 0.117 & {\cellcolor[HTML]{FFCC00}} \color[HTML]{000000} 0.154 \\
16k-24k & {\cellcolor[HTML]{9ACD00}} \color[HTML]{000000} 0.269 & {\cellcolor[HTML]{FF9800}} \color[HTML]{000000} 0.115 & {\cellcolor[HTML]{F6FB00}} \color[HTML]{000000} 0.200 & {\cellcolor[HTML]{B0D800}} \color[HTML]{000000} 0.252 & {\cellcolor[HTML]{FF8C00}} \color[HTML]{F1F1F1} 0.106 & {\cellcolor[HTML]{FF8800}} \color[HTML]{F1F1F1} 0.103 & {\cellcolor[HTML]{FFBA00}} \color[HTML]{000000} 0.141 \\
24k-32k & {\cellcolor[HTML]{B6DB00}} \color[HTML]{000000} 0.247 & {\cellcolor[HTML]{FF1400}} \color[HTML]{F1F1F1} 0.016 & {\cellcolor[HTML]{FFDE00}} \color[HTML]{000000} 0.168 & {\cellcolor[HTML]{ECF600}} \color[HTML]{000000} 0.207 & {\cellcolor[HTML]{FF8600}} \color[HTML]{F1F1F1} 0.101 & {\cellcolor[HTML]{FF8200}} \color[HTML]{F1F1F1} 0.098 & {\cellcolor[HTML]{FF9A00}} \color[HTML]{000000} 0.117 \\
32k-49k & {\cellcolor[HTML]{FFE200}} \color[HTML]{000000} 0.171 & {\cellcolor[HTML]{FF0200}} \color[HTML]{F1F1F1} 0.002 & {\cellcolor[HTML]{FFAA00}} \color[HTML]{000000} 0.128 & {\cellcolor[HTML]{F6FB00}} \color[HTML]{000000} 0.200 & {\cellcolor[HTML]{FF8C00}} \color[HTML]{F1F1F1} 0.106 & {\cellcolor[HTML]{FF9600}} \color[HTML]{000000} 0.114 & {\cellcolor[HTML]{FF8000}} \color[HTML]{F1F1F1} 0.097 \\
49k-65k & {\cellcolor[HTML]{FFFA00}} \color[HTML]{000000} 0.189 & {\cellcolor[HTML]{FF0000}} \color[HTML]{F1F1F1} 0.000 & {\cellcolor[HTML]{FF3400}} \color[HTML]{F1F1F1} 0.040 & {\cellcolor[HTML]{A0D000}} \color[HTML]{000000} 0.264 & {\cellcolor[HTML]{FF7C00}} \color[HTML]{F1F1F1} 0.094 & {\cellcolor[HTML]{FF1400}} \color[HTML]{F1F1F1} 0.016 & {\cellcolor[HTML]{FFBE00}} \color[HTML]{000000} 0.144 \\
\hline
\begin{tabular}[c]{@{}l@{}}{In-Domain}\\ {(0k-8k)}\end{tabular} & {\cellcolor[HTML]{2C9600}} \color[HTML]{F1F1F1} 0.351 & {\cellcolor[HTML]{92C900}} \color[HTML]{000000} 0.275 & {\cellcolor[HTML]{7EBF00}} \color[HTML]{000000} 0.290 & {\cellcolor[HTML]{72B900}} \color[HTML]{F1F1F1} 0.299 & {\cellcolor[HTML]{FFD600}} \color[HTML]{000000} 0.161 & {\cellcolor[HTML]{FEFF00}} \color[HTML]{000000} 0.193 & {\cellcolor[HTML]{D6EB00}} \color[HTML]{000000} 0.224 \\
\begin{tabular}[c]{@{}l@{}}{OOD}\\ {(8k-65k)}\end{tabular} & {\cellcolor[HTML]{9ECF00}} \color[HTML]{000000} 0.266 & {\cellcolor[HTML]{FFD600}} \color[HTML]{000000} 0.162 & {\cellcolor[HTML]{E6F300}} \color[HTML]{000000} 0.211 & {\cellcolor[HTML]{C2E100}} \color[HTML]{000000} 0.238 & {\cellcolor[HTML]{FF9200}} \color[HTML]{000000} 0.111 & {\cellcolor[HTML]{FFA800}} \color[HTML]{000000} 0.127 & {\cellcolor[HTML]{FFD200}} \color[HTML]{000000} 0.159 \\
\begin{tabular}[c]{@{}l@{}}{Long-OOD}\\ {(32k-65k)}\end{tabular} & {\cellcolor[HTML]{FFE800}} \color[HTML]{000000} 0.175 & {\cellcolor[HTML]{FF0200}} \color[HTML]{F1F1F1} 0.002 & {\cellcolor[HTML]{FF8E00}} \color[HTML]{F1F1F1} 0.108 & {\cellcolor[HTML]{E2F100}} \color[HTML]{000000} 0.215 & {\cellcolor[HTML]{FF8800}} \color[HTML]{F1F1F1} 0.103 & {\cellcolor[HTML]{FF7800}} \color[HTML]{F1F1F1} 0.091 & {\cellcolor[HTML]{FF8E00}} \color[HTML]{F1F1F1} 0.108 \\
\begin{tabular}[c]{@{}l@{}}{Full}\\ {(0k-65k)}\end{tabular} & {\cellcolor[HTML]{68B400}} \color[HTML]{F1F1F1} 0.306 & {\cellcolor[HTML]{E2F100}} \color[HTML]{000000} 0.215 & {\cellcolor[HTML]{B6DB00}} \color[HTML]{000000} 0.248 & {\cellcolor[HTML]{9CCE00}} \color[HTML]{000000} 0.266 & {\cellcolor[HTML]{FFB200}} \color[HTML]{000000} 0.135 & {\cellcolor[HTML]{FFD000}} \color[HTML]{000000} 0.158 & {\cellcolor[HTML]{FFFA00}} \color[HTML]{000000} 0.189 \\
\bottomrule

\end{tabular}}
\caption{Comparison of RMT and ARMT on the GR-100+ dataset for Gemma-3-1B-IT model, metric - ROUGE-L. ARMT consistently outperforms RMT model.}
\label{tab:gov_report_rmt_ablation}
\end{center}
\end{table*}

\section{ContractNLI Results}\label{app:cnli}
We also conducted the standard set of experiments on the ContractNLI dataset~\citep{koreeda-manning-2021-contractnli-dataset} from the SCROLLS benchmark~\citep{shaham-etal-2022-scrolls}. The results are presented in~\Cref{tab:cnli_best_with_pretrain}. Overall, the proposed training approach demonstrates robustness across different datasets and can be effectively applied to small real-world datasets without requiring additional synthetic data.
\begin{table*}[t]

\begin{center}
\resizebox{0.9\textwidth}{!}{
\begin{tabular}{lcccccccc}
\toprule
\begin{tabular}[c]{@{}l@{}}\textbf{Model\//}\\ \textbf{Lengths}\end{tabular} & \begin{tabular}[c]{@{}l@{}}\textbf{Base,}\\ \textbf{No Fine-}\\ \textbf{Tuning}\end{tabular} & \begin{tabular}[c]{@{}l@{}}\textbf{Base,}\\ \textbf{CNLI,}\\ \textbf{8k}\end{tabular} & \begin{tabular}[c]{@{}l@{}}\textbf{ARMT,}\\ \textbf{CNLI, 2k}\end{tabular} & \begin{tabular}[c]{@{}l@{}}\textbf{ARMT,}\\ \textbf{CNLI, 4k}\end{tabular} & \begin{tabular}[c]{@{}l@{}}\textbf{ARMT,}\\ \textbf{CNLI, 8k}\end{tabular} & \begin{tabular}[c]{@{}l@{}}\textbf{ARMT,}\\ \textbf{CNLI, 2k,}\\ \textbf{pretrain}\end{tabular} & \begin{tabular}[c]{@{}l@{}}\textbf{ARMT,}\\ \textbf{CNLI, 4k,}\\ \textbf{pretrain}\end{tabular} & \begin{tabular}[c]{@{}l@{}}\textbf{ARMT,}\\ \textbf{CNLI, 8k,}\\ \textbf{pretrain}\end{tabular} \\
\midrule
0k-1k & {\cellcolor[HTML]{FFE800}} \color[HTML]{000000} 0.427 & {\cellcolor[HTML]{7ABD00}} \color[HTML]{F1F1F1} 0.703 & {\cellcolor[HTML]{ACD600}} \color[HTML]{000000} 0.615 & {\cellcolor[HTML]{88C400}} \color[HTML]{000000} 0.678 & {\cellcolor[HTML]{76BB00}} \color[HTML]{F1F1F1} 0.710 & {\cellcolor[HTML]{82C100}} \color[HTML]{000000} 0.689 & {\cellcolor[HTML]{8AC500}} \color[HTML]{000000} 0.675 & {\cellcolor[HTML]{7ABD00}} \color[HTML]{F1F1F1} 0.703 \\
1k-2k & {\cellcolor[HTML]{FF2A00}} \color[HTML]{F1F1F1} 0.089 & {\cellcolor[HTML]{82C100}} \color[HTML]{000000} 0.690 & {\cellcolor[HTML]{8CC600}} \color[HTML]{000000} 0.672 & {\cellcolor[HTML]{80C000}} \color[HTML]{000000} 0.691 & {\cellcolor[HTML]{7CBE00}} \color[HTML]{000000} 0.701 & {\cellcolor[HTML]{8CC600}} \color[HTML]{000000} 0.671 & {\cellcolor[HTML]{6EB700}} \color[HTML]{F1F1F1} 0.724 & {\cellcolor[HTML]{68B400}} \color[HTML]{F1F1F1} 0.737 \\
2k-4k & {\cellcolor[HTML]{FF1200}} \color[HTML]{F1F1F1} 0.046 & {\cellcolor[HTML]{7CBE00}} \color[HTML]{000000} 0.700 & {\cellcolor[HTML]{E6F300}} \color[HTML]{000000} 0.512 & {\cellcolor[HTML]{88C400}} \color[HTML]{000000} 0.680 & {\cellcolor[HTML]{6CB600}} \color[HTML]{F1F1F1} 0.730 & {\cellcolor[HTML]{D4EA00}} \color[HTML]{000000} 0.542 & {\cellcolor[HTML]{7ABD00}} \color[HTML]{F1F1F1} 0.703 & {\cellcolor[HTML]{6CB600}} \color[HTML]{F1F1F1} 0.727 \\
4k-6k & {\cellcolor[HTML]{FF0000}} \color[HTML]{F1F1F1} 0.012 & {\cellcolor[HTML]{40A000}} \color[HTML]{F1F1F1} 0.806 & {\cellcolor[HTML]{FFBC00}} \color[HTML]{000000} 0.347 & {\cellcolor[HTML]{62B100}} \color[HTML]{F1F1F1} 0.747 & {\cellcolor[HTML]{52A900}} \color[HTML]{F1F1F1} 0.776 & {\cellcolor[HTML]{FFC200}} \color[HTML]{000000} 0.359 & {\cellcolor[HTML]{62B100}} \color[HTML]{F1F1F1} 0.747 & {\cellcolor[HTML]{58AC00}} \color[HTML]{F1F1F1} 0.765 \\
6k-8k & {\cellcolor[HTML]{FF6600}} \color[HTML]{F1F1F1} 0.196 & {\cellcolor[HTML]{008000}} \color[HTML]{F1F1F1} 0.922 & {\cellcolor[HTML]{FF8800}} \color[HTML]{F1F1F1} 0.255 & {\cellcolor[HTML]{58AC00}} \color[HTML]{F1F1F1} 0.765 & {\cellcolor[HTML]{168B00}} \color[HTML]{F1F1F1} 0.882 & {\cellcolor[HTML]{FF8800}} \color[HTML]{F1F1F1} 0.255 & {\cellcolor[HTML]{4CA600}} \color[HTML]{F1F1F1} 0.784 & {\cellcolor[HTML]{369B00}} \color[HTML]{F1F1F1} 0.824 \\
\hline
\begin{tabular}[c]{@{}l@{}}{Full}\\ {(0k-8k)}\end{tabular} & {\cellcolor[HTML]{FF3C00}} \color[HTML]{F1F1F1} 0.119 & {\cellcolor[HTML]{76BB00}} \color[HTML]{F1F1F1} 0.710 & {\cellcolor[HTML]{C0E000}} \color[HTML]{000000} 0.579 & {\cellcolor[HTML]{80C000}} \color[HTML]{000000} 0.692 & {\cellcolor[HTML]{70B800}} \color[HTML]{F1F1F1} 0.721 & {\cellcolor[HTML]{B4DA00}} \color[HTML]{000000} 0.599 & {\cellcolor[HTML]{74BA00}} \color[HTML]{F1F1F1} 0.714 & {\cellcolor[HTML]{68B400}} \color[HTML]{F1F1F1} 0.734 \\
\bottomrule
\end{tabular}}
\caption{Results on the ContractNLI (CNLI) dataset for Gemma-3-1B-IT model with ARMT after continuous pretraining, ROUGE-L. ARMT after continuous pretraining shows significantly better results on all splits.}
\label{tab:cnli_best_with_pretrain}
\end{center}
\end{table*}

\section{BABILong Results}\label{app:babilong}
To further validate the proposed ARMT fine-tuning approach, we conducted experiments on the BABILong benchmark~\cite{kuratov2024babilong}, which provides a training set suitable for fine-tuning. Due to the absence of a train set, we did not use other popular long-context benchmarks, such as RULER~\citep{hsieh2024ruler} and LongBench~\citep{bai2024longbench}. We trained both the base Gemma-3-1B-IT model and the ARMT model on sequences of up to 5120 tokens. For ARMT, we used a curriculum with 2, 3, and 5 segments of 1024 tokens each, while the base model was trained directly on sequences of length 5120. All models were trained jointly on tasks QA1-QA5 from the BABILong benchmark. The results are presented in~\Cref{tab:babilong_ablation_qa1_qa3,tab:babilong_ablation_qa4_qa5}. 

While the base model without fine-tuning achieves near-zero performance on long-context inputs, the fine-tuned base model generalizes to context lengths of up to 32k tokens. In contrast, the ARMT model achieves stronger performance on the long-context splits and generalizes to context lengths of up to 64k tokens.

\begin{table*}[t]

\begin{center}
\resizebox{1.0\textwidth}{!}{
\begin{tabular}{l|cccccccc|cccccccc|cccccccc}
\toprule
Task & \multicolumn{8}{c}{QA1} & \multicolumn{8}{c}{QA2} & \multicolumn{8}{c}{QA3} \\
\toprule
\begin{tabular}[c]{@{}l@{}}\textbf{Model\//}\\ \textbf{Length}\end{tabular} & 0k & 1k & 2k & 4k & 8k & 16k & 32k & 64k & 0k & 1k & 2k & 4k & 8k & 16k & 32k & 64k & 0k & 1k & 2k & 4k & 8k & 16k & 32k & 64k \\
\midrule
Gemma-3-1B-IT & {\cellcolor[HTML]{56AB00}} \color[HTML]{F1F1F1} 83 & {\cellcolor[HTML]{CCE600}} \color[HTML]{000000} 60 & {\cellcolor[HTML]{CCE600}} \color[HTML]{000000} 60 & {\cellcolor[HTML]{FFCC00}} \color[HTML]{000000} 40 & {\cellcolor[HTML]{FF2800}} \color[HTML]{F1F1F1} 8 & {\cellcolor[HTML]{FF1400}} \color[HTML]{F1F1F1} 4 & {\cellcolor[HTML]{FF0000}} \color[HTML]{F1F1F1} 0 & {\cellcolor[HTML]{FF0000}} \color[HTML]{F1F1F1} 0 & {\cellcolor[HTML]{FFF400}} \color[HTML]{000000} 48 & {\cellcolor[HTML]{FFF000}} \color[HTML]{000000} 47 & {\cellcolor[HTML]{FF7400}} \color[HTML]{F1F1F1} 23 & {\cellcolor[HTML]{FF2800}} \color[HTML]{F1F1F1} 8 & {\cellcolor[HTML]{FF1800}} \color[HTML]{F1F1F1} 5 & {\cellcolor[HTML]{FF0400}} \color[HTML]{F1F1F1} 1 & {\cellcolor[HTML]{FF0000}} \color[HTML]{F1F1F1} 0 & {\cellcolor[HTML]{FF0000}} \color[HTML]{F1F1F1} 0 & {\cellcolor[HTML]{FFC200}} \color[HTML]{000000} 38 & {\cellcolor[HTML]{FFAE00}} \color[HTML]{000000} 34 & {\cellcolor[HTML]{FF9800}} \color[HTML]{000000} 30 & {\cellcolor[HTML]{FF6000}} \color[HTML]{F1F1F1} 19 & {\cellcolor[HTML]{FF2800}} \color[HTML]{F1F1F1} 8 & {\cellcolor[HTML]{FF4200}} \color[HTML]{F1F1F1} 13 & {\cellcolor[HTML]{FF0000}} \color[HTML]{F1F1F1} 0 & {\cellcolor[HTML]{FF0000}} \color[HTML]{F1F1F1} 0 \\
Gemma-3-1B-IT,
 BABILong, 5k & {\cellcolor[HTML]{008000}} \color[HTML]{F1F1F1} 100 & {\cellcolor[HTML]{008000}} \color[HTML]{F1F1F1} 100 & {\cellcolor[HTML]{008000}} \color[HTML]{F1F1F1} 100 & {\cellcolor[HTML]{008000}} \color[HTML]{F1F1F1} 100 & {\cellcolor[HTML]{048200}} \color[HTML]{F1F1F1} 99 & {\cellcolor[HTML]{148A00}} \color[HTML]{F1F1F1} 96 & {\cellcolor[HTML]{1E8F00}} \color[HTML]{F1F1F1} 94 & {\cellcolor[HTML]{FF0E00}} \color[HTML]{F1F1F1} 3 & {\cellcolor[HTML]{229100}} \color[HTML]{F1F1F1} 93 & {\cellcolor[HTML]{048200}} \color[HTML]{F1F1F1} 99 & {\cellcolor[HTML]{008000}} \color[HTML]{F1F1F1} 100 & {\cellcolor[HTML]{0E8700}} \color[HTML]{F1F1F1} 97 & {\cellcolor[HTML]{2E9700}} \color[HTML]{F1F1F1} 91 & {\cellcolor[HTML]{4CA600}} \color[HTML]{F1F1F1} 85 & {\cellcolor[HTML]{A8D400}} \color[HTML]{000000} 67 & {\cellcolor[HTML]{FF0A00}} \color[HTML]{F1F1F1} 2 & {\cellcolor[HTML]{389C00}} \color[HTML]{F1F1F1} 89 & {\cellcolor[HTML]{188C00}} \color[HTML]{F1F1F1} 95 & {\cellcolor[HTML]{188C00}} \color[HTML]{F1F1F1} 95 & {\cellcolor[HTML]{229100}} \color[HTML]{F1F1F1} 93 & {\cellcolor[HTML]{2E9700}} \color[HTML]{F1F1F1} 91 & {\cellcolor[HTML]{389C00}} \color[HTML]{F1F1F1} 89 & {\cellcolor[HTML]{60B000}} \color[HTML]{F1F1F1} 81 & {\cellcolor[HTML]{FF2E00}} \color[HTML]{F1F1F1} 9 \\
ARMT,
 BABILong, 2k & {\cellcolor[HTML]{008000}} \color[HTML]{F1F1F1} 100 & {\cellcolor[HTML]{008000}} \color[HTML]{F1F1F1} 100 & {\cellcolor[HTML]{0E8700}} \color[HTML]{F1F1F1} 97 & {\cellcolor[HTML]{9ECF00}} \color[HTML]{000000} 69 & {\cellcolor[HTML]{FFDC00}} \color[HTML]{000000} 43 & {\cellcolor[HTML]{FF8E00}} \color[HTML]{F1F1F1} 28 & {\cellcolor[HTML]{FF8A00}} \color[HTML]{F1F1F1} 27 & {\cellcolor[HTML]{FF1400}} \color[HTML]{F1F1F1} 4 & {\cellcolor[HTML]{008000}} \color[HTML]{F1F1F1} 100 & {\cellcolor[HTML]{008000}} \color[HTML]{F1F1F1} 100 & {\cellcolor[HTML]{4CA600}} \color[HTML]{F1F1F1} 85 & {\cellcolor[HTML]{FFC600}} \color[HTML]{000000} 39 & {\cellcolor[HTML]{FF8A00}} \color[HTML]{F1F1F1} 27 & {\cellcolor[HTML]{FF8E00}} \color[HTML]{F1F1F1} 28 & {\cellcolor[HTML]{FF5600}} \color[HTML]{F1F1F1} 17 & {\cellcolor[HTML]{FF1400}} \color[HTML]{F1F1F1} 4 & {\cellcolor[HTML]{0A8500}} \color[HTML]{F1F1F1} 98 & {\cellcolor[HTML]{048200}} \color[HTML]{F1F1F1} 99 & {\cellcolor[HTML]{148A00}} \color[HTML]{F1F1F1} 96 & {\cellcolor[HTML]{C2E100}} \color[HTML]{000000} 62 & {\cellcolor[HTML]{FFC200}} \color[HTML]{000000} 38 & {\cellcolor[HTML]{FF6A00}} \color[HTML]{F1F1F1} 21 & {\cellcolor[HTML]{FF6A00}} \color[HTML]{F1F1F1} 21 & {\cellcolor[HTML]{FF1800}} \color[HTML]{F1F1F1} 5 \\
ARMT,
 BABILong, 3k & {\cellcolor[HTML]{008000}} \color[HTML]{F1F1F1} 100 & {\cellcolor[HTML]{008000}} \color[HTML]{F1F1F1} 100 & {\cellcolor[HTML]{048200}} \color[HTML]{F1F1F1} 99 & {\cellcolor[HTML]{4CA600}} \color[HTML]{F1F1F1} 85 & {\cellcolor[HTML]{C6E300}} \color[HTML]{000000} 61 & {\cellcolor[HTML]{FFE000}} \color[HTML]{000000} 44 & {\cellcolor[HTML]{FF5600}} \color[HTML]{F1F1F1} 17 & {\cellcolor[HTML]{FF4200}} \color[HTML]{F1F1F1} 13 & {\cellcolor[HTML]{048200}} \color[HTML]{F1F1F1} 99 & {\cellcolor[HTML]{048200}} \color[HTML]{F1F1F1} 99 & {\cellcolor[HTML]{289400}} \color[HTML]{F1F1F1} 92 & {\cellcolor[HTML]{7ABD00}} \color[HTML]{F1F1F1} 76 & {\cellcolor[HTML]{FFC200}} \color[HTML]{000000} 38 & {\cellcolor[HTML]{FFAE00}} \color[HTML]{000000} 34 & {\cellcolor[HTML]{FF5C00}} \color[HTML]{F1F1F1} 18 & {\cellcolor[HTML]{FF3C00}} \color[HTML]{F1F1F1} 12 & {\cellcolor[HTML]{188C00}} \color[HTML]{F1F1F1} 95 & {\cellcolor[HTML]{0A8500}} \color[HTML]{F1F1F1} 98 & {\cellcolor[HTML]{188C00}} \color[HTML]{F1F1F1} 95 & {\cellcolor[HTML]{60B000}} \color[HTML]{F1F1F1} 81 & {\cellcolor[HTML]{FF8A00}} \color[HTML]{F1F1F1} 27 & {\cellcolor[HTML]{FF5C00}} \color[HTML]{F1F1F1} 18 & {\cellcolor[HTML]{FF2200}} \color[HTML]{F1F1F1} 7 & {\cellcolor[HTML]{FF2200}} \color[HTML]{F1F1F1} 7 \\
ARMT,
 BABILong, 5k & {\cellcolor[HTML]{008000}} \color[HTML]{F1F1F1} 100 & {\cellcolor[HTML]{008000}} \color[HTML]{F1F1F1} 100 & {\cellcolor[HTML]{008000}} \color[HTML]{F1F1F1} 100 & {\cellcolor[HTML]{008000}} \color[HTML]{F1F1F1} 100 & {\cellcolor[HTML]{0A8500}} \color[HTML]{F1F1F1} 98 & {\cellcolor[HTML]{008000}} \color[HTML]{F1F1F1} 100 & {\cellcolor[HTML]{008000}} \color[HTML]{F1F1F1} 100 & {\cellcolor[HTML]{048200}} \color[HTML]{F1F1F1} 99 & {\cellcolor[HTML]{008000}} \color[HTML]{F1F1F1} 100 & {\cellcolor[HTML]{008000}} \color[HTML]{F1F1F1} 100 & {\cellcolor[HTML]{0A8500}} \color[HTML]{F1F1F1} 98 & {\cellcolor[HTML]{0A8500}} \color[HTML]{F1F1F1} 98 & {\cellcolor[HTML]{148A00}} \color[HTML]{F1F1F1} 96 & {\cellcolor[HTML]{42A100}} \color[HTML]{F1F1F1} 87 & {\cellcolor[HTML]{42A100}} \color[HTML]{F1F1F1} 87 & {\cellcolor[HTML]{50A800}} \color[HTML]{F1F1F1} 84 & {\cellcolor[HTML]{048200}} \color[HTML]{F1F1F1} 99 & {\cellcolor[HTML]{008000}} \color[HTML]{F1F1F1} 100 & {\cellcolor[HTML]{0A8500}} \color[HTML]{F1F1F1} 98 & {\cellcolor[HTML]{188C00}} \color[HTML]{F1F1F1} 95 & {\cellcolor[HTML]{2E9700}} \color[HTML]{F1F1F1} 91 & {\cellcolor[HTML]{42A100}} \color[HTML]{F1F1F1} 87 & {\cellcolor[HTML]{6AB500}} \color[HTML]{F1F1F1} 79 & {\cellcolor[HTML]{9ECF00}} \color[HTML]{000000} 69 \\
\bottomrule
\end{tabular}}
\caption{Results on the BABILong dataset for Gemma-3-1B-IT model with ARMT, tasks QA1-QA3. ARMT outperforms base model.}
\label{tab:babilong_ablation_qa1_qa3}
\end{center}
\end{table*}
\begin{table*}[t]

\begin{center}
\resizebox{1.0\textwidth}{!}{
\begin{tabular}{l|cccccccc|cccccccc|c}
\toprule
Task & \multicolumn{8}{c}{QA4} & \multicolumn{8}{c}{QA5} & Avg. \\
\toprule
\begin{tabular}[c]{@{}l@{}}\textbf{Model\//}\\ \textbf{Length}\end{tabular} & 0k & 1k & 2k & 4k & 8k & 16k & 32k & 64k & 0k & 1k & 2k & 4k & 8k & 16k & 32k & 64k & Avg. (QA1-QA5) \\
\midrule
Gemma-3-1B-IT & {\cellcolor[HTML]{FF9800}} \color[HTML]{000000} 30 & {\cellcolor[HTML]{FF3C00}} \color[HTML]{F1F1F1} 12 & {\cellcolor[HTML]{FF3800}} \color[HTML]{F1F1F1} 11 & {\cellcolor[HTML]{FF2200}} \color[HTML]{F1F1F1} 7 & {\cellcolor[HTML]{FF1E00}} \color[HTML]{F1F1F1} 6 & {\cellcolor[HTML]{FF4600}} \color[HTML]{F1F1F1} 14 & {\cellcolor[HTML]{FF0400}} \color[HTML]{F1F1F1} 1 & {\cellcolor[HTML]{FF0000}} \color[HTML]{F1F1F1} 0 & {\cellcolor[HTML]{98CC00}} \color[HTML]{000000} 70 & {\cellcolor[HTML]{FFC600}} \color[HTML]{000000} 39 & {\cellcolor[HTML]{EAF500}} \color[HTML]{000000} 54 & {\cellcolor[HTML]{FFCC00}} \color[HTML]{000000} 40 & {\cellcolor[HTML]{FF9E00}} \color[HTML]{000000} 31 & {\cellcolor[HTML]{FF6A00}} \color[HTML]{F1F1F1} 21 & {\cellcolor[HTML]{FF2200}} \color[HTML]{F1F1F1} 7 & {\cellcolor[HTML]{FF0000}} \color[HTML]{F1F1F1} 0 & {\cellcolor[HTML]{FF6E00}} \color[HTML]{F1F1F1} 22 \\
Gemma-3-1B-IT,
 BABILong, 5k & {\cellcolor[HTML]{188C00}} \color[HTML]{F1F1F1} 95 & {\cellcolor[HTML]{008000}} \color[HTML]{F1F1F1} 100 & {\cellcolor[HTML]{008000}} \color[HTML]{F1F1F1} 100 & {\cellcolor[HTML]{048200}} \color[HTML]{F1F1F1} 99 & {\cellcolor[HTML]{0E8700}} \color[HTML]{F1F1F1} 97 & {\cellcolor[HTML]{60B000}} \color[HTML]{F1F1F1} 81 & {\cellcolor[HTML]{B2D900}} \color[HTML]{000000} 65 & {\cellcolor[HTML]{FF1800}} \color[HTML]{F1F1F1} 5 & {\cellcolor[HTML]{0E8700}} \color[HTML]{F1F1F1} 97 & {\cellcolor[HTML]{008000}} \color[HTML]{F1F1F1} 100 & {\cellcolor[HTML]{048200}} \color[HTML]{F1F1F1} 99 & {\cellcolor[HTML]{0A8500}} \color[HTML]{F1F1F1} 98 & {\cellcolor[HTML]{0E8700}} \color[HTML]{F1F1F1} 97 & {\cellcolor[HTML]{188C00}} \color[HTML]{F1F1F1} 95 & {\cellcolor[HTML]{2E9700}} \color[HTML]{F1F1F1} 91 & {\cellcolor[HTML]{FF8000}} \color[HTML]{F1F1F1} 25 & {\cellcolor[HTML]{58AC00}} \color[HTML]{F1F1F1} 83 \\
ARMT,
 BABILong, 2k & {\cellcolor[HTML]{008000}} \color[HTML]{F1F1F1} 100 & {\cellcolor[HTML]{008000}} \color[HTML]{F1F1F1} 100 & {\cellcolor[HTML]{289400}} \color[HTML]{F1F1F1} 92 & {\cellcolor[HTML]{A8D400}} \color[HTML]{000000} 67 & {\cellcolor[HTML]{FFE000}} \color[HTML]{000000} 44 & {\cellcolor[HTML]{FFC600}} \color[HTML]{000000} 39 & {\cellcolor[HTML]{FFD600}} \color[HTML]{000000} 42 & {\cellcolor[HTML]{FF1E00}} \color[HTML]{F1F1F1} 6 & {\cellcolor[HTML]{048200}} \color[HTML]{F1F1F1} 99 & {\cellcolor[HTML]{048200}} \color[HTML]{F1F1F1} 99 & {\cellcolor[HTML]{0A8500}} \color[HTML]{F1F1F1} 98 & {\cellcolor[HTML]{46A300}} \color[HTML]{F1F1F1} 86 & {\cellcolor[HTML]{E6F300}} \color[HTML]{000000} 55 & {\cellcolor[HTML]{FFE600}} \color[HTML]{000000} 45 & {\cellcolor[HTML]{FFB800}} \color[HTML]{000000} 36 & {\cellcolor[HTML]{FF0E00}} \color[HTML]{F1F1F1} 3 & {\cellcolor[HTML]{D6EB00}} \color[HTML]{000000} 58 \\
ARMT,
 BABILong, 3k & {\cellcolor[HTML]{008000}} \color[HTML]{F1F1F1} 100 & {\cellcolor[HTML]{008000}} \color[HTML]{F1F1F1} 100 & {\cellcolor[HTML]{229100}} \color[HTML]{F1F1F1} 93 & {\cellcolor[HTML]{74BA00}} \color[HTML]{F1F1F1} 77 & {\cellcolor[HTML]{AED700}} \color[HTML]{000000} 66 & {\cellcolor[HTML]{C2E100}} \color[HTML]{000000} 62 & {\cellcolor[HTML]{FFB800}} \color[HTML]{000000} 36 & {\cellcolor[HTML]{FF3200}} \color[HTML]{F1F1F1} 10 & {\cellcolor[HTML]{048200}} \color[HTML]{F1F1F1} 99 & {\cellcolor[HTML]{048200}} \color[HTML]{F1F1F1} 99 & {\cellcolor[HTML]{048200}} \color[HTML]{F1F1F1} 99 & {\cellcolor[HTML]{289400}} \color[HTML]{F1F1F1} 92 & {\cellcolor[HTML]{94CA00}} \color[HTML]{000000} 71 & {\cellcolor[HTML]{FFB800}} \color[HTML]{000000} 36 & {\cellcolor[HTML]{FF4600}} \color[HTML]{F1F1F1} 14 & {\cellcolor[HTML]{FF0A00}} \color[HTML]{F1F1F1} 2 & {\cellcolor[HTML]{C2E100}} \color[HTML]{000000} 62 \\
ARMT,
 BABILong, 5k & {\cellcolor[HTML]{008000}} \color[HTML]{F1F1F1} 100 & {\cellcolor[HTML]{008000}} \color[HTML]{F1F1F1} 100 & {\cellcolor[HTML]{329900}} \color[HTML]{F1F1F1} 90 & {\cellcolor[HTML]{56AB00}} \color[HTML]{F1F1F1} 83 & {\cellcolor[HTML]{5CAE00}} \color[HTML]{F1F1F1} 82 & {\cellcolor[HTML]{6AB500}} \color[HTML]{F1F1F1} 79 & {\cellcolor[HTML]{6AB500}} \color[HTML]{F1F1F1} 79 & {\cellcolor[HTML]{5CAE00}} \color[HTML]{F1F1F1} 82 & {\cellcolor[HTML]{048200}} \color[HTML]{F1F1F1} 99 & {\cellcolor[HTML]{048200}} \color[HTML]{F1F1F1} 99 & {\cellcolor[HTML]{048200}} \color[HTML]{F1F1F1} 99 & {\cellcolor[HTML]{048200}} \color[HTML]{F1F1F1} 99 & {\cellcolor[HTML]{0A8500}} \color[HTML]{F1F1F1} 98 & {\cellcolor[HTML]{0A8500}} \color[HTML]{F1F1F1} 98 & {\cellcolor[HTML]{0E8700}} \color[HTML]{F1F1F1} 97 & {\cellcolor[HTML]{0E8700}} \color[HTML]{F1F1F1} 97 & {\cellcolor[HTML]{209000}} \color[HTML]{F1F1F1} 94 \\

\bottomrule
\end{tabular}}
\caption{Results on the BABILong dataset for Gemma-3-1B-IT model with ARMT, tasks QA4-QA5. ARMT outperforms base model.}
\label{tab:babilong_ablation_qa4_qa5}
\end{center}
\end{table*}

\section{Long-context Baselines}\label{app:baselines}
We compared the ARMT model with other long-context models, such as Mamba, Mamba-2, DeltaNet, and xLSTM; the results are presented in~\Cref{tab:mt_best_all_baselines}. The ARMT model shows comparable or better generalization on OOD and Long-OOD splits, outperforming other long-context baselines. However, Mamba, Mamba-2, DeltaNet, and xLSTM show better in-domain performance, likely due to their substantially more extensive pretraining and a larger number of parameters compared to ARMT. In contrast, the ARMT model is used only with relatively small language modeling pre-training.

We also conducted experiments with YaRN~\citep{ICLR2024_874a4d89} for context extension. Following the original paper setup, we applied YaRN to the SmolLM-2-360M-IT model and first adapted it on the PG19 dataset~\citep{raecompressive} with 32k token length for 400 steps with a total batch size of 64. After that we fine-tuned the adapted model on the GR-100+ and MT datasets; the results are presented in~\Cref{tab:gov_report_best_smollm_with_yarn,tab:mt_best_smollm_with_yarn}. The ARMT model outperforms the model with YaRN on both MT and GR-100+, while the base model outperforms the model with YaRN on MT dataset.
\begin{table*}[t]

\begin{center}
\resizebox{0.85\textwidth}{!}{
\begin{tabular}{lcccccccc}
\toprule
\begin{tabular}[c]{@{}l@{}}\textbf{Model\//}\\ \textbf{Lengths}\end{tabular} & \begin{tabular}[c]{@{}l@{}}\textbf{Base,}\\ \textbf{MT,}\\ \textbf{8k}\end{tabular} & \begin{tabular}[c]{@{}l@{}}\textbf{ARMT,}\\ \textbf{MT, 8k}\end{tabular} & \begin{tabular}[c]{@{}l@{}}\textbf{ARMT,}\\ \textbf{MT, 8k,}\\ \textbf{pretrain}\end{tabular}  & \begin{tabular}[c]{@{}l@{}}\textbf{Mamba-1.4B,}\\ \textbf{MT, 8k}\end{tabular} & \begin{tabular}[c]{@{}l@{}}\textbf{Mamba-2-1.3B,}\\ \textbf{MT, 8k}\end{tabular} & \begin{tabular}[c]{@{}l@{}}\textbf{DeltaNet-1.3B,}\\ \textbf{MT, 8k}\end{tabular} & \begin{tabular}[c]{@{}l@{}}\textbf{xLSTM-1.4B,}\\ \textbf{MT, 8k}\end{tabular} \\
\midrule
0k-1k & {\cellcolor[HTML]{5CAE00}} \color[HTML]{F1F1F1} 0.795 & {\cellcolor[HTML]{7EBF00}} \color[HTML]{000000} 0.756 & {\cellcolor[HTML]{7EBF00}} \color[HTML]{000000} 0.756 & {\cellcolor[HTML]{309800}} \color[HTML]{F1F1F1} 0.846 & {\cellcolor[HTML]{5CAE00}} \color[HTML]{F1F1F1} 0.795 & {\cellcolor[HTML]{74BA00}} \color[HTML]{F1F1F1} 0.769 & {\cellcolor[HTML]{74BA00}} \color[HTML]{F1F1F1} 0.769 \\
1k-2k & {\cellcolor[HTML]{62B100}} \color[HTML]{F1F1F1} 0.790 & {\cellcolor[HTML]{8AC500}} \color[HTML]{000000} 0.743 & {\cellcolor[HTML]{8AC500}} \color[HTML]{000000} 0.743 & {\cellcolor[HTML]{269300}} \color[HTML]{F1F1F1} 0.857 & {\cellcolor[HTML]{0E8700}} \color[HTML]{F1F1F1} 0.886 & {\cellcolor[HTML]{48A400}} \color[HTML]{F1F1F1} 0.819 & {\cellcolor[HTML]{58AC00}} \color[HTML]{F1F1F1} 0.800 \\
2k-4k & {\cellcolor[HTML]{70B800}} \color[HTML]{F1F1F1} 0.774 & {\cellcolor[HTML]{90C800}} \color[HTML]{000000} 0.736 & {\cellcolor[HTML]{90C800}} \color[HTML]{000000} 0.736 & {\cellcolor[HTML]{56AB00}} \color[HTML]{F1F1F1} 0.802 & {\cellcolor[HTML]{56AB00}} \color[HTML]{F1F1F1} 0.802 & {\cellcolor[HTML]{9ACD00}} \color[HTML]{000000} 0.726 & {\cellcolor[HTML]{80C000}} \color[HTML]{000000} 0.755 \\
4k-6k & {\cellcolor[HTML]{76BB00}} \color[HTML]{F1F1F1} 0.767 & {\cellcolor[HTML]{B0D800}} \color[HTML]{000000} 0.699 & {\cellcolor[HTML]{8CC600}} \color[HTML]{000000} 0.740 & {\cellcolor[HTML]{389C00}} \color[HTML]{F1F1F1} 0.836 & {\cellcolor[HTML]{68B400}} \color[HTML]{F1F1F1} 0.781 & {\cellcolor[HTML]{9ACD00}} \color[HTML]{000000} 0.726 & {\cellcolor[HTML]{46A300}} \color[HTML]{F1F1F1} 0.822 \\
6k-8k & {\cellcolor[HTML]{249200}} \color[HTML]{F1F1F1} 0.859 & {\cellcolor[HTML]{54AA00}} \color[HTML]{F1F1F1} 0.804 & {\cellcolor[HTML]{249200}} \color[HTML]{F1F1F1} 0.859 & {\cellcolor[HTML]{008000}} \color[HTML]{F1F1F1} 0.902 & {\cellcolor[HTML]{008000}} \color[HTML]{F1F1F1} 0.902 & {\cellcolor[HTML]{088400}} \color[HTML]{F1F1F1} 0.891 & {\cellcolor[HTML]{1C8E00}} \color[HTML]{F1F1F1} 0.870 \\
8k-10k & {\cellcolor[HTML]{2A9500}} \color[HTML]{F1F1F1} 0.852 & {\cellcolor[HTML]{82C100}} \color[HTML]{000000} 0.753 & {\cellcolor[HTML]{40A000}} \color[HTML]{F1F1F1} 0.827 & {\cellcolor[HTML]{148A00}} \color[HTML]{F1F1F1} 0.877 & {\cellcolor[HTML]{40A000}} \color[HTML]{F1F1F1} 0.827 & {\cellcolor[HTML]{209000}} \color[HTML]{F1F1F1} 0.864 & {\cellcolor[HTML]{209000}} \color[HTML]{F1F1F1} 0.864 \\
10k-12k & {\cellcolor[HTML]{1C8E00}} \color[HTML]{F1F1F1} 0.868 & {\cellcolor[HTML]{60B000}} \color[HTML]{F1F1F1} 0.791 & {\cellcolor[HTML]{1C8E00}} \color[HTML]{F1F1F1} 0.868 & {\cellcolor[HTML]{148A00}} \color[HTML]{F1F1F1} 0.879 & {\cellcolor[HTML]{0A8500}} \color[HTML]{F1F1F1} 0.890 & {\cellcolor[HTML]{309800}} \color[HTML]{F1F1F1} 0.846 & {\cellcolor[HTML]{0A8500}} \color[HTML]{F1F1F1} 0.890 \\
12k-14k & {\cellcolor[HTML]{64B200}} \color[HTML]{F1F1F1} 0.787 & {\cellcolor[HTML]{88C400}} \color[HTML]{000000} 0.745 & {\cellcolor[HTML]{76BB00}} \color[HTML]{F1F1F1} 0.766 & {\cellcolor[HTML]{6CB600}} \color[HTML]{F1F1F1} 0.777 & {\cellcolor[HTML]{76BB00}} \color[HTML]{F1F1F1} 0.766 & {\cellcolor[HTML]{80C000}} \color[HTML]{000000} 0.755 & {\cellcolor[HTML]{48A400}} \color[HTML]{F1F1F1} 0.819 \\
14k-16k & {\cellcolor[HTML]{78BC00}} \color[HTML]{F1F1F1} 0.764 & {\cellcolor[HTML]{BCDE00}} \color[HTML]{000000} 0.685 & {\cellcolor[HTML]{B2D900}} \color[HTML]{000000} 0.697 & {\cellcolor[HTML]{78BC00}} \color[HTML]{F1F1F1} 0.764 & {\cellcolor[HTML]{78BC00}} \color[HTML]{F1F1F1} 0.764 & {\cellcolor[HTML]{A8D400}} \color[HTML]{000000} 0.708 & {\cellcolor[HTML]{8CC600}} \color[HTML]{000000} 0.742 \\
16k-24k & {\cellcolor[HTML]{4EA700}} \color[HTML]{F1F1F1} 0.812 & {\cellcolor[HTML]{7EBF00}} \color[HTML]{000000} 0.757 & {\cellcolor[HTML]{7EBF00}} \color[HTML]{000000} 0.757 & {\cellcolor[HTML]{108800}} \color[HTML]{F1F1F1} 0.882 & {\cellcolor[HTML]{3C9E00}} \color[HTML]{F1F1F1} 0.833 & {\cellcolor[HTML]{7EBF00}} \color[HTML]{000000} 0.757 & {\cellcolor[HTML]{48A400}} \color[HTML]{F1F1F1} 0.819 \\
24k-32k & {\cellcolor[HTML]{A4D200}} \color[HTML]{000000} 0.713 & {\cellcolor[HTML]{A4D200}} \color[HTML]{000000} 0.713 & {\cellcolor[HTML]{4EA700}} \color[HTML]{F1F1F1} 0.812 & {\cellcolor[HTML]{56AB00}} \color[HTML]{F1F1F1} 0.802 & {\cellcolor[HTML]{7ABD00}} \color[HTML]{F1F1F1} 0.762 & {\cellcolor[HTML]{C8E400}} \color[HTML]{000000} 0.673 & {\cellcolor[HTML]{8AC500}} \color[HTML]{000000} 0.743 \\
32k-49k & {\cellcolor[HTML]{FFEE00}} \color[HTML]{000000} 0.591 & {\cellcolor[HTML]{76BB00}} \color[HTML]{F1F1F1} 0.765 & {\cellcolor[HTML]{60B000}} \color[HTML]{F1F1F1} 0.791 & {\cellcolor[HTML]{68B400}} \color[HTML]{F1F1F1} 0.783 & {\cellcolor[HTML]{96CB00}} \color[HTML]{000000} 0.730 & {\cellcolor[HTML]{FFD800}} \color[HTML]{000000} 0.565 & {\cellcolor[HTML]{FFBA00}} \color[HTML]{000000} 0.530 \\
49k-65k & {\cellcolor[HTML]{FF0000}} \color[HTML]{F1F1F1} 0.317 & {\cellcolor[HTML]{A4D200}} \color[HTML]{000000} 0.713 & {\cellcolor[HTML]{7ABD00}} \color[HTML]{F1F1F1} 0.762 & {\cellcolor[HTML]{FFB600}} \color[HTML]{000000} 0.525 & {\cellcolor[HTML]{E0F000}} \color[HTML]{000000} 0.644 & {\cellcolor[HTML]{FF5E00}} \color[HTML]{F1F1F1} 0.426 & {\cellcolor[HTML]{FF5E00}} \color[HTML]{F1F1F1} 0.426 \\
\hline
In-Domain (0k-8k) & {\cellcolor[HTML]{5AAD00}} \color[HTML]{F1F1F1} 0.797 & {\cellcolor[HTML]{86C300}} \color[HTML]{000000} 0.749 & {\cellcolor[HTML]{76BB00}} \color[HTML]{F1F1F1} 0.767 & {\cellcolor[HTML]{2E9700}} \color[HTML]{F1F1F1} 0.848 & {\cellcolor[HTML]{389C00}} \color[HTML]{F1F1F1} 0.837 & {\cellcolor[HTML]{62B100}} \color[HTML]{F1F1F1} 0.788 & {\cellcolor[HTML]{56AB00}} \color[HTML]{F1F1F1} 0.802 \\
OOD (8k-65k) & {\cellcolor[HTML]{A8D400}} \color[HTML]{000000} 0.709 & {\cellcolor[HTML]{8CC600}} \color[HTML]{000000} 0.741 & {\cellcolor[HTML]{68B400}} \color[HTML]{F1F1F1} 0.783 & {\cellcolor[HTML]{62B100}} \color[HTML]{F1F1F1} 0.788 & {\cellcolor[HTML]{6CB600}} \color[HTML]{F1F1F1} 0.777 & {\cellcolor[HTML]{B6DB00}} \color[HTML]{000000} 0.694 & {\cellcolor[HTML]{9ACD00}} \color[HTML]{000000} 0.724 \\
Long-OOD (32k-65k) & {\cellcolor[HTML]{FF7E00}} \color[HTML]{F1F1F1} 0.463 & {\cellcolor[HTML]{8CC600}} \color[HTML]{000000} 0.741 & {\cellcolor[HTML]{6CB600}} \color[HTML]{F1F1F1} 0.777 & {\cellcolor[HTML]{D0E800}} \color[HTML]{000000} 0.662 & {\cellcolor[HTML]{B8DC00}} \color[HTML]{000000} 0.690 & {\cellcolor[HTML]{FFA000}} \color[HTML]{000000} 0.500 & {\cellcolor[HTML]{FF8E00}} \color[HTML]{F1F1F1} 0.481 \\
Full (0k-65k) & {\cellcolor[HTML]{8CC600}} \color[HTML]{000000} 0.741 & {\cellcolor[HTML]{8AC500}} \color[HTML]{000000} 0.744 & {\cellcolor[HTML]{6CB600}} \color[HTML]{F1F1F1} 0.777 & {\cellcolor[HTML]{50A800}} \color[HTML]{F1F1F1} 0.810 & {\cellcolor[HTML]{5AAD00}} \color[HTML]{F1F1F1} 0.798 & {\cellcolor[HTML]{98CC00}} \color[HTML]{000000} 0.727 & {\cellcolor[HTML]{82C100}} \color[HTML]{000000} 0.752 \\
\bottomrule
\end{tabular}}
\caption{Results on the MT dataset for Gemma-3-1B-IT model with ARMT and for various baselines, EM. ARMT shows better length generalization than Mamba and DeltaNet and do not require training from scratch.}
\label{tab:mt_best_all_baselines}
\end{center}
\end{table*}
\begin{table*}[t]

\begin{center}
\resizebox{0.6\linewidth}{!}{
\begin{tabular}{lccccc}
\toprule
\begin{tabular}[c]{@{}l@{}}\textbf{Model\//}\\ \textbf{Lengths}\end{tabular} & 
\begin{tabular}[c]{@{}l@{}}\textbf{Base,}\\ \textbf{No Fine-}\\ \textbf{Tuning}\end{tabular} & \begin{tabular}[c]{@{}l@{}}\textbf{Base,}\\ \textbf{GR-100+, 8k}\end{tabular} & \begin{tabular}[c]{@{}l@{}}\textbf{Base, YaRN,}\\ \textbf{GR-100+, 8k}\end{tabular} & \begin{tabular}[c]{@{}l@{}}\textbf{ARMT,}\\ \textbf{GR-100+, 8k}\end{tabular} \\
\midrule
0k-1k & {\cellcolor[HTML]{FFCC00}} \color[HTML]{000000} 0.145 & {\cellcolor[HTML]{128900}} \color[HTML]{F1F1F1} 0.349 & {\cellcolor[HTML]{008000}} \color[HTML]{F1F1F1} 0.362 & {\cellcolor[HTML]{229100}} \color[HTML]{F1F1F1} 0.337 \\
1k-2k & {\cellcolor[HTML]{FFAC00}} \color[HTML]{000000} 0.122 & {\cellcolor[HTML]{048200}} \color[HTML]{F1F1F1} 0.358 & {\cellcolor[HTML]{068300}} \color[HTML]{F1F1F1} 0.357 & {\cellcolor[HTML]{54AA00}} \color[HTML]{F1F1F1} 0.302 \\
2k-4k & {\cellcolor[HTML]{FF9E00}} \color[HTML]{000000} 0.113 & {\cellcolor[HTML]{2C9600}} \color[HTML]{F1F1F1} 0.330 & {\cellcolor[HTML]{148A00}} \color[HTML]{F1F1F1} 0.347 & {\cellcolor[HTML]{8EC700}} \color[HTML]{000000} 0.261 \\
4k-6k & {\cellcolor[HTML]{FF8000}} \color[HTML]{F1F1F1} 0.091 & {\cellcolor[HTML]{42A100}} \color[HTML]{F1F1F1} 0.314 & {\cellcolor[HTML]{46A300}} \color[HTML]{F1F1F1} 0.312 & {\cellcolor[HTML]{9ACD00}} \color[HTML]{000000} 0.253 \\
6k-8k & {\cellcolor[HTML]{FF8400}} \color[HTML]{F1F1F1} 0.094 & {\cellcolor[HTML]{6AB500}} \color[HTML]{F1F1F1} 0.286 & {\cellcolor[HTML]{66B300}} \color[HTML]{F1F1F1} 0.289 & {\cellcolor[HTML]{C0E000}} \color[HTML]{000000} 0.225 \\
8k-10k & {\cellcolor[HTML]{FF7C00}} \color[HTML]{F1F1F1} 0.089 & {\cellcolor[HTML]{B6DB00}} \color[HTML]{000000} 0.233 & {\cellcolor[HTML]{ACD600}} \color[HTML]{000000} 0.239 & {\cellcolor[HTML]{C6E300}} \color[HTML]{000000} 0.221 \\
10k-12k & {\cellcolor[HTML]{FF3600}} \color[HTML]{F1F1F1} 0.039 & {\cellcolor[HTML]{FF1C00}} \color[HTML]{F1F1F1} 0.021 & {\cellcolor[HTML]{FFFA00}} \color[HTML]{000000} 0.177 & {\cellcolor[HTML]{D4EA00}} \color[HTML]{000000} 0.211 \\
12k-14k & {\cellcolor[HTML]{FF3600}} \color[HTML]{F1F1F1} 0.039 & {\cellcolor[HTML]{FF3000}} \color[HTML]{F1F1F1} 0.035 & {\cellcolor[HTML]{FF5E00}} \color[HTML]{F1F1F1} 0.067 & {\cellcolor[HTML]{CEE700}} \color[HTML]{000000} 0.215 \\
14k-16k & {\cellcolor[HTML]{FF3600}} \color[HTML]{F1F1F1} 0.039 & {\cellcolor[HTML]{FF4400}} \color[HTML]{F1F1F1} 0.049 & {\cellcolor[HTML]{FF2000}} \color[HTML]{F1F1F1} 0.024 & {\cellcolor[HTML]{D6EB00}} \color[HTML]{000000} 0.210 \\
16k-24k & {\cellcolor[HTML]{FF4200}} \color[HTML]{F1F1F1} 0.048 & {\cellcolor[HTML]{FF3400}} \color[HTML]{F1F1F1} 0.037 & {\cellcolor[HTML]{FF1800}} \color[HTML]{F1F1F1} 0.017 & {\cellcolor[HTML]{CAE500}} \color[HTML]{000000} 0.219 \\
24k-32k & {\cellcolor[HTML]{FF3A00}} \color[HTML]{F1F1F1} 0.042 & {\cellcolor[HTML]{FF4C00}} \color[HTML]{F1F1F1} 0.054 & {\cellcolor[HTML]{FF0400}} \color[HTML]{F1F1F1} 0.003 & {\cellcolor[HTML]{FAFD00}} \color[HTML]{000000} 0.185 \\
32k-49k & {\cellcolor[HTML]{FF3E00}} \color[HTML]{F1F1F1} 0.044 & {\cellcolor[HTML]{FF5E00}} \color[HTML]{F1F1F1} 0.067 & {\cellcolor[HTML]{FF0000}} \color[HTML]{F1F1F1} 0.000 & {\cellcolor[HTML]{FFF000}} \color[HTML]{000000} 0.171 \\
49k-65k & {\cellcolor[HTML]{FF7C00}} \color[HTML]{F1F1F1} 0.088 & {\cellcolor[HTML]{FF7C00}} \color[HTML]{F1F1F1} 0.088 & {\cellcolor[HTML]{FF0000}} \color[HTML]{F1F1F1} 0.000 & {\cellcolor[HTML]{FF8600}} \color[HTML]{F1F1F1} 0.096 \\
\hline
\begin{tabular}[c]{@{}l@{}}{In-Domain}\\ {(0k-8k)}\end{tabular} & {\cellcolor[HTML]{FF9E00}} \color[HTML]{000000} 0.113 & {\cellcolor[HTML]{309800}} \color[HTML]{F1F1F1} 0.327 & {\cellcolor[HTML]{289400}} \color[HTML]{F1F1F1} 0.334 & {\cellcolor[HTML]{7ABD00}} \color[HTML]{F1F1F1} 0.275 \\
\begin{tabular}[c]{@{}l@{}}{OOD}\\ {(8k-65k)}\end{tabular} & {\cellcolor[HTML]{FF4600}} \color[HTML]{F1F1F1} 0.050 & {\cellcolor[HTML]{FF6800}} \color[HTML]{F1F1F1} 0.074 & {\cellcolor[HTML]{FF8000}} \color[HTML]{F1F1F1} 0.091 & {\cellcolor[HTML]{D6EB00}} \color[HTML]{000000} 0.211 \\
\begin{tabular}[c]{@{}l@{}}{Long-OOD}\\ {(32k-65k)}\end{tabular} & {\cellcolor[HTML]{FF4C00}} \color[HTML]{F1F1F1} 0.054 & {\cellcolor[HTML]{FF6400}} \color[HTML]{F1F1F1} 0.072 & {\cellcolor[HTML]{FF0000}} \color[HTML]{F1F1F1} 0.000 & {\cellcolor[HTML]{FFD800}} \color[HTML]{000000} 0.154 \\
\begin{tabular}[c]{@{}l@{}}{Full}\\ {(0k-65k)}\end{tabular} & {\cellcolor[HTML]{FF7000}} \color[HTML]{F1F1F1} 0.080 & {\cellcolor[HTML]{EEF700}} \color[HTML]{000000} 0.192 & {\cellcolor[HTML]{DEEF00}} \color[HTML]{000000} 0.205 & {\cellcolor[HTML]{AAD500}} \color[HTML]{000000} 0.241 \\
\bottomrule
\end{tabular}}

\end{center}
\caption{Results on the GR-100+ dataset for SmolLM-2-360M-IT model with YaRN, ROUGE-L. Model with YaRN shows worse performance than the ARMT model.}
\label{tab:gov_report_best_smollm_with_yarn}
\end{table*}
\begin{table*}[t]

\begin{center}
\resizebox{0.6\linewidth}{!}{
\begin{tabular}{lccccc}
\toprule
\begin{tabular}[c]{@{}l@{}}\textbf{Model\//}\\ \textbf{Lengths}\end{tabular} & 
\begin{tabular}[c]{@{}l@{}}\textbf{Base,}\\ \textbf{No Fine-}\\ \textbf{Tuning}\end{tabular} & \begin{tabular}[c]{@{}l@{}}\textbf{Base,}\\ \textbf{MT, 8k}\end{tabular} & \begin{tabular}[c]{@{}l@{}}\textbf{Base, YaRN,}\\ \textbf{MT, 8k}\end{tabular} & \begin{tabular}[c]{@{}l@{}}\textbf{ARMT,}\\ \textbf{MT, 8k}\end{tabular} \\
\midrule
0k-1k & {\cellcolor[HTML]{FF0000}} \color[HTML]{F1F1F1} 0.000 & {\cellcolor[HTML]{188C00}} \color[HTML]{F1F1F1} 0.795 & {\cellcolor[HTML]{FFCA00}} \color[HTML]{000000} 0.333 & {\cellcolor[HTML]{289400}} \color[HTML]{F1F1F1} 0.769 \\
1k-2k & {\cellcolor[HTML]{FF0000}} \color[HTML]{F1F1F1} 0.000 & {\cellcolor[HTML]{1C8E00}} \color[HTML]{F1F1F1} 0.790 & {\cellcolor[HTML]{FF9C00}} \color[HTML]{000000} 0.257 & {\cellcolor[HTML]{50A800}} \color[HTML]{F1F1F1} 0.705 \\
2k-4k & {\cellcolor[HTML]{FF0000}} \color[HTML]{F1F1F1} 0.000 & {\cellcolor[HTML]{3C9E00}} \color[HTML]{F1F1F1} 0.736 & {\cellcolor[HTML]{FFA000}} \color[HTML]{000000} 0.264 & {\cellcolor[HTML]{54AA00}} \color[HTML]{F1F1F1} 0.698 \\
4k-6k & {\cellcolor[HTML]{FF0000}} \color[HTML]{F1F1F1} 0.000 & {\cellcolor[HTML]{5CAE00}} \color[HTML]{F1F1F1} 0.685 & {\cellcolor[HTML]{FFB800}} \color[HTML]{000000} 0.301 & {\cellcolor[HTML]{7EBF00}} \color[HTML]{000000} 0.630 \\
6k-8k & {\cellcolor[HTML]{FF0000}} \color[HTML]{F1F1F1} 0.000 & {\cellcolor[HTML]{008000}} \color[HTML]{F1F1F1} 0.837 & {\cellcolor[HTML]{FF9800}} \color[HTML]{000000} 0.250 & {\cellcolor[HTML]{2E9700}} \color[HTML]{F1F1F1} 0.761 \\
8k-10k & {\cellcolor[HTML]{FF0000}} \color[HTML]{F1F1F1} 0.000 & {\cellcolor[HTML]{A4D200}} \color[HTML]{000000} 0.568 & {\cellcolor[HTML]{FFA600}} \color[HTML]{000000} 0.272 & {\cellcolor[HTML]{0C8600}} \color[HTML]{F1F1F1} 0.815 \\
10k-12k & {\cellcolor[HTML]{FF0600}} \color[HTML]{F1F1F1} 0.011 & {\cellcolor[HTML]{FF0000}} \color[HTML]{F1F1F1} 0.000 & {\cellcolor[HTML]{FF6A00}} \color[HTML]{F1F1F1} 0.176 & {\cellcolor[HTML]{369B00}} \color[HTML]{F1F1F1} 0.747 \\
12k-14k & {\cellcolor[HTML]{FF0000}} \color[HTML]{F1F1F1} 0.000 & {\cellcolor[HTML]{FF0000}} \color[HTML]{F1F1F1} 0.000 & {\cellcolor[HTML]{FF4000}} \color[HTML]{F1F1F1} 0.106 & {\cellcolor[HTML]{329900}} \color[HTML]{F1F1F1} 0.755 \\
14k-16k & {\cellcolor[HTML]{FF0000}} \color[HTML]{F1F1F1} 0.000 & {\cellcolor[HTML]{FF0000}} \color[HTML]{F1F1F1} 0.000 & {\cellcolor[HTML]{FF0000}} \color[HTML]{F1F1F1} 0.000 & {\cellcolor[HTML]{7EBF00}} \color[HTML]{000000} 0.629 \\
16k-24k & {\cellcolor[HTML]{FF0000}} \color[HTML]{F1F1F1} 0.000 & {\cellcolor[HTML]{FF0000}} \color[HTML]{F1F1F1} 0.000 & {\cellcolor[HTML]{FF0000}} \color[HTML]{F1F1F1} 0.000 & {\cellcolor[HTML]{4AA500}} \color[HTML]{F1F1F1} 0.715 \\
24k-32k & {\cellcolor[HTML]{FF0C00}} \color[HTML]{F1F1F1} 0.020 & {\cellcolor[HTML]{FF0600}} \color[HTML]{F1F1F1} 0.010 & {\cellcolor[HTML]{FF0000}} \color[HTML]{F1F1F1} 0.000 & {\cellcolor[HTML]{3E9F00}} \color[HTML]{F1F1F1} 0.733 \\
32k-49k & {\cellcolor[HTML]{FF0400}} \color[HTML]{F1F1F1} 0.009 & {\cellcolor[HTML]{FF0000}} \color[HTML]{F1F1F1} 0.000 & {\cellcolor[HTML]{FF0000}} \color[HTML]{F1F1F1} 0.000 & {\cellcolor[HTML]{40A000}} \color[HTML]{F1F1F1} 0.730 \\
49k-65k & {\cellcolor[HTML]{FF0000}} \color[HTML]{F1F1F1} 0.000 & {\cellcolor[HTML]{FF0000}} \color[HTML]{F1F1F1} 0.000 & {\cellcolor[HTML]{FF0000}} \color[HTML]{F1F1F1} 0.000 & {\cellcolor[HTML]{FF1200}} \color[HTML]{F1F1F1} 0.030 \\
\hline
\begin{tabular}[c]{@{}l@{}}{In-Domain}\\ {(0k-8k)}\end{tabular} & {\cellcolor[HTML]{FF0000}} \color[HTML]{F1F1F1} 0.000 & {\cellcolor[HTML]{289400}} \color[HTML]{F1F1F1} 0.771 & {\cellcolor[HTML]{FFA800}} \color[HTML]{000000} 0.277 & {\cellcolor[HTML]{4AA500}} \color[HTML]{F1F1F1} 0.714 \\
\begin{tabular}[c]{@{}l@{}}{OOD}\\ {(8k-65k)}\end{tabular} & {\cellcolor[HTML]{FF0200}} \color[HTML]{F1F1F1} 0.005 & {\cellcolor[HTML]{FF2200}} \color[HTML]{F1F1F1} 0.058 & {\cellcolor[HTML]{FF2200}} \color[HTML]{F1F1F1} 0.059 & {\cellcolor[HTML]{76BB00}} \color[HTML]{F1F1F1} 0.643 \\
\begin{tabular}[c]{@{}l@{}}{Long-OOD}\\ {(32k-65k)}\end{tabular} & {\cellcolor[HTML]{FF0200}} \color[HTML]{F1F1F1} 0.005 & {\cellcolor[HTML]{FF0000}} \color[HTML]{F1F1F1} 0.000 & {\cellcolor[HTML]{FF0000}} \color[HTML]{F1F1F1} 0.000 & {\cellcolor[HTML]{FFF600}} \color[HTML]{000000} 0.403 \\
\begin{tabular}[c]{@{}l@{}}{Full}\\ {(0k-65k)}\end{tabular} & {\cellcolor[HTML]{FF0000}} \color[HTML]{F1F1F1} 0.003 & {\cellcolor[HTML]{FFBE00}} \color[HTML]{000000} 0.313 & {\cellcolor[HTML]{FF5200}} \color[HTML]{F1F1F1} 0.137 & {\cellcolor[HTML]{66B300}} \color[HTML]{F1F1F1} 0.668 \\
\bottomrule
\end{tabular}}

\end{center}
\caption{Results on the MT dataset for SmolLM-2-360M-IT model, metric - EM. Model with YaRN shows worse performance than the base model and the ARMT model.}
\label{tab:mt_best_smollm_with_yarn}
\end{table*}

\end{document}